\documentclass[12pt]{article}

\usepackage{amssymb}
\usepackage[T1]{fontenc}
\usepackage{lmodern}
\usepackage{fix-cm}
\usepackage{mathtools}

\usepackage[inline]{enumitem} 

\usepackage{tikz}
\usetikzlibrary{positioning, arrows.meta, shadows, calc}
\usetikzlibrary{arrows.meta, positioning}
\usetikzlibrary{calc}\usetikzlibrary{positioning}

\usetikzlibrary{backgrounds}

\usetikzlibrary{patterns}

\usepackage{tcolorbox}

\usepackage{booktabs,tabularx,threeparttable}
\newcolumntype{Y}{>{\raggedright\arraybackslash}X}

\newtheorem{assumption}{Assumption}
\newtheorem{lemma}{Lemma}
\newtheorem{theorem}{Theorem}
\newtheorem{definition}{Definition}

\newtheorem{corollary}{Corollary}
\newtheorem{proposition}{Proposition}

\newenvironment{sizeddisplay}[1]
 {\par\nopagebreak#1\noindent\ignorespaces}
 {\nopagebreak\ignorespacesafterend}

\newcommand{\ltxlabel}[1]{\ltx@label{#1}}

\newcommand{\Var}{{\mbox{Var}}}

\DeclareMathOperator{\prob}{Pr}

\newcommand{\R}{{\rm I\!R}}

\def\EE{\mathbb{E}}

\newcommand\independent{\protect\mathpalette{\protect\independenT}{\perp}}
\def\independenT#1#2{\mathrel{\rlap{$#1#2$}\mkern2mu{#1#2}}}

\newcounter{lcnt}

\newcommand{\Rnum}[1]{\uppercase\expandafter{\romannumeral #1\relax}}

\def\begar{$$\begin{array}{lll}}
\def\endar{\end{array}$$}
\def\begarlab{\begin{equation} \begin{array}{lll} \label}
\def\endarlab{\end{array} \end{equation}}
\def\argmax{\text{argmax}}
\def\argmin{\text{argmin}}

\def\ds1{{\mathrm{1 \hspace{-2.6pt} I}}}

\def\calA{{\cal A}}

\def\calH{{\cal H}}

\def\calP{{\cal P}}
\def\calQ{{\cal Q}}

\def\calS{{\cal S}}
\def\calT{{\cal T}}

\def\calV{{\cal V}}
\def\calW{{\cal W}}
\def\calX{{\cal X}}

\def\calT{{\cal T}}

\def\given{\, | \,}

\usepackage{xr}

\newcommand{\smallO}{ \scalebox{0.7}{$\mathcal{O}$}}

\def\independenT#1#2{\mathrel{\rlap{$#1#2$}\mkern2mu{#1#2}}}

\usepackage{enumitem}
\usepackage{hyperref}  
\usepackage{cleveref}  
\hypersetup{
    colorlinks=true,
    linkcolor=red,      
    citecolor=blue,     
    urlcolor=blue       
}

\usepackage[ruled,vlined]{algorithm2e} 
\usepackage{algpseudocode} 
\usepackage{float}
\usepackage{graphicx}    
\usepackage{subcaption}
\usepackage[left=0.9in,right=0.9in,top=1.1in,bottom=1.1in]{geometry}
\usepackage{authblk}

\usepackage{natbib}
 \bibpunct[, ]{(}{)}{,}{a}{}{,}%
\title{\bf Offline Dynamic Inventory and Pricing Strategy: Addressing Censored and Dependent Demand}

\author[1]{Korel Gundem}
\author[1]{Zhengling Qi}
\affil[1]{The George Washington University, Washington, DC}

\date{}
\begin{document}

\maketitle

\begin{abstract}
    In this paper, we study the offline sequential feature-based pricing and inventory control problem where the current demand depends on the past demand levels and any demand exceeding the available inventory is lost. Our goal is to leverage the offline dataset, consisting of past prices, ordering quantities, inventory levels, covariates, and censored sales levels, to estimate the optimal pricing and inventory control policy that maximizes long-term profit. While the underlying dynamic without censoring can be modeled by Markov decision process (MDP), the primary obstacle arises from the observed process where demand censoring is present, resulting in missing profit information, the failure of the Markov property, and a non-stationary optimal policy. To overcome these challenges, we first approximate the optimal policy by solving a high-order MDP characterized by the number of consecutive censoring instances, which ultimately boils down to solving a specialized Bellman equation tailored for this problem. Inspired by offline reinforcement learning and survival analysis, we propose two novel data-driven algorithms for solving these Bellman equations and, thus, estimate the optimal policy. Furthermore, we establish finite-sample regret bounds to validate the effectiveness of these algorithms. Finally, we conduct numerical experiments to demonstrate the efficacy of our algorithms in estimating the optimal policy. To the best of our knowledge, this is the first data-driven approach to learning optimal pricing and inventory control policies in a sequential decision-making environment characterized by censored and dependent demand. The implementations of the proposed algorithms are available at \url{https://github.com/gundemkorel/Inventory_Pricing_Control}
\end{abstract}

\baselineskip=21pt


\section{Introduction}
Effective inventory and pricing strategies are central to firm profitability. To identify profit-maximizing policies, firms often rely on online experiments such as A/B tests. Yet in operational settings, such experiments are costly and risky: sub-optimal prices or inventory decisions can erode revenue and disrupt operations. At the same time, firms possess extensive historical data, involving prices, inventory levels, sales, and contextual features, that remain underutilized. Leveraging this fixed observational data for policy design, an approach often termed offline learning, offers a practical alternative that avoids experimental risk while enabling data-driven operational improvement. 

However, offline learning faces two structural challenges in developing inventory and pricing systems. The first challenge is demand censoring. In lost-sales environments, realized sales are the minimum of true demand and available inventory. Any demand above inventory is unobserved, systematically obscuring the revenue potential during stockouts. Therefore, treating observed sales as demand induces biased profitability estimates and can trigger a feedback loop of increasingly sub-optimal pricing and inventory decisions.

The second challenge arises from the interaction between censoring and dependent demand. In many settings, demand is path-dependent, with current customer experiences shaping future demand through mechanisms such as brand loyalty and network effects. If demand were fully observed, the dependent process could be modeled using standard Markov dynamics. Under censoring, however, the key driver of future demand—today’s unobserved demand—remains hidden, violating the Markov property from the retailer’s perspective. Because the retailer cannot observe the true current state, it cannot reliably anticipate its evolution. As a result, policies developed under Markovian assumptions break down in this setting, undermining long-term profit maximization.

Taken together, this implies that any robust framework for policy optimization using historical data must simultaneously address the challenges induced by censoring and the complex, non-Markovian dynamics resulting from its interaction with dependent demand. Failing to do so leaves a significant amount of revenue on the table.

\subsection{Motivating Example}

To illustrate the stakes, consider a major grocery retailer such as Target or Kroger managing ready-to-eat meals and fresh produce. Demand for these products is highly responsive to price as well as external drivers like day-of-week patterns, weather, and in-store promotions. Warm weekends or advertised discounts, for example, reliably generate surges in foot traffic and purchases.
When a high-velocity item (e.g., rotisserie chicken or packaged salads) is discounted during such periods, shelf inventory often depletes well before the end of the day. After a stockout, the retailer observes only sales up to the on-hand inventory, not the true underlying demand. Customers who arrive later either leave empty-handed or switch to competing stores, causing the data to systematically understate the revenue potential in high-demand states, which is the hallmark of demand censoring.

This challenge is amplified by the dependent demand. Grocery demand exhibits inertia: strong sales today—driven by promotions, traffic, or evolving habits—typically signal elevated demand tomorrow. When a stockout occurs, the retailer loses visibility not only into today’s true demand but also into a key signal required to forecast future demand. 
This highlights the need for a unified framework that jointly addresses both censoring and demand dependence.

\subsection{Problem Statement,  Main Results and Contributions}

Most existing offline policy learning methods are typically tailored to address one challenge at a time: some assume that the demand is observable even under censoring, while others account for censoring but assume that demand is independently distributed (given prices) across periods. These approaches are well-suited to their respective modeling assumptions and have advanced the field considerably. However, in practice, as illustrated in the motivating example, censoring and temporal dependence might arise jointly. In such cases, policy learning methods designed under one of these challenges are inadequate, leading to policies that deviate from managerial objectives. Thus, there is a need for a more general framework that simultaneously addresses censoring and demand dependence, providing a unified basis for accurately designing optimal pricing and inventory policies in real-world settings.

Motivated by this, in this paper, we address the problem of offline policy optimization for joint pricing and inventory control while accounting for censored and dependent demand. We consider a firm with access to a historical dataset comprising past prices, inventory decisions, contextual market features, and realized sales, but lacks a known underlying demand process. In particular, this demand process is shaped by two distinct features: it is \emph{censored} and exhibits \emph{temporal dependence}. The firm's objective is to leverage this fixed, observational dataset to learn a feature-based pricing and inventory control policy that maximizes long-term expected profit. In particular, our contributions are threefold.

(i) A General Modeling Framework. We begin by modeling the underlying process where there is no censoring as a standard Markov Decision Process (MDP). We then formally define the observed process experienced by the firm, where demand is censored. A key theoretical insight of our work is demonstrating that this observed process, while not Markovian in the observed state, inherits a high-order Markov property determined by the duration of consecutive censoring. Building on this insight, we characterize the structure of the optimal policy under the observed process by a set of specialized Bellman equations. These equations explicitly account for censoring, providing the theoretical foundation necessary to develop offline policy learning techniques to optimize inventory and pricing decisions under censored and dependent demand.

(ii) Two Practical Algorithms. Bridging theory and practice, we develop two implementable algorithms for policy optimization based on the derived specialized Bellman equations. The first algorithm, named \textit{Censored Fitted Q-Iteration (C-FQI)}, operationalizes our theoretical framework to estimate the optimal pricing and inventory policy from historical logs. Recognizing that historical data is often sparse in critical regions (e.g., long period of stockouts), we further introduce \textit{Pessimistic Censored Fitted Q-Iteration (PC-FQI)}. This algorithm integrates uncertainty quantification to doubt the model in data-poor regions, preventing the dangerous overestimation of profits that plagues standard offline RL methods and ensuring robust performance even under distributional shift. To empirically test our algorithms, we conduct a simulation study and the result confirms the effectiveness of our algorithms in optimizing pricing and inventory control strategies.

(iii) Finite-sample Regret Guarantees. We establish finite-sample regret bounds for the proposed algorithms in censored, dependent environments. Our analysis adopts a layered structure, progressively relaxing assumptions to isolate the specific drivers of sub-optimality. Specifically, we explicitly decompose the total regret into three structural terms: a \emph{statistical} error determined by the model complexity relative to the dataset size; an \emph{unidentifiable regret} arising from blind spots where the optimal policy induces censoring sequences longer than any observed in history; and \emph{alignment penalties} that quantify the profit sacrificed to satisfy operational safety constraints during the high number of sequential stockouts. This decomposition turns theory into a strategic roadmap, helping managers make informed decisions about model complexity, data constraints, and the cost of safe inventory recovery.

\subsection{Literature Review}
Offline learning is valuable when online exploration is costly or infeasible and can also provide warm starts for online methods. Yet, for sequential \emph{joint} pricing and inventory control, existing offline work largely addresses only one component (pricing or inventory), typically in single-period settings and/or without accounting for demand censoring, leaving the dependent-and-censored regime largely unexplored. On the inventory side, \cite{levi2007provably,cheung2019sampling} develop multi-period data-driven inventory control methods based on sample average approximation (SAA), but these approaches are ill-suited for our setting because price is an additional decision that shifts the demand distribution and SAA can overfit in such contexts \citep{feng2022developing} and these works do not model censoring and do not leverage feature information. Feature-based newsvendor variants such as \cite{ban2019big,lin2022data} incorporate covariates but remain single-period and do not model pricing and assume no censoring; additionally, classical newsvendor settings do not feature decision-dependent uncertainty (price affects demand), making them special cases of our problem. On the pricing side, \cite{qi2022offline} and \cite{bu2023offline} study offline pricing with censored demand but in single-period settings without inventory decisions, while \cite{wang2023estimation} and \cite{miao2023personalized} focus on feature-based pricing via causal inference under IV considerations, again without the inventory dimension and censoring in a multi-period control problem. Most closely related, \cite{qin2019data} considers multi-period joint pricing and inventory using offline data but assumes fully observed demand and omits features; likewise, \cite{harsha2021prescriptive} and \cite{liu2023solving} study pricing and newsvendor style problems in single-period settings under fully observed demand. Overall, our work tackles the joint optimization of pricing and inventory decisions in a multi-period setting in a sequential manner while accounting for censored and dependent demand and incorporating contextual features, thereby generalizing these prior works.  

In addition to the works discussed above, Appendix \ref{app: additional lit} provides an overview of closely related offline reinforcement learning (RL) literature, emphasizing the unique methodological challenges posed by censored and dependent demand and summarizing recent pessimism-based RL algorithms that motivate our algorithmic design.



\section{Problem Formulation}\label{sec: problem formulation}
\subsection{Underlying Decision Process and Preliminaries}\label{sec: mf}

We begin by describing an idealized decision-making process for a retailer selling a single product over time, which we call the \textbf{underlying decision process} where the retailer can observe the true demand, even when it exceeds the available inventory. This idealized process, defined by a trajectory of states, actions, and rewards denoted by $\{(S_t, A_t, R_t)\}_{t \ge 0}$, serves as the basis for censored demand setting that we will address later. In particular, at the beginning of each period $t$, the retailer observes the current state $S_t = (X_t, Y_t, D_{t-1})$, which consists of three key components: external features ($X_t$), represented by a $p$-dimensional feature vector (including e.g., market conditions, competitor pricing, seasonal trends) from the space $\mathcal{X} \subseteq \mathbb{R}^{p}$; on-hand Inventory ($Y_t$), the current, non-negative inventory level from the space $\mathcal{Y} \subseteq \mathbb{R}^+$; and lagged Demand ($D_{t-1}$), the total demand from the previous period, $[t-1, t)$, from the space $\mathcal{D} \subseteq \mathbb{R}^+$. The previous demand is included in the state because demand is realized \textit{after} decisions are made, so the demand from period $t-1$ is a crucial piece of information for the decision at period $t$.

Then, given the observed information in the past, the retailer makes two decisions in period $t$, which form the action $A_t = (P_t, O_t)$. Specifically, the price ($P_t$) is referred to as the price set for the product in period $t$, from the space $\mathcal{P} \subseteq \mathbb{R}^+$; and order Quantity ($O_t$) is the amount of new inventory to order for period $t$, from the space $\mathcal{N} \subseteq \mathbb{R}^+$. We let $\mathcal{S} = \mathcal{X} \times \mathcal{Y} \times \mathcal{D}$ and $\mathcal{A} = \mathcal{P} \times \mathcal{N}$ denote the state and action spaces, respectively, and assume that $\mathcal{S} \times \mathcal{A} \subseteq \mathbb{R}^d$,
so that each state–action pair for the retailer can be represented as a $d$-dimensional vector.

The retailer's complete plan for making these pricing and ordering decisions over time is called a policy, $\pi$. A policy is history-dependent if the decision made today is based on  the entire sequence of past states and actions. Alternatively, the policy is stationary if it applies the same mapping from a fixed-length history window at every time step, so the decision at time $t$ depends only on the current history window and not explicitly on $t$ itself.

Following the retailer's action, the period's demand $D_t$ is realized. In a general setting, this demand process can be highly complex, potentially depending on the history of the process. For illustration, we consider a first-order auto-regressive linear model AR(1) for demand process as a running example:
\begin{align}\label{eq:linear_ar_demand}
    D_t &= \theta_0 + \theta_x X_t - \beta P_t + \rho D_{t-1} + \varepsilon^D_t,
\end{align}
where $\varepsilon_t^D$ is some i.i.d. demand shock (following e.g., Gaussian distribution), $\theta_x \in \mathbb{R}^p$, $\beta>0$ captures price sensitivity, and $|\rho|<1$ ensures the stability of the AR(1) process. 

Following the realization of the demand, the retailer's reward/profit $R_t$ can be defined as:
\begin{align}\label{def:immediate_reward}
    R_t = \underbrace{P_t \min(Y_t, D_t)}_{\substack{\text{Sales Revenue}}} - \underbrace{C_{1, t} (D_t - Y_t )^+}_{\substack{\text{Stockout Cost}}} - \underbrace{C_{2, t} O_t}_{\substack{\text{Ordering Cost}}} - \underbrace{C_{3, t} (Y_t - D_t)^+}_{\substack{\text{Inventory Cost}}},
\end{align}
where $C_{1, t}$, $C_{2, t}$, and $C_{3, t}$ are the known per-unit costs for stockouts, ordering, and holding inventory, respectively, the operator $(x)^+ := \max(x, 0)$ denotes the positive part of $x$, and for simplicity, we assume rewards are uniformly bounded by $|R_t| \leq R_{\max}$.

The period $t$ then concludes with the transition to the next state $S_{t+1}$. This starts with the calculation of the next period's inventory, $Y_{t+1}$, via $Y_{t+1} = (Y_t+O_t-D_t)^+.$ Then, next-period exogenous features, $X_{t+1}$, are realized from a conditional distribution $p(X_{t+1} \mid S_t)$, leading to the completion of the period and transition to the next state $S_{t+1} = (X_{t+1},\, Y_{t+1},\, D_t)$.

Overall, the retailer's main objective is to choose a policy, $\pi$, that maximizes long-term profit under the dynamics of this decision process $\{(S_t, A_t, R_t)\}_{t \ge 0}$. We use Markov Decision Process (MDP) to model the underlying decision process $\{(S_t, A_t, R_t)\}_{t \ge 0}$ as it naturally represents evolving system states, connects decisions to probabilistic dynamics, provides a principled basis for policy optimization, and flexibly accommodates nonlinear dependence beyond our running example. Therefore, we assume that
\begin{align}\label{eq: markov property}
(S_{t+1}, R_t)  \independent H_{t-1,\text{MDP}} \given (S_t, A_t)
\quad \text{with time-homogeneous transitions},
\end{align}
where $H_{t-1,\text{MDP}}=\left\{S_j,A_j\right\}_{0\leq j\leq t-1}$.

Under this framework, it is well-established that a stationary and optimal pricing and inventory replenishment policy exists, i.e., $\pi_{\text{oracle}}^*=(\pi_{1,\text{oracle}}^\ast, \pi_{2,\text{oracle}}^\ast)$, that maximizes the expected discounted cumulative rewards (i.e long-term profit), 
\begin{align}\label{eq: oracle optimal}
\pi_{\text{oracle}}^* \in \argmax_{\pi} \,\EE_{\textbf{MDP}}^{\pi}[\sum_{t=0}^{\infty} \gamma^t R_t],
\end{align}
where $\EE_{\textbf{MDP}}^{\pi}$ means that all actions along the trajectory follow $\pi$, the subscript "\textbf{MDP}" is used to differentiate the expectation operator of the underlying MDP from that of the observed process with demand censoring incorporated defined in the following section and $\gamma \in [0, 1)$ is a discount factor.

Moreover, this MDP formulation encompasses many familiar inventory and pricing models as special cases. It can therefore be viewed as a mild generalization rather than a departure from classical setups. For instance, consider a special case where we eliminate exogenous features $X_t$, assume demand is additive (i.e., $D_t = \theta_0  + \beta P_t + \rho D_{t-1}  + \epsilon_t$) with fixed ordering cost $K > 0$, and lost sales. In this setting, \cite{chen2006optimal} proved that the optimal policy $\pi^*_{\text{oracle}}$ admits a stationary $(s, S, p)$ structure. 
We illustrate this connection by highlighting several other such cases in \ref{app: special cases} of Appendix.
With this perspective in place, under the general setting, one popular solution to obtain $\pi_{\text{oracle}}^*$ is to find an optimal Q-function and take point-wise maximum with respect to action for every state \citep{sutton2018reinforcement}.

\subsection{Unidentifiable Oracle Policy}\label{sec:challenges}
In practice, however, offline retail data often lack complete demand information due to censoring. Typically, the retailer observes only sales quantity $Z_t := \min\{Y_t, D_t\} \in \mathcal{Z}$ together with a censoring indicator $\Delta_t := \mathbb{I}(Y_t \ge D_t)$. Consequently, the data are realized through what we term the \emph{observed} process, $\{(W_t,A_t,R_t\mathbb{I}[\Delta_t=1])\}_{t\ge0}$, instead of the underlying decision process $\{(S_t,A_t,R_t)\}_{t\ge0}$, with $ W_t:=(X_t,Y_t,Z_{t-1},\Delta_{t-1})\in\mathcal{W}$. Viewed this way, the observed process adds a layer that generalizes the underlying one: when censoring is absent, the observed process coincides with the underlying process.


Within this generalized and observed layer, two key differences emerge in the presence of censoring. 
First, their state representations differ: the underlying process always includes the true demand $D_t$ in the state $S_t$, whereas the observed process lacks this information when demand is censored, leading to a failure of the Markov property. The following proposition formalizes this idea. Let $H_{0:(t-1)}=\left\{W_j,A_j\right\}_{0\leq j\leq t-1}$.

\begin{proposition}\label{prop: high-order}
Consider the observed process $\{(W_t,A_t,R_t \mathbb{I}[\Delta_t=1])\}_{t\ge0}$. 
If censoring occurs, i.e., $\Delta_{t-1}=0$ for some $t \geq 0$, then the Markov property fails. 
In particular, on the event $\{\Delta_{t-1}=0\}$,
$(W_{t+1},R_t) 
\;\not\!\perp\!\!\!\perp\;
H_{0:(t-1)} \mid (W_t,A_t) $.
\end{proposition}
The proof of Proposition \ref{prop: high-order} is given in Section \ref{proof: proposiiton} of Appendix.
Second, the observed process, unlike the underlying one, does not have complete reward information under censoring. Specifically, the reward in Equation \eqref{def:immediate_reward} depends on $(D_t-Y_t)_+$, which is not directly computable under censoring since $D_t$ is unobserved. Consequently, the Markov property under the observed process fails except in special cases (e.g., i.i.d. demand, $\rho=0$ in  Equation \eqref{eq:linear_ar_demand}, or no censoring) and reward is not observed under censoring. Taken together, they lead to important implications for structural results in both inventory and pricing and MDP theory.

In particular, classical structural results in inventory and pricing theory (e.g. $(s,S,p)$ policies) typically require Markov dynamics with fully observed states and some regularity conditions (e.g., convex costs). With censoring, the problem becomes \emph{partially observed}, so these results do not carry over: any threshold-type structure, if it exists, must be expressed in terms of the observed state $S_t$ rather than $W_t$, and to our knowledge no general $(s,S,p)$-style optimality result holds under censored, dependent demand.\footnote{Even in the linear AR(1) example, updating $D_{t-1}$ involves a truncated distribution that depends on past actions, breaking the monotonicity/sub-modularity arguments used in classical proofs.} Moreover, from an MDP viewpoint, the observed process is generally non-Markov, so the optimal Q-function is not well-defined on the observed state $w$ and standard Bellman equations do not apply.

In summary, well-known structural results from inventory and pricing theory and standard MDP solution methods do not apply directly under this observed process. More fundamentally, the oracle policy $\pi_{\text{oracle}}^{*}$, defined on the underlying state $S_t$ under fully observed demand, may not be learnable from i.i.d.\ trajectories of the censored observed process $\{W_t, A_t, R_t\mathbb{I}[\Delta_{t}=1]\}_{t \geq 0}$ (i.e., not point-identified). Therefore, it is necessary to redefine the learning target. In particular, this means that the best achievable outcome under these conditions is to identify the optimal policy relative to the observed process defined as
\begin{align}\label{def: optimal policy under offline}
\pi^* \in \argmax_{\pi \in \Pi} \EE_{W_0 \sim \nu}\{\EE^{\pi}[\sum_{t=0}^{\infty} \gamma^t R_t\given W_0]\},
\end{align}
where the expectation is taken over the trajectories generated by $\pi$ under the observed process $\{W_t, A_t, R_t\mathbb{I}[\Delta_{t}=1]\}_{t \geq 0}$, with $\Pi$ representing the set of all policies associated with this process.
Here, $W_0$ follows $\nu(\bullet)$ which is some reference distribution over the initial observation and is \textit{predetermined} and known by decision makers. This is a typical assumption in the literature of RL.

In this paper, we demonstrate how offline data can be used to find $\pi^{\ast}$. Before, we impose an assumption on our data-generating process.

\begin{assumption}\label{ass: DGM}
The offline data $ \mathcal{O}_N = \{W_{j,t}, A_{j,t}, W_{j,t+1}, R_{j,t} \mathbb{I}[ \Delta_{j,t}=1]\}^{ 1 \leq j \leq N}_{0 \leq t < T-1}$ is generated by the policy, $\pi^b \in \Pi$, where each trajectory $j$ is an i.i.d. copy from $ \{W_t, A_t, R_t \mathbb{I}[\Delta_t=1] \}_{t \geq 0} $.
\end{assumption}

In RL literature, $\pi^{b}$ is often referred to as the \textit{behavior policy}. The behavior policy may not be known beforehand and can be history-dependent. Using $\mathcal{O}_N$, we aim to develop an algorithm that returns an estimated policy $\widehat{\pi}$ that approaches the optimal policy in terms of the value when the sample size $N$ (and/or $T$) is large.


\section{Characterizing the Optimal Policy under Censoring}

Although the goal is clear, achieving it is far from straightforward compared to traditional settings. The preceding analysis shows that censoring under dependent demand fundamentally alters the problem: the \emph{observed} process is generally non-Markov, and rewards are missing under stockouts. Nevertheless, the observed process is \emph{not} devoid of structure. Our central insight is that it inherits a \emph{latent, high-order Markov property} from the underlying first-order MDP. The effective order is determined by how frequently uncensored periods occur under the governing policy. Building on this observation, in this section, we develop a framework for learning $\pi^*$ from offline data, ranging from characterization of $\pi^*$ to development of algorithms.

For the remainder of the paper, without loss of generality, we assume that there is no censoring at the start of the observed process, i.e., $\Delta_{-1}=1$. We next introduce two definitions that will streamline the subsequent analysis.

\begin{definition}\label{def: definition for set C}
$\mathcal{C}(T',n) = \left\{\mathcal{O}=\{W_t, A_t, R_t \mathbb{I}[\Delta_t=1]\}_{t \geq 0} \, \mid \, C_{T'}(\mathcal{O}) \leq n \right\}$,
where $C_{T'}(\mathcal{O})$ denotes the maximum number of consecutive censoring up to time step $T'$ in the observed process.
\end{definition}

From this definition, note that for a fixed $n$, we have $\mathcal{C}(T_2,n) \subseteq \mathcal{C}(T_1,n)$ for all $T_2 > T_1$, implying that $\mathcal{C}(T',n)$ forms a decreasing sequence of sets with respect to $T'$. This allows us to define the following policy class.

\begin{definition}\label{def: set of policies such that at most n cons censoring observed}
For any $n \geq 0$, define
$
\Pi_{n}=\left\{\pi' \,\middle|\, \prob^{\pi'}\!\left(\lim_{T' \rightarrow \infty} \mathcal{C}(T',n)\right)=1\right\}.
$
\end{definition}

In words, $\Pi_n$ consists of all policies under which any trajectory will, with probability one, contain no more than $n$ consecutive instances of censoring over the entire horizon. By definition, $\Pi_1 \subseteq \Pi_2 \subseteq \cdots \subseteq \Pi_\infty$, where $\Pi_\infty$ represents the set of all feasible policies. Therefore, any policy, and in particular the optimal policy $\pi^*$, must belong to some $\Pi_n$, where $n$ may be finite or infinite.

\subsection{Local Conditional Independence and Finite-Order Structure}\label{sec:local-structure}
Given these definitions, we first take a practical perspective on the potential behavior of the optimal policy before introducing the formal feasibility assumption. Specifically, in real operational settings, no well-designed policy would allow censoring to persist indefinitely. As the number of consecutive censoring increases, firms inevitably intervene through corrective actions. Allowing infinite consecutive censoring would imply never observing demand again, making it impossible to estimate sales potential or update future decisions. Hence, any realistic optimal policy must implicitly limit the length of censoring streaks by triggering a recovery action after a finite number of periods. This motivates our first key assumption, which ensures that the problem is well-posed and learnable from finite data.

\begin{assumption}\label{assump:bounded-n}
There exists a finite $n < \infty$ such that $\pi^\ast \in \Pi_n$.
\end{assumption}

Formally, if the optimal policy $\pi^*$ only belongs to $\Pi_\infty$ (i.e., it allows for unbounded censoring streaks), the dependency structure of the observed process under $\pi^*$ can become arbitrarily complex due to the consistent violation of the Markov property (i.e., the process becomes history-dependent). This makes learning from offline data infeasible. Assumption~\ref{assump:bounded-n} precludes this pathological case.

Given this assumption, we restrict our attention to policy class $\Pi_n$. We now show that this assumption has a profound structural consequence: any policy including $\pi^\ast$ within this class induces a finite-order Markov structure on the observed process, based on which we can characterize the structure of the optimal policy.

Specifically, for integers $a \le b$, we define the history $H_{a:b}\;:=\;\big(A_b,W_b,\,\ldots,\,A_a,W_a\big)$. To characterize how far one must look back to recover the most recent uncensored observation, we define the \emph{most recent uncensored time} within the last $n{+}1$ observations at time $t$ as $
\tau_t(n)\;:=\;\max\big\{\,t'\in\{t,t-1,\ldots,t-n\}\,:\ \Delta_{t'-1}=1\,\big\}$. Because we assume $\Delta_{-1}=1$, $\tau_t(n)$ exists almost surely under any $\pi\in\Pi_n$. Given this, we can establish the following conditional independence property.
\begin{lemma}\label{cor:mdp-order}
Suppose the underlying decision process satisfies \eqref{eq: markov property}. Then, under any $\pi \in \Pi_n$, we have $(W_{t+1},R_t) \independent H_{0:\,\tau_t(n)-1} \mid H_{\tau_t(n):\,t}$, implying an MDP of order at most $n+1$.
\end{lemma}
Intuitively, Lemma~\ref{cor:mdp-order} states that the process resets its history dependence each time an uncensored observation occurs. Once we condition on the history block since that last uncensored anchor ($H_{\tau_t(n):t}$), the entire history prior to that anchor becomes irrelevant for predicting the next state and reward. This finite-order structure is the key to characterize the structure of an optimal policy within $\Pi_n$, which has the same value as that under $\pi^\ast$. The proof of Lemma~\ref{cor:mdp-order} is given in Section \ref{proof: lemma 1} of Appendix.



\subsection{Structure of An Optimal Policy}
Under Assumption~\ref{assump:bounded-n}, with the help of Lemma~\ref{cor:mdp-order}, we can show that there exists an optimal policy $\pi^\ast$, which is stationary under a finite-order MDP.



Firstly, recall $H_{(t-i):t}=(A_t,W_t,\ldots,A_{t-i},W_{t-i})$. For $i\in\{0,\ldots,n\}$, we define
$
H'_{(t-i):t}\;:=\;H_{(t-i):t}\setminus\{A_t\}
\;=\;
(W_t,\,A_{t-1},W_{t-1},\,\ldots,\,A_{t-i},W_{t-i}),
$
so that $H'_{(t-i):t}$ is the relevant history block for a censoring run of length $i$ (i.e.,  in the components of $H_{(t-i):t}$, we have $\{\Delta_{t-j}=0\}_{j=1}^{i}$ and $\Delta_{t-i-1}=1$.

Secondly, by the definition of $\Pi_n$, any policy in this class must select an action that guarantees that the next period is uncensored once a censoring run reaches length $n$. We denote this set of actions as follows:
for any $t\ge0$ and any $\mathbf{h}'_{(t-n):t}$ satisfying $\{\Delta_{t-j}\}_{j=1}^{n}=0$ and $\Delta_{t-n-1}=1$, define
\[
\textup{UP}^{(n)}\!\big(\mathbf{h}'_{(t-n):t}\big)
\;:=\;
\Big\{a\in\mathcal{A}:\ \Pr\!\big(\Delta_t=1\ \big|\ H'_{(t-n):t}=\mathbf{h}'_{(t-n):t},\,A_t=a\big)=1\Big\}.
\]

Next, we introduce the corresponding state–action value functions.
For any $0 \leq i \leq n$, the $i^{\text{th}}$ optimal state–action value function (Q-function) over the policy class $\Pi_n$ is defined as
\begin{align}
Q^{(i)}_\ast\!\big(\mathbf{h}'_{0:i},\,a\big)
&:= \sup_{\pi \in \Pi_n}\,
\EE^{\pi}\!\Bigg[\sum_{t'=i}^{\infty}\gamma^{\,t'-i} R_{t'} \ \Bigg|\ H'_{0:i}=\mathbf{h}'_{0:i},\,A_i=a\Bigg],
\label{eq:Qi-def}
\end{align}
where the components of $\mathbf{h}'_{0:i}$ satisfy $\{\Delta_{i-j}=0\}_{j=1}^i$ and $\Delta_{-1}=1$.

Observe that $Q^{(i)}_\ast$ depends on the history block up to the most recent uncensored time point. This dependence is a direct consequence of the process behaving as an MDP of order at most $n+1$, as established by Lemma~\ref{cor:mdp-order}. We now state the main result establishing the existence and structure of a stationary optimal policy $\pi^\ast$ based on these Q-functions.

\begin{lemma}\label{lemma:optimal-policy-Pin (final)}
There exists a deterministic and stationary policy $\pi^{\ast}$ that maximizes $\EE^{\pi'}[\sum_{t=0}^{\infty} \gamma^t R_t]$ over $\Pi_{n}$ and has the following form:
\begin{equation}\label{eq:policy-mixture-form (Pin)}
\pi^\ast\!\big(H'_{(t-n):t}\big)
\;=\;
\sum_{i=0}^{n}
\pi^{(i)}_{\ast}\!\big(\,\cdot\ \big|\ H'_{(t-i):t}\big)\;
\prod_{j=1}^{i}\mathbb{I}[\Delta_{t-j}=0]\;\cdot\;\mathbb{I}[\Delta_{t-i-1}=1],
\end{equation}
where each $i$-component selects a greedy action with respect to $Q^{(i)}_{\ast}$:
\begin{align}
\pi^{(i)}_{\ast}\!\big(a'\mid H'_{(t-i):t}\big)
&=
\begin{cases}
1, & a'\in\argmax_{a\in\mathcal{A}}\ Q^{(i)}_{\ast}\!\left(H'_{(t-i):t},a\right), \quad i=0,\ldots,n{-}1, \\[6pt]
1, & a'\in\argmax_{a\in\,\textup{UP}^{(n)}(H'_{(t-n):t})}\ Q^{(n)}_{\ast}\!\left(H'_{(t-n):t},a\right), \quad i=n, \\[4pt]
0, & \text{otherwise.}
\end{cases}
\end{align}
Moreover, $Q^{(i)}_\ast(H'_{0:i},\,A_{i}) = Q^{(i)}_{\ast}(H'_{(t-i):t},\,A_{t})$ for all $t\geq i$ and $i \in \{0,\ldots,n\}$.
\end{lemma}

Lemma~\ref{lemma:optimal-policy-Pin (final)} establishes that, within $\Pi_n$, an optimal policy exists that is fully characterized by its corresponding Q-functions. This result follows directly from the finite-order Markov structure derived in Section~\ref{sec:local-structure}. The only difference is that when the censoring run reaches $i=n$, the policy must select an action from $\textup{UP}^{(n)}(H'_{(t-n):t})$ to remain within the class $\Pi_n$. A detailed proof is provided in Appendix~\ref{subsec: proof of optimalpol under C}. Overall, Lemma~\ref{lemma:optimal-policy-Pin (final)} sets a clear path forward, navigating our goal to finding a data-driven approximation of these Q-functions from the offline data, which will, in turn, yield a near-optimal policy.

However, this path is blocked by three fundamental challenges. First, Assumption \ref{assump:bounded-n} enabling feasibility only guarantees that a finite $n$ \textit{exists}, but it does not specify it. Second, the Q-functions themselves, $Q^{(i)}_*$, are non-standard compared to those used in the standard MDP due to the constraint nature of $\Pi_n$. Whether there exist Bellman equations to compute them remains unknown. Finally, even if we knew how to compute these $Q$-functions, our offline data suffers from incomplete reward information due to censoring. In the following sections, we address these challenges in turn.



\subsection{Specialized Bellman Equations and Algorithmic Implications}
\label{subsec:bellman-implications}

We begin by showing a family of specialized Bellman optimality equations that extend the standard first-order MDP Bellman equations to the higher-order, censoring aware setting. 
These equations make the structure of the high-order Q-functions $Q_*^{(i)}$ explicit and provide the foundation for the algorithmic development that follows. 
Before presenting the formal statements, we introduce additional notation and highlight a key structural property of the observed process.


For each $i\in\{0,\ldots, n\}$ and history block $H'_{(t-i):t}$, we define the feasible action set $\mathcal{A}_i$ and the corresponding optimal state-value function $V_\ast^{(i)}$ as
\[
\mathcal{A}_i\!\big(H'_{(t-i):t}\big)
:=\begin{cases}
\calA, & i< n\\[4pt]
\textup{UP}^{( n)}\!\big(H'_{(t-n):t}\big), & i= n
\end{cases},
\qquad
V_\ast^{(i)}\!\big(H'_{(t-i):t}\big)
:=\max_{a\in\mathcal{A}_i\big((H'_{(t-i):t}\big)}
Q^{(i)}_{\pi_{n}^\ast}\big(H'_{(t-i):t},a\big).
\]

By construction of the policy class $\Pi_{n}$, when the current censoring run has length $n$, the only admissible actions are those that \emph{guarantee} an uncensored next state. Hence, any action outside $\textup{UP}^{(n)}$ is infeasible for the $n$-th optimal Q-function, as it would induce $ n{+}1$ consecutive censored periods, violating the membership in $\Pi_{n}$.

With this notation as well as for any $i\in\{0,\ldots,n\}$ and any $a\in\mathcal{A}_i\!\big(H'_{(t-i):t}\big)$, the specialized Bellman optimality equations can now be expressed as
\begin{sizeddisplay}\small
\begin{align}
Q^{(i)}_{\ast}(H'_{(t-i):t},a) = \mathbb{E}\left[ R_t + \gamma \left( \mathbb{I}[\Delta_t=1]V_\ast^{(0)}(W_{t+1}) + \mathbb{I}[\Delta_t=0]V_\ast^{(i+1)}(H'_{(t-i):t+1}) \right) \,\middle|\, H'_{(t-i):t}, A_t=a \right],
\label{eq:bellman-compact}
\end{align}
\end{sizeddisplay}
where the components of $H_{(t-i):t}$ satisfy $\{\Delta_{t-j}=0\}_{j=1}^{i}$ and $\Delta_{t-i-1}=1$ for all $i=0,\ldots,n$.

When $i=n$, the feasible actions are restricted to those in $\textup{UP}^{(n)}(H'_{(t-n):t})$, for which $\Pr(\Delta_t=0 \mid H'_{(t-n:t},A_t)=0$. Therefore, the second continuation term in Equation \eqref{eq:bellman-compact} vanishes, yielding
\begin{align}
    \label{eq:bellman-compact2}
    Q^{(n)}_{\ast}\big(H'_{(t-n):t},a\big)
=\EE\!\Big[
R_t+\gamma\,V_\ast^{(0)}(W_{t+1})
\ \Big|\ H'_{(t-n):t},A_t=a
\Big].
\end{align}

Equations~\eqref{eq:bellman-compact}-\eqref{eq:bellman-compact2} thus provide the dynamic recursion that links Q-functions of order $i$ and $i{+}1$, terminating at the $n$-th level where the process is guaranteed to reset to an uncensored state. This structure generalizes the standard Bellman operator by explicitly conditioning on the length of the current censoring run and constraining feasible actions at the $n$-th level to maintain admissibility within $\Pi_{n}$.


From an algorithmic standpoint, these equations reveal the modularity of learning under censoring: each $Q_*^{(i)}$ can be recursively estimated from offline data by conditioning on the history block $H'_{(t-i):t}$ and truncating whenever an uncensored period appears, after which the estimates can be composed to form a stationary policy greedy with respect to the learned $Q^{(i)}$ functions, yielding a data-driven approximation of $\pi^*$.



\subsection{Estimating the Data-Supported Policy Class}

However, Lemma~\ref{lemma:optimal-policy-Pin (final)} indicates that learning $\pi^*$ requires that our offline data contain \emph{enough} examples of censoring streaks of every length \(i\in\{0,\dots,n\}\) so that all $Q$-functions can be estimated. If the data never exhibit long enough censoring runs, we lack the necessary information to identify the corresponding $Q$-function of the optimal policy. In other words, our ability to recover \(\pi^*\) depends on whether the censoring sequences observed under \(\pi^b\) are sufficiently long to represent all possible dynamics under \(\pi^*\).

To this end, we analyze the data through finite windows of length $K$ and, within each $K$-window, record the longest observed run of consecutive censored periods. This window-based perspective reflects the fact that the statement ``the maximum number of consecutive censoring periods is $n$'' is inherently local. By definition, for any trajectory under any policy $\pi \in \Pi_n$, every window of length $n{+}1$ must contain at least one uncensored period. For example, if $\pi^* \in \Pi_2$, then in any three consecutive periods it is impossible to observe three censored demands in a row; at least one must reveal the true demand. This local structure implies that the relevant dependence in the observed process can always be recovered by looking back at most $n$ periods. So censoring patterns within finite windows of length $K$ empirically quantify the effective order of dependence the data can support.

Given this, we define $n_{K,b}$ and $n_{K,*}$ as the maximum number of consecutive censoring instances observed under $\pi^b$ and $\pi^*$, respectively, within any window of length $K$ where the formal definitions of $n_{K,b}$ and $n_{K,*}$ are given in Definition \ref{def: n_{T',b} and n_{T',*}} of Appendix. To make the problem solvable, we require that the offline data covers the complexity of the optimal policy.

\begin{assumption}[Sufficient Coverage]
\label{assump:sufficient-coverage-main-text}
For a chosen window length $K\in\mathbb{N}$, the data-supported complexity dominates the target: $n \le n_{K,b}$.
\end{assumption}

This assumption guarantees that $\pi^*$ lies within the data-supported policy class $\Pi_{n_{K,b}}$. If this were not true (i.e., $n > n_{K,b}$), the optimal policy would induce censoring sequences longer than any observed in history, rendering the corresponding higher-order Q-functions unidentifiable. When \(n \le n_{K,b}\), however, all necessary censoring patterns are represented in the data, and \(\pi^*\) lies within the data-supported policy class \(\Pi_{n_{K,b}}\) since \(\pi^* \in \Pi_n \subseteq \Pi_{n_{K,b}}\).

The parameter $K$ plays a central role here. It serves as a regularization hyperparameter: it enlarges the search window to satisfy Assumption~\ref{assump:sufficient-coverage-main-text} while simultaneously bounding the modeled dependence order to prevent overfitting to rare, long-run censoring patterns. We provide the theoretical justification for the selection of $K$ in Section \ref{sec: theoretical results}.

Finally, since $n_{K,b}$ is a population-level quantity, we estimate it empirically from the offline trajectories $\{(W_{j,t},\Delta_{j,t})\}_{t=0}^{T-1}$ for $j=1,\dots,N$ by identifying the longest fully censored run within any $K$-window: 
$
\hat n_{K,b}
=
\max_{\substack{j=1,\ldots,N \\ t=0,\ldots,T-K}}\Bigl\{\ell\in\{0,\ldots,K\}:\;
\prod_{k=t}^{t+\ell-1}\bigl(1-\Delta_{j,k}\bigr)=1
\Bigr\}.
$

The estimator $\hat n_{K,b}$ provides a data-driven upper bound on the unknown quantity $n$. Then, under Assumption~\ref{assump:sufficient-coverage-main-text}, we can approximate the optimal policy by learning the family of higher-order Q-functions $\{Q_*^{(i)}\}_{i=0}^{\hat n_{K,b}}$ via specialized Bellman equations. To this end, we develop a tailored FQI–inspired algorithm (\citealp{ernst_tree-based_2005}) to compute a near-optimal policy $\widehat{\pi}$.


\subsection{Censored Reward and Imputation Strategy}
\label{subsec:censored-reward}

To proceed further, note that the specialized Bellman equations require the immediate reward $R_t$ to be known at every time step. However, in our offline data, the reward is only observed when demand is uncensored ($\Delta_t = 1$), and when censoring occurs ($\Delta_t = 0$),  $D_t$ is unobserved. Therefore, the realized profit is partially missing because the stockout penalty cannot be evaluated in the true reward expression, i.e.,
\begin{align}
    R_t = \underbrace{P_t \min(Y_t, D_t)}_{\substack{\text{Sales Revenue}}} - \underbrace{C_{1, t} (D_t - Y_t )^+}_{\substack{\text{Stockout Cost}}} - \underbrace{C_{2, t} O_t}_{\substack{\text{Ordering Cost}}} - \underbrace{C_{3, t} (Y_t - D_t)^+}_{\substack{\text{Inventory Cost}}},
\end{align}
Therefore, the \emph{censored reward problem} must be addressed before any Q-function can be estimated from the offline data. Our goal is to construct an unbiased surrogate reward that replaces unobserved components with consistent estimates. To this end, we define the surrogate reward function for every $t,i \ge 0$ as
\begin{sizeddisplay}\small{
\begin{align}\label{eqn: surrogate outcome}
\widetilde R_t
&=P_t Z_t
- C_{1,t}\Bigl(\EE\!\left[D_t\mid \Delta_t=0,\, H_{(t-i):t}\right]-Y_t\Bigr)\mathbb{I}(\Delta_t = 0)\nonumber\\
&- C_{2,t}O_t  - C_{3,t}(Y_t - D_t)\mathbb{I}(\Delta_t = 1). 
\end{align}}\end{sizeddisplay}
where, in the components of $H_{(t-i):t}$, we have $\{\Delta_{t-j}=0\}_{j=1}^{i}$ and $\Delta_{t-i-1}=1$.
Intuitively, $\widetilde{R}_{t}$ equals the true reward whenever $\Delta_{t}=1$, and when $\Delta_{t}=0$,
it replaces the unobserved term $(D_{t} - Y_{t})$ with its conditional expectation given the observed history up to the most recent time point without censoring.


The next result establishes that the surrogate reward preserves the expected value of the true reward conditional on the observed history, thereby ensuring that optimizing with respect to $\widetilde{R}_t$ is equivalent to optimizing with respect to $R_t$.

\begin{lemma}\label{lm: identification for censored demand-main text} 
For every $t,i \ge 0$, we have almost surely
$
\EE\!\left[(R_{t} -\widetilde{R}_{t})\mathbb{I}[\Delta_{t}=0]\mid H_{(t-i):t}\right]=0, $
where, in components of $H_{(t-i):t}$, we have $\{\Delta_{t-j}=0\}_{j=1}^{i}$ and $\Delta_{t-i-1}=1$.
\end{lemma}
The proof of Lemma~\ref{lm: identification for censored demand-main text} is given in Appendix~\ref{subsec:proof of identification for censored demand }. Lemma~\ref{lm: identification for censored demand-main text} implies that $\EE\!\left[R_t \mid H'_{(t-i):t},A_t=a\right]
=
\EE\!\left[\widetilde R_t \mid H'_{(t-i):t},A_t=a\right]$. Hence, in the specialized Bellman equation \eqref{eq:bellman-compact}, the immediate reward term $R_t$ can be
replaced by $\widetilde R_t$ without changing $Q^{(i)}_\ast$, yielding the equivalent recursion

\begin{equation}
\begin{aligned}
Q^{(i)}_{\ast}(H'_{(t-i):t},a)
&=
\mathbb{E}\Bigl[
\widetilde R_t + \gamma \Bigl(
\mathbb{I}[\Delta_t=1]V_\ast^{(0)}(W_{t+1})
\\
&\qquad
+\mathbb{I}[\Delta_t=0]V_\ast^{(i+1)}(H'_{(t-i):t+1})
\Bigr)
\,\Bigm|\,
H'_{(t-i):t}, A_t=a
\Bigr].
\end{aligned}
\label{eq:bellman-compact-surrogate}
\end{equation}

Therefore, estimating $\widetilde R_t$ is essential for constructing the Bellman targets needed for policy optimization from offline data under censoring. By Equation \eqref{eqn: surrogate outcome}, this reduces to estimating the conditional mean $\EE\!\left[D_t \mid \Delta_t=0, H_{(t-i):t}\right]$. We note that estimating $\EE[D_t \mid \Delta_t=0, H_{(t-i):t}]$ is sufficient for our purposes because, by Lemma~\ref{lm: identification for censored demand-main text}, it enables construction of the surrogate reward $\widetilde R_t$ and hence valid Bellman targets under censoring. 
However, using the same conditional-mean estimator to \emph{impute the missing demand} and then appending this point estimate to the state does not resolve the failure of Markov property: a plug-in imputation collapses the latent demand uncertainty to a single value, whereas the true transition dynamics depend on the full conditional distribution of the censored demand given the observed history. See Appendix~\ref{app: uncensoring} for an explicit counterexample and further discussion.

To estimate $\EE\!\left[D_t \mid \Delta_t=0, H_{(t-i):t}\right]$, we make the following assumption to apply survival analysis tools.


\begin{assumption}[Censored Demand Identification]
\label{ass: inventory and demand independency}\textcolor{white}{smssmsmsmmmmmmmmmmmmmmmmmm}
\begin{enumerate*}[label=(\alph*), ref=\theassumption.(\alph*), itemjoin=\quad]
  \item For every \( t \ge 0 \), \( D_t \independent Y_t \mid (X_t, P_t, O_t) \).
    \label{assump:main:item1}
  \item \( D_{\max} \) is a known upper bound.
    \label{assump:main:item2}
\end{enumerate*}
\end{assumption}


Assumption \ref{ass: inventory and demand independency} \hyperref[assump:main:item1]{ (a)} is reasonable if the demand is intrinsic or inventory levels are unknown to consumers. This assumption, called conditional non-informative censoring, negates any dependency between the demand and inventory levels. If breached, conventional survival analysis methods fail, leading to biased estimates \citep{coemans2022bias}. Assumption \ref{ass: inventory and demand independency} \hyperref[assump:main:item2]{ (b)} imposes a uniform upper bound for the demands across covariates, which is mainly used to simplify the procedure of estimating $\EE\left[D_{t}| \Delta_{t}=0, H_{(t-i):t} \right]$. Under this assumption, we rely on the following lemma inspired by survival analysis in statistics \citep[e.g.,][]{kleinbaum2010survival} to estimate $\EE\left[D_{t}| \Delta_{t}=0, H_{(t-i):t} \right]$.

\begin{lemma}\label{lm: adress censored-main text}
Under Assumption~\ref{ass: inventory and demand independency}, 
$
\EE\!\left[D_{t}\mid \Delta_{t}=0, H_{(t-i):t} \right]
= Y_{t}
+ \int_{Y_{t}}^{D_{\max}}
\frac{\textup{SF}(c\mid H_{(t-i):t}\setminus Y_{t})}
{\textup{SF}(Y_{t}\mid H_{(t-i):t}\setminus Y_{t})}\,dc,
$
where $\textup{SF}(c \mid H_{(t-i):t}\setminus Y_{t})$ denotes the conditional survival function $\Pr(D_t > c \mid H_{(t-i):t}\setminus Y_{t})$.
\end{lemma}
Lemma \ref{lm: adress censored-main text} shows that estimating $\EE\!\left[D_{t}\mid \Delta_{t}=0, H_{(t-i):t} \right]$, and thus $\widetilde{R}_{t}$, requires estimating $\textup{SF}(c \mid H_{(t-i):t}\setminus Y_{t})$. The proof is given in Appendix~\ref{subsec: proof of adress censored}. We therefore propose using the Kaplan-Meier estimator to nonparametrically estimate $\textup{SF}(c \mid H_{(t-i):t}\setminus Y_{t})$. \citep{kaplan1958nonparametric, kleinbaum2010survival}; Kaplan--Meier--type estimators have also been widely used in lost-sales inventory models to learn from censored demand \citep[e.g.,][]{huh2011adaptive, lyu2024ucb}. Then when there are $i$ consecutive censored periods, the estimator for $R_t$  is
\begin{align}\label{eqn: estimated surrogate outcome} \widehat R_t = P_tZ_t - C_{1, t}\left(\widehat\EE\left[D_{t}| \Delta_{t}=0, H_{(t-i):t} \right]-Y_t\right)\mathbb{I}(\Delta_t = 0)-C_{2, t}O_t  - C_{3, t}(Y_t - D_t)\mathbb{I}(\Delta_t = 1), \nonumber \end{align} where we denote the estimator of $\EE\left[D_{t}| \Delta_{t}=0, H_{(t-i):t} \right]$ as $\widehat\EE\left[D_{t}| \Delta_{t}=0, H_{(t-i):t} \right]$.

By imputing the censored reward using the aforementioned methodology along with some abuse of notation, the offline dataset can be transformed from original to augmented as:
\begin{sizeddisplay}\small
\begin{equation}\label{eqn: augdata}
\begin{array}{c}
\substack{\textbf{Offline data }\\ \textbf{(original)}}\\[2pt]
\mathcal{O}_N^{\text{raw}}
= \big\{W_{j,t},A_{j,t},W_{j,t+1},R_{j,t}\,\mathbf{1}[\Delta_{j,t}=1]\big\}_{0 \le t < T}^{1 \le j \le N}
\end{array}
\text{\raisebox{-10pt}{$\boldsymbol{\rightarrow}$}}
\begin{array}{c}
\substack{\textbf{Offline data }\\ \textbf{(augmented)}}\\[2pt]
\mathcal{O}_N
= \big\{W_{j,t},A_{j,t},W_{j,t+1},R_{j,t}^{\ast}\big\}_{0 \le t < T}^{1 \le j \leq N}
\end{array}
\end{equation}
\end{sizeddisplay}
where $R^\ast_{j,t} = \mathbb{I}(\Delta_{j,t}=0)\widehat R_{j,t} + \mathbb{I}(\Delta_{j,t}=1)R_{j,t}$.

The augmented dataset $\mathcal{O}_N$ thus provides a complete reward record suitable for offline estimation of the specialized Bellman equations. This concludes our treatment of the missing reward problem. In the next section, we leverage this augmented dataset to develop a fitted Q-iteration–based algorithm that estimates $\{Q_*^{(i)}\}_{i=0}^{\hat n_{K,b}}$ and computes a data-driven policy $\widehat{\pi}$.


\subsection{Full Algorithms}
\label{subsec:full-algorithms}
Given the imputation strategy in Section~\ref{subsec:censored-reward}, we now work with the augmented offline dataset $\mathcal{O}_N$ that provides complete reward information via $R_t^\ast$. Our objective is to compute a data-driven policy $\widehat{\pi}$ that approximates $\pi^*$. By Lemma~\ref{lemma:optimal-policy-Pin (final)} and the Bellman system in Equation \eqref{eq:bellman-compact}, finding $\pi^*$ is equivalent to estimating its $\hat n_{K,b}{+}1$ constituent Q-functions $\{Q_*^{(i)}\}_{i=0}^{\hat n_{K,b}}$. The recursion decomposes by the current censoring-run length $i$, which suggests partitioning the data accordingly. Therefore, we begin by partitioning the augmented dataset $\mathcal{O}_N$ into $\hat n_{K,b}+1$ subsets, $\{\mathcal{O}_N^{(i)}\}_{i=0}^{\hat n_{K,b}}$, 
where each $\mathcal{O}_N^{(i)}$ contains all transitions preceded by exactly $i$ consecutive censoring. 
Formally, for $i=0,\ldots,\hat n_{K,b}$, let
\begin{sizeddisplay}
\footnotesize
\begin{align*}
\mathcal{O}^{(i)}_N &= \left\{\left(W_{j, t-i}, A_{j, t-i},\cdots, W_{j, t},  A_{j, t}, R^{*}_{j,t}, W_{j, t+1}\right) \mid
\text{$\{\Delta_{j,i}\}_{i =t-i}^{t-1} = 0$ and $\Delta_{j,t-i-1}=1$} \quad 0\leq t \leq T-1, 1 \leq j \leq N\right\}.
\end{align*}
\end{sizeddisplay}With the data partitioned, the remaining part of the FQI-inspired algorithm proceeds iteratively. We begin by initializing all Q-function estimates to zero, $\widehat{Q}_{0}^{(i)} = 0$. Then, at $k$-th iteration, we update each $Q$-function $\widehat{Q}_{k}^{(i)}$ via a two-step process:

First, at $k$-th iteration, for each component $i$ and corresponding subset of data $\mathcal{O}_N^{(i)}$, we construct the target value, $\zeta^{(i,k)}$, using the Q-function estimates from the previous iteration. Essentially, this target is a sample analogue of the right-hand side of the Bellman system for each consecutive censoring case. Specifically, for $i=0,\cdots,\hat{n}_{K,b}$, using $(H^{'}_{j,(t-i):(t+1)}, R^{*}_{j,t}) \in \mathcal{O}_{N}^{(i)}$, we compute the target for the $t$-th transition tuple on the $j$-th trajectory: 
\begin{align}\label{eq: reponse for reg}
\zeta^{(i,k)}_{j, t} 
&=\{ \ R_{j, t} + \gamma\sum_{a' \in \calA}(\widehat{\pi}_{k-1}^{(0)}(a' | W_{j, t+1})\widehat{Q}_{k-1}^{(0)}(W_{j, t+1}, a')) \}\mathbb{I}[\Delta_{j, t} = 1] \nonumber\\
&  + \{ \widehat{R}_{j, t} + \gamma\sum_{a' \in \calA}(\widehat{\pi}_{k-1}^{(i+1)}(a' |H^{'}_{j,(t-i):(t+1)})\widehat{Q}_{k-1}^{(i+1)}(H^{'}_{j,(t-i):(t+1)}, a'))\} \mathbb{I}[\Delta_{j, t} = 0],
\end{align}
where 
$\widehat{\pi}_{k}^{(i)}$, for all $i=0,\cdots,\hat{n}_{K,b}$ and $k=0,\cdots,K-1$, represents the greedy policy with respect to $\widehat{Q}_{k}^{(i)}$. 

With these target values computed, we proceed to the second step: updating the Q-functions. For each component $i=0,\ldots,\hat{n}_{K,b}$ and their corresponding targets, we minimize the Bellman residual. In particular, we fit a model from a pre-specified function class $\calQ^{(i)}$ that maps history-action pairs in $\mathcal{O}^{(i)}_N$ to their corresponding targets:
\begin{align}
	&\widehat Q_{k}^{(i)} \in \underset{Q^{(i)} \in \calQ^{(i)}}{\argmin} \left\{ \frac{1}{|\mathcal{O}^{(i)}_N|}\sum_{t, j} \left(\zeta_{j, t}^{(i,k)} - Q^{(i)}(H^{'}_{j,(t-i):t},A_{j,t})\right)^2 +\lambda J(Q^{(i)}) \right\} \label{eqn: FQI supervised}
\end{align}
where $|\mathcal{O}^{(i)}_N|$ is the cardinality of the set $\mathcal{O}^{(i)}_N$, $J(\cdot)$ is some regularization on the complexity of $Q^{(i)}$ and $\lambda$ is a positive tuning parameter. Observe that under the linear AR(1) demand model in Equation \eqref{eq:linear_ar_demand}, the specialized Bellman equations inherit a linear structure, so each $Q^{(i)}$ lies in a linear class and the regression step reduces to ordinary least squares, justifying linear models in the base case. To accommodate nonlinear structure, we also allow $\mathcal{Q}^{(i)}$ to be a reproducing kernel Hilbert space (RKHS) and estimate via kernel ridge regression, enabling nonlinear approximation while retaining statistical guarantees.

After $K$ iterations, we obtain the final Q-function estimates, $\{\widehat{Q}_K^{(i)}\}_{i=0}^{\hat{n}_{K,b}}$, from which we derive our final policy, $\widehat{\pi}_K$. For $i=0,\dots,\hat{n}_{K,b}-1$, the policy component $\widehat{\pi}_K^{(i)}$ is simply the greedy policy with respect to its corresponding Q-function, $\widehat{Q}_K^{(i)}$.

However, the final policy component, $\widehat{\pi}_K^{(\hat{n}_{K,b})}$, requires special handling to ensure the learned policy remains within the data-supported class $\Pi_{\hat{n}_{K,b}}$. Specifically, when the censoring streak reaches the observed maximum $\hat{n}_{K,b}$, the policy must enforce a reset to an uncensored state to prevent extrapolation into unobserved territory. Ideally, we would select the profit-maximizing action from the set of all actions guaranteeing this reset ($\textup{UP}^{(\hat{n}_{K,b})}$). However, since this feasible set depends on unknown system dynamics, we adopt a practical, conservative approach by assigning a known safe action, $a_{\max}$, at this boundary.

We assume that $a_{\max}$ always exists as in many applications, such as e-commerce platforms, such an action can be always identified. For example, raising prices significantly to reduce demand while placing large orders can prevent stockouts and, in turn, avoid censoring. Under this assumption, we can enforce the required non-extrapolating behavior by defining the final policy component to always select this safe action: $\forall t\geq \hat{n}_{K,b}$:
$
\widehat{\pi}^{(\hat{n}_{K,b})}(A_t =a_{\max}(H'_{t-\hat{n}_{K,b}:t}) \mid H'_{t-\hat{n}_{K,b}:t}) = 1
$.  Overall, we call this complete algorithm Censored-FQI (C-FQI).


A key caveat of FQI-type methods is their susceptibility to overfitting under distributional mismatch between the behavior and target policies, an issue amplified by the max-operator in the Bellman backup \citep{levine2020offline}. When coverage is insufficient, incorporating pessimism becomes essential \citep{buckman2020importance}. Therefore, we adopt the uncertainty-quantifier-based pessimism framework of \citep{jin2021pessimism}, and accordingly, introduce the following definition:
\begin{definition}[$\epsilon$-Uncertainty Quantifier]\label{def:UE}
We say $\{U_k^{(i)}\}_{k=1}^{K}$, where $U_k^{(i)}: \calW^{i+1} \times \calA^{i+1} \rightarrow \mathbb{R}$, is a $\epsilon$-Uncertainty Quantifier (UQ) with respect to the data generating mechanism for the $i^{th}$ supervised learning problem if the event 
\begin{align}\label{def:uncertainty_set}
    \Omega^{(i)} & = \{ | \widehat{Q}_{k+1}^{(i)}(h^{(t-i):t}) - \EE[(R_t + \gamma\max_{a \in \mathcal{A}}\widehat{Q}^{(0)}_{k}(W_{t+1},a) )I[\Delta_{t}=1]  \\
    &\quad\quad + (\widehat{R}_t + \gamma\max_{a \in \mathcal{A}}\widehat{Q}^{(i+1)}_{k}(H^{'}_{(t-i):(t+1)},a))I[\Delta_{t}=0] ]\given H_{(t-i):t} =h^{(t-i):t}] | \leq U_{k+1}^{(i)}(h^{(t-i):t}) \nonumber \\
    &  ~~~~~  \, \, \, \text{where} \, \,  h^{(t-i):t}\in \mathcal{H}^{i+1}=\mathcal{W}^{i+1}\otimes \mathcal{A}^{i+1}, k = 0, \cdots, (K-1) \}\nonumber
\end{align}
satisfies that $\prob{(\Omega^{(i)})} \geq 1-\epsilon$.
\end{definition}
In simple terms, $U_{k}^{(i)}$ quantifies the uniform error which stems from using $\widehat{Q}_{k}^{(i)}$ to approximate the regression target. 
To incorporate the pessimism principle, we make one major change to Equation \eqref{eq: reponse for reg} of C-FQI algorithm, i.e.
\begin{align}\label{eq: response pessimism}
   \widehat{Q}_{k-1}^{(i)}(\bullet) \leftarrow & \widehat{Q}_{k-1}^{(i)}(\bullet) - U_{k-1}^{(i)}(\bullet) \triangleq \widetilde{Q}_{k-1}^{(i)}(\bullet)
\end{align}
for $k=1,\cdots,K$ and $i=0,\cdots,\hat{n}_{K,b}$, and accordingly, the response is denoted as
$\widetilde{\zeta}^{(i,k)}_{j, t}$.
We refer to the resulting algorithm as PC-FQI and denote the final estimated policy produced by PC-FQI as $\widetilde{\pi}_K$. The concise representation of the implementation of these two algorithms is presented in Table \ref{alg:concise}. A more detailed description of the implementation is provided in Table \ref{alg:algo full} in Section \ref{sec: algo} of the Appendix.

\begin{algorithm}[H]\footnotesize
\caption{Censored FQI and Pessimistic Censored FQI}\label{alg:concise}
\SetAlgoLined
\textbf{Data Pre-processing:} Use survival analysis to impute censored reward information\;

\textbf{Estimate $Q^{(i)}_{\ast}$ for $i=0,\cdots,\hat{n}_{K,b}$ by performing FQI for $K$ iterations:}\;

\For{$k = 0$ \KwTo $K-1$}{
    \For{each $i$}{
        \hspace{0.25cm}$R^{\ast} + \gamma\!\max_a\!\widehat{Q}_{k}^{(0)}(\bullet,a)\mathbb{I}[\Delta=1] + \gamma\!\max_a\!\widehat{Q}_{k}^{(i+1)}(\bullet,a)\mathbb{I}[\Delta=0] \xrightarrow{\text{supervised learning}} \widehat{Q}_{k+1}^{(i)}(\bullet)$\;

        \colorbox{red!20}{\parbox{\dimexpr 15cm-2\fboxsep}{%
    \textbf{ $\widehat{Q}_{k+1}^{(i)}(\bullet) \leftarrow \widehat{Q}_{k+1}^{(i)}(\bullet) - U_{k+1}^{(i)}(\bullet) \rightarrow$ use only for Pessimism}\;
}}
}
    }

\textbf{Output:}\\
\For{$i=0$ \KwTo $\hat{n}_{K,b}-1$}{
    Greedy policies $\widehat{\pi}_K^{(i)}$ with respect to $\widehat{Q}_K^{(i)}$\;
}
\colorbox{green!20}{\parbox{\dimexpr 16cm-2\fboxsep}{%
    For $i=\hat{n}_{K,b}$, $\widehat{\pi}_K^{(\hat{n}_{K,b})}$ outputs an action that ensures the next state is uncensored almost surely.}
}
\end{algorithm}

\section{Theoretical Results}\label{sec: theoretical results}
This section develops finite-sample regret guarantees for the policies produced by C-FQI and PC-FQI where the regret of a generic policy $\pi$ is defined as $\textbf{Regret}(\pi)\triangleq   \EE^{\pi^*}[\sum_{t=0}^{\infty} \gamma^t R_t]-\EE^{\pi}[\sum_{t=0}^{\infty} \gamma^t R_t]$. Our theoretical analysis builds upon four complementary groups of assumptions: \emph{feasibility and coverage}, \emph{feasibility--unobservability alignment}, \emph{structural}, and \emph{technical assumptions}. Each category corresponds to a different dimension of the problem, and we provide a managerial interpretation when appropriate. We then conduct a layerwise analysis, starting from an idealized setting and progressively relaxing these assumptions one by one. This strategy isolates the impact of data coverage, operational safety constraints, and model complexity on the resulting regret.


\subsection{Assumptions and Practical Interpretations}


\subsubsection*{A. Feasibility and Coverage Assumptions}

The following two assumptions ensure learnability: the optimal policy cannot generate infinitely long censoring streaks, and the offline data covers the censoring depths that the optimal policy would produce.

\paragraph{Bounded Censoring.}
We exclude degenerate regimes in which demand could remain censored forever. This is what Assumption~\ref{assump:bounded-n} states.

\noindent\textbf{Managerial insight.} In practice, the optimal policy must occasionally trigger a recovery action to avoid never observing true demand.

\paragraph{Sufficient Censoring Coverage.}
The offline data must contain trajectories spanning the censoring depths that would occur under the optimal policy. This is Assumption~\ref{assump:sufficient-coverage-main-text}  we made before. If this fails, some $Q_*^{(i)}$ are unidentifiable from the data, yielding irreducible regret.


\noindent\textbf{Managerial insight.} The historical dataset must include operating conditions (i.e sufficient censoring coverage) comparable to those induced by the optimal policy.

\subsubsection*{B. Feasibility--Unobservability Alignment Assumptions}

The following two assumptions ensure the learning/deployment rules match the unobserved feasible set \(\mathrm{UP}^{(n_{K,b})}\) at the censoring limit.


\paragraph{Termination--Action Domain Validity.}
C-FQI/PC-FQI compute intermediate greedy policies ($\widehat{\pi}_k/\widetilde{\pi}_k$ at $k$-th iteration) over the full action space. However, at the censoring limit $n_{K,b}$, the true optimal policy must select an action from $\mathrm{UP}^{(n_{K,b})}$. A discrepancy arises if the intermediate policies select an action that appears optimal numerically but lies outside $\mathrm{UP}^{(n_{K,b})}$ (i.e., an action that fails to terminate censoring). Since we cannot know $\mathrm{UP}^{(n_{K,b})}$ empirically, we assume that the selected actions lies in $\mathrm{UP}^{(n_{K,b})}$, leading this discrepancy to vanish.

\begin{assumption}[Termination--Action Domain Validity]\label{ass:TAD}
All intermediate policies $\{\widehat{\pi}_k^{(n_{K,b})}\}_{k=0}^{K-1}$ and $\{\widetilde{\pi}_k^{(n_{K,b})}\}_{k=0}^{K-1}$ select actions within $\mathrm{UP}^{(n_{K,b})}$ whenever there is $n_{K,b}$ consecutive censoring.
\end{assumption}
This is primarily an \textbf{algorithmic} condition that keeps the dynamic programming updates aligned with the structural constraint in Equation \eqref{eq:bellman-compact}.

\paragraph{Boundary--Action Alignment.}

For the deployed policy, we instead hard-code a safe backstop action $a_{\max}\in \mathrm{UP}^{(n_{K,b})}$ at depth $n_{K,b}$. Because we cannot verify that this backstop is optimal within $\mathrm{UP}^{(n_{K,b})}$, we impose:
\begin{assumption}[Boundary--Action Alignment]\label{ass:AD}
 $a_{\max}$ maximizes both $\widehat{Q}^{(n_{K,b})}_K$ and $\widetilde{Q}^{(n_{K,b})}_K$.
\end{assumption}

\noindent\textbf{Managerial insight.}
When an intervention is needed to end a censoring streak, the backstop action $a_{\max}$ is the model's value-maximizing action rather than an arbitrary override.

\subsubsection*{C. Structural Assumption}

We impose the standard linear MDP assumption \citep[e.g.,][]{jiang2017contextual, yang2020reinforcement, jin2020provably}.
Here, linear refers to a \emph{functional form} restriction: the conditional reward and transition operator
depend on $(s,a)$ only through a known feature map $\phi(s,a)\in\mathbb{R}^d$ (e.g one may take $\phi$ to include the original coordinates $\phi(s,a)=(s,a)$
or any chosen basis of functions). The role of $d$ is simply the number of features used to parameterize the
MDP.

\begin{assumption}[Linear MDP Representation]\label{ass:linear-mdp}
The decision process $\{(S_t,A_t,R_t)\}_{t\ge0}$ admits a linear structure with respect to a feature map $\phi:\mathcal{S}\times\mathcal{A}\to\mathbb{R}^d$. Specifically, there exist parameter vectors $\theta_r, \theta_g \in \mathbb{R}^d$ such that: 
\begin{align}
\EE[R_t \mid S_t=s, A_t=a] &=\sum_{i=1}^d \phi_i(s,a)\theta_r^{(i)}, \, \text{and } \,
\EE[g(S_{t+1}) \mid S_t=s, A_t=a] = \sum_{i=1}^d \phi_i(s,a)\theta_g^{(i)}
\end{align}
for all bounded measurable functions $g$ and state-action pairs $(s,a)$. Here, $\theta_r^{(i)}$ and $\theta_g^{(i)}$ denote the $i$th coordinate of $\theta_r$ and $\theta_g$, respectively.
\end{assumption}
This holds for our running example (the linear AR(1) demand model in Equation \eqref{eq:linear_ar_demand}).

\subsubsection*{D. Technical Assumptions}
Finally, we impose the standard technical assumptions. These assumptions ensure the statistical consistency of our estimators, the necessary coverage of the offline dataset, and the validity of the uncertainty quantifiers used in Algorithms C-FQI and PC-FQI. Unlike the previous categories, these conditions will remain fixed throughout our relaxation sequence.  First, we make a consistency assumption on the estimator $\hat{n}_{K,b}$:
%
\begin{assumption}\label{ass:consistency}
$\hat{n}_{K,b}$ is a consistent estimator of $n_{K,b}$ such that $\prob(\hat{n}_{K,b} = n_{K,b}) \geq 1 - \omega_{NT}$ for any $K\le T$, where $\omega_{NT}=\smallO(1)$ as $N, T\to\infty$.
\end{assumption}
Such consistency can be justified via extreme-value theory \citep[e.g.,][]{reiss1997statistical, coles2001introduction}.

Next, we make the following two assumptions that are used primarily for facilitating the regret decomposition and its presentation:
\begin{assumption}\label{ass: pi* C - sure event-main-text}$\prob^{\pi^{*}_{\mathcal{C}}}(\mathcal{C}(K,n_{K,b}))=1$, where $\pi^{*}_{\mathcal{C}} \in \argmax_{\pi' \in \Pi}\mathbb{E}^{\pi'}\left[\sum_{t=0}^{\infty}\gamma^t R_t\given \mathcal{C}(K,n_{K,b})\right]$.
\end{assumption}
This, intuitively, guarantees a form of consistency: the optimal policy, when planned under the condition of $\mathcal{C}(K,n_{K,b})$, effectively manages the system dynamics to ensure that the condition is actually satisfied. This implies that staying within the zone implied by $\mathcal{C}(K,n_{K,b})$ is a valid, reachable choice. Notably, if Assumption \ref{assump:sufficient-coverage-main-text} holds (i.e., $\pi^* \in \Pi_{n_{K,b}}$), this condition is automatically satisfied. 
\begin{assumption}[Lower bound on partition sizes]\label{lower bound on partition sizes}
There exist $\rho\in(0,1]$ and an event $\Omega_{\mathrm{str}}:=\{\min_{0\le i\le n_{K,b}} |\mathcal{O}_N^{(i)}| \ge \rho NT\}$ such that $\Pr(\Omega_{\mathrm{str}})\ge 1-\epsilon$.
\end{assumption} 
This assumption is reasonable when the partitioning procedure avoids creating vanishingly small strata.
Furthermore, to guarantee that the optimal policy can be recovered from the offline data, we impose following classic assumption. Let $d_{\nu,i}^{\pi^*}$ and $d_{\nu,i}^{\pi_b}$ denote the generalized discounted visitation measures of the optimal and behavior policies for any history of length $i$ where the formal definition of these is given in Definition~\ref{def: generalized discounted visitation2} of Appendix.

\begin{assumption}\label{ass: general coverage}
There exists a constant $C_{\mathrm{cov}} >0$ such that for any censoring depth $0 \le i \le n_{K,b}$ and history $h^{(i)} \in \mathcal{H}^{i+1}$ (a non-censored start followed by $i$ consecutive censoring), we have $\sup_{h^{(i)} \in \mathcal{H}^{i+1}}  (d_{\nu,i}^{\pi^*}(h^{(i)}) / d_{\nu,i}^{\mu}(h^{(i)})) \le C_{\mathrm{cov}}$.
\end{assumption}

This assumption requires that the offline dataset sufficiently covers the censored observation sequences visited by the optimal policy \(\pi^*\), generalizing the classic offline-RL coverage condition on single-step state--action pairs \citep{xie2021bellman}. Since censoring induces multi-step histories whose distributions depend on the number of consecutive censoring, we impose coverage over entire censoring sequences that \(\pi^*\) may generate. A detailed comparison with the classical notion is deferred to Appendix~\ref{app: remarks}.


\begin{assumption}\label{ass: surrogate outcome}
There exists a constant $C_5(\varepsilon)$ depending on $\varepsilon \in (0, 1)$ such that the event
\begin{sizeddisplay}
\small{
\begin{align*}
\Omega^r:=\sup_{\substack{h^{(t-i):t} \in \mathcal{H}^{i+1} \\ 0 \leq t  \leq T - 1 \\  0 \leq i < n_{K,b}}} 
\bigg| &\EE\left[D_{t} \mid (\Delta_{t}, H_{(t-i):t}) = (0, h^{(t-i):t})\right] - \widehat{\EE}\left[D_{t} \mid (\Delta_{t}, H_{(t-i):t}) = (0, h^{(t-i):t})\right] \bigg| 
\leq C_5(\varepsilon) (NT)^{-\delta}
\end{align*}}
\end{sizeddisplay} happens with
 $\prob(\Omega^r)\geq1-\varepsilon$ for some $\delta > 0$ where in the components of $h^{(t-i):t}$, we have $\left\{\Delta_{t-j}=0\right\}_{j=1}^{i}$ and $\Delta_{t-i-1}=1$.
\end{assumption}By utilizing standard nonparametric techniques such as the Kaplan-Meier estimator, this assumption is validated  with $\delta = \frac{2}{d+5}$, where $d$ represent the dimension of the feature space  \citep{dabrowska1989uniform, khardani2014nonparametric}. For more detailed information, refer to Theorem 3.2 in \cite{khardani2014nonparametric}.

\begin{assumption}[Uncertainty Quantifiers]\label{ass: UQ ass}
   $\{U_k^{(i)}\}_{0 \leq i \leq  n_{K,b}}^{1 \leq k \leq K}$ and $\{\tilde{U}_k^{(i)}\}_{0 \leq i \leq  n_{K,b}}^{1 \leq k \leq K}$ are $\epsilon$-uncertainty quantifiers for C-FQI and PC-FQI, respectively, as given in Definition \ref{def:UE}.
\end{assumption}


In the literature, Assumption~\ref{ass: UQ ass} is known to hold under mild conditions. In particular, under Assumption~\ref{ass:linear-mdp} (Linear MDP), the existence of a valid uncertainty quantifier follows from \cite{jin2021pessimism}. Adapting their construction to our setting, for each iteration \(k\) and partition \(\mathcal{O}_N^{(i)}\), define
$
U_{k}^{(i)}\!\left(h^{(t-i):t}\right)
= \beta_k \Bigl(\phi_i\!\left(h^{(t-i):t}\right)^\top \Lambda_{(i)}^{-1}\phi_i\!\left(h^{(t-i):t}\right)\Bigr)^{1/2},$
where \(\beta_k\) is a constant, \(\Lambda_{(i)}\) is the corresponding regularized covariance matrix, and \(\phi_i(\cdot)\in\mathbb{R}^{d^{(i)}}\) is the feature map induced by \(i\)-step censoring. See Lemma~5.2 in \cite{jin2021pessimism} for details. Existence results for RKHS-based uncertainty quantifiers are provided in \cite{chang2021mitigating}.


\subsection{A Layerwise Regret Analysis}

The remainder of this section examines how relaxing the preceding assumptions affects the regret bound. 
We start from an idealized setting where all assumptions hold exactly and progressively relax some of them one by one to reveal the incremental sources of regret. Finally, we present general regret under the combination of all relaxations.

\subsubsection{Layer 1: Idealized Setting.}

\begin{figure}[ht]
\centering
\resizebox{1.00\linewidth}{!}{%
\begin{tikzpicture}[
    font=\sffamily,
    node distance=0.8cm and 0.3cm, 
         base/.style={
        draw=gray!40,
       thick,
        rounded corners=3pt,
        align=center,
        inner sep=6pt,
        fill=white,
        drop shadow={opacity=0.05, shadow xshift=1pt, shadow yshift=-1pt}
    },
    theorem/.style={
        base,
        top color=blue!5,
        bottom color=blue!10,
        draw=blue!20,
        text width=5cm,
        font=\bfseries\small
    },
    corollary/.style={
        base,
        top color=red!5,
        bottom color=red!10,
        text width=3.8cm, 
        font=\scriptsize, 
        anchor=north
    },
    general/.style={
        base,
        top color=green!5,
        bottom color=green!10,
        draw=orange!20,
        text width=5.5cm,
        font=\bfseries\small
    },
    arrowline/.style={
        draw=gray!60,
        thick,
        rounded corners=4pt,
        >={Stealth[scale=1.0]}
    }
]


\node[theorem] (thm) {Theorem \ref{cor:ideal} \\ Regret under idealized setting};

\coordinate[below=1.0cm of thm] (row_center);

\node[corollary, left=0.15cm of row_center, anchor=north east] (c2) 
    {\textbf{Corollary \ref{cor:relaxed-TAD}}\\ Regret under Relaxation of Assumption \ref{ass:TAD}\\ \textbf{Termination Action Domain}};

\node[corollary, right=0.15cm of row_center, anchor=north west] (c3) 
    {\textbf{Corollary \ref{cor:relaxed-AD}}\\Regret under Relaxation of Assumption \ref{ass:AD}\\ \textbf{Boundary Action Alignment}};

\node[corollary, left=0.3cm of c2] (c1) 
    {\textbf{Corollary \ref{cor:relaxed_coverage}}\\Regret under Relaxation of Assumption \ref{assump:sufficient-coverage-main-text}\\ \textbf{Sufficient Coverage}};

\node[corollary, right=0.3cm of c3] (c4) 
    {\textbf{Corollary \ref{cor:relaxed-structure} (Appendix)}\\Regret under Relaxation of Assumption \ref{ass:linear-mdp}\\ \textbf{Linear MDP Structure}};

\node[theorem, below=3.5cm of row_center, top color=green!5,
        bottom color=green!10] (gen) 
    {Theorem \ref{thm:general_regret}\\ Regret under all relaxations};


\draw[arrowline] (thm.south) -- ++(0,-0.4) coordinate (fork_point);

\draw[arrowline] (c1.north |- fork_point) -- (c4.north |- fork_point);

\draw[arrowline, ->] (c1.north |- fork_point) -- (c1.north);
\draw[arrowline, ->] (c2.north |- fork_point) -- (c2.north);
\draw[arrowline, ->] (c3.north |- fork_point) -- (c3.north);
\draw[arrowline, ->] (c4.north |- fork_point) -- (c4.north);

\coordinate (merge_height) at ($(gen.north) + (0,0.4)$);

\draw[arrowline] (c1.south) -- (c1.south |- merge_height);
\draw[arrowline] (c2.south) -- (c2.south |- merge_height);
\draw[arrowline] (c3.south) -- (c3.south |- merge_height);
\draw[arrowline] (c4.south) -- (c4.south |- merge_height);

\draw[arrowline] (c1.south |- merge_height) -- (c4.south |- merge_height);

\draw[arrowline, ->] (merge_height -| gen.north) -- (gen.north);

\end{tikzpicture}

}
\caption{Structure of the Layerwise Regret Analysis.} 
\label{fig:regret_structure}
\end{figure}

We first establish the benchmark regret rate when all assumptions hold. 
To begin with, we introduce the general regret decomposition which connects the structure of censoring to learnability and holds given $\widehat{\pi} \in \Pi_{n_{K,b}}$: 
\begin{align}\label{eq: challange 3} \textbf{Regret}(\widehat{\pi})\nonumber&=(\underbrace{\EE^{\pi^*}[\sum_{t=0}^{\infty}\gamma^tR_t\given \mathcal{C}(K,n_{K,b})] -\EE^{\widehat{\pi}}[\sum_{t=0}^{\infty}\gamma^tR_t]}_{(i)})\prob^{\pi^*}(\mathcal{C}(K,n_{K,b}))\nonumber\\&+\underbrace{(\EE^{\pi^*}[\sum_{t=0}^{\infty}\gamma^tR_t \given \mathcal{C}^{c}(K,n_{K,b})]-\EE^{\widehat{\pi}}[\sum_{t=0}^{\infty}\gamma^tR_t])\prob^{\pi^*}(\mathcal{C}^{c}(K,n_{K,b}))}_{(ii)},
\end{align} 
where the characterization of $\mathcal{C}(K,n_{K,b})$ is given in Definition~\ref{def: definition for set C} and the details of the derivation is provided in the Section \ref{sec:Alt. regret decomp} of Appendix. We now state our theorem for the idealized setting. Throughout, we use $
d \coloneqq \max_{0\le i\le n_{K,b}} d^{(i)}$
to denote the maximum feature dimension across censoring levels.



\begin{theorem}[Regret under Ideal Conditions]\label{cor:ideal}
Suppose that Assumptions \ref{ass: DGM}-\ref{ass: UQ ass} hold and let $K = \lceil \frac{\ln NT}{2(1-\gamma)} \rceil$. Then for any  $\epsilon \in (0,1)$ and $\varepsilon \in (0, 1)$ such that $2\big((n_{K,b}+3)\epsilon+\varepsilon\big) < 1$, the following regret upper bound holds with probability at least $1-2\big((n_{K,b}+3)\epsilon+\varepsilon\big)$
\footnotesize{\begin{align}
\max(\textup{\textbf{Regret}}(\widehat \pi_K),\ \textup{\textbf{Regret}}(\widetilde \pi_K))
&\lesssim \mathcal{E}_{\text{stat}}(NT, d)\\
&:=\max(
\big(d+\log(1/\epsilon)\big)d
\sqrt{\frac{C_{\mathrm{cov}}}{\rho NT}},\ 
d
\sqrt{\frac{\big(d+\log(1/\epsilon)\big)d}{\rho\,NT}}
)+
C_5(\varepsilon)(NT)^{-\delta}
+
\omega_{NT}.\nonumber
\end{align}
}
\end{theorem}
The proof of Theorem \ref{cor:ideal} is given in Appendix~\ref{app:proof-ideal}. Specifically, it relies on the following key fact, which holds given Assumptions \ref{assump:bounded-n}-\ref{assump:sufficient-coverage-main-text} and for any $\widehat{\pi} \in \Pi_{n_{K,b}}$.
\begin{align}\label{eq: reduced regret} 
\textbf{Regret}(\widehat{\pi})&=\EE^{\pi^*}[\sum_{t=0}^{\infty}\gamma^tR_t\given \mathcal{C}(K,n_{K,b})] -\EE^{\widehat{\pi}}[\sum_{t=0}^{\infty}\gamma^tR_t]= \EE^{\pi^*}[\sum_{t=0}^{\infty}\gamma^tR_t] -\EE^{\widehat{\pi}}[\sum_{t=0}^{\infty}\gamma^tR_t]. \end{align}
Specifically, under the Feasibility and Sufficient Coverage Assumptions (i.e Assumptions \ref{assump:bounded-n} and \ref{assump:sufficient-coverage-main-text}), the term $(ii)$ in Equation \eqref{eq: challange 3} vanishes as $\mathcal{C}(K,n_{K,b})$ becomes an almost sure event under $\pi^*$. Accordingly, the general regret decomposition in Equation \eqref{eq: challange 3} reduces to Equation \eqref{eq: reduced regret}. This provides the basis for Theorem~\ref{cor:ideal}.

With this reduction in place, Theorem~\ref{cor:ideal} establishes the statistical efficiency of our algorithm under ideal conditions for both $\widehat{\pi}_K$ and $\widetilde{\pi}_K$. The bound shows that the leading statistical term $\mathcal{E}_{\text{stat}}(NT,d)$ decays at a rate of $\tilde{O}(1/\sqrt{NT})$ with respect to the total sample size $NT$, while scaling polynomially (in particular, at most quadratically) with the feature dimension $d$ up to logarithmic factors. This convergence rate matches the information-theoretic lower bound for linear MDP up to logarithmic factors, demonstrating that our algorithm is minimax optimal~\citep{jin2021pessimism}. Finally, the selection of $K$ reflects the classical bias-variance trade-off found in FQI: increasing $K$ reduces truncation bias ($\propto \gamma^K$) but amplifies the accumulated statistical error. To preserve the optimal convergence rate of $\tilde{O}(1/\sqrt{NT})$, these errors must be balanced, dictating the logarithmic growth rule $K = \left\lceil \frac{\ln(NT)}{2(1-\gamma)} \right\rceil$.

\noindent\textbf{Managerial insight.}
The bound quantifies the economic value of data. The $\sqrt{NT}$ convergence rate assures managers that the algorithm is data-efficient. It provides a guarantee that the system will not get stuck in suboptimal strategies but will improve as the offline data grows. Furthermore, the dependence on $d$ highlights a strategic choice in model design. While including more features (increasing $d$) allows for capturing richer market dynamics (e.g., seasonality, competitor prices), it requires proportionally more data to learn effectively. Managers must balance the desire for a highly granular model against the availability of historical data to support it. Finally, in Appendix~\ref{app: remarks}, we further discuss the regret bound in Theorem~\ref{cor:ideal}, showing how it reduces in the independent-demand regime and how it compares to related results in the inventory literature under full observability.

\subsubsection{Layer 2: Relaxing Sufficient Coverage.}
We now relax Assumption~\ref{assump:sufficient-coverage-main-text}, which means that 
$n > n_{K,b}$. In other words, under the optimal policy $\pi^{*}$, the system can experience a longer sequence of consecutive censoring events than under the behavior policy. As a result, $\pi^{*}$ belongs to the larger policy class $\Pi_{n}$, where 
$\Pi_{n_{K,b}} \subset \Pi_{n}$ (i.e $\pi^* \notin \Pi_{n_{K,b}}$). Essentially, this relaxation introduces a part of the optimal policy’s behavior that cannot be identified from the offline data. In our regret analysis \eqref{eq: challange 3}, this is captured by the event 
$\mathcal C^{c}(K, n_{K,b})$, which now has a non-zero probability of occurring under $\pi^{*}$. Therefore, the full regret decomposition no longer simplifies to Equation~\eqref{eq: reduced regret} and terms (i) and (ii) must be analyzed separately. As a result, under this relaxation, we have a new, irreducible error term corresponding to term (ii) of the decomposition that is not present in the ideal setting. Specifically,

\begin{corollary}[Regret under Relaxed Coverage]\label{cor:relaxed_coverage}
Suppose all assumptions from Theorem~\ref{cor:ideal} hold, \emph{except} Assumption~\ref{assump:sufficient-coverage-main-text} (Sufficient Coverage). Let Assumption~\ref{ass: monotonic increase in alpha} (stated below) hold and $K = \lceil \frac{\ln NT}{2(1-\gamma)} \rceil$. Then for any  $\epsilon \in (0,1)$ and $\varepsilon \in (0, 1)$ such that $2\big((n_{K,b}+3)\epsilon+\varepsilon\big) < 1$, the following regret bound holds with probability at least $1 -2\big((n_{K,b}+3)\epsilon+\varepsilon\big)$:
\begin{align*}
 \max(\textup{\textbf{Regret}}(\widehat \pi_K),\ \textup{\textbf{Regret}}(\widetilde \pi_K))
&\lesssim
\underbrace{\mathcal{E}_{\text{stat}}(NT, d) \cdot \prob^{\pi^*}\!\big(\mathcal{C}(K,n_{K,b})\big)}_{\text{upper bound of term (i)}}+\underbrace{\mathcal{E}_{\text{cov}}(K, n_{K,b})}_{\text{ upper bound of term (ii)}},
\end{align*}
where $\mathcal{E}_{\text{stat}}(NT, d)$ is the term from the ideal setting in Theorem~\ref{cor:ideal}, and $\mathcal{E}_{\text{cov}}(K, n_{K,b})$ is the \textbf{unidentifiable regret}:
$
\mathcal{E}_{\text{cov}}(K, n_{K,b}) := (K-n_{K,b}+1)(1-\alpha_{\pi^*,1})^{n_{K,b}}\left(\frac{2R_{\max}}{1-\gamma}\right)$.
\end{corollary}
The proof of Corollary \ref{cor:relaxed_coverage} is given in Appendix~\ref{app:proof-relaxed-coverage}. Compared to the ideal bound in Theorem~\ref{cor:ideal}, the regret now consists of two distinct components, which directly correspond to the term (i) (learnable part) and term (ii) (unidentifiable part) from the general regret decomposition in Equation~\eqref{eq: challange 3}. In particular, the new component, $\mathcal{E}_{\text{cov}}(K, n_{K,b})$, is the bound on term (ii) of the regret decomposition. It captures the regret arising from the \textbf{unidentifiable component} of the optimal policy's behavior. Because the offline data contains no trajectories with more than $n_{K,b}$ consecutive censorings, we cannot learn how to act in these scenarios, making it impossible to minimize this term. Our approach, therefore, is to establish an upper bound on its magnitude. This requires a mild regularity assumption on how the optimal policy behaves in response to repeated censoring. Let $\alpha_{\pi^*,j} = \prob^{\pi^*}(\Delta_{t+1}=1 \given \Delta_{t}=\cdots=\Delta_{t-j+1}=0)$ be the probability of becoming uncensored after $j$ consecutive censorings.

\begin{assumption}\label{ass: monotonic increase in alpha}
 We assume $\alpha_{\pi^*,j}$ is non-decreasing in $j$.
\end{assumption}
Assumption \ref{ass: monotonic increase in alpha} is based on the notion that repeated censoring may lead consumers to seek alternative stores, ultimately reducing long-term demand. Therefore, the optimal policy prevents consecutive censoring to a certain extent to strike an optimal balance between inventory and stockout costs.

Under this assumption as detailed in Appendix \ref{sec:Alt. regret decomp}, we can bound term (ii) by $\mathcal{E}_{\text{cov}}(K, n_{K,b})$. This bound reveals a crucial trade-off: it \textbf{decreases exponentially} as $n_{K,b}$ increases, meaning richer offline data that explores longer censoring sequences rapidly shrinks this unlearnable component. However, the bound \textbf{increases linearly} as the analysis horizon, $K$, grows. This is because a larger $K$ simply expands the window of opportunity for the optimal policy $\pi^*$ to exhibit its unobserved behavior (a censoring sequence $> n_{K,b}$), thereby increasing the probability of the unlearnable event $\mathcal{C}^{c}(K,n_{K,b})$.

The first component, $\mathcal{E}_{\text{stat}}(NT, d) \cdot \prob^{\pi^*}\!\big(\mathcal{C}(K,n_{K,b})\big)$, corresponds to term (i) from the regret decomposition. It represents the regret on the trajectories that are supported by the offline data within first $K$ horizon. This implies that we can still leverage the error bound $\mathcal{E}_{\text{stat}}(NT, d)$ from Theorem~\ref{cor:ideal}. In particular, the conditioning event $\mathcal{C}(K, n_{K,b})$ guarantees that for the first $K$ time steps, the optimal policy's behavior is indistinguishable from a learnable policy within our supported class $\Pi_{n_{K,b}}$. Therefore, we propose to approximate the infinite-horizon value of $\pi^*$ with the value of the learnable policy $\pi^*_{n_{K,b}}$. This approximation is valid because for a sufficiently large $K$, the discount factor $\gamma^{K+1}$ makes the rewards from the distant, unidentifiable future (where $\pi^*$ might enter states $i > n_{K,b}$) negligible. 

In summary, under relaxed coverage, the total regret decomposes into: (i) a learnable component $\mathcal{E}_{\text{stat}}(NT, d)$ improved by more and better data; and (ii) an unidentifiable component $\mathcal{E}_{\text{cov}}$ that falls exponentially in $n_{K,b}$ but grows linearly with $K$. We set $K \approx \ln(NT)$ as in the ideal setting to ensure the learnable component (term (i)) converges to zero as data size increases. Although this choice increases the upper bound on the unidentifiable regret (term (ii)), the rate is only logarithmical and thus incremental. 

\subsubsection{Layer 3: Relaxing Termination–Action Domain Validity (TAD).} We now relax Assumption~\ref{ass:TAD}. Recall that this assumption addresses a subtle mismatch: during our iterative algorithms, intermediate policies select actions that are greedy over the \emph{entire} action space $\mathcal{A}$. However, the optimal policy $\pi^{\ast}_{n_{K,b}}$ selects from the \emph{restricted} feasible set $\mathrm{UP}^{(n_{K,b})}$—the set of actions that guarantee termination of censoring. Because $\mathrm{UP}^{(n_{K,b})}$ is unknown in the offline setting, an intermediate policy's greedy action may lie outside this feasible set. Therefore, when this assumption fails, the accumulated loss from these infeasible actions emerges in the regret bound, which are quantified as a Termination Action Penalty (TAP).

\begin{corollary}[Regret under Relaxed TAD]\label{cor:relaxed-TAD}
Suppose all assumptions from Theorem~\ref{cor:ideal} hold, \emph{except} Assumption~\ref{ass:TAD} and let $K = \lceil \frac{\ln NT}{2(1-\gamma)} \rceil$. Then for any  $\epsilon \in (0,1)$ and $\varepsilon \in (0, 1)$ such that $2\big((n_{K,b}+3)\epsilon+\varepsilon\big) < 1$, the following regret bound holds with probability at least $1 -2\big((n_{K,b}+3)\epsilon+\varepsilon\big)$:
\begin{align*}
\max(\textup{\textbf{Regret}}(\widehat \pi_K),\ \textup{\textbf{Regret}}(\widetilde \pi_K))
\;\lesssim\;
\mathcal{E}_{\text{stat}}(NT, d)
\;+\;
+
\max\bigl(\mathrm{TAP}(\widehat{\pi}_K),\mathrm{TAP}(\widetilde{\pi}_K)\bigr),
\end{align*}
where 
$
\mathrm{TAP}(\pi)
:=
\frac{1-\gamma^{n_{K,b}}}{(1-\gamma)^2}
\EE_{H^{(I)} \sim d_{\nu,I}^{\pi}}
\!\left[
\EE^{\pi}\!\left[\textup{TAD}^{\pi}(H^{(I)},A^{(I)}) \mid H^{(I)}\right]
\prod_{j=1}^{I}\mathbb{I}[\Delta^{(j)}=0]\,
\mathbb{I}[\Delta^{(0)}=1]
\right],
$
and the definition of TAD for C-FQI and PC-FQI can be found in Definition ~\hyperlink{total_action_discrepancy_ii}{\ref{def:action_discrepancy}-(ii)} and~\hyperlink{total_action_discrepancy_iii}{-(iii)} of the Appendix, respectively. 
\end{corollary}
The proof of Corollary \ref{cor:relaxed-TAD} is given in Appendix~\ref{app:proof-relaxed-TAD}. In this bound, $d_{\nu,I}^{ \pi}$ denotes the generalized discounted visitation measure under $\pi$ and $I$ is the random variable whose characterizations are given in Definitions \ref{def: generalized discounted visitation} and \ref{def: random var. I} of Appendix, respectively and $H^{(I)}=(W^{(0)},A^{(0)},\cdots W^{(I)})$. 

\noindent\textbf{Managerial insight.}
TAP captures the operational cost of allowing the learning algorithm to take actions that \emph{fail to resolve} consecutive stockouts even when a feasible resolution exists. This corresponds to unsafe exploration: temporary policies that increase stockouts to gather information, which is risky in practice. When Assumption~\ref{ass:TAD} fails, \(\mathrm{TAP}\) precisely quantifies the resulting performance loss.

\subsubsection{Layer 4: Relaxing Boundary–Action Alignment.}
We now relax Assumption~\ref{ass:AD}. This assumption concerns the behavior of deploying the learned policy produced by our algorithms specifically at the \emph{boundary} of the data-supported censoring depth. When the system reaches $n_{K,b}$ consecutive censoring periods (i.e., $i=n_{K,b}$), the optimal policy selects the action that maximizes the true $Q$–function over the unknown $\mathrm{UP}^{(n_{K,b})}$, ensuring the end of censoring streak. To guarantee termination of censoring in the testing environment, the algorithm applies a conservative safeguard action $a_{\max} \in \mathrm{UP}^{(n_{K,b})}$. However, even though $a_{\max}$ is \emph{feasible}, it may not maximize the learned $Q$–functions $\widehat{Q}^{(n_{K,b})}_K$ or $\widetilde{Q}^{(n_{K,b})}_K$. This creates the Boundary–Action Discrepancy (AD). Assumption~\ref{ass:AD} was imposed specifically to eliminate this source of mismatch. If this assumption fails, the resulting discrepancy appears in the regret bound as \textbf{Boundary Action Penalty} ($\mathrm{BAP}$).

\begin{corollary}[Regret under Relaxed AD]\label{cor:relaxed-AD}
Suppose all assumptions from Theorem~\ref{cor:ideal} hold, except Assumption~\ref{ass:AD} and let $K = \lceil \frac{\ln NT}{2(1-\gamma)} \rceil$. Then for any  $\epsilon \in (0,1)$ and $\varepsilon \in (0, 1)$ such that $2\big((n_{K,b}+3)\epsilon+\varepsilon\big) < 1$, the following regret bound holds with probability at least $1 -2\big((n_{K,b}+3)\epsilon+\varepsilon\big)$:
\begin{align*}
\max(\textup{\textbf{Regret}}(\widehat \pi_K),\ \textup{\textbf{Regret}}(\widetilde \pi_K))
\;\lesssim\;
\mathcal{E}_{\text{stat}}(NT,d)
\;+\;
\max(\mathrm{BAP}(\widehat \pi_K),\mathrm{BAP}(\widetilde \pi_K)),
\end{align*}
where $\mathrm{BAP}(\pi)
:=
\frac{1-\gamma^{n_{K,b}}}{(1-\gamma)^2}
\EE_{H^{(I)} \sim d_{\nu,I}^{\pi}}
\!\left[
\textup{AD}^{\pi}(H^{(I)})\,\mathbb{I}[I=n_{K,b}]
\right],$
and the definition of AD for C-FQI and PC-FQI can be found in Definition ~\hyperlink{AD_tilde_pi}{\ref{def:Qhat_action_discrepancy}-(i)} and~\hyperlink{AD_hat_pi}{-(ii)} of the Appendix, respectively.
\end{corollary}
The proof of Corollary \ref{cor:relaxed-AD} is given in Appendix~\ref{app:proof-relaxed-AD}. Specifically, as indicated by the term $\mathbb{I}[I=n_{K,b}]$, this penalty is only incurred at the specific moment the censoring streak $I$ hits the maximum supported depth $n_{K,b}$. $\mathrm{BAP}$ thus isolates the cost of enforcing a safe but suboptimal action $a_{\max}$ precisely at the boundary where the algorithm's knowledge ends. Intuitively, AD quantifies the loss incurred when the emergency action taken at depth $n_{K,b}$ is not the profit-maximizing one, even though it is operationally safe.

\noindent\textbf{Managerial insight.}
AD captures the performance loss arising from an overly conservative but corrective action at the censoring boundary.
Operationally, this reflects situations where the firm takes a safe action that guarantees uncensored next period but may not be profit-maximizing under current conditions.
When Assumption~\ref{ass:AD} fails, $\mathrm{BAP}$ quantifies exactly how much profit is sacrificed due to such misaligned boundary responses.

\subsubsection{Summary: The General Regret Bound.}

We conclude our theoretical analysis by synthesizing the relaxations from Layers 2 through 5. We note that the discussion on the relaxation of Layer 5 is given in Appendix~\ref{sec: layer 5}. At a high level, Layer~5 shows that moving from the linear models to RKHS setting preserves the almost the same regret decomposition, with the statistical term, denoted by $\overline{\mathcal{E}}_{\text{stat}}$ in this setting, now governed by the kernel’s effective dimension $d_*$ rather than a fixed dimension $d$. The exact characterization of $d_*$ is provided in Definition \ref{def:effective_dim} of Appendix~\ref{appendix: key lemmas}. By combining the generalized structural capacity of RKHS (Layer 5), the implications of insufficient censoring coverage (Layer 2), and the potential alignment discrepancies in intermediate and boundary actions (Layers 3 and 4), we arrive at the most general regret bound for our Algorithms.

\begin{theorem}\label{thm:general_regret}
Suppose that Assumptions \ref{ass: DGM}, \ref{ass:consistency}-\ref{ass: monotonic increase in alpha} hold and the Linear MDP, Sufficient Coverage, TAD, and AD assumptions are relaxed with $K = \lceil \frac{\ln NT}{2(1-\gamma)} \rceil$. Then for any  $\epsilon \in (0,1)$ and $\varepsilon \in (0, 1)$ such that $2\big((n_{K,b}+3)\epsilon+\varepsilon\big) < 1$, the following regret bound holds with probability at least $1 -2\big((n_{K,b}+3)\epsilon+\varepsilon\big)$:
\begin{align}\label{eq:master_bound}
\max(\textup{\textbf{Regret}}(\widehat \pi_K),\ \textup{\textbf{Regret}}(\widetilde \pi_K))\;\lesssim\; 
&\underbrace{\prob^{\pi^*}(\mathcal{C}(K, n_{K,b}))}_{\text{Learnability Probability}} \cdot \left[ 
\underbrace{\overline{\mathcal{E}}_{\text{stat}}(NT, d_*)}_{\text{Statistical \& Structural Error}} \nonumber \right. \\
&\left.+ \underbrace{\max(\mathrm{TAP}(\widehat \pi_K),\mathrm{TAP}(\widetilde \pi_K)) + \max(\mathrm{BAP}(\widehat \pi_K),\mathrm{BAP}(\widetilde \pi_K))}_{\text{Alignment Discrepancy}} 
\right] \nonumber \\
&+ \underbrace{\mathcal{E}_{\text{cov}}(K, n_{K,b})}_{\text{Unidentifiable Regret}},
\end{align}
where $
\overline{\mathcal{E}}_{\text{stat}}(NT, d_*) :=
\max(
d_*\big(d_*+\log(1/\epsilon)\big)
\sqrt{\frac{C_{\mathrm{cov}}}{\rho NT}},\;
\sqrt{d_*\big(d_*+\log(1/\epsilon)\big)}\;
\sqrt{\frac{1+d_*}{\rho\,NT}}
)
+ C_5(\varepsilon)(NT)^{-\delta} + \omega_{NT}.$
\end{theorem}
The proof of Theorem \ref{thm:general_regret} is given in Appendix~\ref{app:proof-general-theorem}. Equation \eqref{eq:master_bound} serves as the master regret for the proposed Algorithms. Observe that there is a slight difference in the structure of the leading statistical term $\overline{\mathcal{E}}_{\text{stat}}$ compared to the one in idealized setting $\mathcal{E}_{\text{stat}}$. This comes from the change of dimension $d$ with effective dimension of kernel $d_*$ and slight change in the regret bound of C-FQI for this generalized setting. Overall, this bound demonstrates that minimizing regret is not solely a problem of gathering more samples ($NT$), but a structural balancing act. It tells the firm that success depends on three pillars: (1) Data Quantity \& Coverage (driving down $\overline{\mathcal{E}}_{\text{stat}}$ and $\mathcal{E}_{\text{cov}}$), (2) Model Fit (choosing the kernel/features to manage $d_*$), and (3) Operational Consistency (ensuring the algorithm's recommended safety actions during stockouts align with profit maximization and minimizing $\mathrm{BAP}$). Neglecting any single pillar, for example,  ignoring the unobservable long-term stockout risks (high $\mathcal{E}_{\text{cov}}$), will impose a hard floor on the achievable performance. Moreover, the bound's structural terms—\textbf{Unidentifiable Regret} ($\mathcal{E}_{\text{cov}}$) and \textbf{Alignment Penalties} ($\mathrm{TAP}, \mathrm{BAP}$)—quantify the irreducible cost of dependent and censored demand and do not necessarily vanish with sample size.

\section{Numerical Results}\label{sec: numerical results}
In this section, we present results from numerical experiments that validate the effectiveness of the proposed algorithms. 
We report the main findings in the body of the paper and defer technical details to the appendix. Specifically, the Appendix~\ref{sec: numerical details} provides the definitions of the state and action variables that characterize the environment, the procedure for generating the offline dataset with censored observations, and additional implementation details.

We consider two scenarios with two different behavior policies to generate offline dataset. In particular, we use a uniform behavior policy, $\pi^b$, over the action space in the first scenario while, in the second scenario, we employ $\pi^{\ast}$ as the behavior policy, where we approximate $\pi^\ast$ by numerically solving the specialized Bellman optimality equations (Eqs.~\eqref{eq:bellman-compact}--\eqref{eq:bellman-compact2}) via Monte Carlo method, where the expectations in the dynamic programming recursion are approximated using large-sample averages generated directly from the simulation environment. 
For comparison, we consider $\pi^\ast$ and the oracle policy, $\pi^{\ast}_{\text{oracle}}$, given in Equation \eqref{eq: oracle optimal}. To approximate the oracle policy, we implement an online RL algorithm known as Proximal Policy Optimization (PPO) \citep{schulman2017proximal} in the environment where demand is observable throughout the entire horizon.   

\begin{figure}[!htp]
  \centering
  \begin{subfigure}[b]{0.32\textwidth}
    \centering
    \includegraphics[width=\textwidth]{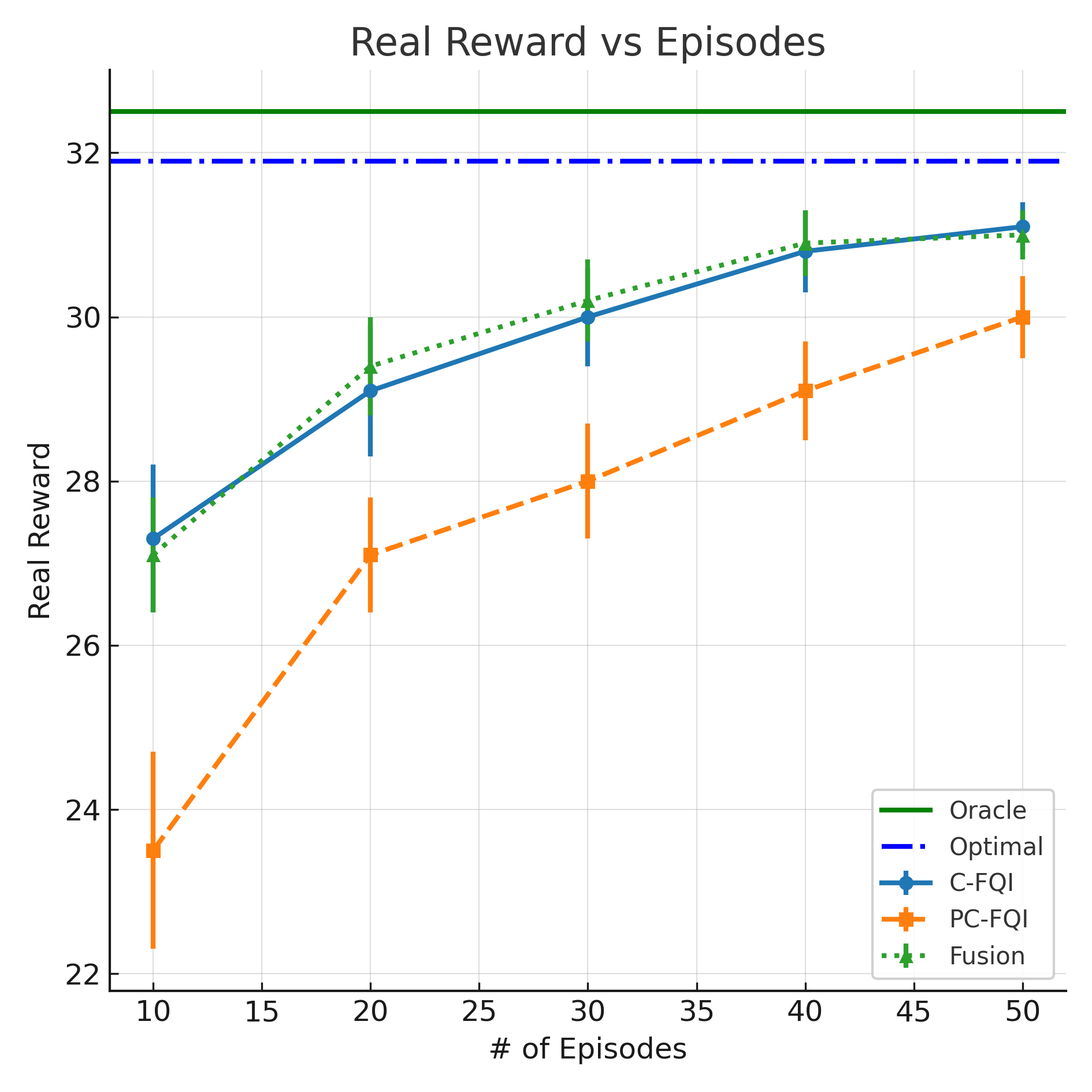}
    \captionsetup{font=small}
\subcaption{}
    \label{fig:fqi_plot}
  \end{subfigure}
  \hspace{0.0001\textwidth}
  \begin{subfigure}[b]{0.32\textwidth}
    \centering
    \includegraphics[width=\textwidth]{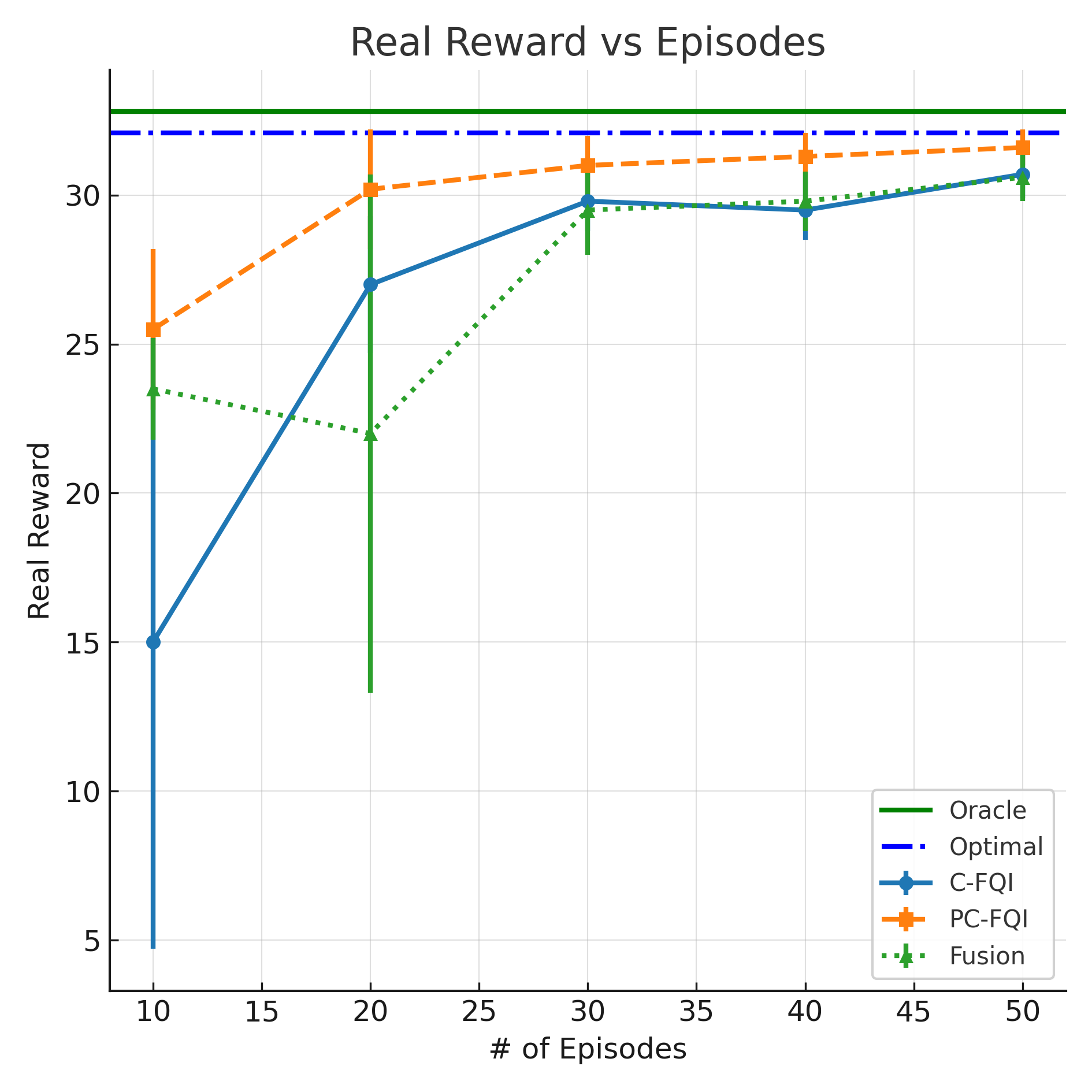}
    \captionsetup{font=small}
\subcaption{}
    \label{fig:pess_plot}
  \end{subfigure}
  \caption{Convergence of the proposed algorithms to the optimal policy as the sample size increases: (a) Results under a uniform behavior policy; (b) Results under the optimal behavior policy. The error bars represent 0.95 confidence intervals.}
  \label{fig:comparison}
\end{figure}

The Figures \ref{fig:fqi_plot}
 and \ref{fig:pess_plot} illustrate the performance of our algorithms under uniform and optimal behavior policies, respectively. These results clearly demonstrate that our proposed algorithms' performance converges to that of the optimal policy as the sample size increases. Additionally, the figures reveal an important insight: there is a cost associated with censoring, evident in the gap between the performance of the oracle and the optimal one. This indicates that censoring diminishes the potential to achieve higher rewards, as anticipated.

In the first simulation, where the behavior policy is uniform, C-FQI outperforms PC-FQI as depicted in Figure \ref{fig:fqi_plot}. This is consistent with both theoretical expectations and our algorithmic design. When the data distribution diverges from that induced by the optimal policy, sufficient exploration is necessary to achieve higher performance—a task that C-FQI accomplishes effectively. Notably, C-FQI tends to explore more as it does not incurs penalties for exploration, whereas PC-FQI discourages exploration by penalizing both states and actions that are infrequent in the training data using the uncertainty quantifier. The fusion of these two algorithms performs comparably to C-FQI. Initially, it adopts a pessimistic approach and refrains from extensive exploration during the first four iterations when the uncertainty is high. However, after the fourth iteration, it begins to explore more, allowing it to catch up with C-FQI.

In the second simulation, where the behavior policy is near optimal, we observe the opposite case: PC-FQI outperforms C-FQI as shown in Figure \ref{fig:pess_plot}. This outcome also aligns with theoretical expectations and the structure of proposed algorithms. Since the offline data distribution matches the optimal policy, avoiding infrequent states and actions in the offline data is crucial for better performance, which PC-FQI achieves effectively. In contrast, C-FQI encourages exploration, which can lead to divergence from the optimal policy distribution and, consequently, worse performance. Finally, the fusion of these two algorithms performs similarly to C-FQI, as it explores more than PC-FQI but not as extensively as C-FQI.

\section{Conclusion}\label{sec: conclusion}


In this paper, we study an offline sequential feature-based pricing and inventory control problem in which current demand depends on past demand levels and any demand exceeding available inventory is censored. Using an offline dataset of past prices, ordering quantities, inventory levels, covariates, and censored sales, our goal is to learn an optimal joint pricing and ordering policy that maximizes long-run profit. The central difficulty is that demand censoring creates missing profit information and breaks the Markov property in the observed state, leading to a non-stationary optimal policy under the observed process. To address this, we show the observed dynamics admit a high-order Markov structure indexed by the length of consecutive censoring and characterize the optimal policy via specialized Bellman equations that explicitly account for consecutive censoring. Building on this foundation, we propose two novel algorithms C-FQI and its uncertainty-aware counterpart, PC-FQI that synthesizes the developed modeling framework with offline reinforcement learning and survival analysis to effectively estimate optimal policy from historical data. We further provide finite-sample regret guarantees, decomposing the performance gap into interpretable components.

To conclude, we highlight several avenues for future research. Our current focus is on a single agent; however, supply chains typically involve multiple echelons where decisions interact significantly. Investigating competition and cooperation among echelons under incomplete information presents a valuable area for exploration. Additionally, our model assumes simultaneous pricing and ordering at the start of each period. In reality, firms often adjust prices dynamically after observing initial demand signals—a strategy known as responsive pricing. Incorporating responsive pricing into our framework represents an intriguing direction, offering deeper insights into how firms can adapt strategies intra-period to further optimize performance.

\newpage

\appendix

\section{Organization of the Appendix}\label{app:org}

The Appendix collects all proofs, supporting lemmas, and supplementary discussions that underlie the theoretical and algorithmic developments in the main paper. Below we provide a roadmap so readers can locate the components relevant to their interests.

\paragraph{Appendix~\ref{app: additional lit}:  Additional Literature Review.} Appendix~\ref{app: additional lit} places our contribution within the broader offline reinforcement-learning literature. It explains why classical offline RL methods fail in the presence of censored and dependent demand, outlines the construction of the high-order MDP and the associated Bellman equations, and motivates the development of C-FQI and PC-FQI as principled adaptations of fitted Q-iteration with reward imputation and pessimism.

\paragraph{Appendix~\ref{app: uncensoring}: Why does not Na\"{\i}ve ``Uncensoring'' the State Fix the Structure?}
Appendix~\ref{app: uncensoring} provides a dedicated discussion explaining why simply imputing latent demand (or rewards) does not restore the Markov property in the observational state under temporally dependent demand. This section directly supports the modeling choice of censoring-aware, history-indexed value functions and specialized Bellman equations developed in Sections~\ref{sec:local-structure}--\ref{subsec:censored-reward}.

\paragraph{Appendix~\ref{sec: layer 5}: Layer~5 --- Relaxing Linear MDP Structure.}
Appendix~\ref{sec: layer 5} presents the final extension beyond the linear MDP condition by allowing nonlinear dynamics modeled in a reproducing kernel Hilbert space (RKHS). 

\paragraph{Appendix~\ref{sec: algo}: Algorithmic Details.}
Appendix~\ref{sec: algo} provides a consolidated algorithm table of the proposed algorithms, C-FQI and PC-FQI.

\paragraph{Appendix~\ref{sec: numerical details}: Details of Numerical Results.}
Appendix~\ref{sec: numerical details} contains experimental details (i.e., implementation choices)

\paragraph{Appendix~\ref{app: additional mat}: Additional Material}

Appendix~\ref{app: additional mat} complements the main theoretical development by situating the proposed framework within classical inventory models, clarifying the structural novelty introduced by censoring and temporal dependence, and positioning the regret guarantees relative to both the pricing and inventory-control and offline reinforcement-learning literatures.





\paragraph{Appendix \ref{sec:Alt. regret decomp}: Regret Decomposition.}
Appendix~\ref{sec:Alt. regret decomp} presents the regret decomposition that serves as the starting point for all subsequent theoretical derivations in the paper.

\paragraph{Appendix \ref{app:proof-ideal}: Proof of Theorem~\ref{cor:ideal} (Idealized Setting).}
Appendix~\ref{app:proof-ideal} contains the complete proof of the regret guarantee in the idealized regime (i.e., Layer 1). Concretely, it utilizes the regret decomposition derived in Appendix~\ref{sec:Alt. regret decomp} under the idealized setting.

\paragraph{Appendix \ref{app:proof-relaxed-coverage}--\ref{app:proof-relaxed-structure}: Proofs for Layered Relaxations (Corollaries~\ref{cor:relaxed_coverage}--\ref{cor:relaxed-structure}).}
Appendices~\ref{app:proof-relaxed-coverage} through~\ref{app:proof-relaxed-structure} provide the proofs for the intermediate corollaries corresponding to Layers~2--5 in the layered regret analysis.
Appendix~\ref{app:proof-relaxed-coverage} proves Corollary~\ref{cor:relaxed_coverage} (Layer 2), which relaxes sufficient censoring coverage and introduces the unidentifiable regret component.
Appendix~\ref{app:proof-relaxed-TAD} proves Corollary~\ref{cor:relaxed-TAD} (Layer~3), which relaxes Assumption~\ref{ass:TAD} and yields the Total Action Discrepancy (TAD) penalty.
Appendix~~\ref{app:proof-relaxed-AD} proves Corollary~\ref{cor:relaxed-AD} (Layer~4), where Assumption~\ref{ass:AD} is relaxed.
Appendix~~\ref{app:proof-relaxed-structure} proves Corollary~\ref{cor:relaxed-structure} (Layer 5), which relaxes Linear MDP assumption (i.e., Assumption \ref{ass:linear-mdp}). Overall, they complete the sequence of relaxations needed before synthesizing the master theorem.
Together, these appendices isolate each added source of complexity.

\paragraph{Appendix \ref{app:proof-general-theorem}: Proof of Theorem~\ref{thm:general_regret} (General Regret Bound).}
Appendix~\ref{app:proof-general-theorem} synthesizes the relaxations established in Appendices~\ref{app:proof-relaxed-coverage}--\ref{app:proof-relaxed-structure} and derives the full general regret bound.

\paragraph{Appendix~\ref{appendix: key lemmas}--\ref{app: technical lemmas}: Key and Technical Lemmas.}
Appendix~\ref{appendix: key lemmas} collects the main auxiliary lemmas used throughout the proofs.
Appendix~\ref{app: technical lemmas} provides additional technical lemmas that are referenced across multiple proof sections.

\paragraph{Appendix \ref{app: Proof of Propositions and Lemmas}: Proof of Propositions and Lemmas.}
Appendix~\ref{app: Proof of Propositions and Lemmas} contains the complete proofs of the propositions and auxiliary lemmas stated in Appendix~\ref{appendix: key lemmas}--\ref{app: technical lemmas}. While these lemmas are summarized and invoked throughout the main theoretical developments, their detailed technical proofs are deferred to Appendix~\ref{app: Proof of Propositions and Lemmas} to preserve readability of the primary arguments.

\section{Additional Literature}\label{app: additional lit}

\textbf{Policy Learning through Offline Reinforcement Learning.}
The field of offline policy learning within the reinforcement learning (RL) framework has garnered significant attention in both computer science and statistics. Traditional offline RL approaches are typically built on the framework of MDPs \citep{bertsekas1995dynamic, sutton2018reinforcement}. Under this framework, it is well known that an optimal policy can be derived based on the optimal Q-function, obtained through solving the Bellman equation which relies on the Markov property \citep[e.g.,][]{ernst2005tree, riedmiller2005neural, jin2021pessimism, kumar2020conservative}. However, in our problem setting, existing offline RL methods fall short due to two primary challenges: the failure of the Markov property caused by the censored demand and the absence of reward information at censored data points.

Firstly, the failure of the Markov property as a result of the censored demand prevents us from characterizing the optimal policy using the standard Q-function, the cornerstone of conventional offline RL algorithms. In classical MDP settings, an optimal stationary policy always exists and it takes the greedy action with respect to the optimal Q-function. But, in our case, the structure of the optimal policy becomes unknown, potentially non-stationary and non-Markovian. We demonstrate that identifying the optimal policy is nearly equivalent to solving for an optimal policy within a specific high-order MDP, characterized by multiple optimal Q-functions that depend on consecutive censoring instances and can be obtained through an alternative form of Bellman equation tailored for our problem setting.

Secondly, the most existing offline RL algorithms assume that the offline data contains complete reward information throughout the observation period. In our scenario, however, reward information is missing at censored data points, which complicates the direct application of standard methods. To overcome this, we employ advanced survival analysis techniques to impute the missing reward information by generalizing the idea proposed in \cite{qi2022offline}.

With the alternative form of Bellman equation catered for our problem setting and imputed rewards, we adapt the idea of FQI algorithm to our specific context, leading to the development of the Censored Fitted Q-Iteration (C-FQI) algorithm. This algorithm is designed to approximate the aforementioned Bellman equation from the offline data with imputed reward information, and accordingly, output a near-optimal and data-driven policy.

Furthermore, a typical challenge in offline reinforcement learning is insufficient data coverage, which stems from the static nature of historical data and the lack of exploration. This issue arises in our problem as well and often leads to distribution shift, where the optimal policy differs significantly from the data available for training \citep{wang2020statistical,levine2020offline}. Recent research has aimed to address this challenge by adopting a pessimistic approach in policy optimization, minimizing exploration of rarely encountered state-action pairs within the batch data. This approach has been shown to mitigate the risks associated with insufficient data coverage and has led to the development of new algorithms based on pessimism principle \citep{kumar2020conservative, fujimoto2019off, liu2020provably, rashidinejad2021bridging, jin2021pessimism, zanette2021provable, zhan2022offline, fu2022offline}. In our work, we incorporate the pessimism principle into the C-FQI algorithm to address extrapolation error and possibly enhance its performance in pricing and inventory control tasks, leading to the development of PC-FQI. This approach allows us to extend the applicability of our method to real-world scenarios where the offline data coverage is often limited.

\section{Why does not Naive Uncensoring the State Fix the Structure?}\label{app: uncensoring}

A crucial question, central to the challenge of censored data, is whether imputation can effectively ``uncensor'' the problem and simplify it to a standard MDP. To investigate this, one might propose creating a plug-in state by replacing the observed sales, $Z_{t-1}$, with an imputed demand, $\widehat{D}_{t-1}$, whenever a stockout occurs:
\[
\widetilde W_t \;:=\; \big(X_t,\,Y_t,\,\widetilde Z_{t-1},\,\Delta_{t-1}\big),
\qquad
\widetilde Z_{t-1} \;:=\; 
\begin{cases}
D_{t-1}, & \Delta_{t-1}=1,\\
\widehat D_{t-1}, & \Delta_{t-1}=0.
\end{cases}
\]
While this approach is tempting, it is fundamentally flawed. The core reason is that a single plug-in number $\widehat D_{t-1}$ is \emph{not} a sufficient statistic for the future dynamics under censoring; the transition to next state and reward depend on the \emph{full posterior distribution} of the latent $D_{t-1}$, not just its point estimate. In other words, imputation addresses the \textit{statistical} issue of a missing data but fails to solve the deeper \textit{structural} problem of a non-Markovian process. This failure, which we detail and illustrate below, is precisely what necessitates the high-order framework developed in this paper.

To illustrate, assume the linear AR(1) demand model with Gaussian error:
\[
D_t \;=\; \theta_0+\theta_x X_t - \beta P_t + \rho D_{t-1} + \varepsilon_t^D,\qquad \varepsilon_t^D\sim\mathcal{N}(0,\sigma_D^2),
\]
and suppose $t\!-\!1$ is censored ($\Delta_{t-1}=0$) while $t\!-\!2$ is uncensored ($\Delta_{t-2}=1$), so $D_{t-2}=Z_{t-2}$ is known. Define the AR(1) predictor for $D_{t-1}$ given $W_{t-1}, A_{t-1}$:
\[
m_{t-1}:=\theta_0+\theta_x X_{t-1}-\beta P_{t-1}+\rho D_{t-2}.
\]
Under censoring at $t\!-\!1$, we only know $D_{t-1}>Y_{t-1}$, so the posterior distribution is \emph{truncated normal}:
\[
D_{t-1} \,\big|\, W_{t-1},A_{t-1},\,\Delta_{t-1}=0 \;\sim\; \mathcal{N}(m_{t-1},\sigma_D^2)\ \text{ truncated to }(Y_{t-1},\infty).
\]
Consequently, the \emph{predictive} distribution of $D_t$ given $H_{(t-1):t}$ is a normal mean–variance mixture:
\[
D_t \,\big|\, H_{(t-1):t}
\ \stackrel{d}{=}\ \theta_0+\theta_x X_t - \beta P_t + \rho D_{t-1} + \varepsilon_t^D,
\]
with
\[
\mu_t=\EE[D_t \mid H_{(t-1):t}] \;=\; \theta_0+\theta_x X_t - \beta P_t + \rho\,\EE[D_{t-1}\mid W_{t-1},A_{t-1},\,\Delta_{t-1}=0],
\]
\[
\sigma_t=\Var(D_t \mid H_{(t-1):t}) \;=\; \rho^2\,\Var(D_{t-1}\mid W_{t-1},A_{t-1},\,\Delta_{t-1}=0)\;+\;\sigma_D^2.
\]
A plug-in state $\widetilde W_t$ that stores only $\widehat D_{t-1}$ reproduces $\mu_t$ but \emph{not} $\sigma_t$; Hence 
\begin{equation}
\big(W_{t+1},R_t\big) \not\independent   (W_{t-1},A_{t-1})\ \Big|\ (\widetilde W_t, A_t),
\end{equation}
This implies the plug-in state is informationally incomplete. To recover the missing variance information, one must look back to the history through the latest time point, with no censoring which, in this example, is $(W_{t-1},A_{t-1})$. The plug-in state therefore fails to ensure first order conditional independence as in the case of underlying decision process $\{(S_t,A_t, R_t)\}_{t\geq 0}$. Therefore, the process remains fundamentally non-Markovian even if the plug-in state is used. This reinforces the necessity of characterization of the observed process.

Furthermore, this approach breaks the stationary-environment assumption required by all off-policy algorithms, a classic pitfall in off-policy learning. To make this concrete, consider that our imputation model is trained on a static dataset generated by a specific \textbf{behavior policy} (e.g., a ``cautious'' manager who keeps inventory high and rarely has stockouts). The goal of an off-policy algorithm, however, is to find a new, potentially very different \textbf{target policy} (e.g., an ``aggressive'' manager who keeps inventory low, leading to more frequent stockouts). When the algorithm evaluates this aggressive policy, it generates histories that are \textbf{out-of-distribution} for the imputation model. The model, trained on ``cautious'' data, will produce biased and unreliable estimates when confronted with the frequent, consecutive stockouts generated by the aggressive policy. This means the very definition of plug-in state, $\widetilde{W}_t$, becomes systematically wrong under the new policy. The rules of the game are no longer fixed; the state transition probabilities effectively change as the learned policy evolves. This violates the core assumption of off-policy optimization and makes convergence to a true optimum impossible.

\section{Layer 5: Relaxing Linear MDP Structure.}\label{sec: layer 5}
Finally, we relax Assumption~\ref{ass:linear-mdp}. Since operational dynamics can be nonlinear, we extend our analysis to kernel-based nonlinear function approximation in an RKHS, adopting the theoretical framework established by \cite{chang2021mitigating}. Algorithmically, this replaces the linear least-squares step in the FQI updates with kernel ridge regression under a kernel $k(\cdot,\cdot)$.

In the RKHS setting, the fixed feature dimension $d$ is replaced by a capacity measure of the kernel class. Following the analysis in \cite{chang2021mitigating}, we quantify this capacity using the \emph{effective dimension} of a kernel, denoted as $d_*$, where the exact characterization of this term is provided in Definition \ref{def:effective_dim} of Appendix~\ref{appendix: key lemmas}. Intuitively, $d_*$ captures the "effective dimension" of the feature space accessible by the kernel given $NT$ samples. As derived in \cite{chang2021mitigating}, the convergence rate of the uncertainty quantifier in the RKHS setting depends on this effective dimension rather than a static dimension. This generalization modifies the regret bound in the idealized setting as follows.
\begin{corollary}[Regret under General Structure]\label{cor:relaxed-structure}
Suppose all assumptions from Theorem~\ref{cor:ideal} hold, except Assumption~\ref{ass:linear-mdp} and let $K = \lceil \frac{\ln NT}{2(1-\gamma)} \rceil$. Furthermore, assume that the dynamics belong to a reproducing kernel Hilbert space (RKHS) characterized by a kernel $k(\cdot,\cdot)$. Then for any  $\epsilon \in (0,1)$ and $\varepsilon \in (0, 1)$ such that $2\big((n_{K,b}+3)\epsilon+\varepsilon\big) < 1$, the following regret bound holds with probability at least $1 -2\big((n_{K,b}+3)\epsilon+\varepsilon\big)$:
\begin{align*}
\max(\textup{\textbf{Regret}}(\widehat \pi_K),\ \textup{\textbf{Regret}}(\widetilde \pi_K))
\;\lesssim\;
\overline{\mathcal{E}}_{\text{stat}}(NT, d_*),
\end{align*}
where \begin{small}
\begin{equation*}
\overline{\mathcal{E}}_{\text{stat}}(NT, d_*) :=
\max\left(
d_*\big(d_*+\log(1/\epsilon)\big)
\sqrt{\frac{C_{\mathrm{cov}}}{\rho NT}},\;
\sqrt{d_*\big(d_*+\log(1/\epsilon)\big)}\;
\sqrt{\frac{1+d_*}{\rho\,NT}}
\right)
+ C_5(\varepsilon)(NT)^{-\delta} + \omega_{NT}
\end{equation*}
\end{small}
\end{corollary}

This bound replaces the $\mathcal{E}_{\text{stat}}(NT, d)$ term from the ideal setting in Theorem~\ref{cor:ideal} with $\overline{\mathcal{E}}_{\text{stat}}(NT, d_*)$. The key change is the substitution of the finite dimension $d$ with the effective dimension $d_*$, which is defined through the eigenspectrum of the kernel (via Mercer’s theorem) relative to the offline data distribution and the sample size \citep{chang2021mitigating}. Intuitively, $d_*$ measures how many effective degrees of freedom of the RKHS are learnable from $NT$ samples: kernels with faster spectral decay yield smaller $d_*$ at a given sample size, while richer kernels induce a larger $d_*$. Importantly, in the finite-dimensional linear-kernel case, $d_*$ is bounded by the (data-dependent) rank quantity and the bound recovers the linear-structure behavior (up to logarithmic factors). Finally, the choice of $K = \left\lceil \frac{\ln NT}{2(1-\gamma)} \right\rceil$ continues to reflect the same bias-variance trade-off as in the linear case: it makes the truncation bias $\gamma^K$ comparable to the leading statistical term (which decays at rate $(NT)^{-1/2}$), ensuring that the regret is asymptotically dominated by $\overline{\mathcal{E}}_{\text{stat}}(NT, d_*)$. Finally, it is worth emphasizing that the leading term in $\overline{\mathcal{E}}_{\mathrm{stat}}(NT,d_*)$ attains the state-of-the-art scaling for kernelized offline RL: up to logarithmic factors and problem-dependent constants (e.g., horizon and coverage), it behaves as $\tilde{O}\!\big(d_*^{\,2}/\sqrt{NT}\big)$, which is consistent with the effective-dimension rates established in \citet{chang2021mitigating}.

\textbf{Managerial insight.}
Relaxing the structural assumption sharpens the trade-off between model flexibility and data requirements through $d_*$. A simple representation (small $d_*$) is easier to learn and provides reliable improvements with limited data, but may miss nonlinear effects such as localized price sensitivities or regime changes. A more expressive kernel can capture these nuances, yet it effectively increases $d_*$ and therefore demands more historical data to separate signal from noise. Since $d_*$ is defined through the kernel spectrum relative to the offline data distribution, the same modeling choice can behave very differently across firms and markets: if the historical data concentrates on a low-dimensional set of operating regimes, the learnable complexity (and thus the achievable accuracy) is correspondingly limited. Practically, this suggests that firms with smaller datasets should favor simpler kernels or stronger regularization, while firms with large transaction logs can deploy richer kernel-based model, provided their data spans the operating regimes where decisions will be made.

\section{Algorithm}\label{sec: algo}

In this section, we present the details of our proposed algorithms in Table \ref{alg:algo full}.
\begin{sizeddisplay}
{\footnotesize}
\begin{algorithm}[!ht]
    \caption{Censored Fitted Q-Iteration and Pessimistic Censored Fitted Q-Iteration}
    \label{alg:algo full}
    \begin{algorithmic}[1] 
        \item[\textbf{Input:}] Observed data $\mathcal{O}_N = \{W_{j, t}, A_{j, t}, W_{j, t+1}\}_{0 \leq t < T-1, 1 \leq j \leq N}$.
        \item[\textbf{Step 1:}] Input $ \{W_{j, t}, A_{j, t}\}_{0 \leq t < T-1, 1 \leq j \leq N}$ and apply Kaplan-Meier method to estimate the conditional survival function $\textup{SF}$. Then, compute $\{\widehat{R}_{j, t}\}_{1 \leq j \leq N, 0 \leq t < T}$ based on the methods outlined in Section \ref{subsec:censored-reward}.
        \item[\textbf{Step 2:}] Partition the data in the following way $\forall i=0,\cdots,n_{K,b}$
        \begin{itemize}
            \item  $\begin{aligned}[t]
\mathcal{O}^{(i)}_N &= \left\{\left(W_{j, t-i}, A_{j, t-i},\cdots, W_{j, t},  A_{j, t}, R^{*}_{j,t}, W_{j, t+1}\right) \mid
\text{$\{\Delta_i\}_{i =t-i}^{t-1} = 0$ and $\Delta_{t-i-1}=1$},\right.\\
&\left. \quad\quad 0 \leq t < T-1, 1 \leq j \leq N\right\},
\end{aligned}$
            
      \end{itemize}

        \item[\textbf{Step 3:}] Initialize $\forall i=0,\cdots,n_{K,b}$
        \begin{itemize}
            \item $\widehat{Q}^{(i)}_0(H_{0:i}) \gets 0$,$\quad\quad$ 
            \item $\Tilde{U}^{(i)}_{0}(H_{0:i}) \gets 0$
        \end{itemize}
        \item[\textbf{Step 4 (FQI):}] For $k=0,\ldots, K-1$ and for each $i=0,\cdots,n_{K,b}$
        \begin{itemize}
            \item Compute $\zeta_{j,t}^{i,k}$ as shown in \eqref{eq: reponse for reg} 
            \item$
\begin{aligned}[t]
    \widehat{Q}^{(i)}_{k+1} \gets \argmin_{Q^{(i)} \in \calQ^{(i)}} \Bigg\{ & \sum_{\mathcal{O}^{(i)}_N} \Big[ Q^{(i)}(H_{j,(t-i):t}) - \zeta_{j,t}^{i,k}\Big]^2
    + \lambda J(Q^{(i)}) \Bigg\}
\end{aligned}
$
                \end{itemize}
                
            \item[\textbf{Step 4' (P-FQI):}] For $k=0,\ldots, K-1$ and for each $i=0,\cdots,n_{K,b}$
            \begin{itemize}
                 \item Compute $\widetilde{\zeta}_{j,t}^{i,k}$ as shown in \eqref{eq: response pessimism}
            \item$\begin{aligned}[t]
                    \widehat{Q}^{(i)}_{k+1}\gets \argmin_{Q^{(i)} \in \calQ^{(i)}} \Bigg\{ &\sum_{\mathcal{O}^{(i)}_N}\Big[Q^{(i)}(H_{j,(t-i):t})- \widetilde{\zeta}_{j,t}^{i,k}\Big]^2 +\lambda J(Q^{(i)})\Bigg\}
                \end{aligned}$
                 \item Compute $\Tilde{U}_{k+1}^{(i)}$ that satisfies Definition \ref{def:UE}.
            \item $\widetilde{Q}_{k+1}^{(i)} \gets \left(\widehat{Q}_{k+1}^{(i)} - \Tilde{U}_{k+1}^{(i)}\right)$
     \end{itemize}
      \item[\textbf{Output (FQI):}] \text{ } 
     \begin{itemize}
       
        \item $\begin{aligned}[t]
                    \text{For } i=0,\cdots,n_{K,b}-1\text{: } \widehat{\pi}_{K}^{(i)}(a | H^{'}_{(t-i):t}) =\begin{cases}
     1 & \text{if} \; a'=    \argmax_{a \in \calA}\widehat{Q}_{K}^{(i)}(H^{'}_{(t-i):t},a)\\
     0 & \text{otherwise}
    \end{cases}
                \end{aligned}$  
        \item For $i=n_{K,b}$: $\widehat{\pi}_{K}^{(n_{K,b})}(a_{\max}(H^{'}_{(t-n_{K,b}):t})| H^{'}_{(t-n_{K,b}):t})=1$
        \end{itemize}
        \item[\textbf{Output (P-FQI):}] \text{ }
        \begin{itemize}
        
        \item $\begin{aligned}[t]
                    \text{For } i=0,\cdots,n_{K,b}-1\text{: } \widetilde{\pi}_{K}^{(i)}(a |  H^{'}_{(t-i):t}) = \begin{cases}
     1 & \text{if} \; a'=    \argmax_{a \in \calA}\widetilde{Q}_{K}^{(i)}(H^{'}_{(t-i):t},a)\\
     0 & \text{otherwise}
    \end{cases}
                \end{aligned}$  
        \item For $i=n_{K,b}$\text{: }  $\widetilde{\pi}_{K}^{(n_{K,b})}(a_{\max}(H^{'}_{(t-n_{K,b}):t}) | H^{'}_{(t-n_{K,b}):t})=a_{\max}(H^{'}_{(t-n_{K,b}):t})$
        \end{itemize}
        \end{algorithmic}
\end{algorithm}
\end{sizeddisplay}

\section{Details of Numerical Results}\label{sec: numerical details}
The environment in which our algorithms are tested is characterized by specific features: the interest rate ($IR$) and the state of the economy ($SE$), denoted as $X_t = (IR_t, SE_t)$. The capacity of the inventory is set to 25, i.e., $I_{t} \in \{0, \cdots, 25\}$, and the maximum order quantity at any given time is 15, i.e., $O_t \in \{0, \cdots, 15\}$. For simplicity, we only allow discrete price actions such that $P_t \in \{4, 4.25, 4.5\}$. For the reward function, we set the stock-out cost $C_{1,t}=2$, the ordering cost $C_{2,t}=3$, and the holding cost $C_{3,t}=1$ for all $ t\geq 0$. 


In data generation phase, the interest rate is modeled using an autoregressive process, while the state of the economy is represented as a discrete variable indicating either an expanding or contracting economy. The demand process is also modeled as an autoregressive process influenced by various covariates, including the interest rate, the economic state, and the price. When generating the demand, we truncate it to ensure that the consecutive censoring is limited to 3 instances, i.e., $n_{T,b}$ is 3. 
In the numerical study, we consider two scenarios with two different behavior policies to generate actions. In particular, we use a uniform behavior policy, $\pi^b$, over the action space in the first scenario while, in the second scenerio, we employ $\pi^{\ast}$ as the behavior policy, where we approximate $\pi^\ast$ by a Monte Carlo method. For each experiment, we consider a $T = 50$ planning horizon of 50 episodes, executed independently 50 times. Training of our algorithms starts with 5 episodes (i.e., $5 \times 50$ observations) and increases in increments of 5 up to 50 episodes. For each episode level $(5, 10, \ldots, 50)$, we run $K=10$ iterations of the algorithms. The resulting policy is tested in the environment over 500 episodes. 
We then compute the average reward for these 500 episodes, and finally, report the overall mean across the 50 independent runs as a performance evaluation of our algorithm. In addition to running C-FQI and PC-FQI on these offline data sets, we also consider a hybrid approach that combines both algorithms. Specifically, this fusion approach runs PC-FQI for the first 4 iterations, followed by C-FQI for the subsequent 6 iterations. In each iteration of our algorithms, we utilize kernel ridge regression.  For hyperparameter tuning, we define a parameter grid that includes the regularization parameter $\lambda$, the kernel options, and the kernel bandwidth. Hyperparameter selection is performed via 5-fold cross-validation.

\section{Additional Material}\label{app: additional mat}

\subsection{Special Cases of the Proposed Modeling Framework}\label{app: special cases}

    \paragraph{Case 1: Classical Lost-Sales with Linear Demand (Reduction \& Policy).}
Under our linear demand model \ref{eq:linear_ar_demand}, we can set $\rho=0$ with and remove $X_t$. Thus $D_t=\alpha-\beta P_t+\varepsilon_t^D$, $\beta>0$. Assume only ordering decisions is made as well as lost sales, zero lead time, and stationary \emph{per-unit} costs (no fixed setup cost). Then the problem reduces to the textbook lost-sales model with linear demand. Accordingly, $\pi^*_{\text{oracle}}$ matches classical structures: Scenario (i) \emph{single-period}: the \emph{critical-fractile} order with price  chosen to maximize single-period expected profit; and Scenario (ii) \emph{multi-period, fixed price}: an \emph{order-up-to (base-stock)} policy \citep[e.g.,][]{porteus2002foundations}. 

\paragraph{Case 2: I.I.D.\ Demand with Pricing and Fixed Ordering Cost (Reduction \& Policy).} In the same linear demand model, 
eliminating $X_t$ and setting $\rho=0$ makes demand i.i.d.\ and linear in price. With a \emph{fixed} ordering cost and convex holding/shortage costs, the problem becomes the classic joint pricing--inventory model with lost sales; the optimal policy, $\pi^*_{\text{oracle}}$, admits a \emph{static $(s,S,p)$} structure \citep{chen2004coordinating}.

\paragraph{Case 3: Markovian Demand without Pricing (Reduction \& Policy).}
Omit pricing ($P_t$) while only retaining demand in the state via $D_{t-1}$. With zero lead time and per-unit costs, the problem reduces to lost-sales inventory control with \emph{Markovian} demand. Under standard convexity/regularity conditions, the optimal policy, $\pi^*_{\text{oracle}}$, is a \emph{state-dependent $(s(D_{t-1}),S(D_{t-1}))$ policy} (thresholds depending on the demand state) \citep[see][]{sethi1997optimality}.

\paragraph{Reference Price}
Furthermore, a key advantage of this framework is its ability to accommodate rich demand structures. First, any finite-order temporal dependence can be encoded by augmenting the state with appropriate lags of demand. Second, nonlinear effects of prices, inventory, and covariates are naturally allowed, since the MDP is defined through an arbitrary transition kernel and reward function rather than a specific linear specification.
Additionally, incorporating covariates (e.g., macroeconomic indicators within $X_t$) allows dependence across periods even when prices are controlled, generalizing models that implicitly assume demand is i.i.d.\ conditional on price \citep{petruzzi2002dynamic,araman2009dynamic,feng2014dynamic,chen2017efficient}. 
The framework also accommodates behavioral features such as \emph{reference price} effects \citep[e.g.,][]{chen2020data}. 
For example, under a commonly studied reference-price model where consumers compare the current price to a perceived reference price, demand can be specified as
\[
D_t = b - a p_t + c (r_t - p_t), \quad 
r_t = \alpha r_{t-1} + (1 - \alpha) p_{t-1},
\]
where $r_t$ represents the evolving reference price. 
In this case, augmenting the state by $r_t$ (or equivalently $p_{t-1}$) preserves the first-order MDP representation, demonstrating that such behavioral extensions are naturally accommodated within our framework.

\paragraph{Why is our modeling framework necessary?}
Our framework addresses two distinct consequences of censoring: (i) missing reward components under stockouts (since profit depends on $(D_t-Y_t)^+$), and (ii) loss of Markovian structure under \emph{dependent} demand because censoring hides the latent demand shocks that drive future demand.  If demand is independent across periods (e.g., $\rho=0$ in the AR(1) model), then the unobserved demand history carries no predictive information about the future, and the observed process remains Markov even under censoring; in this regime, the high-order MDP and specialized Bellman equations are unnecessary, but reward imputation is still required to recover $(D_t-Y_t)^+$. If censoring is absent, then demand is fully observed and rewards are observed; both difficulties vanish and the problem reduces to a standard MDP setting where classical FQI applies. This highlights that our high-order/Bellman construction is needed precisely for the interaction of \emph{censoring and temporal dependence}, whereas imputation only targets the separate issue of \emph{missing reward} under stockouts.

\subsection{Further Remarks on Theoretical Results}\label{app: remarks}

\paragraph{The difference between Classical Coverage and Censoring Coverage,} Recall that each $Q^{(i)}$ is defined over distinct history blocks consisting of the most recent $(i{+}1)$ observations. In particular, $Q^{(i)}$ depends on a history containing exactly $i-1$ consecutive censored periods. Thus, learning the optimal policy under censoring requires reliably estimating all such $Q^{(i)}$ functions from data. If the offline dataset never exhibits more than $n$ consecutive censored periods, then the dataset never contains the history blocks required to define $Q^{(n)}$. In this case, the corresponding Bellman equations are structurally unidentifiable: the transitions needed to form valid Bellman targets simply do not exist in the data. This exact phenomenon is what our notion of \emph{censoring coverage} is designed to capture. It does not measure how often particular state--action pairs are visited; instead, it measures whether the dataset contains sufficiently long censoring runs to support all layers of the high-order MDP induced by the optimal policy. This is formalized as \textbf{Sufficient Censoring Coverage} in Assumption~\ref{assump:sufficient-coverage-main-text}. In contrast to classical coverage notions—where the state space is fixed and one only requires sufficient visitation density—our High-Order MDP formulation reveals that the effective state space itself expands with censoring depth. Each additional censored period creates a new structural state indexed by $i$. As a result, we must learn a specific Q-function $Q^{(i)}$ for every censoring depth $i$.

To illustrate suppose the optimal policy may induce up to $n=5$ consecutive stockouts during high-demand surges. This means that $Q^{(5)}$ is required to optimally act after five consecutive lost-sales periods. If the historical data only ever exhibits runs of length at most $n_{K,b}=3$, then:
\[
Q^{(4)} \ \text{and} \ Q^{(5)} \ \text{are structurally undefined in the dataset},
\]
because the corresponding history blocks never appear. This failure is not a lack of statistical density—it is a lack of structural support. Classical coverage metrics cannot detect this failure mode.

Consequently, classical coverage in offline RL is defined for a \emph{fixed} state--action space: once the MDP is specified, the question is whether the dataset visits the state--action regions used by the target/optimal policy often enough to estimate values accurately. Our setting is different because censoring changes the \emph{effective order} of the control problem: each additional consecutive censored period creates a new “depth” regime with its own state representation and associated $Q^{(i)}$. Therefore, before classical coverage is even meaningful, the data must contain censoring runs as deep as those induced by the optimal policy; otherwise the corresponding regimes (and the relevant Bellman objects) are structurally absent and the optimal policy is not identifiable from the dataset. Once these censoring depths are present (censoring coverage), the remaining question becomes the classical one—within each supported depth, does the data adequately cover the states and actions that the optimal policy would take

\paragraph{Reduction of Theorem~\ref{cor:ideal} under Independent Demand.}
Consider the case where demand is independent across periods (e.g., setting $\rho=0$ in the AR(1) model from Eq.~\eqref{eq:linear_ar_demand}). In this regime, the unobserved demand history provides no information about the future. This implies that the observed process satisfies the Markov property even under censoring, and thus, the estimation error for the censoring depth ($\omega_{NT}$) vanishes from the regret bound, as the "memory" of the process is effectively zero ($n_{K,b}=0$). Consequently, the regret bound in Theorem~\ref{cor:ideal} collapses to:
\begin{sizeddisplay}
    \small
\begin{align*}
\max(\textup{\textbf{Regret}}(\widehat \pi_K),\ \textup{\textbf{Regret}}(\widetilde \pi_K))
&\lesssim \mathcal{E}_{\text{stat}}(NT, d)\\
&:=\max\left(
\big(d+\log(1/\epsilon)\big)d
\sqrt{\frac{C_{\mathrm{cov}}}{\rho NT}},\ 
d
\sqrt{\frac{\big(d+\log(1/\epsilon)\big)d}{\rho\,NT}}
\right)+
C_5(\varepsilon)(NT)^{-\delta}
\nonumber
\end{align*}\end{sizeddisplay}
This confirms that our result thus can be interpreted as the finite-sample regret guarantee for the problem of offline multi-stage inventory and pricing control with censored, independent demand. Furthermore, if there were no censoring, then ''Imputation Error'' (i.e., $C_5(\varepsilon)(NT)^{-\delta}$) would vanish as well and the bound would reduce to classical offline reinforcement learning case.

\paragraph{Comparison of Theorem~\ref{cor:ideal} with Inventory Literature.}
The regret bound in Theorem~\ref{cor:ideal} consists of a statistical term and two censoring-specific terms: the depth estimation error $\omega_{NT}$ and the reward-imputation term $C_5(\epsilon)(NT)^{-\delta}$. 
When demand is fully observed, as in the offline setting of \cite{xie2024vc} and in the backlogging model of the online setting in \cite{gong2024bandits}, both censoring-related terms vanish. 
In this regime, our bound reduces to $\tilde{O}(1/\sqrt{NT})$, matching the standard $\tilde{O}(1/\sqrt{N_{\mathrm{sample}}})$ statistical rate.
We note that \cite{gong2024bandits} study an \emph{online} control problem, whereas our setting is \emph{offline}, so the notions of regret differ; nevertheless, their fully observed model yields a per-period convergence rate of $\tilde{O}(1/\sqrt{T})$, consistent with our $\tilde{O}(1/\sqrt{NT})$ scaling.

Furthermore, \cite{agrawal2019learning} study an \emph{online} periodic lost-sales inventory problem with \emph{censored} but \textbf{i.i.d.} demand across periods. In contrast, we study \emph{offline} joint pricing--inventory control with censored and \emph{dependent} demand.  Under the \emph{identical} conditions to their regime—inventory-only (fixed price), i.i.d.\ demand (no dependence), the unobserved demand history provides no information about the future. This implies that the observed process satisfies the Markov property even under censoring, and thus, the estimation error for the censoring depth ($\omega_{NT}$) vanishes from the regret bound, as the "memory" of the process is effectively zero ($n_{K,b}=0$). Consequently, the regret bound in Theorem~\ref{cor:ideal} collapses to:
\begin{sizeddisplay}
    \small
\begin{align*}
\max(\textup{\textbf{Regret}}(\widehat \pi_K),\ \textup{\textbf{Regret}}(\widetilde \pi_K))
&\lesssim \mathcal{E}_{\text{stat}}(NT, d)\\
&:=\max\left(
\big(d+\log(1/\epsilon)\big)d
\sqrt{\frac{C_{\mathrm{cov}}}{\rho NT}},\ 
d
\sqrt{\frac{\big(d+\log(1/\epsilon)\big)d}{\rho\,NT}}
\right)+
C_5(\varepsilon)(NT)^{-\delta}
\nonumber
\end{align*}\end{sizeddisplay}
This is consistent with the $1/\sqrt{T}$ average-cost decay implied by their $\tilde{O}(L\sqrt{T}+D)$ cumulative-regret bound in \cite{agrawal2019learning}.

\paragraph{Comparison of Theorem~\ref{thm:general_regret} with Standard Offline RL.} 
It is crucial to distinguish our result from standard sample complexity bounds in pessimistic offline RL (e.g., \cite{jin2021pessimism, xie2021bellman, shi2022pessimistic}). Standard bounds typically scale as $\tilde{O}(1/\sqrt{NT})$ and vanish asymptotically provided the dataset covers the optimal policy's state-action distribution. In contrast, our bound in Theorem~\ref{thm:general_regret} includes structural terms—specifically the \textbf{Unidentifiable Regret} ($\mathcal{E}_{\text{cov}}$) and \textbf{Alignment Penalties} ($\mathrm{TAP}, \mathrm{BAP}$)—that do not necessarily vanish with sample size. These terms quantify the unique "cost of dependent and censored demand": the irreducible risk incurred when the optimal policy generates censoring sequences longer than those observed in history, or when operational safety constraints force deviations from greedy value maximization. These are unique components arising from the specific problem structure. However, under specific structural conditions, such as the ideal setting in Theorem \ref{cor:ideal} and the relaxed linear MDP assumption in Corollary \ref{cor:relaxed-structure}, our bounds recover the rates found in standard analyses.

\section{Regret Decomposition}\label{sec:Alt. regret decomp}
In this section, we present the derivation of the regret decomposition for a policy $\pi$, defined as follows:
\begin{align}
	\text{Regret}(\pi) =\calV(\pi^\ast) - \calV(\pi)\text{,}
\end{align}
where $\pi^*$ is the optimal policy defined in Equation \eqref{def: optimal policy under offline}. Firstly, we make several observations and state the necessary assumptions for deriving the regret decomposition.

 

Recall that the final estimated policies produced by C-FQI and Pessimistic C-FQI belong to $\Pi_{\hat{n}_{K,b}}$.
This shows that both policies, $\widehat{\pi}_K$ and $\widetilde{\pi}_K$, are implicitly a function of the offline data through $\hat{n}_{K,b}$. Therefore, when considering value of the estimated final policies, data dependency needs to be taken into account. In addition, we have the following observation,

$$\calV(\pi)=\EE\left[\EE^{\pi}\left[\sum_{t=0}^{\infty}\gamma^{t}R_{t}\given \mathcal{O}_N\right]\right].$$

Next, we make the following assumption.
\begin{assumption}\label{ass: pi* C - sure event}$\prob^{\pi^{*}_{\mathcal{C}}}(\mathcal{C}(K,n_{K,b}))=1$, where $\pi^{*}_{\mathcal{C}} \in \argmax_{\pi' \in \Pi}\mathbb{E}^{\pi'}\left[\sum_{t=0}^{\infty}\gamma^t R_t\given \mathcal{C}(K,n_{K,b})\right]$
\end{assumption}

Now, we are ready to state the Lemma outlining the regret decomposition:
\begin{lemma}\label{lm: regret decom}
    Under Assumptions \ref{ass: DGM}, \ref{ass:consistency} and \ref{ass: pi* C - sure event}  the regret of any policy, $\pi \in\Pi_{n_{K,b}}$ is given by:
  \begin{align}
    \textup{Regret}(\pi) &\leq\left(\sup_{\pi' \in \Pi_{n_{K,b}}} \EE^{\pi'}\left[\sum_{t=0}^{\infty}\gamma^tR_t\right] -\EE^{\pi} \left[ \sum_{t=0}^{\infty} \gamma^t R_t  \right] +2R_{\max}\frac{\gamma^{K+1}}{1-\gamma}\right)\prob^{\pi^*}(\mathcal{C}(K, n_{K,b})) \nonumber \\
    &\quad +(K-n_{K,b}+1)(1-\alpha_{\pi^*,1})^{n_{K,b}}\left(\frac{2R_{\max}}{1-\gamma}\right)\nonumber\\
    &\quad+\frac{4R_{\max}\sqrt{K-1}}{1-\gamma}\omega_{NT}\nonumber
\end{align}
where $\mathcal{C}(K,n_{K,b})$ and $\Pi_{n_{K,b}}$ are characterized in Definitions \ref{def: definition for set C} and \ref{def: set of policies such that at most n cons censoring observed} respectively. 
\end{lemma}

\textit{Proof:}

First, we propose to decompose the regret using the law of iterated expectations as follows:
\begin{align}
    \text{Regret}(\pi) &= \calV(\pi^*) - \calV(\pi) \nonumber\\
    &= \EE^{\pi^*} \left[ \sum_{t=0}^{\infty} \gamma^t R_t \right] - \EE_{\mathcal{O}_N} \left[\EE^{\pi} \left[\sum_{t=0}^{\infty} \gamma^t R_t \given \mathcal{O}_N \right]\right] \nonumber \\
    &= \EE_{\mathcal{O}_N} \left[\EE^{\pi^*} \left[ \sum_{t=0}^{\infty} \gamma^t R_t \given \mathcal{O}_N \right] - \EE^{\pi} \left[\sum_{t=0}^{\infty} \gamma^t R_t \given \mathcal{O}_N \right]\right] \nonumber\\
    &= \EE_{\mathcal{O}_N} \left\{\EE^{\pi^*} \left[ \sum_{t=0}^{\infty} \gamma^t R_t \mid \mathcal{C}(K, \hat{n}_{K,b}), \mathcal{O}_N \right] \prob^{\pi^*}(\mathcal{C}(K, \hat{n}_{K,b}) \mid \mathcal{O}_N) \nonumber \right.\\
    &\quad \left.+ \EE^{\pi^*} \left[ \sum_{t=0}^{\infty} \gamma^t R_t \mid \mathcal{C}^{c}(K, \hat{n}_{K,b}), \mathcal{O}_N \right] \prob^{\pi^*}(\mathcal{C}^{c}(K, \hat{n}_{K,b}) \mid \mathcal{O}_N) \nonumber \right.\\
    &\quad \left.- \EE^{\pi} \left[ \sum_{t=0}^{\infty} \gamma^t R_t \mid \mathcal{C}(K, \hat{n}_{K,b}), \mathcal{O}_N \right] \prob^{\pi}(\mathcal{C}(K, \hat{n}_{K,b}) \mid \mathcal{O}_N) \nonumber \right.\\
    &\quad \left.- \EE^{\pi} \left[ \sum_{t=0}^{\infty} \gamma^t R_t \mid \mathcal{C}^{c}(K, \hat{n}_{K,b}), \mathcal{O}_N \right] \prob^{\pi}(\mathcal{C}^{c}(K, \hat{n}_{K,b}) \mid \mathcal{O}_N)\right\} \nonumber\\
    &= \sum_{m=0}^{K} \prob^{\pi^b}(\hat{n}_{K,b}=m) \left\{\EE^{\pi^*} \left[ \sum_{t=0}^{\infty} \gamma^t R_t \mid \mathcal{C}(K, m)\right] \prob^{\pi^*}(\mathcal{C}(K, m)) \nonumber \right.\\
    &\quad \left.+ \EE^{\pi^*} \left[ \sum_{t=0}^{\infty} \gamma^t R_t \mid \mathcal{C}^{c}(K, m) \right] \prob^{\pi^*}(\mathcal{C}^{c}(K, m)) \nonumber \right.\\
    &\quad \left.- \EE^{\pi} \left[ \sum_{t=0}^{\infty} \gamma^t R_t \mid \mathcal{C}(K, m)\right] \prob^{\pi}(\mathcal{C}(K, m) )\nonumber \right.\\
    &\quad \left.- \EE^{\pi} \left[ \sum_{t=0}^{\infty} \gamma^t R_t \mid \mathcal{C}^{c}(K, m)\right] \prob^{\pi}(\mathcal{C}^{c}(K, m)) \right\}, \nonumber
\end{align}
where we leverage the fact that only $\hat{n}_{K,b}$ is dependent on the offline data, whereas both $\pi^\ast$ and $\pi$ are independent of it, with the independence of $\pi$ contingent upon a fixed $\hat{n}_{K,b}$.

Following this, we decompose the summation into two parts as follows:
\begin{align}
    \text{Regret}(\pi) &=\prob^{\pi^b}(\hat{n}_{K,b}=n_{K,b})\left\{\EE^{\pi^*} \left[ \sum_{t=0}^{\infty} \gamma^t R_t \mid \mathcal{C}(K, n_{K,b})\right] \prob^{\pi^*}(\mathcal{C}(K, n_{K,b})) \nonumber \right.\\
    &\left.\quad + \EE^{\pi^*} \left[ \sum_{t=0}^{\infty} \gamma^t R_t \mid \mathcal{C}^{c}(K, n_{K,b}) \right] \prob^{\pi^*}(\mathcal{C}^{c}(K, n_{K,b})) \nonumber \right.\\
    &\left.\quad - \EE^{\pi} \left[ \sum_{t=0}^{\infty} \gamma^t R_t \mid \mathcal{C}(K, n_{K,b}) \right] \prob^{\pi}(\mathcal{C}(K, n_{K,b}) ) \nonumber \right.\\
    &\left.\quad - \EE^{\pi} \left[ \sum_{t=0}^{\infty} \gamma^t R_t \mid \mathcal{C}^{c}(K, n_{K,b}) \right] \prob^{\pi}(\mathcal{C}^{c}(K, n_{K,b}))\right\}\nonumber\\
    &+\sum_{m\neq n_{K,b}}\prob^{\pi^b}(\hat{n}_{K,b}=m)\left\{\EE^{\pi^*} \left[ \sum_{t=0}^{\infty} \gamma^t R_t \mid \mathcal{C}(K, m) \right] \prob^{\pi^*}(\mathcal{C}(K, m)) \nonumber \right.\\
    &\left.\quad + \EE^{\pi^*} \left[ \sum_{t=0}^{\infty} \gamma^t R_t \mid \mathcal{C}^{c}(K, m)\right] \prob^{\pi^*}(\mathcal{C}^{c}(K, m)) \nonumber \right.\\
    &\left.\quad - \EE^{\pi} \left[ \sum_{t=0}^{\infty} \gamma^t R_t \mid \mathcal{C}(K, m)\right] \prob^{\pi}(\mathcal{C}(K, m)) \nonumber \right.\\
    &\left.\quad - \EE^{\pi} \left[ \sum_{t=0}^{\infty} \gamma^t R_t \mid \mathcal{C}^{c}(K, m)\right] \prob^{\pi}(\mathcal{C}^{c}(K, m))\right\}.
\end{align}

Consider the first part where $\hat{n}_{K,b}=n_{K,b}$. $\mathcal{C}(K, n_{K,b})$ is a sure event under $\pi$ as $\pi \in \Pi_{n_{K,b}}$. Next, consider the second part based on which we define $a_m$ and $b_m$ as 
$$a_m=\prob^{\pi^b}(\hat{n}_{K,b}=m)$$
\begin{align*}
    b_m&=\EE^{\pi^*} \left[ \sum_{t=0}^{\infty} \gamma^t R_t \mid \mathcal{C}(K, m) \right] \prob^{\pi^*}(\mathcal{C}(K, m))  + \EE^{\pi^*} \left[ \sum_{t=0}^{\infty} \gamma^t R_t \mid \mathcal{C}^{c}(K, m)\right] \prob^{\pi^*}(\mathcal{C}^{c}(K, m)) \nonumber \\
    &\quad - \EE^{\pi} \left[ \sum_{t=0}^{\infty} \gamma^t R_t \mid \mathcal{C}(K, m)\right] \prob^{\pi}(\mathcal{C}(K, m))  - \EE^{\pi} \left[ \sum_{t=0}^{\infty} \gamma^t R_t \mid \mathcal{C}^{c}(K, m)\right] \prob^{\pi}(\mathcal{C}^{c}(K, m))
\end{align*}

Following this, we note that $ \sum_{m \neq n_{K,b}} a_m \leq \omega_{NT} $, and therefore, $ \sum_{m \neq n_{K,b}} a_m^2 \leq \omega_{NT}^2 $, as implied by Assumption \ref{ass:consistency}. Additionally, since $ b_m \leq \frac{4R_{\max}}{1 - \gamma} $, we have $ \sum_{m \neq n_{K,b}} b_m^2 \leq (K-1)\left(\frac{4R_{\max}}{1 - \gamma}\right)^2 $, given that $ |R_t| \leq R_{\max} $ for all $ t \geq 0 $. Then, by applying the Cauchy-Schwarz inequality, it follows that

$$\sum_{m\neq n_{K,b}}a_m b_m \leq \frac{4R_{\max}\sqrt{K-1}}{1-\gamma}\omega_{NT}$$

Therefore, the upper bound on the regret is expressed as:
\begin{align}\label{eq:general regret decomposition}
    \text{Regret}(\pi) &\leq\prob^{\pi^b}(\hat{n}_{K,b}=n_{K,b})\left\{\left(\EE^{\pi^*} \left[ \sum_{t=0}^{\infty} \gamma^t R_t \mid \mathcal{C}(K, n_{K,b})\right]-\EE^{\pi} \left[ \sum_{t=0}^{\infty} \gamma^t R_t  \right] \right)\prob^{\pi^*}(\mathcal{C}(K, n_{K,b})) \nonumber \right.\\
    &\left.\quad + \left(\EE^{\pi^*} \left[ \sum_{t=0}^{\infty} \gamma^t R_t \mid \mathcal{C}^{c}(K, n_{K,b}) \right] -\EE^{\pi} \left[ \sum_{t=0}^{\infty} \gamma^t R_t \right]\right)\prob^{\pi^*}(\mathcal{C}^{c}(K, n_{K,b})) \right\}\nonumber\\
    &\quad+\frac{4R_{\max}\sqrt{K-1}}{1-\gamma}\omega_{NT},
\end{align}

Next we focus on bounding the second term in Equation \eqref{eq:general regret decomposition}:
$$\left(\EE^{\pi^*} \left[ \sum_{t=0}^{\infty} \gamma^t R_t \mid \mathcal{C}^{c}(K, n_{K,b}) \right] -\EE^{\pi} \left[ \sum_{t=0}^{\infty} \gamma^t R_t \right]\right)\prob^{\pi^*}(\mathcal{C}^{c}(K, n_{K,b}))$$ This term essentially corresponds to the unlearnable portion of the optimal policy. Recall that the probability of observing censoring in the next time step under policy $\pi$ given the last $n$ periods are censored is denoted by:
\begin{align*}
    \prob^{\pi}(\Delta_{t+1}=0\given\Delta_{t}=\cdots=\Delta_{t-n+1}=0)=1-\alpha_{\pi,n} \text{   } \forall t \geq 0 \text{.}
\end{align*}
Let $\mathcal{B}_i$ represent the set containing all trajectories with sub-sequences that exhibit more than $n_{K,b}$ consecutive instances of censoring within the first $K$ horizon starting from the $i$-th decision point. Formally,
$$\mathcal{B}_i = \left\{ \{W_t, A_t \}_{t \geq 0} \mid \exists k > n_{K,b} \ \text{such that} \ \forall j \in \{i, \ldots, i+k-1\}, \Delta_j = 0, \text{ and } \ i+k-1 < K \right\}
$$

Under these definitions, $\prob^{\pi}(\mathcal{B}_i)$ can be bounded from above as follows:
\begin{align*}
    \prob^{\pi}(\mathcal{B}_i) &=\prob^{\pi}(\Delta_{i-1}=1,\Delta_{i}=0,\cdots,\Delta_{i+n_{K,b}}=0)\\
    &\leq \prob^{\pi}(\Delta_{i}=0,\cdots,\Delta_{i+n_{K,b}}=0)\\
    &=\prob^{\pi}(\Delta_{i}=0)\prob^{\pi}(\Delta_{i+1}=0\given \Delta_{i}=0)\cdots\prob^{\pi}(\Delta_{i+n_{K,b}}=0\given \Delta_{i+n_{K,b}-1}=\cdots=\Delta_{i}=0)\\
    &\leq \prod_{j=1}^{n_{K,b}}(1-\alpha_{\pi,j})\text{.}
\end{align*}

Under Assumption \ref{ass: monotonic increase in alpha}, the probability of observing a non-censored state under the optimal policy increases as the number of consecutive censoring in the  previous periods increases. This implies $1-\alpha_{\pi^*,j}$ decreases as $j$ increases. Therefore,
\begin{align}\label{eq:bound B_i}
    \prob^{\pi^*}(\mathcal{B}_i)\leq (1-\alpha_{\pi^*,1})^{n_{K,b}}.
\end{align}

Furthermore, it follows from Definition \ref{def: definition for set C} and the definition of $\mathcal{B}_i$ that $\mathcal{C}^{c}(K,n_{K,b}) = \cup_{i=0}^{K-n_{K,b}}\mathcal{B}_i$. Therefore, for any policy $\pi$, we have:
\begin{align}\label{eq: bound on C^c for any policy}
    \prob^{\pi}(\mathcal{C}^{c}(K,n_{K,b}))\leq \sum_{i=0}^{K-n_{K,b}} \prob^{\pi}(\mathcal{B}_i)\leq (K-n_{K,b}+1)\prod_{j=1}^{n_{K,b}}(1-\alpha_{\pi,j}).
\end{align}

Given Equations \eqref{eq:bound B_i} and \eqref{eq: bound on C^c for any policy}, it can be concluded that:
\begin{align}\label{eq: bound on prob(g^c) under pi^*}
    \prob^{\pi^*}(\mathcal{C}^{c}(K,n_{K,b}))\leq (K-n_{K,b}+1)(1-\alpha_{\pi^*,1})^{n_{K,b}}\text{.}
\end{align}

Next, we use Equation \eqref{eq: bound on prob(g^c) under pi^*} and the fact that $|R_{t}|\leq R_{\max}$ $\forall t\geq 0$ to obtain the following:
$$\left(\EE^{\pi^*} \left[ \sum_{t=0}^{\infty} \gamma^t R_t \mid \mathcal{C}^{c}(K, n_{K,b}) \right] -\EE^{\pi} \left[ \sum_{t=0}^{\infty} \gamma^t R_t \right]\right)\prob^{\pi^*}(\mathcal{C}^{c}(K, n_{K,b})) \leq (K-n_{K,b}+1)(1-\alpha_{\pi^*,1})^{n_{K,b}}\left(\frac{2R_{\max}}{1-\gamma}\right)$$

Plugging this to Equation \eqref{eq:general regret decomposition} yields:
\begin{align}
    \text{Regret}(\pi) &\leq\prob^{\pi^b}(\hat{n}_{K,b}=n_{K,b})\left\{\left(\EE^{\pi^*} \left[ \sum_{t=0}^{\infty} \gamma^t R_t \mid \mathcal{C}(K, n_{K,b})\right]-\EE^{\pi} \left[ \sum_{t=0}^{\infty} \gamma^t R_t  \right] \right)\prob^{\pi^*}(\mathcal{C}(K, n_{K,b})) \right\}\nonumber \\
    &\quad +(K-n_{K,b}+1)(1-\alpha_{\pi^*,1})^{n_{K,b}}\left(\frac{2R_{\max}}{1-\gamma}\right)\nonumber\\
    &\quad+\frac{4R_{\max}\sqrt{K-1}}{1-\gamma}\omega_{NT}
\end{align}
Observe that by definition, $\pi^*$ is the optimal policy that maximizes $\mathbb{E}^{\pi}\left[\sum_{t=0}^{\infty}\gamma^t R_t\right]$. However, it does not necessarily remain optimal for $\mathbb{E}^{\pi}\left[\sum_{t=0}^{\infty}\gamma^t R_t\given \mathcal{C}(K,n_{K,b})\right]$ as the transition probabilities are altered once conditioned on $\mathcal{C}(K,n_{K,b})$. We denote by $\pi^*_{\mathcal{C}}$ the optimal policy under the event $\mathcal{C}(K,n_{K,b})$ (i.e $\pi^{*}_{\mathcal{C}} \in \argmax_{\pi^{'} \in \Pi}\mathbb{E}^{\pi^{'}}\left[\sum_{t=0}^{\infty}\gamma^t R_t\given \mathcal{C}(K,n_{K,b})\right]$). Given the optimality of $\pi^*_{\mathcal{C}}$, it follows that $\EE^{\pi^{'}}\left[\sum_{t=0}^{\infty}\gamma^tR_t\given \mathcal{C}(K,n_{K,b})\right]\leq \EE^{\pi_{\mathcal{C}}^*}\left[\sum_{t=0}^{\infty}\gamma^tR_t\given \mathcal{C}(K,n_{K,b})\right]$ $\forall \pi^{'} \in \Pi$.

Consequently, the upper bound on the regret can be further expressed as:
\begin{align}
    \text{Regret}(\pi) &\leq\left(\EE^{\pi_{\mathcal{C}}^*} \left[ \sum_{t=0}^{\infty} \gamma^t R_t \mid \mathcal{C}(K, n_{K,b})\right]-\EE^{\pi} \left[ \sum_{t=0}^{\infty} \gamma^t R_t  \right] \right)\prob^{\pi^*}(\mathcal{C}(K, n_{K,b})) \nonumber \\
    &\quad +(K-n_{K,b}+1)(1-\alpha_{\pi^*,1})^{n_{K,b}}\left(\frac{2R_{\max}}{1-\gamma}\right)\nonumber\\
    &\quad+\frac{4R_{\max}\sqrt{K-1}}{1-\gamma}\omega_{NT}
\end{align}

Consider the following term $\forall \pi \in \Pi_{n_{K,b}}$:
\begin{align}
    \EE^{\pi_{\mathcal{C}}^*}\left[\sum_{t=0}^{\infty}\gamma^tR_t\given \mathcal{C}(K,n_{K,b})\right] 
    &=\EE^{\pi_{\mathcal{C}}^*}\left[\sum_{t=0}^{K}\gamma^tR_t\given \mathcal{C}(K,n_{K,b})\right] + \EE^{\pi_{\mathcal{C}}^*}\left[\sum_{t=K+1}^{\infty}\gamma^tR_t\given \mathcal{C}(K,n_{K,b})\right]\nonumber \\
    &\leq \EE^{\pi_{\mathcal{C}}^*}\left[\sum_{t=0}^{K}\gamma^tR_t\given \mathcal{C}(K,n_{K,b})\right]+\EE^{\pi}\left[\sum_{t=K+1}^{\infty}\gamma^tR_t\given \mathcal{C}(K,n_{K,b})\right] \nonumber\\
    &+2R_{\max}\frac{\gamma^{K+1}}{1-\gamma}\nonumber\\
    &=\EE^{\pi_{\mathcal{C}}^*}\left[\sum_{t=0}^{K}\gamma^tR_t \right]+\EE^{\pi}\left[\sum_{t=K+1}^{\infty}\gamma^tR_t\right] \nonumber\\
    &+2R_{\max}\frac{\gamma^{K+1}}{1-\gamma},
\end{align}
where the last equality follows from the fact that $\mathcal{C}(K, n_{K,b})$ is a sure event under $\pi^{\ast}_{\mathcal{C}}$, as implied by Assumption \ref{ass: pi* C - sure event}, and from the condition that $\pi \in \Pi_{n_{K,b}}$.

Next, note that $\pi_\mathcal{C}^{\ast}$ results in at most $n_{K,b}$ consecutive censorings within the first $K$ steps, while $\pi$ results in at most $n_{K,b}$ consecutive censoring over the entire horizon. Based on this, we define a policy, denoted by $\pi_{n_{K,b}}$, as $\pi^{\ast}{\mathcal{C}}\mathbb{I}[t \leq K] + \pi\mathbb{I}[t > K]$. This policy ensures at most $n_{K,b}$ consecutive censoring across the entire horizon. Therefore, 
\begin{align*}
    \EE^{\pi_{\mathcal{C}}^*}\left[\sum_{t=0}^{K}\gamma^tR_t \right]+\EE^{\pi}\left[\sum_{t=K+1}^{\infty}\gamma^tR_t\right]&=\EE^{\pi_{n_{K,b}}}\left[\sum_{t=0}^{K}\gamma^tR_t \right]\\
    &\leq \sup_{\pi' \in \Pi_{n_{K,b}}} \EE^{\pi'}\left[\sum_{t=0}^{\infty}\gamma^tR_t\right]. 
\end{align*}
This implies
\begin{align}
    \EE^{\pi_{\mathcal{C}}^*}\left[\sum_{t=0}^{\infty}\gamma^tR_t\given \mathcal{C}(K,n_{K,b})\right] \leq\sup_{\pi' \in \Pi_{n_{K,b}}} \EE^{\pi'}\left[\sum_{t=0}^{\infty}\gamma^tR_t\right] + 2R_{\max}\frac{\gamma^{K+1}}{1-\gamma}.
\end{align}

Therefore, the upper bound on the regret becomes the following:
\begin{align}
    \text{Regret}(\pi) &\leq\left(\sup_{\pi' \in \Pi_{n_{K,b}}} \EE^{\pi'}\left[\sum_{t=0}^{\infty}\gamma^tR_t\right] -\EE^{\pi} \left[ \sum_{t=0}^{\infty} \gamma^t R_t  \right] +2R_{\max}\frac{\gamma^{K+1}}{1-\gamma}\right)\prob^{\pi^*}(\mathcal{C}(K, n_{K,b})) \nonumber \\
    &\quad +(K-n_{K,b}+1)(1-\alpha_{\pi^*,1})^{n_{K,b}}\left(\frac{2R_{\max}}{1-\gamma}\right)\nonumber\\
    &\quad+\frac{4R_{\max}\sqrt{K-1}}{1-\gamma}\omega_{NT}.
\end{align}
This concludes the proof of Lemma \ref{lm: regret decom}.

\textbf{Remark}: Observe that if Assumptions~\ref{assump:bounded-n} and \ref{assump:sufficient-coverage-main-text} holds (i.e., $\pi^* \in \Pi_{n_{K,b}}$), then $\mathbb{P}^{\pi^*}(\mathcal{C}(K,n_{K,b})) = 1$. Accordingly, the optimal policy learnable overall, and thus, Equation \eqref{eq:general regret decomposition} reduces to the following:

\begin{align}\label{eq:general regret decomposition with coverage}
    \text{Regret}(\pi) &\leq\left\{\left(\EE^{\pi^*} \left[ \sum_{t=0}^{\infty} \gamma^t R_t \right]-\EE^{\pi} \left[ \sum_{t=0}^{\infty} \gamma^t R_t  \right] \right) \right\}\nonumber\\
    &\quad+\frac{4R_{\max}\sqrt{K-1}}{1-\gamma}\omega_{NT},
\end{align}
We will use this result in the next section.

\section{Proof of Theorem~\ref{cor:ideal}}\label{app:proof-ideal}

This section provide a proof of Theorem~\ref{cor:ideal} via three steps. In Step~1, we analyze PC-FQI under idealized setting where Assumptions~\ref{ass: DGM}-\ref{ass: UQ ass} hold. We first invoke the special form of Lemma~\ref{lm: regret decom} which is valid specifically under Assumptions~\ref{assump:bounded-n} and \ref{assump:sufficient-coverage-main-text}, and we bound resulting term using Lemma~\ref{lm: performance difference} (i.e., performance-difference lemma) together with the upper/lower bounds on estimated $Q$-functions. Step~2 repeats the same derivation for C-FQI, replacing $\widetilde{\pi}_K$ by $\widehat{\pi}_K$ and using the analogous advantage decomposition and $Q$-function bounds. Finally, Step~3 combines the two bounds via a union bound to obtain the final result.

\subsection{Step 1: Regret for PC-FQI under Idealized Setting}\label{PC-FQI under Idealized Setting}

Recall that in the idealized setting, Assumptions~\ref{assump:bounded-n} and \ref{assump:sufficient-coverage-main-text} hold. Moreover, by definition, ($\widetilde{\pi}_K \in \Pi_{n_{K,b}}$). Therefore, we may invoke the reduced form of Lemma~\ref{lm: regret decom}, which is valid under these two assumptions. Specifically, this corresponds to the decomposition given in Equation \eqref{eq:general regret decomposition with coverage} in the remark of Lemma~\ref{lm: regret decom}.
\begin{sizeddisplay}
{\footnotesize}
 \begin{align}\label{pesseq: fqi regret 1_app}
    \textup{\textbf{Regret}}(\widetilde{\pi}_K) &\leq\underbrace{\EE^{\pi^*} \left[ \sum_{t=0}^{\infty} \gamma^t R_t \right]-\mathbb{E}^{\tilde{\pi}_K}\left[ \sum_{t=0}^{\infty} \gamma^t R_t \right]
}_{\textstyle i(\widetilde{\pi}_K)} \nonumber\\
    &\quad+\frac{4R_{\max}\sqrt{K-1}}{1-\gamma}\omega_{NT},
\end{align}
\end{sizeddisplay}
Thus, it remains to upper bound the term $i(\widetilde{\pi}_K)$.

Under Assumptions~\ref{assump:bounded-n} and \ref{assump:sufficient-coverage-main-text}, the optimal policy satisfies $\pi^* \in \Pi_n \subseteq \Pi_{n_{K,b}}$. Since $\pi^*$ maximizes the value globally and is a valid candidate within the restricted class $\Pi_{n_{K,b}}$, for simplicity, let $\pi^{\ast}_{n_{K,b}} = \pi^*$, where  $\pi^{\ast}_{n_{K,b}}\in \argmax_{\pi' \in \Pi_{n_{K,b}}}\EE^{\pi'}\!\left[\sum_{t=0}^{\infty}\gamma^tR_t\right]$.  Given this and the fact that $\pi^{\ast}_{n_{K,b}}$ and $\widetilde{\pi}_K$ both belong to $\Pi_{n_{K,b}}$, Lemma~\ref{lm: performance difference} yields
\begin{sizeddisplay}
    {\footnotesize}{\begin{align}\label{eq: pes regret 1}
         i(\widetilde{\pi}_K)
         &=\EE^{\pi^{\ast}}\left[\sum_{t=0}^{\infty}\gamma^tR_t\right]  -\EE^{\tilde{\pi}_K}\left[\sum_{t=0}^{\infty}\gamma^tR_t\right]\nonumber\\
         &=\EE^{\pi^{\ast}_{n_{K,b}}}\left[\sum_{t=0}^{\infty}\gamma^tR_t\right]  -\EE^{\tilde{\pi}_K}\left[\sum_{t=0}^{\infty}\gamma^tR_t\right]\\
         &=\frac{1-\gamma^{n_{K,b}}}{(1-\gamma)^2}\EE_{(W^{(0)},A^{(0)},\cdots,W^{(I)}, A^{(I)}) \sim d_{\nu,I}^{ \tilde{\pi}_K} \times \tilde{\pi}_K
    }\left[-A^{(I)}_{\pi^{\ast}_{n_{K,b}}}(W^{(0)},A^{(0)},\cdots,W^{(I)},A^{(I)})\prod_{j=1}^{I}\left\{\mathbb{I}[\Delta^{(j)}=0]\right\}\mathbb{I}[\Delta^{(0)}=1]\right]\nonumber.  
    \end{align}}
\end{sizeddisplay}

Assume that by conditioning on $I=i$ and $(W^{(0)},A^{(0)},\cdots,W^{(i)})=(w^{(0)},a^{(0)},\cdots,w^{(i)})$ such that $\Delta^{(0)}=1 $  and $\Delta^{(j)}=0 $ $\forall j=1,\cdots,i$, the expectation above can be upper bounded as follows: $\forall i=0,\cdots,n_{K,b}-1$,
\begin{sizeddisplay}
{\footnotesize}
\begin{align}\label{eqn:advantage ineq pessimistic}
    &-A^{(i)}_{\pi^{\ast}_{n_{K,b}}}(w^{(0)},a^{(0)},\cdots, w^{(i)},\widetilde{\pi}^{(i)}(w^{(i)},\cdots,w^{(0)},a^{(0)}))\nonumber\\
&=Q^{(i)}_{\pi^{\ast}_{n_{K,b}}}(w^{(0)},a^{(0)},\cdots, w^{(i)},\pi^{(i)}_{n_{K,b},*}(w^{(i)},\cdots,w^{(0)},a^{(0)}) - Q^{(i)}_{\pi^{\ast}_{n_{K,b}}}(w^{(0)},a^{(0)},\cdots, w^{(i)},\widetilde{\pi}^{(i)}_K(w^{(i)},\cdots,w^{(0)},a^{(0)}))\nonumber\\
&\leq Q^{(i)}_{\pi^{\ast}_{n_{K,b}}}(w^{(0)},a^{(0)},\cdots, w^{(i)},\pi^{(i)}_{n_{K,b},*}(w^{(i)},\cdots,w^{(0)},a^{(0)})-\widetilde{Q}^{(i)}_K(w^{(0)},a^{(0)},\cdots, w^{(i)},\pi^{(i)}_{n_{K,b},*}(w^{(i)},\cdots,w^{(0)},a^{(0)}))\nonumber\\
&+\widetilde{Q}^{(i)}_K(w^{(0)},a^{(0)},\cdots, w^{(i)},\widetilde{\pi}^{(i)}_K(s^{(i)},\cdots,w^{(0)},a^{(0)})-Q^{(i)}_{\pi^{\ast}_{n_{K,b}}}(w^{(0)},a^{(0)},\cdots, w^{(i)},\widetilde{\pi}^{(i)}_K(w^{(i)},\cdots,w^{(0)},a^{(0)}),
\end{align}
\end{sizeddisplay}
where the first inequality comes from the optimality of $\widetilde{\pi}_K$ on $\widetilde{Q}^{(i)}_K$, $\forall i=0,\cdots,n_{K,b}-1$

For $i=n_{K,b}$,  the policy $\widetilde{\pi}_K$ is no longer optimal on $\widetilde{Q}^{(n)}_K$ by construction as it outputs $a_{\max}$. This implies that Equation \eqref{eqn:advantage ineq pessimistic} does not hold when $i=n_{K,b}$. Therefore, we quantify the optimality gap as:
\begin{sizeddisplay}
{\footnotesize}
\begin{align}\label{eq:Qhat bound}
    &\textup{AD}^{\tilde{\pi}_K}(w^{(0)},a^{(0)},\cdots,w^{(n_{K,b})}) \\
    &=\max_{a\in\calA}\widetilde{Q}_{K}^{(n_{K,b})}(w^{(0)},a^{(0)},\cdots,w^{(n_{K,b})},a) - \widetilde{Q}_{K}^{(n_{K,b})}(w^{(0)},a^{(0)},\cdots,w^{(n_{K,b})},\widetilde{\pi}_K^{(n_{K,b})}(w^{(n_{K,b})},\cdots,w^{(0)},a^{(0)})).\nonumber
\end{align}
\end{sizeddisplay}
It is worth noting that this constant disappears under Assumption \ref{ass:AD}, and thus, under idealized setting. In particular, Assumption \ref{ass:AD} implies that $a_{\max}$ is the action that maximizes the corresponding $Q$-function, so the constant term evaluates to zero. Accordingly, we will enforce this simplification at the end of the derivation, since Assumption \ref{ass:AD} holds in the idealized setting.

Therefore, by using quantified optimality gap, we can derive Equation \eqref{eqn:advantage ineq pessimistic}  when $i=n_{K,b}$ as follows:
\begin{sizeddisplay}
{\footnotesize}
\begin{align}\label{eqn:pessimistic advantage ineq when i=n}
&-A^{(n_{K,b})}_{\pi^{\ast}_{n_{K,b}}}(w^{(0)},a^{(0)},\cdots, w^{(n_{K,b})},\widetilde{\pi}_{K}^{(n_{K,b})}(w^{(n_{K,b})},\cdots,w^{(0)},a^{(0)}))\nonumber\\
&= Q^{(n_{K,b})}_{\pi^{\ast}_{n_{K,b}}}(w^{(0)},a^{(0)},\cdots, w^{(n_{K,b})},\pi^{(n_{K,b})}_{n_{K,b},\ast}(w^{(n_{K,b})},\cdots,w^{(0)},a^{(0)})-\max_{a\in\calA}\widetilde{Q}^{(n_{K,b})}_K(w^{(0)},a^{(0)},\cdots, w^{(n_{K,b})},a)\nonumber\\
&+\max_{a\in\calA}\widetilde{Q}^{(n_{K,b})}_K(w^{(0)},a^{(0)},\cdots, w^{(n_{K,b})},a)-Q^{(n_{K,b})}_{\pi^{\ast}_{n_{K,b}}}(w^{(0)},a^{(0)},\cdots, w^{(n_{K,b})},\widetilde{\pi}^{(n_{K,b})}_K(w^{(n_{K,b})},\cdots,w^{(0)},a^{(0)}))\nonumber\\
&\leq Q^{(n_{K,b})}_{\pi^{\ast}_{n_{K,b}}}(w^{(0)},a^{(0)},\cdots, w^{(n_{K,b})},\pi^{(n_{K,b})}_{n_{K,b},\ast}(w^{(n_{K,b})},\cdots,w^{(0)},a^{(0)})\nonumber\\
&-\widetilde{Q}^{(n_{K,b})}_K(w^{(0)},a^{(0)},\cdots, w^{(n_{K,b})},\pi^{(n_{K,b})}_{n_{K,b},\ast}(w^{(n_{K,b})},\cdots,w^{(0)},a^{(0)}))\nonumber\\
&+\widetilde{Q}^{(n_{K,b})}_K(w^{(0)},a^{(0)},\cdots, w^{(n_{K,b})},\widetilde{\pi}^{(n_{K,b})}_K(w^{(n_{K,b})},\cdots,w^{(0)},a^{(0)}))\nonumber\\
&-Q^{(n_{K,b})}_{\pi^{\ast}_{n_{K,b}}}(w^{(0)},a^{(0)},\cdots, w^{(n_{K,b})},\widetilde{\pi}^{(n_{K,b})}_K(w^{(n_{K,b})},\cdots,w^{(0)},a^{(0)})) \nonumber\\
&+ \textup{AD}^{\widetilde{\pi}_K}(w^{(0)},a^{(0)},\cdots, w^{(n_{K,b})}),
\end{align}
\end{sizeddisplay}
where first inequality comes from \eqref{eq:Qhat bound} and the fact that 
{\footnotesize \begin{align}
\max_{a\in\calA}\widetilde{Q}^{(n_{K,b})}_K(w^{(0)},a^{(0)},\cdots, w^{(n_{K,b})},a) \geq \widetilde{Q}^{(n_{K,b})}_K(w^{(0)},a^{(0)},\cdots, w^{(n_{K,b})},\widetilde{\pi}^{(n_{K,b})}_*(w^{(n_{K,b})},\cdots,w^{(0)},a^{(0)}))\nonumber
\end{align}}

Equations \eqref{eqn:advantage ineq pessimistic} and \eqref{eqn:pessimistic advantage ineq when i=n} combined yield the following $\forall i =0,\cdots,n_{K,b}$:
\begin{sizeddisplay}
{\footnotesize}
\begin{align}\label{eq: pessimism combined upper bound on Advantage i=0,..,n}
   &-A^{(i)}_{\pi^{\ast}_{n_{K,b}}}(w^{(0)},a^{(0)},\cdots, w^{(i)},\widetilde{\pi}^{(i)}(w^{(i)},\cdots,w^{(0)},a^{(0)}))\nonumber\\
    &\leq Q^{(i)}_{\pi^{\ast}_{n_{K,b}}}(w^{(0)},a^{(0)},\cdots, w^{(i)},\pi^{(i)}_{n_{K,b},\ast}(w^{(i)},\cdots,w^{(0)},a^{(0)})-\widetilde{Q}^{(i)}_K(w^{(0)},a^{(0)},\cdots, w^{(i)},\pi^{(i)}_{n_{K,b},\ast}(w^{(i)},\cdots,w^{(0)},a^{(0)}))\nonumber\\
&+\widetilde{Q}^{(i)}_K(w^{(0)},a^{(0)},\cdots, w^{(i)},\widetilde{\pi}^{(i)}_K(w^{(i)},\cdots,w^{(0)},a^{(0)})-Q^{(i)}_{\pi^{\ast}_{n_{K,b}}}(w^{(0)},a^{(0)},\cdots, w^{(i)},\widetilde{\pi}^{(i)}_K(w^{(i)},\cdots,w^{(0)},a^{(0)}))\nonumber\\
&+\textup{AD}^{\tilde{\pi}_K}(w^{(0)},a^{(0)},\cdots, w^{(i)})\mathbb{I}[i=n_{K,b}].
\end{align}
\end{sizeddisplay}
To upper bound the right hand side of Equation \eqref{eq: pessimism combined upper bound on Advantage i=0,..,n}, we use Lemmas \ref{lm:upper bound on Qs pessimistic} and \ref{lm:lower bound on Qs pessimistic}, which provide an upper and a lower bound on the difference between $Q^{(i)}_{\pi^{\ast}_{n_{K,b}}}$ and $\widetilde{Q}^{(i)}_K$ on the event $\Omega_P \cap\Omega^r$ that holds $\forall i=0,\cdots,n_{K,b}$ simultaneously with probability at least $1 -(n_{K,b}+1)\epsilon-\varepsilon$:
 \begin{sizeddisplay}
{\footnotesize}\begin{align*}
   &Q^{(i)}_{\pi^{\ast}_{n_{K,b}}}(w^{(0)},a^{(0)},\cdots, w^{(i)},a^{(i)})-\widetilde{Q}^{(i)}_K(w^{(0)},a^{(0)},\cdots, w^{(i)},a^{(i)})\\&\leq \gamma^{K-1}\left(\frac{2-\gamma}{1-\gamma}R_{\max}\right)+ \frac{1}{1-\gamma
    }C_5(\varepsilon)(NT)^{-\delta}\nonumber\\
   &+\widetilde{\text{UQ}}(w^{(0)},a^{(0)},\cdots, w^{(i)},a^{(i)}) 
\end{align*}\end{sizeddisplay}
and
\begin{sizeddisplay}
{\footnotesize}\begin{align*}
    &Q^{(i)}_{\pi^{\ast}_{n_{K,b}}}(w^{(0)},a^{(0)},\cdots, w^{(i)},a^{(i)})-\widetilde{Q}^{(i)}_K(w^{(0)},a^{(0)},\cdots, w^{(i)},a^{(i)})\\
    &\geq  -\gamma^{K-1}\left(\frac{2-\gamma}{1-\gamma}R_{\max}\right)-\frac{1}{1-\gamma
    }C_5(\varepsilon)(NT)^{-\delta}-\textup{TAD}^{\tilde \pi_K}(w^{(0)},a^{(0)}\cdots,w^{(i)}).
\end{align*}\end{sizeddisplay}
Therefore, it can be shown $\forall i=0,\cdots,n_{K,b}$ that:
{\footnotesize
\begin{align}\label{eq:pessimism final bound on Advantage when i=0..,n}
    &-A^{(i)}_{\pi^{\ast}_{n_{K,b}}}(w^{(0)},a^{(0)},\cdots,\widetilde{\pi}_K^{(i)}(w^{(i)},\cdots,w^{(0)},a^{(0)})) \nonumber\\
     &\leq \widetilde{\text{UQ}}(w^{(0)},a^{(0)},\cdots, w^{(i)},\pi^{(i)}_{n_{K,b},\ast}(w^{(i)},\cdots,w^{(0)},a^{(0)}))\nonumber\\
     &+\textup{TAD}^{\tilde \pi_K}(w^{(0)},a^{(0)},\cdots, w^{(i)},\widetilde{\pi}_K^{(i)}(w^{(i)},\cdots,w^{(0)},a^{(0)}))\nonumber\\
&+ 2\gamma^{K-1}\left(\frac{2-\gamma}{1-\gamma}R_{\max}\right)+\frac{2}{1-\gamma
    }C_5(\varepsilon)(NT)^{-\delta} \nonumber\\
&+\textup{AD}^{\tilde{\pi}_K}(w^{(0)},a^{(0)},\cdots, w^{(i)})\mathbb{I}[i=n_{K,b}].
\end{align}
}

Consequently, an upper bound on the term $i(\widetilde{\pi}_K)$ from Equation \eqref{pesseq: fqi regret 1_app} can be established based on Equation \eqref{eq:pessimism final bound on Advantage when i=0..,n} as follows:
\begin{sizeddisplay}
{\footnotesize}
\begin{align}\label{eq: pes bound on term (i)}
    i(\widetilde{\pi}_K)&=\EE^{\pi^{\ast}_{n_{K,b}}}\left[\sum_{t=0}^{\infty}\gamma^tR_t\right]  -\EE^{\tilde{\pi}_K}\left[\sum_{t=0}^{\infty}\gamma^tR_t\right]\nonumber\\
    &=\frac{1-\gamma^{n_{K,b}}}{(1-\gamma)^2}\EE_{(W^{(0)},A^{(0)},\cdots,W^{(I)}, A^{(I)}) \sim d_{\nu,I}^{\tilde{\pi}_K} \times \tilde{\pi}_K}\left[-A^{(I)}_{\pi^{\ast}_{n_{K,b}}}(W^{(0)},A^{(0)},\cdots,W^{(I)},A^{(I)})\prod_{j=1}^{I}\left\{\mathbb{I}[\Delta^{(j)}=0]\right\}\mathbb{I}[\Delta^{(0)}=1]\right]\nonumber\\
    &\leq \frac{1-\gamma^{n_{K,b}}}{(1-\gamma)^2}\EE_{(W^{(0)},A^{(0)},\cdots,W^{(I)}, A^{(I)}) \sim d_{\nu,I}^{\tilde{\pi}_K} }\left[\left(\EE^{\pi^{\ast}_{n_{K,b}}}\left[\widetilde{\text{UQ}}(W^{(0)},A^{(0)},\cdots W^{(I)},A^{(I)})\given (W^{(0)},A^{(0)},\cdots W^{(I)})\right]  \right.\right.\nonumber\\
    &\left.\left.+\EE^{\tilde{\pi}_K}\left[\textup{TAD}^{\tilde{\pi}_K}(W^{(0)},A^{(0)},\cdots W^{(I)},A^{(I)})\given(W^{(0)},A^{(0)},\cdots W^{(I)})\right]+ 2\gamma^{K-1}\left(\frac{2-\gamma}{1-\gamma}R_{\max}\right)+\frac{2}{1-\gamma
    }C_5(\varepsilon)(NT)^{-\delta} \right.\right.\nonumber\\
    &\left.\left. +\textup{AD}^{\tilde{\pi}_K}(W^{(0)},A^{(0)},\cdots W^{(I)})\mathbb{I}[I=n_{K,b}]
  \right)\prod_{j=1}^{I}\left\{\mathbb{I}[\Delta^{(j)}=0]\right\}\mathbb{I}[\Delta^{(0)}=1]\right],
\end{align}
\end{sizeddisplay}
which holds with probability at least $1 -(n_{K,b}+1)\epsilon-\varepsilon$.

Given Equation \eqref{eq: pes bound on term (i)}, the components related to the alignment discrepancies are exactly:
\begin{sizeddisplay}
{\scriptsize}
\begin{align*}
    \mathrm{TAP}(\widetilde \pi_K)
&:=
\frac{1-\gamma^{n_{K,b}}}{(1-\gamma)^2}
\EE_{H^{(I)} \sim d_{\nu,I}^{\tilde \pi_K}}
\!\left[
\EE^{\tilde \pi_K}\!\left[\textup{TAD}^{\tilde \pi_K}(W^{(0)},A^{(0)},\cdots,W^{(I)},A^{(I)}) \mid (W^{(0)},A^{(0)},\cdots,W^{(I)})\right]
\prod_{j=1}^{I}\mathbb{I}[\Delta^{(j)}=0]\,
\mathbb{I}[\Delta^{(0)}=1]
\right],
\end{align*}
\end{sizeddisplay}
and
\begin{sizeddisplay}
{\footnotesize}
\begin{align*}
   \mathrm{BAP}(\widetilde \pi_K)
&:=
\frac{1-\gamma^{n_{K,b}}}{(1-\gamma)^2}
\EE_{H^{(I)} \sim d_{\nu,I}^{\tilde \pi_K}}
\!\left[
\textup{AD}^{\tilde \pi_K}(W^{(0)},A^{(0)},\cdots,W^{(I)})\,\mathbb{I}[I=n_{K,b}]
\right]. 
\end{align*}
\end{sizeddisplay}
Here, $\textup{TAD}$ captures the boundary misalignment incurred at censoring depth $n_{K,b}$. Across iterations, it accumulates the suboptimality (measured under $Q^{(n_{K,b})}_{\pi^{\ast}_{n_{K,b}}}$) of the boundary action selected by $\widetilde{\pi}_k$ relative to the boundary action prescribed by $\pi^{\ast}_{n_{K,b}}$. Concretely, at the boundary and each iteration $k$, we compare
\begin{sizeddisplay}\small
\begin{align*}
&Q_{\pi^{\ast}_{n_{K,b}}}^{(n_{K,b})}\!\Big(W_0,A_0,\cdots,W_{n_{K,b}},\widetilde{\pi}^{(n_{K,b})}_{k}(W_{n_{K,b}},\cdots,W_0,A_0)\Big)-Q_{\pi^{\ast}_{n_{K,b}}}^{(n_{K,b})}\!\Big(W_0,A_0,\cdots,W_{n_{K,b}},\pi_{n_{K,b},*}^{(n_{K,b})}(W_{n_{K,b}},\cdots,W_0,A_0)\Big).
\end{align*}
\end{sizeddisplay}
Under Assumption~\ref{ass:TAD}, the action outputted by $\widetilde{\pi}^{(n_{K,b})}_k$ is in the set $\textup{UP}^{(n_{K,b})}$. Since $\pi^{\ast}_{n_{K,b}}$ also selects an action in $\textup{UP}^{(n_{K,b})}$ that is \emph{optimal} with respect to $Q_{\pi^{\ast}_{n_{K,b}}}^{(n_{K,b})}$, the above difference is non-positive. Therefore,
\[
\mathrm{TAP}(\widetilde \pi_K) \le 0,
\]

Furthermore, Assumption \ref{ass:AD} ensures the action $a_{\max}$ outputted by $\widetilde{\pi}_K^{(n_{K,b})}$ maximizes the estimated Q-function.
\begin{sizeddisplay}
{\footnotesize}
\begin{align}
    &\textup{AD}^{\hat{\pi}_K}(w^{(0)},a^{(0)},\cdots,w^{(n_{K,b})}) \\
    &=\max_{a\in\calA}\widetilde{Q}_{K}^{(n_{K,b})}(w^{(0)},a^{(0)},\cdots,w^{(n_{K,b})},a) - \widetilde{Q}_{K}^{(n_{K,b})}(w^{(0)},a^{(0)},\cdots,w^{(n_{K,b})},\widetilde{\pi}_K^{(n_{K,b})}(w^{(n_{K,b})},\cdots,w^{(0)},a^{(0)}))\nonumber\\ \nonumber
    &=0
\end{align}
\end{sizeddisplay} 
This implies that
\begin{align}\label{eq: alingment penalties pess}
     \mathrm{BAP}(\widetilde{\pi}_K) = 0.
\end{align}

Given these, Equation \eqref{eq: pes bound on term (i)} reduces to the following:
\begin{sizeddisplay}
{\footnotesize}
\begin{align}\label{eq: pess bound on term (i) 2nd}
    i(\widetilde{\pi}_K)
    &\leq \frac{1-\gamma^{n_{K,b}}}{(1-\gamma)^2}\EE_{(W^{(0)},A^{(0)},\cdots,W^{(I)}, A^{(I)}) \sim d_{\nu,I}^{\tilde{\pi}_K} }\left[\left(\EE^{\pi^{\ast}_{n_{K,b}}}\left[\widetilde{\text{UQ}}(W^{(0)},A^{(0)},\cdots W^{(I)},A^{(I)})\given (W^{(0)},A^{(0)},\cdots W^{(I)})\right]  \right.\right.\nonumber\\
    &\left.\left.+ 2\gamma^{K-1}\left(\frac{2-\gamma}{1-\gamma}R_{\max}\right) +\frac{2}{1-\gamma
    }C_5(\varepsilon)(NT)^{-\delta}
  \right)\prod_{j=1}^{I}\left\{\mathbb{I}[\Delta^{(j)}=0]\right\}\mathbb{I}[\Delta^{(0)}=1]\right],
\end{align}
\end{sizeddisplay}
which holds with probability at least $1 -(n_{K,b}+1)\epsilon-\varepsilon$

We now bound the contribution of $\widetilde{\text{UQ}}$ in Equation \eqref{eq: pess bound on term (i) 2nd}.
Throughout, we assume bounded rewards $|R_t|\le R_{\max}$ and $\|\phi_i(\cdot)/d^{(i)}  \|_2\le 1$ for all $i\in\{0,\ldots,n_{K,b}\}$. Given this, recall the characterization of $\widetilde{\text{UQ}}$
\begin{sizeddisplay}
    \scriptsize
\begin{align*}
    \widetilde{\text{UQ}}(w^{(0)},a^{(0)},\ldots,w^{(i)},a^{(i)})
=
\sum_{k=0}^{K-1}\gamma^k\,
\EE^{\pi^\ast_{n_{K,b}}}\!\left[2
\tilde{U}_{K-k}^{(M_{i+k})}\!\big(H_{(i+k-M_{i+k}):(i+k)}\big)
\ \Big|\ (W_0,A_0\ldots,W_i,A_i)=(w^{(0)},a^{(0)},\ldots,w^{(i)},a^{(i)})
\right],
\end{align*}
\end{sizeddisplay}
where $M_t
:=\min\Big\{n_{K,b},\ \max\{v\ge 0:\ \Delta_{t-v}=\cdots=\Delta_{t-1}=0,\ \Delta_{t-v-1}=1\}\Big\}$

Next, we derive an upper bound on $\tilde{U}_{k}^{(i)}$ which is defined for each depth $i$ and iteration index $k$ as follows
\begin{align}\label{eq: pess U_single_def}
\tilde{U}_{k}^{(i)}\!\big(h^{(t-i):t}\big)
\;:=\;
\tilde{\beta}_k\left[\phi_i(h^{(t-i):t})^\top(\Lambda^{(i)})^{-1}\phi_i(h^{(t-i):t})\right]^{1/2},
\end{align}
where 
\begin{align}\label{eq:Lambda_def}
\Lambda^{(i)}
\;:=\;
\sum_{\tau=1}^{|\mathcal{O}_N^{(i)}|} \phi_i(h^{(t-i):t})\phi_i(h^{(t-i):t})^\top
\;+\;\lambda I.
\end{align}
and thus, for each $i$, $U_{k}^{(i)}$ in Equation~\eqref{eq: pess U_single_def} satisfies Definition~\ref{def:UE} as implied by \citet{jin2021pessimism,chang2021mitigating}.

Given this, we fix $i$, invoke Assumption~\ref{ass: general coverage} and apply Jensen's inequality which yields
\begin{align}
\EE_{H^{(i)}\sim d^{\pi^\ast}}\!\big[\tilde{U}_{k}^{(i)}(H^{(i)})\big]
&\;=\;
\tilde{\beta}_k\,\EE_{H^{(i)}\sim d^{\pi^\ast}}\!\left[\sqrt{\phi_i(H^{(i)})^\top (\Lambda^{(i)})^{-1}\phi_i(H^{(i)})}\right] \nonumber \\
&\;\le\;
\tilde{\beta}_k\, \sqrt{C_{\mathrm{cov}}\EE_{H^{(i)}\sim d^{\mu}}\!\left[\phi_i(H^{(i)})^\top (\Lambda^{(i)})^{-1}\phi_i(H^{(i)})\right]}, \label{eq:measure_jensen_bound}
\end{align}
where $H^{(i)}$ is the generic random variable involving history block with length $i$.

We now bound the $\EE_{H^{(i)}\sim d^{\mu}}\!\left[\phi_i(H^{(i)})^\top (\Lambda^{(i)})^{-1}\phi_i(H^{(i)})\right]$ using Lemma~\ref{lemma: aux mitigating}. Let
\[
\Sigma_i^\mu \;:=\; \mathbb{E}_{H^{(i)}\sim d_i^\mu}\!\big[\phi_i(H^{(i)})\phi_i(H^{(i)})^\top\big],
\qquad
r^{(i)} \;:=\; \mathrm{rank}(\Sigma_i^\mu) \leq d^{(i)}.
\]
On the high-probability event $\Omega_{MP}$ from Lemma~\ref{lemma: aux mitigating}, we have for all $h$ and for all $i \in \{0,\cdots,n_{K,b}\}$,
\[
\phi_i(h)^\top (\Lambda^{(i)})^{-1}\phi_i(h)
\;\le\;
c_1\big(r^{(i)}+\log(c_2/\epsilon)\big)\,
\phi_i(h)^\top \big(|\mathcal{O}_N^{(i)}|\,\Sigma_i^\mu+\lambda I\big)^{-1}\phi_i(h).
\]
Then, we have
\begin{align}
\mathbb{E}_{H^{(i)}\sim d_i^{\mu}}\!\left[\phi_i(H^{(i)})^\top (\Lambda^{(i)})^{-1}\phi_i(H^{(i)})\right]
&\le
c_1\big(r^{(i)}+\log(c_2/\epsilon)\big)\,
\mathbb{E}_{H^{(i)}\sim d_i^{\mu}}\!\left[\phi_i(H^{(i)})^\top \big(|\mathcal{O}_N^{(i)}|\,\Sigma_i^\mu+\lambda I\big)^{-1}\phi_i(H^{(i)})\right] \nonumber\\
&=
c_1\big(r^{(i)}+\log(c_2/\epsilon)\big)\,
\mathrm{tr}\!\left(\Sigma_i^\mu\big(|\mathcal{O}_N^{(i)}|\,\Sigma_i^\mu+\lambda I\big)^{-1}\right).
\label{eq:trace_step}
\end{align}
where the last uses the standard identity $\mathbb{E}[x^\top A x]=\mathrm{tr}(A\,\mathbb{E}[xx^\top])$ with $x=\phi_i(H^{(i)})$ and $A=(|\mathcal{O}_N^{(i)}|\Sigma_i^\mu+\lambda I)^{-1}$.

Next, we use the spectral decomposition of $\Sigma_i^\mu$ with eigenvalues $\{\mu^{(i)}_j\}_{j=1}^{r^{(i)}}$ to upper bound the second term in Equation~\eqref{eq:trace_step}. This yields
\[
\mathrm{tr}\!\left(\Sigma_i^\mu\big(|\mathcal{O}_N^{(i)}|\,\Sigma_i^\mu+\lambda I\big)^{-1}\right)
=
\sum_{j=1}^{r^{(i)}}\frac{\mu_j^{(i)}}{|\mathcal{O}_N^{(i)}|\mu_j^{(i)}+\lambda}
\;\le\;
\sum_{j=1}^{r^{(i)}}\frac{1}{|\mathcal{O}_N^{(i)}|}
=
\frac{r^{(i)}}{|\mathcal{O}_N^{(i)}|},
\]
where the inequality follows from $\frac{\mu_j}{|\mathcal{O}_N^{(i)}|\mu_j+\lambda}\le \frac{\mu_j}{|\mathcal{O}_N^{(i)}|\mu_j}=\frac{1}{|\mathcal{O}_N^{(i)}|}$ (since $\lambda\ge 0$ and $\mu_j>0$ for $j\le r^{(i)}$).

Applying this to Equation~\eqref{eq:trace_step} brings
\begin{align}
\mathbb{E}_{H^{(i)}\sim d_i^{\mu}}\!\left[\phi_i(H^{(i)})^\top (\Lambda^{(i)})^{-1}\phi_i(H^{(i)})\right]
\;\le\;
c_1\big(r^{(i)}+\log(c_2/\epsilon)\big)\,\frac{r^{(i)}}{|\mathcal{O}_N^{(i)}|},
\label{eq:muZ_bound}
\end{align}
which holds with probability at least $1-\epsilon$ on the event $\Omega_{MP}$.

Then by substituting Equation \eqref{eq:muZ_bound} into Equation \eqref{eq:measure_jensen_bound}, we have
\begin{align}
\mathbb{E}_{H^{(i)}\sim d^{\pi^\ast}}\!\big[\tilde{U}_{k}^{(i)}(H^{(i)})\big]
&\leq
\tilde{\beta}_k\,
\sqrt{
C_{\mathrm{cov}}\,
c_1\big(r^{(i)}+\log(c_2/\epsilon)\big)\,
\frac{r^{(i)}}{|\mathcal{O}_N^{(i)}|}
}\nonumber\\
&\leq\tilde{\beta}_k\,
\sqrt{
C_{\mathrm{cov}}\,
c_1\big(d^{(i)}+\log(c_2/\epsilon)\big)\,
\frac{d^{(i)}}{|\mathcal{O}_N^{(i)}|}
}
\label{eq:final_U_bound}
\end{align}
which holds with probability at least $1-\epsilon$ simultaneously for each $i \in \{0,\cdots,n_{K,b}\}$. In the second inequality, we use the fact that $r^{(i)}\leq d^{(i)}$ for all $i \in \{0,\cdots,n_{K,b}\}$.

Next, we invoke Assumption~\ref{lower bound on partition sizes} (Uniform stratum mass / lower bound on
partition sizes): there exists a constant $\rho\in(0,1]$ and an event $\Omega_{\mathrm{str}}$ such that
$\Pr(\Omega_{\mathrm{str}})\ge 1-\epsilon$ and, on $\Omega_{\mathrm{str}}$,
\[
\min_{0\le i\le n_{K,b}}|\mathcal{O}_N^{(i)}|\ \ge\ \rho\,NT.
\]

Accordingly, by a union bound, we have 
\begin{align}
\Pr(\widetilde{\Omega}_P):=\Pr(\Omega_P \cap\Omega^r \cap \Omega_{\mathrm{str}}\cap \Omega_{MP}) \ \ge\ 1-(n_{K,b}+3)\epsilon -\varepsilon.
\end{align}

Next, we apply Equation \eqref{eq:final_U_bound} to the definition of $\widetilde{\text{UQ}}$ on the event on $\widetilde{\Omega}_P$. This brings
\begin{sizeddisplay}\small
\begin{align}\label{eq: pess UQ_bound_cov}
\widetilde{\text{UQ}}(w^{(0)},a^{(0)},\ldots,w^{(i)},a^{(i)})
&\leq
\sum_{k=0}^{K-1}\gamma^k\; 
\;\tilde{\beta}_{K-k}\EE_{M_{i+k} \sim \pi^*_{n_{K,b}}}\left[\sqrt{
C_{\mathrm{cov}}\,
c_1\big(d^{(M_{i+k})}+\log(c_2/\epsilon)\big)\,
\frac{d^{(M_{i+k})}}{|\mathcal{O}_N^{(M_{i+k})}|}
}\right]\nonumber\\
&\leq
\frac{\;\tilde{\beta}_{\max}}{1-\gamma}\sqrt{\frac{\max_{0\leq j \leq n_{K,b}}C_{\mathrm{cov}}\,
c_1\big(d^{(j)}+\log(c_2/\epsilon)\big)d^{(j)}}{\min_{0\le j\le n_{K,b}}|\mathcal{O}_N^{(j)}|}}\nonumber\\
&\leq
\frac{\tilde{\beta}_{\max}}{1-\gamma}\sqrt{\frac{C_{\mathrm{cov}}\,
c_1\big(d+\log(c_2/\epsilon)\big)d}{\rho NT}}
\end{align}\end{sizeddisplay}
where we use the fact that
$\min_{j}|\mathcal{O}_N^{(j)}|\ge \rho NT$, $d := \max_{0\leq i \leq n_{K,b}}d^{(i)}$ and  $\tilde{\beta}_{\max}:=\max_{0\le \ell\le K-1}\tilde{\beta}_\ell$.

Then, plugging Equation \eqref{eq: pess UQ_bound_cov} into Equation \eqref{eq: pess  bound on term (i) 2nd}, we can obtain
\begin{align}\label{eq: pess i_bound_plug_cov}
i(\widetilde{\pi}_K)
&\leq
\frac{1-\gamma^{n_{K,b}}}{(1-\gamma)^2}
\left(\frac{\tilde{\beta}_{\max}}{1-\gamma}\sqrt{\frac{C_{\mathrm{cov}}\,
c_1\big(d+\log(c_2/\epsilon)\big)d}{\rho NT}}
+
2\gamma^{K-1}\frac{2-\gamma}{1-\gamma}R_{\max}+
\frac{2}{1-\gamma
    }C_5(\varepsilon)(NT)^{-\delta}
\right)\nonumber\\
&\le
\frac{1}{(1-\gamma)^2}
\left(\frac{\tilde{\beta}_{\max}}{1-\gamma}\sqrt{\frac{C_{\mathrm{cov}}\,
c_1\big(d+\log(c_2/\epsilon)\big)d}{\rho NT}}+
2\gamma^{K-1}\frac{2-\gamma}{1-\gamma}R_{\max}
+
\frac{2}{1-\gamma
    }C_5(\varepsilon)(NT)^{-\delta}
\right),
\end{align}
which holds with probability at least $ 1-(n_{K,b}+3)\epsilon -\varepsilon.$

To state the final rate, we choose $K$ as follows.
\begin{align}
K \;=\; \left\lceil \frac{\log(NT)}{2(1-\gamma)} \right\rceil. \nonumber
\end{align}
This implies that $\sqrt{K-1} < \sqrt{\frac{\log(NT)}{2(1-\gamma)}}$ and  $\gamma^{K-1}\le (NT)^{-1/2}$. Therefore, the truncation contribution satisfies
\[
2\gamma^{K-1}\frac{2-\gamma}{1-\gamma}R_{\max}
\;\le\;
\frac{c_0\,R_{\max}}{1-\gamma}\cdot (NT)^{-1/2}
\]
for an absolute constant $c_0>0$. Furthermore, by following \citet{chang2021mitigating}, we choose $\tilde{\beta}_{\max}$ on the order of
\[
\tilde{\beta}_{\max}
\;\asymp\;
\sqrt{\,\Big(d+\log(c_2/\epsilon)\Big)d}
\qquad \text{(up to additional logarithmic factors and constants)}.
\]
and set the regularization parameter $\lambda$ used in $\Lambda^{(i)}$ to $1$.

Then, on event $\widetilde{\Omega}_P$, given Equations~\eqref{pesseq: fqi regret 1_app} and \eqref{eq: pess i_bound_plug_cov} as well as the selection of $K$ and $\tilde{\beta}_{\max}$, we have 
\begin{align}
\textup{\textbf{Regret}}(\widetilde{\pi}_K)
\;\le\;&
\frac{1}{(1-\gamma)^2}
\left(
\frac{\tilde{\beta}_{\max}}{1-\gamma}\sqrt{\frac{C_{\mathrm{cov}}\,
c_1\big(d+\log(c_2/\epsilon)\big)d}{\rho NT}}
+
\frac{c_0\,R_{\max}}{1-\gamma}\,(NT)^{-1/2}
+
\frac{2}{1-\gamma}C_5(\varepsilon)(NT)^{-\delta}
\right)\nonumber\\
&+\frac{4R_{\max}}{1-\gamma}
\sqrt{\left\lceil \frac{\log(NT)}{2(1-\gamma)} \right\rceil}\;\omega_{NT}\nonumber\\
\;\leq\;&
\frac{1}{(1-\gamma)^3}
\left(\big(d+\log(c_2/\epsilon)\big)d
\sqrt{\frac{C_{\mathrm{cov}}\,c_1}{\rho NT}}\;
+
\frac{c_0R_{\max}}{\sqrt{NT}}
+
2C_5(\varepsilon)(NT)^{-\delta}
\right)\nonumber\\
&+\frac{4R_{\max}}{1-\gamma}
\sqrt{\left\lceil \frac{\log(NT)}{2(1-\gamma)} \right\rceil}\;\omega_{NT}
\end{align}

To simplify the expression, we consider the regime in which the truncation term is dominated by the statistical term, namely without loss of generality, we assume
\begin{align}\label{eq: truncation pfqi}
   c_0R_{\max}\;\le\;\big(d+\log(c_2/\epsilon)\big)d\sqrt{\frac{C_{\mathrm{cov}}\,c_1}{\rho}}.
\end{align}

Under this condition, the truncation contribution is absorbed into the leading statistical term. Given this, we further use $\lesssim$ notation which hides absolute constants and logarithmic factors
(e.g., $\gamma$, $R_{\max}$), while keeping $\sqrt{\log(1/\epsilon)}$ explicit. This yields the final bound on $i(\widetilde{\pi}_K)$ and $\textup{\textbf{Regret}}(\widetilde{\pi}_K)$ as follows:
\begin{align}\label{eq: pess i_final_cov_simplified}
i(\widetilde{\pi}_K)
\;\lesssim\;&\big(d+\log(1/\epsilon)\big)d
\sqrt{\frac{C_{\mathrm{cov}}\,}{\rho NT}}\;
+
C_5(\varepsilon)(NT)^{-\delta}
\end{align}
and
\begin{align}\label{eq: pess final regret}
\textup{\textbf{Regret}}(\widetilde{\pi}_K)
\;\lesssim\;&\big(d+\log(1/\epsilon)\big)d
\sqrt{\frac{C_{\mathrm{cov}}\,}{\rho NT}}\;
+
C_5(\varepsilon)(NT)^{-\delta}
\;+\;
\omega_{NT},
\end{align}
which holds with probability at least $1-(n_{K,b}+3)\epsilon -\varepsilon$.

\subsection{Step 2: Regret for C-FQI under Idealized Setting}\label{C-FQI under Idealized Setting}

The proof for $\widehat{\pi}_K$ follows the same steps as $\widetilde{\pi}_K$. Recall that in the idealized setting, Assumptions~\ref{assump:bounded-n} and \ref{assump:sufficient-coverage-main-text} hold. Moreover, by definition, ($\widehat{\pi}_K \in \Pi_{n_{K,b}}$). Therefore, we may invoke the reduced form of Lemma~\ref{lm: regret decom}, which is valid under these two assumptions. Specifically, this corresponds to the decomposition given in Equatoin \eqref{eq:general regret decomposition with coverage} in the proof of Lemma~\ref{lm: regret decom}.
\begin{sizeddisplay}
{\footnotesize}
 \begin{align}\label{eq: fqi regret 1_app}
    \textup{\textbf{Regret}}(\widehat{\pi}_K)&\leq\underbrace{\EE^{\pi^*} \left[ \sum_{t=0}^{\infty} \gamma^t R_t \right]-\mathbb{E}^{\hat{\pi}_K}\left[ \sum_{t=0}^{\infty} \gamma^t R_t \right]
}_{\textstyle i(\widehat{\pi}_K)} \nonumber\\
    &\quad+\frac{4R_{\max}\sqrt{K-1}}{1-\gamma}\omega_{NT},
\end{align}
\end{sizeddisplay}
Thus, it remains to upper bound the term $i(\widehat{\pi})$.

Assumptions~\ref{assump:bounded-n} and \ref{assump:sufficient-coverage-main-text}, the global optimal policy satisfies $\pi^* \in \Pi_n \subseteq \Pi_{n_{K,b}}$. Since $\pi^*$ maximizes the value globally and is a valid candidate within the restricted class $\Pi_{n_{K,b}}$, it must coincide with the $\pi^{\ast}_{n_{K,b}}$ where  $\pi^{\ast}_{n_{K,b}}\in \argmax_{\pi' \in \Pi_{n_{K,b}}}\EE^{\pi'}\!\left[\sum_{t=0}^{\infty}\gamma^tR_t\right]$. Thus, $\pi^{\ast}_{n_{K,b}} = \pi^*$. Given this and the fact that $\pi^{\ast}_{n_{K,b}}$ and $\widehat{\pi}_K$ both belong to $\Pi_{n_{K,b}}$, Lemma~\ref{lm: performance difference} yields
\begin{sizeddisplay}
    {\footnotesize}{\begin{align}\label{eq: regret 1_app}
         i(\widehat{\pi}_K)
         &=\EE^{\pi^{\ast}}\left[\sum_{t=0}^{\infty}\gamma^tR_t\right]  -\EE^{\hat{\pi}_K}\left[\sum_{t=0}^{\infty}\gamma^tR_t\right]\nonumber\\
         &=\EE^{\pi^{\ast}_{n_{K,b}}}\left[\sum_{t=0}^{\infty}\gamma^tR_t\right]  -\EE^{\hat{\pi}_K}\left[\sum_{t=0}^{\infty}\gamma^tR_t\right]\\
         &=\frac{1-\gamma^{n_{K,b}}}{(1-\gamma)^2}\EE_{(W^{(0)},A^{(0)},\cdots,W^{(I)}, A^{(I)}) \sim d_{\nu,I}^{ \hat{\pi}_K} \times \hat{\pi}_K
    }\left[-A^{(I)}_{\pi^{\ast}_{n_{K,b}}}(W^{(0)},A^{(0)},\cdots,W^{(I)},A^{(I)})\prod_{j=1}^{I}\left\{\mathbb{I}[\Delta^{(j)}=0]\right\}\mathbb{I}[\Delta^{(0)}=1]\right].\nonumber  
    \end{align}}
\end{sizeddisplay}

Assume that by conditioning on $I=i$ and the sequence $(W^{(0)}, A^{(0)}, \dots, W^{(i)}) = (w^{(0)}, a^{(0)}, \dots, w^{(i)})$, where $\Delta^{(0)} = 1$ and $\Delta^{(j)} = 0$ for all $j = 1, \dots, i$, the expectation above can be upper bounded for all $i = 0, \dots, n_{K,b}-1$ as follows:

\begin{sizeddisplay} {\footnotesize} \begin{align}\label{eqn:fqi advantage ineq} &-A^{(i)}_{\pi^{\ast}_{n_{K,b}}}(w^{(0)},a^{(0)},\cdots, w^{(i)},\widehat{\pi}^{(i)}(w^{(i)},\cdots,w^{(0)},a^{(0)}))\nonumber\\ &=Q^{(i)}_{\pi^{\ast}_{n_{K,b}}}(w^{(0)},a^{(0)},\cdots, w^{(i)},\pi^{(i)}_{n_{K,b},*}(w^{(i)},\cdots,w^{(0)},a^{(0)}) - Q^{(i)}_{\pi^{\ast}_{n_{K,b}}}(w^{(0)},a^{(0)},\cdots, w^{(i)},\widehat{\pi}^{(i)}_K(w^{(i)},\cdots,w^{(0)},a^{(0)})\nonumber\\ &\leq Q^{(i)}_{\pi^{\ast}_{n_{K,b}}}(w^{(0)},a^{(0)},\cdots, w^{(i)},\pi^{(i)}_{n_{K,b},*}(w^{(i)},\cdots,w^{(0)},a^{(0)})-\widehat{Q}^{(i)}_K(w^{(0)},a^{(0)},\cdots, w^{(i)},\pi^{(i)}_{n_{K,b},*}(w^{(i)},\cdots,w^{(0)},a^{(0)}))\nonumber\\ &+\widehat{Q}^{(i)}_K(w^{(0)},a^{(0)},\cdots, w^{(i)},\widehat{\pi}^{(i)}_K(w^{(i)},\cdots,w^{(0)},a^{(0)})-Q^{(i)}_{\pi^{\ast}_{n_{K,b}}}(w^{(0)},a^{(0)},\cdots, w^{(i)},\widehat{\pi}^{(i)}_K(w^{(i)},\cdots,w^{(0)},a^{(0)}), \end{align} \end{sizeddisplay}
where the inequality stems from the optimality of $\widehat{\pi}_K$ on $\widehat{Q}^{(i)}_K$ $\forall i = 0, \dots, n_{K,b}-1$. 

For $i=n_{K,b}$, the policy $\widehat{\pi}_K$ no longer maintains its optimality with respect to $\widehat{Q}^{(n_{K,b})}_K$ as it picks $a_{\max}$. Consequently, the inequality specified in Equation \eqref{eqn:fqi advantage ineq} is not applicable when $i=n_{K,b}$. To quantify the optimality gap, let

\begin{sizeddisplay}
{\footnotesize}
\begin{align}\label{eq:fqi Qhat bound}
    &\textup{AD}^{\hat{\pi}_K}(w^{(0)},a^{(0)},\cdots,w^{(n_{K,b})}) \\
    &=\max_{a\in\calA}\widehat{Q}_{K}^{(n_{K,b})}(w^{(0)},a^{(0)},\cdots,w^{(n_{K,b})},a) - \widehat{Q}_{K}^{(n_{K,b})}(w^{(0)},a^{(0)},\cdots,w^{(n_{K,b})},\widehat{\pi}_K^{(n_{K,b})}(w^{(n_{K,b})},\cdots,w^{(0)},a^{(0)})).\nonumber
\end{align}
\end{sizeddisplay} 
It is worth noting that this constant disappears under Assumption \ref{ass:AD}, and thus, under idealized setting. In particular, Assumption \ref{ass:AD} implies that $a_{\max}$ is the action that maximizes the corresponding $Q$-function, so the constant term evaluates to zero. Accordingly, we will enforce this simplification at the end of the derivation, since Assumption \ref{ass:AD} holds in the idealized setting.

Hence, with this optimality gap, we can upper bound $-A^{(n_{K,b})}_{\pi^{\ast}_{n_{K,b}}}$ as follows:

\begin{sizeddisplay}
{\footnotesize}
\begin{align}\label{eqn:fqi advantage ineq when i=n}
&-A^{(n_{K,b})}_{\pi^{\ast}_{n_{K,b}}}(w^{(0)},a^{(0)},\cdots, w^{(n_{K,b})},\widehat{\pi}_{K}^{(n_{K,b})}(w^{(n_{K,b})},\cdots,w^{(0)},a^{(0)}))\nonumber\\
&= Q^{(n_{K,b})}_{\pi^{\ast}_{n_{K,b}}}(w^{(0)},a^{(0)},\cdots, w^{(n_{K,b})},\pi^{(n_{K,b})}_{n_{K,b},\ast}(w^{(n_{K,b})},\cdots,w^{(0)},a^{(0)})-\max_{a\in\calA}\widehat{Q}^{(n_{K,b})}_K(w^{(0)},a^{(0)},\cdots, w^{(n_{K,b})},a)\nonumber\\
&+\max_{a\in\calA}\widehat{Q}^{(n_{K,b})}_K(w^{(0)},a^{(0)},\cdots, w^{(n_{K,b})},a)-Q^{(n_{K,b})}_{\pi^{\ast}_{n_{K,b}}}(w^{(0)},a^{(0)},\cdots, w^{(n_{K,b})},\widehat{\pi}^{(n_{K,b})}_K(w^{(n_{K,b})},\cdots,w^{(0)},a^{(0)}))\nonumber\\
&\leq Q^{(n_{K,b})}_{\pi^{\ast}_{n_{K,b}}}(w^{(0)},a^{(0)},\cdots, w^{(n_{K,b})},\pi^{(n_{K,b})}_{n_{K,b},\ast}(w^{(n_{K,b})},\cdots,w^{(0)},a^{(0)})\nonumber\\
&-\widehat{Q}^{(n_{K,b})}_K(w^{(0)},a^{(0)},\cdots, w^{(n_{K,b})},\pi^{(n_{K,b})}_{n_{K,b},\ast}(w^{(n_{K,b})},\cdots,w^{(0)},a^{(0)}))\nonumber\\
&+\widehat{Q}^{(n_{K,b})}_K(w^{(0)},a^{(0)},\cdots, w^{(n_{K,b})},\widehat{\pi}^{(n_{K,b})}_K(w^{(n_{K,b})},\cdots,w^{(0)},a^{(0)}))\nonumber\\
&-Q^{(n_{K,b})}_{\pi^{\ast}_{n_{K,b}}}(w^{(0)},a^{(0)},\cdots, w^{(n_{K,b})},\widehat{\pi}^{(n_{K,b})}_K(w^{(n_{K,b})},\cdots,w^{(0)},a^{(0)})) \nonumber\\
&+ \textup{AD}^{\hat{\pi}_K}(w^{(0)},a^{(0)},\cdots, w^{(n_{K,b})}),
\end{align}
\end{sizeddisplay}
where first inequality comes from Equation \eqref{eq:fqi Qhat bound} and the fact that 
{\footnotesize \begin{align}
\max_{a\in\calA}\widehat{Q}^{(n_{K,b})}_K(w^{(0)},a^{(0)},\cdots, w^{(n_{K,b})},a) \geq \widehat{Q}^{(n_{K,b})}_K(w^{(0)},a^{(0)},\cdots, w^{(n_{K,b})},\widehat{\pi}^{(n_{K,b})}_*(w^{(n_{K,b})},\cdots,w^{(0)},a^{(0)})).\nonumber
\end{align}}
By combining Equations \eqref{eqn:fqi advantage ineq} and \eqref{eqn:fqi advantage ineq when i=n}, we have the following $\forall i=0,\cdots,n_{K,b}$:
\begin{sizeddisplay}
{\footnotesize}
\begin{align}\label{eq: fqi combined upper bound on Advantage i=0,..,n}
   &-A^{(i)}_{\pi^{\ast}_{n_{K,b}}}(w^{(0)},a^{(0)},\cdots, w^{(i)},\widehat{\pi}^{(i)}(w^{(i)},\cdots,w^{(0)},a^{(0)}))\nonumber\\
    &\leq Q^{(i)}_{\pi^{\ast}_{n_{K,b}}}(w^{(0)},a^{(0)},\cdots, w^{(i)},\pi^{(i)}_{n_{K,b},\ast}(w^{(i)},\cdots,w^{(0)},a^{(0)})-\widehat{Q}^{(i)}_K(w^{(0)},a^{(0)},\cdots, w^{(i)},\pi^{(i)}_{n_{K,b},\ast}(w^{(i)},\cdots,w^{(0)},a^{(0)}))\nonumber\\
&+\widehat{Q}^{(i)}_K(w^{(0)},a^{(0)},\cdots, w^{(i)},\widehat{\pi}^{(i)}_K(w^{(i)},\cdots,w^{(0)},a^{(0)})-Q^{(i)}_{\pi^{\ast}_{n_{K,b}}}(w^{(0)},a^{(0)},\cdots, w^{(i)},\widehat{\pi}^{(i)}_K(w^{(i)},\cdots,w^{(0)},a^{(0)}))\nonumber\\
&+\textup{AD}^{\hat{\pi}_K}(w^{(0)},a^{(0)},\cdots, w^{(i)})\mathbb{I}[i=n_{K,b}].
\end{align}
\end{sizeddisplay}
Recall that 
\begin{sizeddisplay}
{\footnotesize} \begin{align}
\overline{Q}^{(i)}_K(w^{(0)},a^{(0)},\cdots, w^{(i)},a^{(i)})=\widehat{Q}^{(i)}_K(w^{(0)},a^{(0)},\cdots, w^{(i)},a^{(i)})-C^{(i)}(w^{(0)},a^{(0)},\cdots,w^{(i)}),
\end{align}\end{sizeddisplay}
where the exact definition of $\overline{Q}_{j}$ $\forall j =0,\cdots,K$ is given in Equation \eqref{def: tilde Q} and the characterization of $C^{(i)}(w^{(0)},a^{(0)},\cdots,w^{(i)})$ is given in Definition \ref{def: price of pessimisim}.

Next, we add and subtract $C^{(i)}(w^{(0)},a^{(0)},\cdots,w^{(i)})$ to the right hand side of Equation \eqref{eq: fqi combined upper bound on Advantage i=0,..,n} to obtain $\overline{Q}^{(i)}$. This implies the following:
\begin{sizeddisplay}
{\footnotesize}
\begin{align}\label{eq: fqi combined upper bound on Advantage i=0,..,n, overline}
    &-A^{(i)}_{\pi^{\ast}_{n_{K,b}}}(w^{(0)},a^{(0)},\cdots, w^{(i)},\widehat{\pi}^{(i)}(w^{(i)},\cdots,w^{(0)},a^{(0)}))\nonumber\\
    &\leq Q^{(i)}_{\pi^{\ast}_{n_{K,b}}}(w^{(0)},a^{(0)},\cdots, w^{(i)},\pi^{(i)}_{n_{K,b},\ast}(w^{(i)},\cdots,w^{(0)},a^{(0)})-\overline{Q}^{(i)}_K(w^{(0)},a^{(0)},\cdots, w^{(i)},\pi^{(i)}_{n_{K,b},\ast}(w^{(i)},\cdots,w^{(0)},a^{(0)}))\nonumber\\
&+\overline{Q}^{(i)}_K(w^{(0)},a^{(0)},\cdots, w^{(i)},\widehat{\pi}^{(i)}_K(w^{(i)},\cdots,w^{(0)},a^{(0)})-Q^{(i)}_{\pi^{\ast}_{n_{K,b}}}(w^{(0)},a^{(0)},\cdots, w^{(i)},\widehat{\pi}^{(i)}_K(w^{(i)},\cdots,w^{(0)},a^{(0)}))\nonumber\\
&+\textup{AD}^{\hat{\pi}_K}(w^{(0)},a^{(0)},\cdots, w^{(i)})\mathbb{I}[i=n_{K,b}].
\end{align}
\end{sizeddisplay}

To upper bound the right hand side of Equation \eqref{eq: fqi combined upper bound on Advantage i=0,..,n, overline}, we use Lemmas \ref{lm:upper bound on Qs} and \ref{lm:lower bound on Qs}, which provide an upper and a lower bound on the difference between $Q^{(i)}_{\pi^{\ast}_{n_{K,b}}}$ and $\overline{Q}^{(i)}_K$ on the event $\Omega \cap\Omega^r$ that holds $\forall i=0,\cdots,n_{K,b}$ simultaneously with probability at least $1 -(n_{K,b}+1)\epsilon-\varepsilon$:
{\footnotesize
\begin{align*}
&Q^{(i)}_{\pi^{\ast}_{n_{K,b}}}(w^{(0)},a^{(0)},\cdots, w^{(i)},a^{(i)})-\overline{Q}^{(i)}_K(w^{(0)},a^{(0)},\cdots, w^{(i)},a^{(i)})\\
&\leq \gamma^{K-1}\left(\frac{2-\gamma}{1-\gamma}R_{\max}\right)+ \frac{1}{1-\gamma
    }C_5(\varepsilon)(NT)^{-\delta}\nonumber\\
     & + \textup{UQ}(w^{(0)},a^{(0)}\cdots,w^{(i)},a^{(i)}) +C^{(i)}(w^{(0)},a^{(0)}\cdots,w^{(i)})\end{align*}}
     and
{\footnotesize
\begin{align*}
&Q^{(i)}_{\pi^{\ast}_{n_{K,b}}}(w^{(0)},a^{(0)},\cdots, w^{(i)},a^{(i)})-\overline{Q}^{(i)}_K(w^{(0)},a^{(0)},\cdots, w^{(i)},a^{(i)})\\
    &\geq -\gamma^{K-1}\left(\frac{2-\gamma}{1-\gamma}R_{\max}\right)-\frac{1}{1-\gamma
    }C_5(\varepsilon)(NT)^{-\delta}-\textup{TAD}^{\hat \pi_K}(w^{(0)},a^{(0)}\cdots,w^{(i)},a^{(i)}).
\end{align*}}

Hence, we can show that $\forall i=0,\cdots,n_{K,b}$:{\footnotesize\begin{align}\label{eq:fqi final bound on Advantage when i=0..,n}
    &-A^{(i)}_{\pi^{\ast}_{n_{K,b}}}(w^{(0)},a^{(0)},\cdots,\widehat{\pi}_K^{(i)}(w^{(i)},\cdots,w^{(0)},a^{(0)})) \nonumber\\
     &\leq \text{UQ}(w^{(0)},a^{(0)},\cdots, w^{(i)},\pi^{(i)}_{n_{K,b},\ast}(w^{(i)},\cdots,w^{(0)},a^{(0)}))\nonumber\\
     &+\textup{TAD}^{\hat \pi_K}(w^{(0)},a^{(0)},\cdots, w^{(i)},\widehat{\pi}^{(i)}_{K}(w^{(i)},\cdots,w^{(0)},a^{(0)}))\nonumber\\
&+C^{(i)}(w^{(0)},a^{(0)},\cdots, w^{(i)})+ 2\gamma^{K-1}\left(\frac{2-\gamma}{1-\gamma}R_{\max}\right)+\frac{2}{1-\gamma
    }C_5(\varepsilon)(NT)^{-\delta} \nonumber\\
&+\textup{AD}^{\hat{\pi}_K}(w^{(0)},a^{(0)},\cdots, w^{(i)})\mathbb{I}[i=n_{K,b}].
\end{align}}

Consequently, an upper bound on the term $i(\widehat{\pi})$ from Equation \eqref{eq: regret 1_app} can be formed based on Equation \eqref{eq:fqi final bound on Advantage when i=0..,n} as follows:
\begin{sizeddisplay}
{\footnotesize}
\begin{align}\label{eq: bound on term (i)}
    i(\widehat{\pi}_K)
    &=\frac{1-\gamma^{n_{K,b}}}{(1-\gamma)^2}\EE_{(W^{(0)},A^{(0)},\cdots,W^{(I)}, A^{(I)}) \sim d_{\nu,I}^{\hat{\pi}_K} \times \
    \hat{\pi}_K}\left[-A^{(I)}_{\pi^{\ast}_{n_{K,b}}}(W^{(0)},A^{(0)},\cdots,W^{(I)},A^{(I)})\prod_{j=1}^{I}\left\{\mathbb{I}[\Delta^{(j)}=0]\right\}\mathbb{I}[\Delta^{(0)}=1]\right]\nonumber\\
    &\leq \frac{1-\gamma^{n_{K,b}}}{(1-\gamma)^2}\EE_{(W^{(0)},A^{(0)},\cdots,W^{(I)}, A^{(I)}) \sim d_{\nu,I}^{\hat{\pi}_K} }\left[\left(\EE^{\pi^{\ast}_{n_{K,b}}}\left[\text{UQ}(W^{(0)},A^{(0)},\cdots W^{(I)},A^{(I)})\given (W^{(0)},A^{(0)},\cdots W^{(I)})\right]  \right.\right.\nonumber\\
    &\left.\left.+ \EE^{\hat{\pi}_K}\left[ \textup{TAD}^{\hat \pi_K}(W^{(0)},A^{(0)},\cdots W^{(I)},A^{(I)})\given(W^{(0)},A^{(0)},\cdots W^{(I)})\right]+C^{(I)}(W^{(0)},A^{(0)},\cdots W^{(I)})+ 2\gamma^{K-1}\left(\frac{2-\gamma}{1-\gamma}R_{\max}\right) \right.\right.\nonumber\\
    &\left.\left. +\frac{2}{1-\gamma
    }C_5(\varepsilon)(NT)^{-\delta}+\textup{AD}^{\hat{\pi}_K}(W^{(0)},A^{(0)},\cdots W^{(I)})\mathbb{I}[I=n_{K,b}]
  \right)\prod_{j=1}^{I}\left\{\mathbb{I}[\Delta^{(j)}=0]\right\}\mathbb{I}[\Delta^{(0)}=1]\right],
\end{align}
\end{sizeddisplay}
which holds with probability at least $1 -(n_{K,b}+1)\epsilon-\varepsilon$.

Given Equation \eqref{eq: bound on term (i)}, the components related to the alignment discrepancies are exactly:
\begin{sizeddisplay}
\scriptsize
\begin{align*}
    \mathrm{TAP}(\widehat \pi_K)
&:=
\frac{1-\gamma^{n_{K,b}}}{(1-\gamma)^2}
\EE_{H^{(I)} \sim d_{\nu,I}^{\tilde \pi_K}}
\!\left[
\EE^{\tilde \pi_K}\!\left[\textup{TAD}^{\tilde \pi_K}(W^{(0)},A^{(0)},\cdots,W^{(I)},A^{(I)}) \mid (W^{(0)},A^{(0)},\cdots,W^{(I)})\right]
\prod_{j=1}^{I}\mathbb{I}[\Delta^{(j)}=0]\,
\mathbb{I}[\Delta^{(0)}=1]
\right],
\end{align*}
\end{sizeddisplay}
and
\begin{sizeddisplay}
{\footnotesize}
\begin{align*}
   \mathrm{BAP}(\widehat \pi_K)
&:=
\frac{1-\gamma^{n_{K,b}}}{(1-\gamma)^2}
\EE_{H^{(I)} \sim d_{\nu,I}^{\tilde \pi_K}}
\!\left[
\textup{AD}^{\tilde \pi_K}(W^{(0)},A^{(0)},\cdots,W^{(I)})\,\mathbb{I}[I=n_{K,b}]
\right]. 
\end{align*}
\end{sizeddisplay}
Here, $\textup{TAD}$ captures the boundary misalignment incurred at censoring depth $n_{K,b}$. Across iterations, it accumulates the suboptimality (measured under $Q^{(n_{K,b})}_{\pi^{\ast}_{n_{K,b}}}$) of the boundary action selected by $\widehat{\pi}_k$ relative to the boundary action prescribed by $\pi^{\ast}_{n_{K,b}}$. Concretely, at the boundary and each iteration $k$, we compare
\begin{sizeddisplay}\small
\begin{align*}
&Q_{\pi^{\ast}_{n_{K,b}}}^{(n_{K,b})}\!\Big(W_0,A_0,\cdots,W_{n_{K,b}},\widehat{\pi}^{(n_{K,b})}_{k}(W_{n_{K,b}},\cdots,W_0,A_0)\Big) \\
&\hspace{3em}-Q_{\pi^{\ast}_{n_{K,b}}}^{(n_{K,b})}\!\Big(W_0,A_0,\cdots,W_{n_{K,b}},\pi_{n_{K,b},*}^{(n_{K,b})}(W_{n_{K,b}},\cdots,W_0,A_0)\Big).
\end{align*}
\end{sizeddisplay}
Under Assumption~\ref{ass:TAD}, the action outputted by $\widehat{\pi}^{(n_{K,b})}_k$ is in the set $\textup{UP}^{(n_{K,b})}$. Since $\pi^{\ast}_{n_{K,b}}$ also selects an action in $\textup{UP}^{(n_{K,b})}$ that is \emph{optimal} with respect to $Q_{\pi^{\ast}_{n_{K,b}}}^{(n_{K,b})}$, the above difference is non-positive. Therefore,
\[
\mathrm{TAP}(\widehat \pi_K) \leq 0,
\]

Furthermore, Assumption \ref{ass:AD} ensures the action $a_{\max}$ outputted by $\widehat{\pi}_K^{(n_{K,b})}$ maximizes the estimated Q-function.
\begin{sizeddisplay}
{\footnotesize}
\begin{align}
    &\textup{AD}^{\hat{\pi}_K}(w^{(0)},a^{(0)},\cdots,w^{(n_{K,b})}) \\
    &=\max_{a\in\calA}\widehat{Q}_{K}^{(n_{K,b})}(w^{(0)},a^{(0)},\cdots,w^{(n_{K,b})},a) - \widehat{Q}_{K}^{(n_{K,b})}(w^{(0)},a^{(0)},\cdots,w^{(n_{K,b})},\widehat{\pi}_K^{(n_{K,b})}(w^{(n_{K,b})},\cdots,w^{(0)},a^{(0)}))\nonumber\\ \nonumber
    &=0
\end{align}
\end{sizeddisplay} 
This implies that
\begin{align}\label{eq: alingment penalties fqi}
     \mathrm{BAP}(\widehat{\pi}_K) = 0.
\end{align}

Given these, Equation \eqref{eq: bound on term (i)} reduces to the following:
\begin{sizeddisplay}
{\footnotesize}
\begin{align}\label{eq: bound on term (i) 2nd}
    i(\widehat{\pi}_K)
    &\leq \frac{1-\gamma^{n_{K,b}}}{(1-\gamma)^2}\EE_{(W^{(0)},A^{(0)},\cdots,W^{(I)}, A^{(I)}) \sim d_{\nu,I}^{\hat{\pi}_K} }\left[\left(\EE^{\pi^{\ast}_{n_{K,b}}}\left[\text{UQ}(W^{(0)},A^{(0)},\cdots W^{(I)},A^{(I)})\given (W^{(0)},A^{(0)},\cdots W^{(I)})\right]  \right.\right.\nonumber\\
    &\left.\left.+ C^{(I)}(W^{(0)},A^{(0)},\cdots W^{(I)})+ 2\gamma^{K-1}\left(\frac{2-\gamma}{1-\gamma}R_{\max}\right) +\frac{2}{1-\gamma
    }C_5(\varepsilon)(NT)^{-\delta}
  \right)\prod_{j=1}^{I}\left\{\mathbb{I}[\Delta^{(j)}=0]\right\}\mathbb{I}[\Delta^{(0)}=1]\right],
\end{align}
\end{sizeddisplay}
which holds with probability at least $1 -(n_{K,b}+1)\epsilon-\varepsilon$.

We now bound the contribution of $\textup{UQ}$ and $C^{(I)}$ in Equation \eqref{eq: bound on term (i) 2nd}.
Throughout, without loss of generality, we assume $|R_t|\le R_{\max}$ and $\|\phi_i(\cdot)/d^{(i)}  \|_2\le 1$ for all $i\in\{0,\ldots,n_{K,b}\}$. Given this, recall the characterization of $\text{UQ}$ and $C^{(I)}$ as follows. 
\begin{sizeddisplay}
    \scriptsize\begin{align*}
        \text{UQ}(w^{(0)},a^{(0)},\ldots,w^{(i)},a^{(i)})
=
\sum_{k=0}^{K-1}\gamma^k\,
\EE^{\pi^\ast_{n_{K,b}}}\!\left[U_{K-k}^{(M_{i+k})}\!\big(H_{(i+k-M_{i+k}):(i+k)}\big)
\ \Big|\ (W_0,A_0\ldots,W_i,A_i)=(w^{(0)},a^{(0)},\ldots,w^{(i)},a^{(i)})
\right],
    \end{align*}
\end{sizeddisplay}

\begin{sizeddisplay}
    \scriptsize
\begin{align}
C^{(i)}(w^{(0)},a^{(0)},\ldots,w^{(i)})
&=
\max_{a\in\mathcal{A}}
\sum_{k=0}^{K-1}\gamma^{k}\,
\EE^{\pi_{K-1},\cdots,\pi_{K-k}}\!\left[
U_{K-k}^{(M_{i+k})}\big(H_{i+k}^{(M_{i+k})}\big)
\ \Big|\ (W_0,\ldots,W_i,A_i)=(w^{(0)},\ldots,w^{(i)},a)
\right],\nonumber
\end{align}
\end{sizeddisplay}
where $M_t
:=\min\Big\{n_{K,b},\ \max\{v\ge 0:\ \Delta_{t-v}=\cdots=\Delta_{t-1}=0,\ \Delta_{t-v-1}=1\}\Big\}$

Next, we derive an upper bound on ${U}_{k}^{(i)}$ which is defined for each depth $i$ and iteration index $k$ as follows
\begin{align}\label{eq: U_single_def}
U_{k}^{(i)}\!\big(h^{(t-i):t}\big)
\;:=\;
\beta_k\left[\phi_i(h^{(t-i):t})^\top(\Lambda^{(i)})^{-1}\phi_i(h^{(t-i):t})\right]^{1/2},
\end{align}
where 
\begin{align}
\Lambda^{(i)}
\;:=\;
\sum_{\tau=1}^{|\mathcal{O}_N^{(i)}|} \phi_i(h^{(t-i):t})\phi_i(h^{(t-i):t})^\top
\;+\;\lambda I.
\end{align}

Next, we rely on Lemma~\ref{lemma: Matrix bound} to relate the empirical covariance $\widehat{\Sigma}^{(i)}:= \frac{1}{|\mathcal{O}_N^{(i)}|}\sum_{\tau=1}^{|\mathcal{O}_N^{(i)}|} \phi_i(h_\tau^{(t-i):t})\phi_i(h_\tau^{(t-i):t})^\top$ to the population covariance $\Sigma_i^\mu$. Specifically, under the event $\Omega_M$ from Lemma~\ref{lemma: Matrix bound} which holds with probability at least $1-\epsilon$, the regularized design matrix $\Lambda^{(i)} = |\mathcal{O}_N^{(i)}| \widehat{\Sigma}^{(i)} + \lambda I$ satisfies the following for all $i \in \{0,\dots, n_{K,b}\}$:
\[
\widehat{\Sigma}^{(i)} \succeq \tfrac12\Sigma_i^\mu 
\]
Given this for the sake of simplicity, we assume $\Sigma_i^\mu \succeq \sigma I$ for all $i\in\{0,\ldots,n_{K,b}\}.$
This, in turn, leads to the following
$$\widehat{\Sigma}^{(i)} \succeq \tfrac12\Sigma_i^\mu \succeq \tfrac{\sigma}{2}I,$$
which further implies
\begin{align}\label{eq: fqi lambda bound}
    \Lambda^{(i)} = |\mathcal{O}_N^{(i)}|\widehat{\Sigma}^{(i)}+\lambda I \succeq \Big(\tfrac{\sigma}{2}|\mathcal{O}_N^{(i)}|+\lambda\Big)I.
\end{align}

Given this, we fix $i$ and utilize Equation \eqref{eq: fqi lambda bound} which yields, for any policy $\pi \in \Pi$, the following
\begin{align}\label{eq: bound on uq fqi}
\EE_{H^{(i)}\sim d^{\pi}}\!\big[\tilde{U}_{k}^{(i)}(H^{(i)})\big]
&=
\beta_k\, \EE_{H^{(i)}\sim d^{\pi}}\!\left[\sqrt{\phi_i(H^{(i)})^\top (\Lambda^{(i)})^{-1}\phi_i(H^{(i)})}\right] \nonumber \\
&\;\le\;
\beta_k\, \sqrt{\EE_{H^{(i)}\sim d^{\pi}}\!\left[\phi_i(H^{(i)})^\top (\Lambda^{(i)})^{-1}\phi_i(H^{(i)})\right]}\nonumber\\
&\leq \beta_k \sqrt{\big\|(\Lambda^{(i)})^{-1}\big\|_{\mathrm{op}}\,
\EE_{H^{(i)}\sim d^{\pi}}\!\left[\|\phi_i(H^{(i)})\|_2^2\ \right] }\nonumber\\
&\leq \beta_k \sqrt{\frac{(d^{(i)})^2}{\tfrac{\sigma}{2}|\mathcal{O}_N^{(i)}|+\lambda}} =  \beta_k \frac{d^{(i)}}{\sqrt{\tfrac{\sigma}{2}|\mathcal{O}_N^{(i)}|+\lambda} },
\end{align}
where $H^{(i)}$ is the generic random variable involving history block with length $i$. In the first step we use Jensen's inequality and in the second step we use $x^\top A x \le \|A\|_{\mathrm{op}}\|x\|_2^2$ and lastly we utilize Equation \eqref{eq: fqi lambda bound} to bound the operator norm.

Next, we invoke Assumption~\ref{lower bound on partition sizes} (Uniform stratum mass / lower bound on
partition sizes): there exists a constant $\rho\in(0,1]$ and an event $\Omega_{\mathrm{str}}$ such that
$\Pr(\Omega_{\mathrm{str}})\ge 1-\epsilon$ and, on $\Omega_{\mathrm{str}}$,
\[
\min_{0\le i\le n_{K,b}}|\mathcal{O}_N^{(i)}|\ \ge\ \rho\,NT.
\]

Accordingly, by a union bound, we have 
\begin{align}
\Pr(\widetilde{\Omega}):=\Pr(\Omega \cap\Omega^r \cap \Omega_{\mathrm{str}}\cap \Omega_M) \ \ge\ 1-(n_{K,b}+3)\epsilon -\varepsilon.
\end{align}

Given this, we apply Equation \eqref{eq: bound on uq fqi} to the definition of ${\text{UQ}}$ on the event on $\widetilde{\Omega}$. This brings
\begin{align}\label{eq:UQ_bound_cov}
\textup{UQ}(w^{(0)},a^{(0)},\ldots,w^{(i)},a^{(i)})
&\le
\sum_{k=0}^{K-1}\gamma^k\;
\;\beta_{K-k}\EE_{M_{i+k} \sim \pi^*_{n_{K,b}}}\left[\frac{d^{(M_{i+k})}}{\sqrt{\tfrac{\sigma}{2}|\mathcal{O}_N^{(M_{i+k})}|+\lambda}}\right]
\nonumber\\
&\leq \frac{\;\beta_{\max}}{1-\gamma}\frac{\max_{0\le j\le n_{K,b}}d^{(j)}}{\sqrt{\tfrac{\sigma}{2}\min_{0\le j\le n_{K,b}}|\mathcal{O}_N^{(j)}| +\lambda}}\nonumber\\
&\leq \frac{\;\beta_{\max}}{1-\gamma}\frac{d}{\sqrt{\tfrac{\sigma}{2}\rho NT +\lambda}}
\end{align}
and similarly for $C^{(i)}$, we have
\begin{align}\label{eq:C_bound_cov}
C^{(i)}(w^{(0)},a^{(0)},\ldots,w^{(i)})
&\leq\frac{\;\beta_{\max}}{1-\gamma}\frac{d}{\sqrt{\tfrac{\sigma}{2}\rho NT +\lambda}}
\end{align}
where we use the fact that
$\min_{j}|\mathcal{O}_N^{(j)}|\ge \rho NT$, $d := \max_{0\leq i \leq n_{K,b}}d^{(i)}$ and  $\beta_{\max}:=\max_{0\le \ell\le K-1}\beta_\ell$.

Then, plugging Equations~\eqref{eq:UQ_bound_cov} and \eqref{eq:C_bound_cov} into Equation \eqref{eq: bound on term (i) 2nd}, we can obtain
\begin{align}\label{eq:i_bound_plug_cov}
i(\widehat{\pi}_K)
&\le
\frac{1-\gamma^{n_{K,b}}}{(1-\gamma)^2}
\Bigg(
\frac{\beta_{\max}}{1-\gamma}\frac{d}{\sqrt{\tfrac{\sigma}{2}\rho NT +\lambda}}
+
\frac{\beta_{\max}}{1-\gamma}\frac{d}{\sqrt{\tfrac{\sigma}{2}\rho NT +\lambda}}
+
2\gamma^{K-1}\frac{2-\gamma}{1-\gamma}R_{\max}
+
\frac{2}{1-\gamma}C_5(\varepsilon)(NT)^{-\delta}
\Bigg)\nonumber\\
&\le
\frac{1}{(1-\gamma)^2}
\left(
\frac{2\beta_{\max}}{1-\gamma}\frac{d}{\sqrt{\tfrac{\sigma}{2}\rho NT +\lambda}}
+
2\gamma^{K-1}\frac{2-\gamma}{1-\gamma}R_{\max}
+
\frac{2}{1-\gamma}C_5(\varepsilon)(NT)^{-\delta}
\right),
\end{align}
which holds with probability at least $1-(n_{K,b}+3)\epsilon-\varepsilon$.

To state the final rate, we choose $K$ as follows
\begin{align}\label{eq:K_choice_cov}
K \;=\; \left\lceil \frac{\log(NT)}{2(1-\gamma)} \right\rceil.
\end{align}
This implies that $\sqrt{K-1}<\sqrt{\frac{\log(NT)}{2(1-\gamma)}}$ and $\gamma^{K-1}\le (NT)^{-1/2}$.
Therefore, the truncation contribution satisfies
\[
2\gamma^{K-1}\frac{2-\gamma}{1-\gamma}R_{\max}
\;\le\;
\frac{c_0\,R_{\max}}{1-\gamma}\cdot (NT)^{-1/2}
\]
for an absolute constant $c_0>0$.

Furthermore, by following \citet{chang2021mitigating}, we choose $\tilde{\beta}_{\max}$ on the order of
\[
{\beta}_{\max}
\;\asymp\;
\sqrt{\,\Big(d+\log(1/\epsilon)\Big)d}
\qquad
\]
and set the regularization parameter $\lambda$ used in $\Lambda^{(i)}$ to $1$.

Then, on event $\Omega_P$, given Equations~\eqref{eq: fqi regret 1_app} and \eqref{eq:i_bound_plug_cov}
as well as the selection of $K$ and $\beta_{\max}$, we have
\begin{align}
\textup{\textbf{Regret}}(\widehat{\pi}_K)
\;\le\;&
\frac{1}{(1-\gamma)^3}
\Bigg(
2\,\sqrt{\big(d+\log(1/\epsilon)\big)\,d}\;\frac{d}{\sqrt{\tfrac{\sigma}{2}\rho NT + \lambda}}
\;+\;
\frac{c_0 R_{\max}}{\sqrt{NT}}
\;+\;
2\,C_5(\varepsilon)\,(NT)^{-\delta}
\Bigg),\nonumber\\
&\;+\;
\frac{4R_{\max}}{1-\gamma}
\sqrt{\left\lceil \frac{\log(NT)}{2(1-\gamma)} \right\rceil}\;\omega_{NT}.
\end{align}

To simplify the expression, we consider the regime in which the truncation term is dominated by the statistical term, namely assume
\begin{align}\label{eq: truncation fqi}
    2 d \sqrt{d + \log(1/\epsilon)} \;\ge\; c_0 R_{\max} \sqrt{\frac{\sigma \rho}{2}}.
\end{align}

Under this condition, the truncation contribution is absorbed into the leading statistical term.
Given this, we further use $\lesssim$ notation which hides absolute constants and logarithmic factors
(e.g., $\gamma$, $R_{\max}$ and $\sigma$), while keeping $\sqrt{\log(1/\epsilon)}$ explicit.

This yields the final bound on $i(\widehat{\pi}_K)$ and $\textup{\textbf{Regret}}(\widehat{\pi}_K)$ as follows:
\begin{align}\label{eq:i_final_cov_simplified}
i(\widehat{\pi}_K)
\;\lesssim\;&d
\sqrt{\frac{\big(d+\log(1/\epsilon)\big)d}{\rho\,NT}}
\;+\;
C_5(\varepsilon)(NT)^{-\delta}
\end{align}
and
\begin{align}\label{eq: final regret}
\textup{\textbf{Regret}}(\widehat{\pi}_K)
\;\lesssim\;&
d
\sqrt{\frac{\big(d+\log(1/\epsilon)\big)d}{\rho\,NT}}
\;+\;
C_5(\varepsilon)(NT)^{-\delta}
\;+\;
\omega_{NT},
\end{align}
which holds with probability at least $1-(n_{K,b}+3)\epsilon-\varepsilon$.



\subsection{Step 3: Combined Regret Bound}

We now combine the results from the previous two steps to derive the final bound.
First, recall that under the event
$\widetilde{\Omega}_P$,
Equation~\eqref{eq: pess final regret} yields
\begin{sizeddisplay}
{\footnotesize}
 \begin{align}\label{eq: combined regret tilde}
\textup{\textbf{Regret}}(\widetilde{\pi}_K)
\;\lesssim\;&\big(d+\log(1/\epsilon)\big)d
\sqrt{\frac{C_{\mathrm{cov}}\,}{\rho NT}}\;
+
C_5(\varepsilon)(NT)^{-\delta}
\;+\;
\omega_{NT},
\end{align}
\end{sizeddisplay}
which holds with probability at least $ 1-(n_{K,b}+3)\epsilon -\varepsilon$.

Similarly, under the event
$\widetilde{\Omega}$,
Equation~\eqref{eq: final regret} gives
\begin{sizeddisplay}
{\footnotesize}
 \begin{align}\label{eq: combined regret hat}
   \textup{\textbf{Regret}}(\widehat{\pi}_K)
\;\lesssim\;&
d
\sqrt{\frac{\big(d+\log(1/\epsilon)\big)d}{\rho\,NT}}
\;+\;
C_5(\varepsilon)(NT)^{-\delta}
\;+\;
\omega_{NT},
\end{align}
\end{sizeddisplay}
which holds with probability at least $ 1-(n_{K,b}+3)\epsilon -\varepsilon$.

Define the simultaneous event
\begin{align}\label{eq: simultaneous event}
\Omega_{\mathrm{all}}
\;:=\;
\widetilde{\Omega}_P \cap \widetilde{\Omega}.
\end{align}
By a union bound,
\begin{align}\label{eq: prob Omega all}
\Pr(\Omega_{\mathrm{all}})
\;\ge\;
1
-
\Pr(\widetilde{\Omega}_P^{c})
-
\Pr(\widetilde{\Omega}^{c})
\;\ge\;
1-2\big((n_{K,b}+3)\epsilon+\varepsilon\big),
\end{align}
and on $\Omega_{\mathrm{all}}$ both regret bounds in
Equations~\eqref{eq: combined regret tilde} and~\eqref{eq: combined regret hat}
hold simultaneously.

We define the statistical error term $\mathcal{E}_{\text{stat}}(NT, d)$ by
\begin{align}\label{eq: def stat error}
\mathcal{E}_{\text{stat}}(NT, d) :=
\max\left(
\big(d+\log(1/\epsilon)\big)d
\sqrt{\frac{C_{\mathrm{cov}}}{\rho NT}},\ 
d
\sqrt{\frac{\big(d+\log(1/\epsilon)\big)d}{\rho\,NT}}
\right)
+
C_5(\varepsilon)(NT)^{-\delta}
+
\omega_{NT}
\end{align}

Therefore, on $\Omega_{\mathrm{all}}$,
\begin{align}\label{eq: max regret final}
\max\!\big(\textup{\textbf{Regret}}(\widehat{\pi}_K),\ \textup{\textbf{Regret}}(\widetilde{\pi}_K)\big)
\;\lesssim\;
\mathcal{E}_{\text{stat}}(NT, d),
\end{align}
which holds with probability at least
$1-2\big((n_{K,b}+3)\epsilon+\varepsilon\big)$. This concludes the proof of Theorem~\ref{cor:ideal}.

\section{Proof of Corollary~\ref{cor:relaxed_coverage}}\label{app:proof-relaxed-coverage}

This section provides a proof of Corollary~\ref{cor:relaxed_coverage} by extending the idealized setting to the setting where Assumption~\ref{assump:sufficient-coverage-main-text} (sufficient coverage / bounded censoring depth under $\pi^\ast$) is relaxed, so that the optimal policy may induce censoring streaks longer than those supported by the offline data (i.e., $n>n_{K,b}$ is possible). The proof follows the same three-step structure as Theorem~\ref{cor:ideal}: in Step~1 (PC-FQI) and Step~2 (C-FQI), we invoke the general form of Lemma~\ref{lm: regret decom} (which does \emph{not} rely on Assumption~\ref{assump:sufficient-coverage-main-text}) to decompose regret into (i) a \emph{learnable-class} term comparing the learned policy to the best policy in the data-supported policy class $\Pi_{n_{K,b}}$, plus (ii) an \emph{unidentifiable} remainder capturing the trajectories that produce more than $n_{K,b}$ consecutive censoring; we denote this by $\mathcal{E}_{\text{cov}}(K,n_{K,b})$. The learnable-class term is then bounded using the same way as in the idealized setting. Finally, Step~3 combines the two bounds via a union bound.

\subsection{Step 1: Regret for PC-FQI under the Relaxation of Assumption~\ref{assump:sufficient-coverage-main-text}}\label{Step 1: Regret for PC-FQI under the Relaxation of coverage}

In this section, we relax Assumption~\ref{assump:sufficient-coverage-main-text} (\emph{Sufficient Coverage}). Consequently, it is possible that $n > n_{K,b}$, meaning the optimal policy $\pi^*$ may induce censoring streaks longer than those supported by the offline dataset. In this case, we invoke Lemma~\ref{lm: regret decom}, which does not rely on Assumption~\ref{assump:sufficient-coverage-main-text}, together with the fact that the learned policy $\widetilde{\pi}_K \in \Pi_{n_{K,b}}$. Therefore, by Lemma~\ref{lm: regret decom}, the regret of $\widetilde{\pi}_K$ satisfies
\begin{sizeddisplay}
{\footnotesize}
 \begin{align}\label{eq: corr regret 1_app}
    \textup{\textbf{Regret}}(\widetilde{\pi}_K) 
    &\leq\left(\underbrace{\sup_{\pi' \in \Pi_{n_{K,b}}} \EE^{\pi'}\!\left[\sum_{t=0}^{\infty}\gamma^tR_t\right] -\EE^{\tilde{\pi}_K} \!\left[ \sum_{t=0}^{\infty} \gamma^t R_t  \right]}_{i(\tilde{\pi}_K)} 
    +2R_{\max}\frac{\gamma^{K+1}}{1-\gamma}\right)\prob^{\pi^*}(\mathcal{C}(K, n_{K,b}))  \nonumber\\
    &\quad +(K-n_{K,b}+1)(1-\alpha_{\pi^*,1})^{n_{K,b}}\left(\frac{2R_{\max}}{1-\gamma}\right)\nonumber\\
    &\quad+\frac{4R_{\max}\sqrt{K-1}}{1-\gamma}\omega_{NT}. 
\end{align}
\end{sizeddisplay}

Consider the term $i(\widetilde{\pi}_K)$:
\begin{align}\label{eq: pess relaxed coverage}
    i(\widetilde{\pi}_K)
    &=\sup_{\pi' \in \Pi_{n_{K,b}}} \EE^{\pi'}\!\left[\sum_{t=0}^{\infty}\gamma^tR_t\right] -\EE^{\tilde{\pi}_K} \!\left[ \sum_{t=0}^{\infty} \gamma^t R_t  \right]\nonumber\\
    &=\EE^{\pi^*_{n_{K,b}}}\!\left[\sum_{t=0}^{\infty}\gamma^tR_t\right] -\EE^{\tilde{\pi}_K} \!\left[ \sum_{t=0}^{\infty} \gamma^t R_t  \right],
\end{align}
where $\pi^*_{n_{K,b}} \in \argmax_{\pi' \in \Pi_{n_{K,b}}} \EE^{\pi'}\!\left[\sum_{t=0}^{\infty}\gamma^tR_t\right]$.

Observe that \(i(\widetilde{\pi}_K)\) has the same structure as the term analyzed in Section~\ref{PC-FQI under Idealized Setting} of the idealized setting (see Equation~\eqref{pesseq: fqi regret 1_app}), except that \(\pi^*\) is replaced by \(\pi^*_{n_{K,b}}\). Under Assumption~\ref{assump:sufficient-coverage-main-text}, we have \(\pi^* \in \Pi_n \subseteq \Pi_{n_{K,b}}\) and
\[
\pi^*_{n_{K,b}} \in \argmax_{\pi' \in \Pi_{n_{K,b}}} \EE^{\pi'}\!\left[\sum_{t=0}^{\infty}\gamma^t R_t\right].
\]
It follows that \(\pi^*\) is also optimal over \(\Pi_{n_{K,b}}\), and hence \(\pi^*\) and \(\pi^*_{n_{K,b}}\) are equivalent. However, with the relaxation of Assumption~\ref{assump:sufficient-coverage-main-text}, this no longer holds, and the equivalence between \(\pi^*\) and \(\pi^*_{n_{K,b}}\) fail. As a result, we have unlearnable part of $\pi^*$ which is the term $(K-n_{K,b}+1)(1-\alpha_{\pi^*,1})^{n_{K,b}}\left(\frac{2R_{\max}}{1-\gamma}\right)$ in Equation \eqref{eq: corr regret 1_app}.

However, Lemma~\ref{lm: performance difference} and the analysis of the idealized setting in Section~\ref{PC-FQI under Idealized Setting} require only that both policies lie in $\Pi_{n_{K,b}}$. This condition is satisfied here, as both $\pi^*_{n_{K,b}}$ and $\widetilde{\pi}_K$ in Equation~\eqref{eq: pess relaxed coverage} belong to $\Pi_{n_{K,b}}$. Therefore, the same bound applies. Specifically, by invoking Equation \eqref{eq: pess i_final_cov_simplified}, we have with probability at least $1-(n_{K,b}+3)\epsilon -\varepsilon$,
\begin{align}\label{eq: corr_i_bound}
i(\widetilde{\pi}_K)
\;\lesssim\;&\big(d+\log(1/\epsilon)\big)d
\sqrt{\frac{C_{\mathrm{cov}}\,}{\rho NT}}\;
+
C_5(\varepsilon)(NT)^{-\delta}
\end{align}

Additionally, we choose $K$ as in the idealized setting:
\begin{align}
K \;=\; \left\lceil \frac{\log(NT)}{2(1-\gamma)} \right\rceil.\nonumber
\end{align}
This implies that $\sqrt{K-1} < \sqrt{\frac{\log(NT)}{2(1-\gamma)}}$ and $\gamma^{K-1}\le (NT)^{-1/2}$ (up to a constant factor $\gamma^{-1}$). Therefore, the truncation term in Equation~\eqref{eq: corr regret 1_app} satisfies
\[
2R_{\max}\frac{\gamma^{K+1}}{1-\gamma}
\;\le\;
\frac{c_0\,R_{\max}}{1-\gamma}\cdot (NT)^{-1/2}
\]
for an absolute constant $c_0>0$. Again, to simplify further, we consider the same regime in which the truncation term is dominated by the statistical term (i.e., Equation~\eqref{eq: truncation pfqi})

Under this condition, the truncation contribution is subsumed by the leading statistical term in Equation~\eqref{eq: corr_i_bound}. Finally, for the sake of clarity, we assume without loss of generality that for the chosen $K$, there exists a constant $c_1 \in (0,1]$ such that $\prob^{\pi^\ast}\!\big(\mathcal{C}(K,n_{K,b})\big) \ge c_1$. This lower bound ensures that any additive term of order $\omega_{NT}$ can be absorbed into $\prob^{\pi^\ast}(\mathcal{C}(K,n_{K,b}))$ up to a constant factor $1/c_1$. This condition is imposed primarily to simplify the exposition.

Given these, we plug Equation \eqref{eq: corr_i_bound} into Equation \eqref{eq: corr regret 1_app} and this yields, with probability at least $1-(n_{K,b}+3)\epsilon -\varepsilon$,
\begin{sizeddisplay}
{\footnotesize}
 \begin{align}\label{eq: corr regret 1_plug}
    \textup{\textbf{Regret}}(\widetilde{\pi}_K) 
    &\lesssim\left(\big(d+\log(1/\epsilon)\big)d
\sqrt{\frac{C_{\mathrm{cov}}\,}{\rho NT}}\;
+
C_5(\varepsilon)(NT)^{-\delta}+\omega_{NT}\right)\prob^{\pi^*}(\mathcal{C}(K, n_{K,b}))  \nonumber\\
    &\quad +(K-n_{K,b}+1)(1-\alpha_{\pi^*,1})^{n_{K,b}}\left(\frac{2R_{\max}}{1-\gamma}\right),
\end{align}
\end{sizeddisplay}
where $\lesssim$ hides absolute constants and logarithmic factors, while keeping $\sqrt{\log(1/\epsilon)}$ explicit.

\subsection{Step 2: Regret for C-FQI under Regret for PC-FQI under the Relaxation of Assumption~\ref{assump:sufficient-coverage-main-text}}\label{Step 2: Regret for C-FQI under the Relaxation of coverage}
In this section, we continue to work in the setting where Assumption~\ref{assump:sufficient-coverage-main-text} (\emph{Sufficient Coverage}) is not relaxed. As before, it is possible that $n>n_{K,b}$, so the optimal policy $\pi^\ast$ may induce censoring streaks longer than those supported by the offline dataset. Since the learned policy $\widehat{\pi}_K\in\Pi_{n_{K,b}}$, we can again invoke Lemma~\ref{lm: regret decom} as it does not rely on sufficient coverage). Therefore, by Lemma~\ref{lm: regret decom}, the regret of $\widehat{\pi}_K$ can be written as
\begin{sizeddisplay}
{\footnotesize}
 \begin{align}\label{eq: corr regret 2_app}
    \textup{\textbf{Regret}}(\widehat{\pi}_K) 
    &\leq\left(\underbrace{\sup_{\pi' \in \Pi_{n_{K,b}}} \EE^{\pi'}\!\left[\sum_{t=0}^{\infty}\gamma^tR_t\right] -\EE^{\hat{\pi}_K} \!\left[ \sum_{t=0}^{\infty} \gamma^t R_t  \right]}_{i(\hat{\pi}_K)} 
    +2R_{\max}\frac{\gamma^{K+1}}{1-\gamma}\right)\prob^{\pi^*}(\mathcal{C}(K, n_{K,b}))  \nonumber\\
    &\quad +(K-n_{K,b}+1)(1-\alpha_{\pi^*,1})^{n_{K,b}}\left(\frac{2R_{\max}}{1-\gamma}\right)\nonumber\\
    &\quad+\frac{4R_{\max}\sqrt{K-1}}{1-\gamma}\omega_{NT}. 
\end{align}
\end{sizeddisplay}

Consider the term $i(\widehat{\pi}_K)$:
\begin{align}\label{eq: fqi relaxed coverage}
    i(\widehat{\pi}_K)
    &=\sup_{\pi' \in \Pi_{n_{K,b}}} \EE^{\pi'}\!\left[\sum_{t=0}^{\infty}\gamma^tR_t\right] -\EE^{\hat{\pi}_K} \!\left[ \sum_{t=0}^{\infty} \gamma^t R_t  \right]\nonumber\\
    &=\EE^{\pi^*_{n_{K,b}}}\!\left[\sum_{t=0}^{\infty}\gamma^tR_t\right] -\EE^{\hat{\pi}_K} \!\left[ \sum_{t=0}^{\infty} \gamma^t R_t  \right],
\end{align}
where $\pi^*_{n_{K,b}} \in \argmax_{\pi' \in \Pi_{n_{K,b}}} \EE^{\pi'}\!\left[\sum_{t=0}^{\infty}\gamma^tR_t\right]$.

Observe that \(i(\widehat{\pi}_K)\) has the same structure as the term analyzed in Step~\ref{C-FQI under Idealized Setting} of the idealized setting (see Equation~\eqref{eq: fqi regret 1_app}), except that \(\pi^*\) is replaced by \(\pi^*_{n_{K,b}}\). Under Assumption~\ref{assump:sufficient-coverage-main-text}, we have \(\pi^* \in \Pi_n \subseteq \Pi_{n_{K,b}}\) and
\[
\pi^*_{n_{K,b}} \in \argmax_{\pi' \in \Pi_{n_{K,b}}} \EE^{\pi'}\!\left[\sum_{t=0}^{\infty}\gamma^t R_t\right].
\]
It follows that \(\pi^*\) is also optimal over \(\Pi_{n_{K,b}}\), and hence \(\pi^*\) and \(\pi^*_{n_{K,b}}\) are equivalent. However, with the relaxation of Assumption~\ref{assump:sufficient-coverage-main-text}, this no longer holds, and the equivalence between \(\pi^*\) and \(\pi^*_{n_{K,b}}\) fail. As a result, we have unlearnable part of $\pi^*$ which is the term $(K-n_{K,b}+1)(1-\alpha_{\pi^*,1})^{n_{K,b}}\left(\frac{2R_{\max}}{1-\gamma}\right)$ in Equation \eqref{eq: corr regret 2_app}.

However, Lemma~\ref{lm: performance difference} and the analysis of the idealized setting in Section~\ref{C-FQI under Idealized Setting} require only that both policies lie in $\Pi_{n_{K,b}}$. This condition is satisfied here, as both $\pi^*_{n_{K,b}}$ and $\widehat{\pi}_K$ in Equation~\eqref{eq: fqi relaxed coverage} belong to $\Pi_{n_{K,b}}$. Therefore, the same bound applies. Specifically, by invoking Equation \eqref{eq:i_final_cov_simplified}, we have with probability at least $1-(n_{K,b}+3)\epsilon -\varepsilon$,
\begin{align}\label{eq: corr_i_hat_bound}
i(\widehat{\pi}_K)
\;\lesssim\;&d
\sqrt{\frac{\big(d+\log(1/\epsilon)\big)d}{\rho\,NT}}
\;+\;
C_5(\varepsilon)(NT)^{-\delta}
\end{align}

Additionally, we choose $K$ as in the idealized setting:
\begin{align}
K \;=\; \left\lceil \frac{\log(NT)}{2(1-\gamma)} \right\rceil.\nonumber
\end{align}
This implies that $\sqrt{K-1} < \sqrt{\frac{\log(NT)}{2(1-\gamma)}}$ and $\gamma^{K-1}\le (NT)^{-1/2}$ (up to a constant factor $\gamma^{-1}$). Therefore, the truncation term in Equation~\eqref{eq: corr regret 1_app} satisfies
\[
2R_{\max}\frac{\gamma^{K+1}}{1-\gamma}
\;\le\;
\frac{c_0\,R_{\max}}{1-\gamma}\cdot (NT)^{-1/2}
\]
for an absolute constant $c_0>0$. Again, to simplify further, we consider the same regime in which the truncation term is dominated by the statistical term (i.e., Equation \eqref{eq: truncation fqi}).

Under this condition, the truncation contribution is subsumed by the leading statistical term in Equation~\eqref{eq: corr_i_bound}. Finally, for the sake of clarity, we assume without loss of generality that for the chosen $K$, there exists a constant $c_1 \in (0,1]$ such that $\prob^{\pi^\ast}\!\big(\mathcal{C}(K,n_{K,b})\big) \ge c_1$. This lower bound ensures that any additive term of order $\omega_{NT}$ can be absorbed into $\prob^{\pi^\ast}(\mathcal{C}(K,n_{K,b}))$ up to a constant factor $1/c_1$. This condition is imposed primarily to simplify the exposition.

Given this, we plug Equation \eqref{eq: corr_i_hat_bound} into Equation \eqref{eq: corr regret 2_app} and this yields, with probability at least $1-(n_{K,b}+3)\epsilon-\varepsilon$,
\begin{sizeddisplay}
{\footnotesize}
 \begin{align}\label{eq: corr regret 2_plug}
    \textup{\textbf{Regret}}(\widehat{\pi}_K) 
    &\lesssim\left(d
\sqrt{\frac{\big(d+\log(1/\epsilon)\big)d}{\rho\,NT}}
\;+\;
C_5(\varepsilon)(NT)^{-\delta}+\omega_{NT}
    \right)\prob^{\pi^*}(\mathcal{C}(K, n_{K,b}))  \nonumber\\
    &\quad +(K-n_{K,b}+1)(1-\alpha_{\pi^*,1})^{n_{K,b}}\left(\frac{2R_{\max}}{1-\gamma}\right)
\end{align}
\end{sizeddisplay}

\subsection{Step 3: Combined Regret Bound under Relaxed Coverage}\label{subsec: Combined Regret Bound under Relaxed Coverage}

We now combine the results from the previous two steps to derive the final bound.
First, recall that under the event
$\widetilde{\Omega}_P$,
Equation~\eqref{eq: corr regret 1_plug} yields
\begin{sizeddisplay}
{\footnotesize}
 \begin{align}\label{eq: combined cor1 pfqi regret tilde}
   \textup{\textbf{Regret}}(\widetilde{\pi}_K) 
    &\lesssim\left(\big(d+\log(1/\epsilon)\big)d
\sqrt{\frac{C_{\mathrm{cov}}\,}{\rho NT}}\;
+
C_5(\varepsilon)(NT)^{-\delta}+\omega_{NT}\right)\prob^{\pi^*}(\mathcal{C}(K, n_{K,b}))  \nonumber\\
    &\quad +(K-n_{K,b}+1)(1-\alpha_{\pi^*,1})^{n_{K,b}}\left(\frac{2R_{\max}}{1-\gamma}\right)
\end{align}
\end{sizeddisplay}
which holds with probability at least $ 1-(n_{K,b}+3)\epsilon -\varepsilon$.

Similarly, under the event
$\widetilde{\Omega}$,
Equation~\eqref{eq: corr regret 2_plug} gives
\begin{sizeddisplay}
{\footnotesize}
 \begin{align}\label{eq: cor1 fqi combined regret hat}
   \textup{\textbf{Regret}}(\widehat{\pi}_K) 
    &\lesssim\left(d
\sqrt{\frac{\big(d+\log(1/\epsilon)\big)d}{\rho\,NT}}
\;+\;
C_5(\varepsilon)(NT)^{-\delta}+\omega_{NT}
    \right)\prob^{\pi^*}(\mathcal{C}(K, n_{K,b}))  \nonumber\\
    &\quad +(K-n_{K,b}+1)(1-\alpha_{\pi^*,1})^{n_{K,b}}\left(\frac{2R_{\max}}{1-\gamma}\right)
\end{align}
\end{sizeddisplay}
which holds with probability at least $ 1-(n_{K,b}+3)\epsilon -\varepsilon$.

Recall the definition of the simultaneous event from Equation~\eqref{eq: simultaneous event}
\begin{align*}
\Omega_{\mathrm{all}}
\;:=\;
\widetilde{\Omega}_P \cap \widetilde{\Omega}.
\end{align*}
By a union bound,
\begin{align}
\Pr(\Omega_{\mathrm{all}})
\;\ge\;
1
-
\Pr(\widetilde{\Omega}_P^{c})
-
\Pr(\widetilde{\Omega}^{c})
\;\ge\;
1-2\big((n_{K,b}+3)\epsilon+\varepsilon\big),
\end{align}
and on $\Omega_{\mathrm{all}}$ both regret bounds in
Equations~\eqref{eq: combined cor1 pfqi regret tilde} and~\eqref{eq: cor1 fqi combined regret hat}
hold simultaneously.

Furthermore, recall the definition of statistical error term in the idealized setting from Equation~\eqref{eq: def stat error} (i.e., $\mathcal{E}_{\text{stat}}(NT, d)$) as
\begin{align}
\mathcal{E}_{\text{stat}}(NT, d) :=
\max\left(
\big(d+\log(1/\epsilon)\big)d
\sqrt{\frac{C_{\mathrm{cov}}}{\rho NT}},\ 
d
\sqrt{\frac{\big(d+\log(1/\epsilon)\big)d}{\rho\,NT}}
\right)
+
C_5(\varepsilon)(NT)^{-\delta}
+
\omega_{NT}
\end{align}

Analogous to this, we define the unidentifiable regret term
\begin{align}\label{eq: def Ecov}
\mathcal{E}_{\text{cov}}(K, n_{K,b})
\;:=\;
(K-n_{K,b}+1)(1-\alpha_{\pi^*,1})^{n_{K,b}}\left(\frac{2R_{\max}}{1-\gamma}\right).
\end{align}

Then, on $\Omega_{\mathrm{all}}$,
\begin{align}\label{eq: max regret final cor1}
\max\!\big(\textup{\textbf{Regret}}(\widehat{\pi}_K),\ \textup{\textbf{Regret}}(\widetilde{\pi}_K)\big)
\;\lesssim\;
\mathcal{E}_{\text{stat}}(NT, d)\prob^{\pi^*}\!\big(\mathcal{C}(K,n_{K,b})\big)+\mathcal{E}_{\text{cov}}(K, n_{K,b}),
\end{align}
which holds with probability at least
$1-2\big((n_{K,b}+3)\epsilon+\varepsilon\big)$. This concludes the proof of Corollary~\ref{cor:relaxed_coverage}.

\section{Proof of Corollary~\ref{cor:relaxed-TAD}}\label{app:proof-relaxed-TAD}

This section proves Corollary~\ref{cor:relaxed-TAD} in the setting where Assumption~\ref{ass:TAD} (termination-action alignment) is relaxed while the remaining idealized assumptions stay in force. The argument mirrors the proof structure of Theorem~\ref{cor:ideal} and proceeds in three steps. In Step~1 (PC-FQI) and Step~2 (C-FQI), we begin from the same reduced regret decomposition provided by Lemma~\ref{lm: regret decom} under Assumptions~\ref{assump:bounded-n} and~\ref{assump:sufficient-coverage-main-text}, and we upper bound the resulting learnable-class term via Lemma~\ref{lm: performance difference} and the bounds on Q-functions. The sole change relative to the idealized proof is that, without Assumption~\ref{ass:TAD}, the termination-discrepancy contribution cannot be dropped; it is explicitly retained through the \emph{termination alignment penalty} $\mathrm{TAP}(\pi)$, while the remaining components are controlled exactly as before and yield the same term $\mathcal{E}_{\text{stat}}(NT,d)$. In Step~3, we combine two bounds to obtain the final result.

\subsection{Step 1: Regret for PC-FQI under the Relaxation of Assumption~\ref{ass:TAD}}\label{Step 1: Regret for PC-FQI under the Relaxation of tad}

In this section, we relax Assumption~\ref{ass:TAD} (\emph{Termination-Action Domain Validity}), while all other assumptions remain in force.
In particular, Assumptions~\ref{assump:bounded-n} and~\ref{assump:sufficient-coverage-main-text} hold, and by construction $\widetilde{\pi}_K \in \Pi_{n_{K,b}}$.
Therefore, we may invoke the reduced form of Lemma~\ref{lm: regret decom} that is valid under Assumptions~\ref{assump:bounded-n} and~\ref{assump:sufficient-coverage-main-text}. This is exactly the same decomposition used at the beginning of the proof of Theorem~\ref{cor:ideal} in Section~\ref{app:proof-ideal} (see the decomposition corresponding to Equation~\eqref{eq:general regret decomposition with coverage} in the proof of Lemma~\ref{lm: regret decom}).
\begin{sizeddisplay}
{\footnotesize}
 \begin{align}\label{eq: tadrel_regret_decomp_pc}
    \textup{\textbf{Regret}}(\widetilde{\pi}_K)
    &\leq
    \underbrace{
    \EE^{\pi^*} \left[ \sum_{t=0}^{\infty} \gamma^t R_t \right]
    -\mathbb{E}^{\tilde{\pi}_K}\left[ \sum_{t=0}^{\infty} \gamma^t R_t \right]
    }_{\textstyle i(\widetilde{\pi}_K)}
    \;+\;
    \frac{4R_{\max}\sqrt{K-1}}{1-\gamma}\omega_{NT}.
\end{align}
\end{sizeddisplay}
Thus, it remains to upper bound the term $i(\widetilde{\pi}_K)$ while accounting for the relaxation of Assumption~\ref{ass:TAD}.

Since Assumptions~\ref{assump:bounded-n} and \ref{assump:sufficient-coverage-main-text} holds in this section, the global optimal policy satisfies $\pi^* \in \Pi_n \subseteq \Pi_{n_{K,b}}$. Since $\pi^*$ maximizes the value globally and is a valid candidate within the restricted class $\Pi_{n_{K,b}}$, it must coincide with the $\pi^{\ast}_{n_{K,b}}$ where  $\pi^{\ast}_{n_{K,b}}\in \argmax_{\pi' \in \Pi_{n_{K,b}}}\EE^{\pi'}\!\left[\sum_{t=0}^{\infty}\gamma^tR_t\right]$. Thus, $\pi^{\ast}_{n_{K,b}} = \pi^*$. Given this and the fact that $\pi^{\ast}_{n_{K,b}}$ and $\widetilde{\pi}_K$ both belong to $\Pi_{n_{K,b}}$, Lemma~\ref{lm: performance difference} yields
\begin{sizeddisplay}
{\footnotesize}
\begin{align}\label{eq: tadrel_i_start}
    i(\widetilde{\pi}_K) &=
    \EE^{\pi^{\ast}}\left[\sum_{t=0}^{\infty}\gamma^tR_t\right]
    -\EE^{\tilde{\pi}_K}\left[\sum_{t=0}^{\infty}\gamma^tR_t\right]\nonumber\\
    &=
    \EE^{\pi^{\ast}_{n_{K,b}}}\left[\sum_{t=0}^{\infty}\gamma^tR_t\right]
    -\EE^{\tilde{\pi}_K}\left[\sum_{t=0}^{\infty}\gamma^tR_t\right]\nonumber\\
    &=
    \frac{1-\gamma^{n_{K,b}}}{(1-\gamma)^2}
    \EE_{(W^{(0)},A^{(0)},\cdots,W^{(I)}, A^{(I)}) \sim d_{\nu,I}^{\tilde{\pi}_K} \times \tilde{\pi}_K}
    \Bigg[
    -A^{(I)}_{\pi^{\ast}_{n_{K,b}}}(W^{(0)},A^{(0)},\cdots,W^{(I)},A^{(I)})
    \prod_{j=1}^{I}\mathbb{I}[\Delta^{(j)}=0]\,
    \mathbb{I}[\Delta^{(0)}=1]
    \Bigg]\nonumber\\
    &\leq
    \frac{1-\gamma^{n_{K,b}}}{(1-\gamma)^2}
    \EE_{(W^{(0)},A^{(0)},\cdots,W^{(I)}, A^{(I)}) \sim d_{\nu,I}^{\tilde{\pi}_K}}
    \Bigg[
    \Big(
    \EE^{\pi^{\ast}_{n_{K,b}}}\!\big[\widetilde{\textup{UQ}}(W^{(0)},A^{(0)},\cdots,W^{(I)},A^{(I)})\mid (W^{(0)},A^{(0)},\cdots,W^{(I)})\big] \nonumber\\
    &\hspace{6em}
    +
    \EE^{\tilde{\pi}_K}\!\big[\textup{TAD}^{\tilde{\pi}_K}(W^{(0)},A^{(0)},\cdots,W^{(I)},A^{(I)})\mid (W^{(0)},A^{(0)},\cdots,W^{(I)})\big]+
    2\gamma^{K-1}\Big(\frac{2-\gamma}{1-\gamma}R_{\max}\Big)\nonumber\\
    &\hspace{6em}
    +\frac{2}{1-\gamma
    }C_5(\varepsilon)(NT)^{-\delta} 
    +\textup{AD}^{\tilde{\pi}_K}(W^{(0)},A^{(0)},\cdots,W^{(I)})\mathbb{I}[I=n_{K,b}]
    \Big)\prod_{j=1}^{I}\mathbb{I}[\Delta^{(j)}=0]\,
    \mathbb{I}[\Delta^{(0)}=1]
    \Bigg],
\end{align}
\end{sizeddisplay}
Observe that this is exactly the same bound as Equation~\eqref{eq: pes bound on term (i)} from Section~\ref{PC-FQI under Idealized Setting}.

Given Equation \eqref{eq: tadrel_i_start}, the alignment-related contributions are precisely
\begin{sizeddisplay}
{\scriptsize}
\begin{align*}
    \mathrm{TAP}(\widetilde \pi_K)
&:=
\frac{1-\gamma^{n_{K,b}}}{(1-\gamma)^2}
\EE_{H^{(I)} \sim d_{\nu,I}^{\tilde \pi_K}}
\!\left[
\EE^{\tilde \pi_K}\!\left[\textup{TAD}^{\tilde \pi_K}(W^{(0)},A^{(0)},\cdots,W^{(I)},A^{(I)}) \mid (W^{(0)},A^{(0)},\cdots,W^{(I)})\right]
\prod_{j=1}^{I}\mathbb{I}[\Delta^{(j)}=0]\,
\mathbb{I}[\Delta^{(0)}=1]
\right],
\end{align*}
\end{sizeddisplay}
and
\begin{sizeddisplay}
{\footnotesize}
\begin{align*}
   \mathrm{BAP}(\widetilde \pi_K)
&:=
\frac{1-\gamma^{n_{K,b}}}{(1-\gamma)^2}
\EE_{H^{(I)} \sim d_{\nu,I}^{\tilde \pi_K}}
\!\left[
\textup{AD}^{\tilde \pi_K}(W^{(0)},A^{(0)},\cdots,W^{(I)})\,\mathbb{I}[I=n_{K,b}]
\right]. 
\end{align*}
\end{sizeddisplay}

We first note that Assumption~\ref{ass:AD} continues to hold, and therefore the boundary-action optimality gap vanishes exactly as in the idealized case (i.e., Equation~\eqref{eq: alingment penalties pess} in Section~\ref{app:proof-ideal}). In particular, under  Assumption~\ref{ass:AD}, we have
\begin{sizeddisplay}
{\footnotesize}
\begin{align}
    &\textup{AD}^{\hat{\pi}_K}(w^{(0)},a^{(0)},\cdots,w^{(n_{K,b})}) \\
    &=\max_{a\in\calA}\widetilde{Q}_{K}^{(n_{K,b})}(w^{(0)},a^{(0)},\cdots,w^{(n_{K,b})},a) - \widetilde{Q}_{K}^{(n_{K,b})}(w^{(0)},a^{(0)},\cdots,w^{(n_{K,b})},\widetilde{\pi}_K^{(n_{K,b})}(w^{(n_{K,b})},\cdots,w^{(0)},a^{(0)}))\nonumber\\ \nonumber
    &=0
\end{align}
\end{sizeddisplay} 
which implies
\begin{align}\label{eq: tadrel_BAP_zero_pc}
\mathrm{BAP}(\widetilde{\pi}_K)=0.
\end{align}

Next, recall that in the idealized proof, Assumption~\ref{ass:TAD} was used to guarantee that the termination action selected by $\widetilde{\pi}^{(n_{K,b})}_k$ lies in $\textup{UP}^{(n_{K,b})}$, implying that the boundary misalignment is non-positive.
With Assumption~\ref{ass:TAD} relaxed, we can no longer assert this sign, and thus the term $\mathrm{TAP}(\widetilde{\pi}_K)$ must be retained.

Using Equations \eqref{eq: tadrel_i_start} and \eqref{eq: tadrel_BAP_zero_pc} and keeping the $\mathrm{TAP}(\widetilde{\pi}_K)$ contribution yields
\begin{sizeddisplay}
{\footnotesize}
\begin{align}\label{eq: tadrel_i_split_pc}
    i(\widetilde{\pi}_K)
    &\leq
    \frac{1-\gamma^{n_{K,b}}}{(1-\gamma)^2}
    \EE_{H^{(I)} \sim d_{\nu,I}^{\tilde{\pi}_K}}
    \Bigg[
    \Big(
    \EE^{\pi^{\ast}_{n_{K,b}}}\left[\widetilde{\text{UQ}}(W^{(0)},A^{(0)},\cdots W^{(I)},A^{(I)})\given (W^{(0)},A^{(0)},\cdots W^{(I)})\right]
    +2\gamma^{K-1}\Big(\frac{2-\gamma}{1-\gamma}R_{\max}\Big)
    \nonumber\\
    &+\frac{2}{1-\gamma
    }C_5(\varepsilon)(NT)^{-\delta} 
    \Big)
    \prod_{j=1}^{I}\mathbb{I}[\Delta^{(j)}=0]\,
    \mathbb{I}[\Delta^{(0)}=1]
    \Bigg]
    \;+\;
    \mathrm{TAP}(\widetilde{\pi}_K).
\end{align}
\end{sizeddisplay}

Observe that the first term on the right-hand side of \eqref{eq: tadrel_i_split_pc} (i.e., the part excluding $\mathrm{TAP}(\widetilde{\pi}_K)$) is exactly the same object bounded in the idealized proof after imposing $\mathrm{BAP}(\widetilde{\pi}_K)=0$ and dropping the $\mathrm{TAP}(\widetilde{\pi}_K)$ contribution (see Equation~\eqref{eq: pess bound on term (i) 2nd}). Therefore, the same bound as in Equation~\eqref{eq: pess i_final_cov_simplified} from Section~\ref{PC-FQI under Idealized Setting} applies directly. Hence, with probability at least $1-(n_{K,b}+3)\epsilon -\varepsilon$,
\begin{sizeddisplay}
{\footnotesize}
\begin{align}\label{eq: tadrel_i_final_pc}
i(\widetilde{\pi}_K)
\;\lesssim\;&\big(d+\log(1/\epsilon)\big)d
\sqrt{\frac{C_{\mathrm{cov}}\,}{\rho NT}}\;
+
C_5(\varepsilon)(NT)^{-\delta}
    \;+\;
    \mathrm{TAP}(\widetilde{\pi}_K).
\end{align}
\end{sizeddisplay}
Combining Equation \eqref{eq: tadrel_regret_decomp_pc} with Equation \eqref{eq: tadrel_i_final_pc} and keeping the same choice of $K$ as in the idealized setting (i.e., Equation~\eqref{eq:K_choice_cov}) yield that, with probability at least $1-(n_{K,b}+3)\epsilon -\varepsilon$,
\begin{sizeddisplay}
{\footnotesize}
 \begin{align}\label{eq: tadrel_regret_final_pc}
    \textup{\textbf{Regret}}(\widetilde{\pi}_K)
    &\lesssim
   \big(d+\log(1/\epsilon)\big)d
\sqrt{\frac{C_{\mathrm{cov}}\,}{\rho NT}}\;
+
C_5(\varepsilon)(NT)^{-\delta}
    \;+\;
    \mathrm{TAP}(\widetilde{\pi}_K)+\omega_{NT}.
\end{align}
\end{sizeddisplay}
Observe that this is exactly the same bound derived in Equation~\eqref{eq: pess final regret} in the proof of Theorem~\ref{cor:ideal} with an additional term $\mathrm{TAP}(\widetilde{\pi}_K)$ as a result of the relaxation of Assumption~\ref{ass:TAD}

\subsection{Step 2: Regret for C-FQI under the Relaxation of Assumption~\ref{ass:TAD}}\label{Step 2: Regret for C-FQI under the Relaxation of tad}

In this section, we derive the analogous regret bound for $\widehat{\pi}_K$ under the same relaxation of Assumption~\ref{ass:TAD} (\emph{Termination-Action Domain Validity}), while all other assumptions remain in force.
In particular, Assumptions~\ref{assump:bounded-n} and~\ref{assump:sufficient-coverage-main-text} hold, and by construction $\widehat{\pi}_K \in \Pi_{n_{K,b}}$.
Therefore, we may again invoke the reduced form of Lemma~\ref{lm: regret decom} that is valid under Assumptions~\ref{assump:bounded-n} and~\ref{assump:sufficient-coverage-main-text}. This is exactly the same decomposition used at the beginning of the proof of Theorem~\ref{cor:ideal} in Section~\ref{PC-FQI under Idealized Setting} (see the decomposition corresponding to Equation~\eqref{eq:general regret decomposition with coverage} in the proof of Lemma~\ref{lm: regret decom}).
\begin{sizeddisplay}
{\footnotesize}
 \begin{align}\label{eq: tadrel_regret_decomp_c}
    \textup{\textbf{Regret}}(\widehat{\pi}_K)
    &\leq
    \underbrace{
    \EE^{\pi^*}\left[\sum_{t=0}^{\infty}\gamma^tR_t\right]
    -\EE^{\hat{\pi}_K}\left[\sum_{t=0}^{\infty}\gamma^tR_t\right]
    }_{\textstyle i(\widehat{\pi}_K)}
    \;+\;
    \frac{4R_{\max}\sqrt{K-1}}{1-\gamma}\omega_{NT}.
\end{align}
\end{sizeddisplay}
Thus, it remains to upper bound the term $i(\widehat{\pi}_K)$ while accounting for the relaxation of Assumption~\ref{ass:TAD}.

Since Assumptions~\ref{assump:bounded-n} and~\ref{assump:sufficient-coverage-main-text} hold in this section, the global optimal policy satisfies $\pi^\ast\in \Pi_n\subseteq \Pi_{n_{K,b}}$.
Since $\pi^\ast$ maximizes the value globally and is a valid candidate within the restricted class $\Pi_{n_{K,b}}$, it coincides with $\pi^\ast_{n_{K,b}}$, where
$\pi^\ast_{n_{K,b}}\in \argmax_{\pi'\in\Pi_{n_{K,b}}}\EE^{\pi'}[\sum_{t=0}^{\infty}\gamma^tR_t]$. Thus, $\pi^\ast_{n_{K,b}}=\pi^\ast$.
Given this and the fact that $\pi^\ast_{n_{K,b}}$ and $\widehat{\pi}_K$ both belong to $\Pi_{n_{K,b}}$, Lemma~\ref{lm: performance difference} yields
\begin{sizeddisplay}
{\footnotesize}
\begin{align}\label{eq: tadrel_i_start_c}
    i(\widehat{\pi}_K)
    &=
    \EE^{\pi^{\ast}}\left[\sum_{t=0}^{\infty}\gamma^tR_t\right]
    -\EE^{\hat{\pi}_K}\left[\sum_{t=0}^{\infty}\gamma^tR_t\right]\nonumber\\
    &=
    \EE^{\pi^{\ast}_{n_{K,b}}}\left[\sum_{t=0}^{\infty}\gamma^tR_t\right]
    -\EE^{\hat{\pi}_K}\left[\sum_{t=0}^{\infty}\gamma^tR_t\right]\nonumber\\
    &=
    \frac{1-\gamma^{n_{K,b}}}{(1-\gamma)^2}
    \EE_{(W^{(0)},A^{(0)},\cdots,W^{(I)},A^{(I)})\sim d_{\nu,I}^{\hat{\pi}_K}\times \hat{\pi}_K}
    \Bigg[
    -A^{(I)}_{\pi^\ast_{n_{K,b}}}(W^{(0)},A^{(0)},\cdots,W^{(I)},A^{(I)})
    \prod_{j=1}^{I}\mathbb{I}[\Delta^{(j)}=0]\,
    \mathbb{I}[\Delta^{(0)}=1]
    \Bigg]\nonumber\\
    &\le
    \frac{1-\gamma^{n_{K,b}}}{(1-\gamma)^2}
    \EE_{(W^{(0)},A^{(0)},\cdots,W^{(I)},A^{(I)})\sim d_{\nu,I}^{\hat{\pi}_K}}
    \Bigg[
    \Big(
    \EE^{\pi^\ast_{n_{K,b}}}\!\big[\textup{UQ}(W^{(0)},A^{(0)},\cdots,W^{(I)},A^{(I)})\mid (W^{(0)},A^{(0)},\cdots,W^{(I)})\big]\nonumber\\
    &\hspace{6em}
    +
    \EE^{\hat{\pi}_K}\!\big[\textup{TAD}^{\hat{\pi}_K}(W^{(0)},A^{(0)},\cdots,W^{(I)},A^{(I)})\mid (W^{(0)},A^{(0)},\cdots,W^{(I)})\big]
    +C^{(I)}(W^{(0)},A^{(0)},\cdots,W^{(I)})\nonumber\\
    &\hspace{6em}
    +2\gamma^{K-1}\Big(\frac{2-\gamma}{1-\gamma}R_{\max}\Big)
    +\frac{2}{1-\gamma
    }C_5(\varepsilon)(NT)^{-\delta}\nonumber\\
    &\hspace{6em}
    +\textup{AD}^{\hat{\pi}_K}(W^{(0)},A^{(0)},\cdots,W^{(I)})\mathbb{I}[I=n_{K,b}]
    \Big)
    \prod_{j=1}^{I}\mathbb{I}[\Delta^{(j)}=0]\,
    \mathbb{I}[\Delta^{(0)}=1]
    \Bigg].
\end{align}
\end{sizeddisplay}
Observe that this is exactly the same bound as Equation~\eqref{eq: bound on term (i)} from Section~\ref{C-FQI under Idealized Setting}.

Given Equation~\eqref{eq: tadrel_i_start_c}, the alignment-related contributions are precisely
\begin{sizeddisplay}
{\scriptsize}
\begin{align*}
    \mathrm{TAP}(\widehat{\pi}_K)
&:=
\frac{1-\gamma^{n_{K,b}}}{(1-\gamma)^2}
\EE_{H^{(I)} \sim d_{\nu,I}^{\hat{\pi}_K}}
\!\left[
\EE^{\hat{\pi}_K}\!\left[\textup{TAD}^{\hat{\pi}_K}(W^{(0)},A^{(0)},\cdots,W^{(I)},A^{(I)}) \mid (W^{(0)},A^{(0)},\cdots,W^{(I)})\right]
\prod_{j=1}^{I}\mathbb{I}[\Delta^{(j)}=0]\,
\mathbb{I}[\Delta^{(0)}=1]
\right],
\end{align*}
\end{sizeddisplay}
and
\begin{sizeddisplay}
{\footnotesize}
\begin{align*}
   \mathrm{BAP}(\widehat{\pi}_K)
&:=
\frac{1-\gamma^{n_{K,b}}}{(1-\gamma)^2}
\EE_{H^{(I)} \sim d_{\nu,I}^{\hat{\pi}_K}}
\!\left[
\textup{AD}^{\hat{\pi}_K}(W^{(0)},A^{(0)},\cdots,W^{(I)})\,\mathbb{I}[I=n_{K,b}]
\right]. 
\end{align*}
\end{sizeddisplay}

We first note that Assumption~\ref{ass:AD} continues to hold, and therefore the boundary-action optimality gap vanishes exactly as in the idealized case (i.e., Equation~\eqref{eq: alingment penalties fqi} in Section~\ref{app:proof-ideal}). In particular, under Assumption~\ref{ass:AD}, we have
\begin{sizeddisplay}
{\footnotesize}
\begin{align}
    &\textup{AD}^{\hat{\pi}_K}(w^{(0)},a^{(0)},\cdots,w^{(n_{K,b})}) \\
    &=\max_{a\in\calA}\widehat{Q}_{K}^{(n_{K,b})}(w^{(0)},a^{(0)},\cdots,w^{(n_{K,b})},a)
    -\widehat{Q}_{K}^{(n_{K,b})}(w^{(0)},a^{(0)},\cdots,w^{(n_{K,b})},\widehat{\pi}_K^{(n_{K,b})}(w^{(n_{K,b})},\cdots,w^{(0)},a^{(0)}))\nonumber\\ \nonumber
    &=0,
\end{align}
\end{sizeddisplay}
which implies
\begin{align}\label{eq: tadrel_BAP_zero_c}
\mathrm{BAP}(\widehat{\pi}_K)=0.
\end{align}

Next, recall that in the idealized proof, Assumption~\ref{ass:TAD} was used to guarantee that the termination action selected by $\widehat{\pi}^{(n_{K,b})}_k$ lies in $\textup{UP}^{(n_{K,b})}$, implying that the boundary misalignment is non-positive.
With Assumption~\ref{ass:TAD} relaxed, we can no longer assert this sign, and thus the term $\mathrm{TAP}(\widehat{\pi}_K)$ must be retained.

Using Equations~\eqref{eq: tadrel_i_start_c} and~\eqref{eq: tadrel_BAP_zero_c} and keeping the $\mathrm{TAP}(\widehat{\pi}_K)$ contribution yields
\begin{sizeddisplay}
{\scriptsize}
\begin{align}\label{eq: tadrel_i_split_c}
    i(\widehat{\pi}_K)
    &\leq
    \frac{1-\gamma^{n_{K,b}}}{(1-\gamma)^2}
    \EE_{H^{(I)}\sim d_{\nu,I}^{\hat{\pi}_K}}
    \Bigg[
    \Big(
    \EE^{\pi^\ast_{n_{K,b}}}\!\left[\textup{UQ}(W^{(0)},A^{(0)},\cdots,W^{(I)},A^{(I)})\given (W^{(0)},A^{(0)},\cdots,W^{(I)})\right]
    +C^{(I)}(W^{(0)},A^{(0)},\cdots,W^{(I)})\nonumber\\
    &\hspace{6em}
    +2\gamma^{K-1}\Big(\frac{2-\gamma}{1-\gamma}R_{\max}\Big)
    +\frac{2}{1-\gamma
    }C_5(\varepsilon)(NT)^{-\delta}
    \Big)
    \prod_{j=1}^{I}\mathbb{I}[\Delta^{(j)}=0]\,
    \mathbb{I}[\Delta^{(0)}=1]
    \Bigg]
    \;+\;
    \mathrm{TAP}(\widehat{\pi}_K).
\end{align}
\end{sizeddisplay}

Observe that the first term on the right-hand side of Equation \eqref{eq: tadrel_i_split_c} (i.e., the part excluding $\mathrm{TAP}(\widehat{\pi}_K)$) is exactly the same object bounded in the idealized proof after imposing $\mathrm{BAP}(\widehat{\pi}_K)=0$ and dropping the $\mathrm{TAP}(\widehat{\pi}_K)$ contribution (see Equation~\eqref{eq: bound on term (i) 2nd} in Section~\ref{C-FQI under Idealized Setting}). Therefore, the same bound as in Equation~\eqref{eq:i_final_cov_simplified} from Section~\ref{C-FQI under Idealized Setting} applies directly. Hence, with probability at least $1-(n_{K,b}+3)\epsilon -\varepsilon$,
\begin{sizeddisplay}
{\footnotesize}
\begin{align}\label{eq: tadrel_i_final_c}
   i(\widehat{\pi}_K)
\;\lesssim\;&d
\sqrt{\frac{\big(d+\log(1/\epsilon)\big)d}{\rho\,NT}}
\;+\;
C_5(\varepsilon)(NT)^{-\delta}
    \;+\;
    \mathrm{TAP}(\widehat{\pi}_K).
\end{align}
\end{sizeddisplay}
Combining Equation~\eqref{eq: tadrel_regret_decomp_c} with Equation~\eqref{eq: tadrel_i_final_c} and keeping the same choice of $K$ as in the idealized setting (i.e., Equation~\eqref{eq:K_choice_cov}) yield that, with probability at least $1-(n_{K,b}+3)\epsilon -\varepsilon$,
\begin{sizeddisplay}
{\footnotesize}
 \begin{align}\label{eq: tadrel_regret_final_c}
    \textup{\textbf{Regret}}(\widehat{\pi}_K)
    &\lesssim
   d
\sqrt{\frac{\big(d+\log(1/\epsilon)\big)d}{\rho\,NT}}
\;+\;
C_5(\varepsilon)(NT)^{-\delta}
    \;+\;
    \mathrm{TAP}(\widehat{\pi}_K)+\omega_{NT}.
\end{align}
\end{sizeddisplay}
Observe that this is exactly the same bound derived in Equation~\eqref{eq: final regret} in the proof of Theorem~\ref{cor:ideal} with an additional term $\mathrm{TAP}(\widehat{\pi}_K)$ as a result of the relaxation of Assumption~\ref{ass:TAD}.

\subsection{Step 3: Combined Regret Bound under the Relaxation of Assumption~\ref{ass:TAD}}\label{Step 3: Combined Regret Bound under the Relaxation of tad}

We now combine the results of Section~\ref{Step 1: Regret for PC-FQI under the Relaxation of tad} and Section~\ref{Step 2: Regret for C-FQI under the Relaxation of tad} to obtain a simultaneous regret bound under the relaxed setting where Assumption~\ref{ass:TAD} (\emph{Termination-Action Domain Validity}) is not assumed, while all other assumptions remain in force.

Recall that in the idealized setting (Theorem~\ref{cor:ideal}), we defined the statistical error term as
\begin{align}\label{eq: tadrel_def_stat_error_ideal}
\mathcal{E}_{\text{stat}}(NT, d) :=
\max\left(
\big(d+\log(1/\epsilon)\big)d
\sqrt{\frac{C_{\mathrm{cov}}}{\rho NT}},\ 
d
\sqrt{\frac{\big(d+\log(1/\epsilon)\big)d}{\rho\,NT}}
\right)
+
C_5(\varepsilon)(NT)^{-\delta}
+
\omega_{NT}
\end{align}
We keep exactly the same definition here.

From Section~\ref{Step 1: Regret for PC-FQI under the Relaxation of tad} (Equation~\eqref{eq: tadrel_regret_final_pc}) and Section~\ref{Step 2: Regret for C-FQI under the Relaxation of tad} (Equation~\eqref{eq: tadrel_regret_final_c}), we have that with probability at least $1-(n_{K,b}+3)\epsilon -\varepsilon$,
\begin{align}\label{eq: tadrel_pc_compact}
 \textup{\textbf{Regret}}(\widetilde{\pi}_K)
    &\lesssim
   \big(d+\log(1/\epsilon)\big)d
\sqrt{\frac{C_{\mathrm{cov}}\,}{\rho NT}}\;
+
C_5(\varepsilon)(NT)^{-\delta}
    \;+\;
    \mathrm{TAP}(\widetilde{\pi}_K)+\omega_{NT}
\end{align}
and similarly,
\begin{align}\label{eq: tadrel_c_compact}
    \textup{\textbf{Regret}}(\widehat{\pi}_K)
    &\lesssim
   d
\sqrt{\frac{\big(d+\log(1/\epsilon)\big)d}{\rho\,NT}}
\;+\;
C_5(\varepsilon)(NT)^{-\delta}
    \;+\;
    \mathrm{TAP}(\widehat{\pi}_K)+\omega_{NT}..
\end{align}

Recall the definition of the simultaneous event $\Omega_{\mathrm{all}}$ from Equation \eqref{eq: simultaneous event}. By a union bound,
\begin{align}\label{eq: tadrel_union_bound}
\Pr(\Omega_{\mathrm{all}})
\;\ge\;
1-2\big((n_{K,b}+3)\epsilon+\varepsilon\big),
\end{align}
and on $\Omega_{\mathrm{all}}$ both bounds \eqref{eq: tadrel_pc_compact}--\eqref{eq: tadrel_c_compact}  hold simultaneously.
Therefore, 
\begin{align*}
\max\!\Big(\textup{\textbf{Regret}}(\widehat \pi_K),\ \textup{\textbf{Regret}}(\widetilde \pi_K)\Big)
&\lesssim
\mathcal{E}_{\text{stat}}(NT, d)
\;+\;
\max\!\Big(\mathrm{TAP}(\widehat{\pi}_K),\mathrm{TAP}(\widetilde{\pi}_K)\Big).
\end{align*}
which holds with probability at least $1-2\big((n_{K,b}+3)\epsilon+\varepsilon\big)$. This concludes the proof of Corollary~\ref{cor:relaxed-TAD}.

\section{Proof of Corollary~\ref{cor:relaxed-AD}}\label{app:proof-relaxed-AD}

This section proves Corollary~\ref{cor:relaxed-AD} in the setting where Assumption~\ref{ass:AD} is relaxed while all other idealized assumptions remain in force. The proof follows the same three-step template used in the idealized analysis. In Step~1 (PC-FQI) and Step~2 (C-FQI), we start from the same reduced regret decomposition from Lemma~\ref{lm: regret decom}, valid under Assumptions~\ref{assump:bounded-n} and~\ref{assump:sufficient-coverage-main-text}, and we control the learnable-class term via Lemma~\ref{lm: performance difference} together with bounds on Q-function estimates. Relative to the idealized proof, the only substantive change is at the censoring boundary: without Assumption~\ref{ass:AD}, the action $a_{\max}$ is not guaranteed to maximize the estimated $Q$-function when consecutive censoring is $n_{K,b}$, so the boundary optimality gap no longer vanishes as it was in idealized setting. This gap is isolated and carried through the analysis as the $\mathrm{BAP}(\pi)$. Consequently, both Step~1 and Step~2 yield regret bounds that match the term $\mathcal{E}_{\text{stat}}(NT,d)$ plus an additional additive $\mathrm{BAP}(\pi)$ term. In Step~3 we combine the two bounds to obtain the final result.

\subsection{Step 1: Regret for PC-FQI under the Relaxation of Assumption~\ref{ass:AD}}\label{Step 1: Regret for PC-FQI under the Relaxation of ad}

In this section, we relax Assumption~\ref{ass:AD}, while all other assumptions remain in force.
In particular, Assumptions~\ref{assump:bounded-n} and~\ref{assump:sufficient-coverage-main-text} hold, and by construction $\widetilde{\pi}_K \in \Pi_{n_{K,b}}$.
Therefore, we may invoke the reduced form of Lemma~\ref{lm: regret decom} that is valid under Assumptions~\ref{assump:bounded-n} and~\ref{assump:sufficient-coverage-main-text}. This is exactly the same decomposition used at the beginning of the proof of Theorem~\ref{cor:ideal} in Section~\ref{app:proof-ideal} (see the decomposition corresponding to Equation~\eqref{eq:general regret decomposition with coverage} in the proof of Lemma~\ref{lm: regret decom}).
\begin{sizeddisplay}
{\footnotesize}
 \begin{align}\label{eq: adrel_regret_decomp_pc}
    \textup{\textbf{Regret}}(\widetilde{\pi}_K)
    &\leq
    \underbrace{
    \EE^{\pi^*} \left[ \sum_{t=0}^{\infty} \gamma^t R_t \right]
    -\mathbb{E}^{\tilde{\pi}_K}\left[ \sum_{t=0}^{\infty} \gamma^t R_t \right]
    }_{\textstyle i(\widetilde{\pi}_K)}
    \;+\;
    \frac{4R_{\max}\sqrt{K-1}}{1-\gamma}\omega_{NT}.
\end{align}
\end{sizeddisplay}
Thus, it remains to upper bound the term $i(\widetilde{\pi}_K)$ while accounting for the relaxation of Assumption~\ref{ass:AD}.

Since Assumptions~\ref{assump:bounded-n} and \ref{assump:sufficient-coverage-main-text} holds in this section, the global optimal policy satisfies $\pi^* \in \Pi_n \subseteq \Pi_{n_{K,b}}$. Since $\pi^*$ maximizes the value globally and is a valid candidate within the restricted class $\Pi_{n_{K,b}}$, it must coincide with the $\pi^{\ast}_{n_{K,b}}$ where  $\pi^{\ast}_{n_{K,b}}\in \argmax_{\pi' \in \Pi_{n_{K,b}}}\EE^{\pi'}\!\left[\sum_{t=0}^{\infty}\gamma^tR_t\right]$. Thus, $\pi^{\ast}_{n_{K,b}} = \pi^*$. Given this and the fact that $\pi^{\ast}_{n_{K,b}}$ and $\widetilde{\pi}_K$ both belong to $\Pi_{n_{K,b}}$, Lemma~\ref{lm: performance difference} yields
\begin{sizeddisplay}
{\footnotesize}
\begin{align}\label{eq: adrel_i_start}
    i(\widetilde{\pi}_K) &=
    \EE^{\pi^{\ast}}\left[\sum_{t=0}^{\infty}\gamma^tR_t\right]
    -\EE^{\tilde{\pi}_K}\left[\sum_{t=0}^{\infty}\gamma^tR_t\right]\nonumber\\
    &=
    \EE^{\pi^{\ast}_{n_{K,b}}}\left[\sum_{t=0}^{\infty}\gamma^tR_t\right]
    -\EE^{\tilde{\pi}_K}\left[\sum_{t=0}^{\infty}\gamma^tR_t\right]\nonumber\\
    &=
    \frac{1-\gamma^{n_{K,b}}}{(1-\gamma)^2}
    \EE_{(W^{(0)},A^{(0)},\cdots,W^{(I)}, A^{(I)}) \sim d_{\nu,I}^{\tilde{\pi}_K} \times \tilde{\pi}_K}
    \Bigg[
    -A^{(I)}_{\pi^{\ast}_{n_{K,b}}}(W^{(0)},A^{(0)},\cdots,W^{(I)},A^{(I)})
    \prod_{j=1}^{I}\mathbb{I}[\Delta^{(j)}=0]\,
    \mathbb{I}[\Delta^{(0)}=1]
    \Bigg]\nonumber\\
    &\leq
    \frac{1-\gamma^{n_{K,b}}}{(1-\gamma)^2}
    \EE_{(W^{(0)},A^{(0)},\cdots,W^{(I)}, A^{(I)}) \sim d_{\nu,I}^{\tilde{\pi}_K}}
    \Bigg[
    \Big(
    \EE^{\pi^{\ast}_{n_{K,b}}}\!\big[\widetilde{\textup{UQ}}(W^{(0)},A^{(0)},\cdots,W^{(I)},A^{(I)})\mid (W^{(0)},A^{(0)},\cdots,W^{(I)})\big] \nonumber\\
    &\hspace{6em}
    +
    \EE^{\tilde{\pi}_K}\!\big[\textup{TAD}^{\tilde{\pi}_K}(W^{(0)},A^{(0)},\cdots,W^{(I)},A^{(I)})\mid (W^{(0)},A^{(0)},\cdots,W^{(I)})\big]+
    2\gamma^{K-1}\Big(\frac{2-\gamma}{1-\gamma}R_{\max}\Big)\nonumber\\
    &\hspace{6em}
    +\frac{2}{1-\gamma
    }C_5(\varepsilon)(NT)^{-\delta} 
    +\textup{AD}^{\tilde{\pi}_K}(W^{(0)},A^{(0)},\cdots,W^{(I)})\mathbb{I}[I=n_{K,b}]
    \Big)\prod_{j=1}^{I}\mathbb{I}[\Delta^{(j)}=0]\,
    \mathbb{I}[\Delta^{(0)}=1]
    \Bigg].
\end{align}
\end{sizeddisplay}
 Observe that this is exactly the same bound as Equation~\eqref{eq: pes bound on term (i)} from Section~\ref{PC-FQI under Idealized Setting}.

Given Equation \eqref{eq: tadrel_i_start}, the alignment-related contributions are precisely
\begin{sizeddisplay}
{\scriptsize}
\begin{align*}
    \mathrm{TAP}(\widetilde \pi_K)
&:=
\frac{1-\gamma^{n_{K,b}}}{(1-\gamma)^2}
\EE_{H^{(I)} \sim d_{\nu,I}^{\tilde \pi_K}}
\!\left[
\EE^{\tilde \pi_K}\!\left[\textup{TAD}^{\tilde \pi_K}(W^{(0)},A^{(0)},\cdots,W^{(I)},A^{(I)}) \mid (W^{(0)},A^{(0)},\cdots,W^{(I)})\right]
\prod_{j=1}^{I}\mathbb{I}[\Delta^{(j)}=0]\,
\mathbb{I}[\Delta^{(0)}=1]
\right],
\end{align*}
\end{sizeddisplay}
and
\begin{sizeddisplay}
{\footnotesize}
\begin{align*}
   \mathrm{BAP}(\widetilde \pi_K)
&:=
\frac{1-\gamma^{n_{K,b}}}{(1-\gamma)^2}
\EE_{H^{(I)} \sim d_{\nu,I}^{\tilde \pi_K}}
\!\left[
\textup{AD}^{\tilde \pi_K}(W^{(0)},A^{(0)},\cdots,W^{(I)})\,\mathbb{I}[I=n_{K,b}]
\right]. 
\end{align*}
\end{sizeddisplay}

We first note that Assumption~\ref{ass:TAD} continues to hold and it was used to guarantee that the termination action selected by $\widetilde{\pi}^{(n_{K,b})}_k$ lies in $\textup{UP}^{(n_{K,b})}$, implying that  $\mathrm{TAP}(\widetilde \pi_K)\leq 0$ (see Section~\ref{PC-FQI under Idealized Setting}).
With Assumption~\ref{ass:TAD} still in force, we can use this fact and drop the term $\mathrm{TAP}(\widetilde{\pi}_K)$ from Equation~\eqref{eq: adrel_i_start}.

However, due to the relaxation of Assumption~\ref{ass:AD}, the boundary-action optimality gap does not vanish as in the idealized case (i.e., Equation~\eqref{eq: alingment penalties pess} in Section~\ref{app:proof-ideal}). In particular, the relaxation of Assumption~\ref{ass:AD} leads to
\begin{sizeddisplay}
{\footnotesize}
\begin{align}
    &\textup{AD}^{\hat{\pi}_K}(w^{(0)},a^{(0)},\cdots,w^{(n_{K,b})}) \\
    &=\max_{a\in\calA}\widetilde{Q}_{K}^{(n_{K,b})}(w^{(0)},a^{(0)},\cdots,w^{(n_{K,b})},a) - \widetilde{Q}_{K}^{(n_{K,b})}(w^{(0)},a^{(0)},\cdots,w^{(n_{K,b})},\widetilde{\pi}_K^{(n_{K,b})}(w^{(n_{K,b})},\cdots,w^{(0)},a^{(0)}))\nonumber\\ \nonumber
    &\geq0
\end{align}
\end{sizeddisplay} 
This is because the action $a_{\max}$ which is outputted by $\widetilde{\pi}_K^{(n_{K,b})}$ is not necessarily the action that maximizes estimated Q-function $\widetilde{Q}_{K}^{(n_{K,b})}$ when there is $n_{K,b}$ consecutive censoring. As a result
\begin{align}\label{eq: adrel_BAP_nonzero_pc}
\mathrm{BAP}(\widetilde{\pi}_K) \geq 0.
\end{align}

Using Equations \eqref{eq: adrel_i_start} and \eqref{eq: adrel_BAP_nonzero_pc} yields
\begin{sizeddisplay}
{\footnotesize}
\begin{align}\label{eq: adrel_i_split_pc}
    i(\widetilde{\pi}_K)
    &\leq
    \frac{1-\gamma^{n_{K,b}}}{(1-\gamma)^2}
    \EE_{H^{(I)} \sim d_{\nu,I}^{\tilde{\pi}_K}}
    \Bigg[
    \Big(
    \EE^{\pi^{\ast}_{n_{K,b}}}\left[\widetilde{\text{UQ}}(W^{(0)},A^{(0)},\cdots W^{(I)},A^{(I)})\given (W^{(0)},A^{(0)},\cdots W^{(I)})\right]
    +2\gamma^{K-1}\Big(\frac{2-\gamma}{1-\gamma}R_{\max}\Big)
    \nonumber\\
    &+\frac{2}{1-\gamma
    }C_5(\varepsilon)(NT)^{-\delta} 
    \Big)
    \prod_{j=1}^{I}\mathbb{I}[\Delta^{(j)}=0]\,
    \mathbb{I}[\Delta^{(0)}=1]
    \Bigg]
    \;+\;
    \mathrm{BAP}(\widetilde{\pi}_K).
\end{align}
\end{sizeddisplay}

Observe that the first term on the right-hand side of \eqref{eq: adrel_i_split_pc} (i.e., the part excluding $\mathrm{BAP}(\widetilde{\pi}_K)$) is exactly the same object bounded in the idealized setting after imposing $\mathrm{BAP}(\widetilde{\pi}_K)=0$ and dropping the $\mathrm{TAP}(\widetilde \pi_K)$ contribution (see Equation~\eqref{eq: pess bound on term (i) 2nd}). Therefore, the same bound as in Equation~\eqref{eq: pess i_final_cov_simplified} from Section~\ref{PC-FQI under Idealized Setting} applies directly with an addition of the term  $\mathrm{BAP}(\widetilde{\pi}_K)$. Hence, with probability at least $1-(n_{K,b}+3)\epsilon -\varepsilon$,
\begin{sizeddisplay}
{\footnotesize}
\begin{align}\label{eq: adrel_i_final_pc}
   i(\widetilde{\pi}_K)
\;\lesssim\;&\big(d+\log(1/\epsilon)\big)d
\sqrt{\frac{C_{\mathrm{cov}}\,}{\rho NT}}\;
+
C_5(\varepsilon)(NT)^{-\delta}
    \;+\;
    \mathrm{BAP}(\widetilde{\pi}_K).
\end{align}
\end{sizeddisplay}
Combining Equation \eqref{eq: adrel_regret_decomp_pc} with Equation \eqref{eq: adrel_i_final_pc} and keeping the same choice of $K$ as in the idealized setting (i.e., Equation~\eqref{eq:K_choice_cov}) yield that, with probability at least $1-(n_{K,b}+3)\epsilon -\varepsilon$,
\begin{sizeddisplay}
{\footnotesize}
 \begin{align}\label{eq: adrel_regret_final_pc}
    \textup{\textbf{Regret}}(\widetilde{\pi}_K)
    &\lesssim \big(d+\log(1/\epsilon)\big)d
\sqrt{\frac{C_{\mathrm{cov}}\,}{\rho NT}}\;
+
C_5(\varepsilon)(NT)^{-\delta}
    \;+\;
    \mathrm{BAP}(\widetilde{\pi}_K)+\omega_{NT}.
\end{align}
\end{sizeddisplay}
Observe that this is exactly the same bound derived in Equation~\eqref{eq: pess final regret} in the proof of Theorem~\ref{cor:ideal} with an additional term $\mathrm{BAP}(\widetilde{\pi}_K)$ as a result of the relaxation of Assumption~\ref{ass:AD}.

\subsection{Step 2: Regret for C-FQI under the Relaxation of Assumption~\ref{ass:AD}}\label{Step 2: Regret for C-FQI under the Relaxation of ad}

In this section, we relax Assumption~\ref{ass:AD}, while all other assumptions remain in force.
In particular, Assumptions~\ref{assump:bounded-n} and~\ref{assump:sufficient-coverage-main-text} hold, and by construction $\widehat{\pi}_K \in \Pi_{n_{K,b}}$.
Therefore, we may invoke the reduced form of Lemma~\ref{lm: regret decom} that is valid under Assumptions~\ref{assump:bounded-n} and~\ref{assump:sufficient-coverage-main-text}. This is exactly the same decomposition used at the beginning of the proof of Theorem~\ref{cor:ideal} in Section~\ref{app:proof-ideal} (see the decomposition corresponding to Equation~\eqref{eq:general regret decomposition with coverage} in the proof of Lemma~\ref{lm: regret decom}).
\begin{sizeddisplay}
{\footnotesize}
 \begin{align}\label{eq: adrel_regret_decomp_c}
    \textup{\textbf{Regret}}(\widehat{\pi}_K)
    &\leq
    \underbrace{
    \EE^{\pi^*} \left[ \sum_{t=0}^{\infty} \gamma^t R_t \right]
    -\mathbb{E}^{\hat{\pi}_K}\left[ \sum_{t=0}^{\infty} \gamma^t R_t \right]
    }_{\textstyle i(\widehat{\pi}_K)}
    \;+\;
    \frac{4R_{\max}\sqrt{K-1}}{1-\gamma}\omega_{NT}.
\end{align}
\end{sizeddisplay}
Thus, it remains to upper bound the term $i(\widehat{\pi}_K)$ while accounting for the relaxation of Assumption~\ref{ass:AD}.

Since Assumptions~\ref{assump:bounded-n} and \ref{assump:sufficient-coverage-main-text} holds in this section, the global optimal policy satisfies $\pi^* \in \Pi_n \subseteq \Pi_{n_{K,b}}$. Since $\pi^*$ maximizes the value globally and is a valid candidate within the restricted class $\Pi_{n_{K,b}}$, it must coincide with the $\pi^{\ast}_{n_{K,b}}$ where  $\pi^{\ast}_{n_{K,b}}\in \argmax_{\pi' \in \Pi_{n_{K,b}}}\EE^{\pi'}\!\left[\sum_{t=0}^{\infty}\gamma^tR_t\right]$. Thus, $\pi^{\ast}_{n_{K,b}} = \pi^*$. Given this and the fact that $\pi^{\ast}_{n_{K,b}}$ and $\widehat{\pi}_K$ both belong to $\Pi_{n_{K,b}}$, Lemma~\ref{lm: performance difference} yields
\begin{sizeddisplay}
{\footnotesize}
\begin{align}\label{eq: adrel_i_start_c}
    i(\widehat{\pi}_K) &=
    \EE^{\pi^{\ast}}\left[\sum_{t=0}^{\infty}\gamma^tR_t\right]
    -\EE^{\hat{\pi}_K}\left[\sum_{t=0}^{\infty}\gamma^tR_t\right]\nonumber\\
    &=
    \EE^{\pi^{\ast}_{n_{K,b}}}\left[\sum_{t=0}^{\infty}\gamma^tR_t\right]
    -\EE^{\hat{\pi}_K}\left[\sum_{t=0}^{\infty}\gamma^tR_t\right]\nonumber\\
    &=
    \frac{1-\gamma^{n_{K,b}}}{(1-\gamma)^2}
    \EE_{(W^{(0)},A^{(0)},\cdots,W^{(I)}, A^{(I)}) \sim d_{\nu,I}^{\hat{\pi}_K} \times \hat{\pi}_K}
    \Bigg[
    -A^{(I)}_{\pi^{\ast}_{n_{K,b}}}(W^{(0)},A^{(0)},\cdots,W^{(I)},A^{(I)})
    \prod_{j=1}^{I}\mathbb{I}[\Delta^{(j)}=0]\,
    \mathbb{I}[\Delta^{(0)}=1]
    \Bigg]\nonumber\\
    &\leq
    \frac{1-\gamma^{n_{K,b}}}{(1-\gamma)^2}
    \EE_{(W^{(0)},A^{(0)},\cdots,W^{(I)}, A^{(I)}) \sim d_{\nu,I}^{\hat{\pi}_K}}
    \Bigg[
    \Big(
    \EE^{\pi^{\ast}_{n_{K,b}}}\!\big[\textup{UQ}(W^{(0)},A^{(0)},\cdots,W^{(I)},A^{(I)})\mid (W^{(0)},A^{(0)},\cdots,W^{(I)})\big] \nonumber\\
    &\hspace{6em}
    +
    \EE^{\hat{\pi}_K}\!\big[\textup{TAD}^{\hat{\pi}_K}(W^{(0)},A^{(0)},\cdots,W^{(I)},A^{(I)})\mid (W^{(0)},A^{(0)},\cdots,W^{(I)})\big]+
    2\gamma^{K-1}\Big(\frac{2-\gamma}{1-\gamma}R_{\max}\Big)\nonumber\\
    &\hspace{6em}
    +\frac{2}{1-\gamma
    }C_5(\varepsilon)(NT)^{-\delta} 
    +\textup{AD}^{\hat{\pi}_K}(W^{(0)},A^{(0)},\cdots,W^{(I)})\mathbb{I}[I=n_{K,b}]
    \Big)\prod_{j=1}^{I}\mathbb{I}[\Delta^{(j)}=0]\,
    \mathbb{I}[\Delta^{(0)}=1]
    \Bigg].
\end{align}
\end{sizeddisplay}
Observe that this is exactly the same bound as the corresponding inequality used in the C-FQI part of the idealized proof in Section~\ref{C-FQI under Idealized Setting}.

Given Equation \eqref{eq: adrel_i_start_c}, the alignment-related contributions are precisely
\begin{sizeddisplay}
{\scriptsize}
\begin{align*}
    \mathrm{TAP}(\widehat \pi_K)
&:=
\frac{1-\gamma^{n_{K,b}}}{(1-\gamma)^2}
\EE_{H^{(I)} \sim d_{\nu,I}^{\hat \pi_K}}
\!\left[
\EE^{\hat \pi_K}\!\left[\textup{TAD}^{\hat \pi_K}(W^{(0)},A^{(0)},\cdots,W^{(I)},A^{(I)}) \mid (W^{(0)},A^{(0)},\cdots,W^{(I)})\right]
\prod_{j=1}^{I}\mathbb{I}[\Delta^{(j)}=0]\,
\mathbb{I}[\Delta^{(0)}=1]
\right],
\end{align*}
\end{sizeddisplay}
and
\begin{sizeddisplay}
{\footnotesize}
\begin{align*}
   \mathrm{BAP}(\widehat \pi_K)
&:=
\frac{1-\gamma^{n_{K,b}}}{(1-\gamma)^2}
\EE_{H^{(I)} \sim d_{\nu,I}^{\hat \pi_K}}
\!\left[
\textup{AD}^{\hat \pi_K}(W^{(0)},A^{(0)},\cdots,W^{(I)})\,\mathbb{I}[I=n_{K,b}]
\right]. 
\end{align*}
\end{sizeddisplay}

We first note that Assumption~\ref{ass:TAD} continues to hold and it was used to guarantee that the termination action selected by $\widehat{\pi}^{(n_{K,b})}_k$ lies in $\textup{UP}^{(n_{K,b})}$, implying that $\mathrm{TAP}(\widehat \pi_K)\leq 0$ (see the C-FQI discussion in the idealized setting in Section~\ref{C-FQI under Idealized Setting}).
With Assumption~\ref{ass:TAD} still in force, we can use this fact and drop the term $\mathrm{TAP}(\widehat{\pi}_K)$ from Equation~\eqref{eq: adrel_i_start_c}.

However, due to the relaxation of Assumption~\ref{ass:AD}, the boundary-action optimality gap does not vanish as in the idealized case. In particular, the relaxation of Assumption~\ref{ass:AD} leads to
\begin{sizeddisplay}
{\footnotesize}
\begin{align}
    &\textup{AD}^{\hat{\pi}_K}(w^{(0)},a^{(0)},\cdots,w^{(n_{K,b})}) \\
    &=\max_{a\in\calA}\widehat{Q}_{K}^{(n_{K,b})}(w^{(0)},a^{(0)},\cdots,w^{(n_{K,b})},a) - \widehat{Q}_{K}^{(n_{K,b})}(w^{(0)},a^{(0)},\cdots,w^{(n_{K,b})},\widehat{\pi}_K^{(n_{K,b})}(w^{(n_{K,b})},\cdots,w^{(0)},a^{(0)}))\nonumber\\
    &\geq 0.\nonumber
\end{align}
\end{sizeddisplay} 
This is because the action $a_{\max}$ which is outputted by $\widehat{\pi}_K^{(n_{K,b})}$ is not necessarily the action that maximizes estimated Q-function $\widehat{Q}_{K}^{(n_{K,b})}$ at censoring depth $n_{K,b}$. As a result,
\begin{align}\label{eq: adrel_BAP_nonzero_c}
\mathrm{BAP}(\widehat{\pi}_K) \geq 0.
\end{align}

Using Equations \eqref{eq: adrel_i_start_c} and \eqref{eq: adrel_BAP_nonzero_c} yields
\begin{sizeddisplay}
{\footnotesize}
\begin{align}\label{eq: adrel_i_split_c}
    i(\widehat{\pi}_K)
    &\leq
    \frac{1-\gamma^{n_{K,b}}}{(1-\gamma)^2}
    \EE_{H^{(I)} \sim d_{\nu,I}^{\hat{\pi}_K}}
    \Bigg[
    \Big(
    \EE^{\pi^{\ast}_{n_{K,b}}}\left[\text{UQ}(W^{(0)},A^{(0)},\cdots W^{(I)},A^{(I)})\given (W^{(0)},A^{(0)},\cdots W^{(I)})\right]
    +2\gamma^{K-1}\Big(\frac{2-\gamma}{1-\gamma}R_{\max}\Big)
    \nonumber\\
    &\qquad\qquad\qquad\qquad\qquad\qquad
    +\frac{2}{1-\gamma
    }C_5(\varepsilon)(NT)^{-\delta} 
    \Big)
    \prod_{j=1}^{I}\mathbb{I}[\Delta^{(j)}=0]\,
    \mathbb{I}[\Delta^{(0)}=1]
    \Bigg]
    \;+\;
    \mathrm{BAP}(\widehat{\pi}_K).
\end{align}
\end{sizeddisplay}

Observe that the first term on the right-hand side of Equation \eqref{eq: adrel_i_split_c} (i.e., the part excluding $\mathrm{BAP}(\widehat{\pi}_K)$) is exactly the same object bounded in the idealized setting for C-FQI after imposing $\mathrm{BAP}(\widehat{\pi}_K)=0$ and dropping the $\mathrm{TAP}(\widehat \pi_K)$ contribution (see Equation~\eqref{eq: bound on term (i) 2nd}).  Therefore, the same bound as in Equation~\eqref{eq:i_final_cov_simplified} from Section~\ref{C-FQI under Idealized Setting} applies directly  with an addition of the term  $\mathrm{BAP}(\widehat{\pi}_K)$. Hence, with probability at least  $1 - \varepsilon - \epsilon_s- (n_{K,b}+1)\epsilon$,
\begin{sizeddisplay}
{\footnotesize}
\begin{align}\label{eq: adrel_i_final_c}
     i(\widehat{\pi}_K)
\;\lesssim\;&d
\sqrt{\frac{\big(d+\log(1/\epsilon)\big)d}{\rho\,NT}}
\;+\;
C_5(\varepsilon)(NT)^{-\delta}
    \;+\;
    \mathrm{BAP}(\widehat{\pi}_K).
\end{align}
\end{sizeddisplay}
Combining Equation \eqref{eq: adrel_regret_decomp_c} with Equation \eqref{eq: adrel_i_final_c} and keeping the same choice of $K$ as in the idealized setting (i.e., Equation~\eqref{eq:K_choice_cov}) yield that, with probability at least $1-(n_{K,b}+3)\epsilon -\varepsilon$,
\begin{sizeddisplay}
{\footnotesize}
 \begin{align}\label{eq: adrel_regret_final_c}
    \textup{\textbf{Regret}}(\widehat{\pi}_K)
    &\lesssim
  d
\sqrt{\frac{\big(d+\log(1/\epsilon)\big)d}{\rho\,NT}}
\;+\;
C_5(\varepsilon)(NT)^{-\delta}
    \;+\;
    \mathrm{BAP}(\widehat{\pi}_K)+\omega_{NT}.
\end{align}
\end{sizeddisplay}
Observe that this is exactly the same bound derived in Equation~\eqref{eq: final regret} in the proof of Theorem~\ref{cor:ideal} with an additional term $\mathrm{BAP}(\widehat{\pi}_K)$ as a result of the relaxation of Assumption~\ref{ass:AD}.
\subsection{Step 3: Combined Regret Bound under the Relaxation of Assumption~\ref{ass:AD}}\label{Step 3: Combined regret under ad relaxation}

We now combine the results of Section~\ref{Step 1: Regret for PC-FQI under the Relaxation of ad} (PC-FQI) and Section~\ref{Step 2: Regret for C-FQI under the Relaxation of ad} (C-FQI) under the relaxed setting where Assumption~\ref{ass:AD} is not assumed, while all other assumptions remain in force.

In the idealized setting (Theorem~\ref{cor:ideal}), we defined the statistical error term as
\begin{align}\label{eq: def stat error (ideal_adrel)}
\mathcal{E}_{\text{stat}}(NT, d) :=
\max\left(
\big(d+\log(1/\epsilon)\big)d
\sqrt{\frac{C_{\mathrm{cov}}}{\rho NT}},\ 
d
\sqrt{\frac{\big(d+\log(1/\epsilon)\big)d}{\rho\,NT}}
\right)
+
C_5(\varepsilon)(NT)^{-\delta}
+
\omega_{NT}.
\end{align}
We keep exactly the same definition here.

From Section~\ref{Step 1: Regret for PC-FQI under the Relaxation of ad} (i.e., Equation~\eqref{eq: adrel_regret_final_pc}) and Section~\ref{Step 2: Regret for C-FQI under the Relaxation of ad} (i.e., Equation~\eqref{eq: adrel_regret_final_c}), we have that with probability at least $1-(n_{K,b}+3)\epsilon -\varepsilon$,
\begin{align}\label{eq: adrel_pc_compact}
\textup{\textbf{Regret}}(\widetilde{\pi}_K)
    &\lesssim \big(d+\log(1/\epsilon)\big)d
\sqrt{\frac{C_{\mathrm{cov}}\,}{\rho NT}}\;
+
C_5(\varepsilon)(NT)^{-\delta}
    \;+\;
    \mathrm{BAP}(\widetilde{\pi}_K)+\omega_{NT}
\end{align}
and similarly,
\begin{align}\label{eq: adrel_c_compact}
\textup{\textbf{Regret}}(\widehat{\pi}_K)
    &\lesssim
  d
\sqrt{\frac{\big(d+\log(1/\epsilon)\big)d}{\rho\,NT}}
\;+\;
C_5(\varepsilon)(NT)^{-\delta}
    \;+\;
    \mathrm{BAP}(\widehat{\pi}_K)+\omega_{NT}.
\end{align}

Furthermore, recall the definition of the simultaneous event $\Omega_{\mathrm{all}}$ from Equation \eqref{eq: simultaneous event}. Then by a union bound, we have
\begin{align}\label{eq: adrel_union_bound}
\Pr(\Omega_{\mathrm{all}})
\;\ge\;
1-2\big((n_{K,b}+3)\epsilon+\varepsilon\big),
\end{align}
and on $\Omega_{\mathrm{all}}$ both bounds in Equations \eqref{eq: adrel_pc_compact}--\eqref{eq: adrel_c_compact} hold simultaneously. Therefore,
\begin{align*}
\max\!\Big(\textup{\textbf{Regret}}(\widehat \pi_K),\ \textup{\textbf{Regret}}(\widetilde \pi_K)\Big)
\;\lesssim\;
\mathcal{E}_{\text{stat}}(NT,d)
\;+\;
\max\!\Big(\mathrm{BAP}(\widehat \pi_K),\mathrm{BAP}(\widetilde \pi_K)\Big).
\end{align*}
which holds with probability at least $1-2\big((n_{K,b}+3)\epsilon+\varepsilon\big)$. This concludes the proof of Corollary~\ref{cor:relaxed-AD}.

\section{Proof of Corollary~\ref{cor:relaxed-structure}}\label{app:proof-relaxed-structure}

This section proves Corollary~\ref{cor:relaxed-structure}, where we relax the Linear MDP assumption (Assumption~\ref{ass:linear-mdp}) to Reproducing Kernel Hilbert Space (RKHS) setting. The proof mirrors the idealized proof structure and proceeds in three steps: Step~1 establishes the regret bound for PC-FQI, Step~2 establishes the analogous bound for C-FQI, and Step~3 combines the two bounds into a single statement. In Steps~1--2, we start from the same reduced regret decomposition from Lemma~\ref{lm: regret decom}, which is valid under Assumptions~\ref{assump:bounded-n} and~\ref{assump:sufficient-coverage-main-text}. The learnable-class term is bounded exactly as in the linear case. Since Assumptions~\ref{ass:TAD} and~\ref{ass:AD} remain in force in this corollary, the alignment penalties are dropped as in the idealized setting. The only substantive modification occurs in the contribution of the uncertainty-quantifier: we replace the uncertainty quantifier defined for the linear MDP case with the uncertainty quantifier defined for RKHS. We adapt the RKHS concentration arguments used in \citet{chang2021mitigating} to our case. 


\subsection{Step 1: Regret for PC-FQI under the Relaxation of Assumption~\ref{ass:linear-mdp}}
\label{Step-1-Regret-RKHS}

In this section, we relax Assumption~\ref{ass:linear-mdp} (Linear MDP). Instead, we assume that the relevant value-function classes at each censoring depth belong to a Reproducing Kernel Hilbert Space (RKHS) associated with a positive semi-definite kernel
$k_i(\cdot,\cdot)$ on length-$i$ history blocks. All other assumptions from the ideal setting (Theorem~\ref{cor:ideal}) remain in force.

Given this, Assumptions~\ref{assump:bounded-n} and~\ref{assump:sufficient-coverage-main-text} hold and $\widetilde{\pi}_K \in \Pi_{n_{K,b}}$ by construction.
Therefore, we may invoke the reduced form of Lemma~\ref{lm: regret decom} that is valid under Assumptions~\ref{assump:bounded-n} and~\ref{assump:sufficient-coverage-main-text}. This is exactly the same decomposition used at the beginning of the proof of Theorem~\ref{cor:ideal} in Section~\ref{PC-FQI under Idealized Setting} (see the decomposition corresponding to Equation~\eqref{eq:general regret decomposition with coverage} in the proof of Lemma~\ref{lm: regret decom}).
\begin{sizeddisplay}
{\footnotesize}
 \begin{align}\label{eq: rkhs_regret_decomp_pc}
    \textup{\textbf{Regret}}(\widetilde{\pi}_K)
    &\leq
    \underbrace{
    \EE^{\pi^*} \left[ \sum_{t=0}^{\infty} \gamma^t R_t \right]
    -\mathbb{E}^{\tilde{\pi}_K}\left[ \sum_{t=0}^{\infty} \gamma^t R_t \right]
    }_{\textstyle i(\widetilde{\pi}_K)}
    \;+\;
    \frac{4R_{\max}\sqrt{K-1}}{1-\gamma}\omega_{NT}.
\end{align}
\end{sizeddisplay}
Thus, it remains to upper bound the term $i(\widetilde{\pi}_K)$ while accounting for the relaxation of Assumption~\ref{ass:linear-mdp}.

Since Assumptions~\ref{assump:bounded-n} and \ref{assump:sufficient-coverage-main-text} hold in this section, the optimal policy satisfies $\pi^* \in \Pi_n \subseteq \Pi_{n_{K,b}}$. Since $\pi^*$ maximizes the value globally and is a valid candidate within the restricted class $\Pi_{n_{K,b}}$, it must coincide with the $\pi^{\ast}_{n_{K,b}}$ where  $\pi^{\ast}_{n_{K,b}}\in \argmax_{\pi' \in \Pi_{n_{K,b}}}\EE^{\pi'}\!\left[\sum_{t=0}^{\infty}\gamma^tR_t\right]$. Thus, without loss of generality, we let $\pi^{\ast}_{n_{K,b}} = \pi^*$ as in the idealized setting. Given this and the fact that $\pi^{\ast}_{n_{K,b}}$ and $\widetilde{\pi}_K$ both belong to $\Pi_{n_{K,b}}$, Lemma~\ref{lm: performance difference} yields
\begin{sizeddisplay}
{\footnotesize}
\begin{align}\label{eq: rkhs_i_start}
    i(\widetilde{\pi}_K) &=
    \EE^{\pi^{\ast}}\left[\sum_{t=0}^{\infty}\gamma^tR_t\right]
    -\EE^{\tilde{\pi}_K}\left[\sum_{t=0}^{\infty}\gamma^tR_t\right]\nonumber\\
    &=
    \EE^{\pi^{\ast}_{n_{K,b}}}\left[\sum_{t=0}^{\infty}\gamma^tR_t\right]
    -\EE^{\widetilde{\pi}_K}\left[\sum_{t=0}^{\infty}\gamma^tR_t\right]\nonumber\\
    &=
    \frac{1-\gamma^{n_{K,b}}}{(1-\gamma)^2}
    \EE_{(W^{(0)},A^{(0)},\cdots,W^{(I)}, A^{(I)}) \sim d_{\nu,I}^{\tilde{\pi}_K} \times \tilde{\pi}_K}
    \Bigg[
    -A^{(I)}_{\pi^{\ast}_{n_{K,b}}}(W^{(0)},A^{(0)},\cdots,W^{(I)},A^{(I)})
    \prod_{j=1}^{I}\mathbb{I}[\Delta^{(j)}=0]\,
    \mathbb{I}[\Delta^{(0)}=1]
    \Bigg]\nonumber\\
    &\leq
    \frac{1-\gamma^{n_{K,b}}}{(1-\gamma)^2}
    \EE_{(W^{(0)},A^{(0)},\cdots,W^{(I)}, A^{(I)}) \sim d_{\nu,I}^{\tilde{\pi}_K}}
    \Bigg[
    \Big(
    \EE^{\pi^{\ast}_{n_{K,b}}}\!\big[\widetilde{\textup{UQ}}(W^{(0)},A^{(0)},\cdots,W^{(I)},A^{(I)})\mid (W^{(0)},A^{(0)},\cdots,W^{(I)})\big] \nonumber\\
    &\hspace{6em}
    +
    \EE^{\tilde{\pi}_K}\!\big[\textup{TAD}^{\tilde{\pi}_K}(W^{(0)},A^{(0)},\cdots,W^{(I)},A^{(I)})\mid (W^{(0)},A^{(0)},\cdots,W^{(I)})\big]
    +2\gamma^{K-1}\Big(\frac{2-\gamma}{1-\gamma}R_{\max}\Big)\nonumber\\
    &\hspace{6em}
    +\frac{2}{1-\gamma
    }C_5(\varepsilon)(NT)^{-\delta}
    +\textup{AD}^{\tilde{\pi}_K}(W^{(0)},A^{(0)},\cdots,W^{(I)})\mathbb{I}[I=n_{K,b}]
    \Big)\prod_{j=1}^{I}\mathbb{I}[\Delta^{(j)}=0]\,
    \mathbb{I}[\Delta^{(0)}=1]
    \Bigg].
\end{align}
\end{sizeddisplay}
which holds with probability at least $1 -(n_{K,b}+1)\epsilon-\varepsilon$ (i.e., on the event  $\Omega_P \cap\Omega^r$). Observe that this is exactly the same bound as Equation~\eqref{eq: pes bound on term (i)} from Section~\ref{PC-FQI under Idealized Setting}.

Given Equation \eqref{eq: rkhs_i_start}, the alignment-related contributions are precisely
\begin{sizeddisplay}
{\scriptsize}
\begin{align*}
    \mathrm{TAP}(\widetilde \pi_K)
&:=
\frac{1-\gamma^{n_{K,b}}}{(1-\gamma)^2}
\EE_{H^{(I)} \sim d_{\nu,I}^{\tilde \pi_K}}
\!\left[
\EE^{\tilde \pi_K}\!\left[\textup{TAD}^{\tilde \pi_K}(W^{(0)},A^{(0)},\cdots,W^{(I)},A^{(I)}) \mid (W^{(0)},A^{(0)},\cdots,W^{(I)})\right]
\prod_{j=1}^{I}\mathbb{I}[\Delta^{(j)}=0]\,
\mathbb{I}[\Delta^{(0)}=1]
\right],
\end{align*}
\end{sizeddisplay}
and
\begin{sizeddisplay}
{\footnotesize}
\begin{align*}
   \mathrm{BAP}(\widetilde \pi_K)
&:=
\frac{1-\gamma^{n_{K,b}}}{(1-\gamma)^2}
\EE_{H^{(I)} \sim d_{\nu,I}^{\tilde \pi_K}}
\!\left[
\textup{AD}^{\tilde \pi_K}(W^{(0)},A^{(0)},\cdots,W^{(I)})\,\mathbb{I}[I=n_{K,b}]
\right]. 
\end{align*}
\end{sizeddisplay}

Recall that under Assumption~\ref{ass:TAD} and Assumption~\ref{ass:AD}, we have
$\mathrm{TAP}(\widetilde{\pi}_K) \leq 0$ and $\mathrm{BAP}(\widetilde{\pi}_K)=0$
as shown in Section~\ref{PC-FQI under Idealized Setting}.
Therefore, Equation \eqref{eq: rkhs_i_start} reduces to the following:
\begin{sizeddisplay}
{\footnotesize}
\begin{align}\label{eq: rkhs_bound_on_term_i_2nd}
    i(\widetilde{\pi}_K)
    &\leq
    \frac{1-\gamma^{n_{K,b}}}{(1-\gamma)^2}
    \EE_{(W^{(0)},A^{(0)},\cdots,W^{(I)}, A^{(I)}) \sim d_{\nu,I}^{\tilde{\pi}_K}}
    \Bigg[
    \Big(
    \EE^{\pi^{\ast}_{n_{K,b}}}\!\left[\widetilde{\text{UQ}}(W^{(0)},A^{(0)},\cdots,W^{(I)},A^{(I)})\given (W^{(0)},A^{(0)},\cdots,W^{(I)})\right]\nonumber\\
    &\hspace{6em}
    + 2\gamma^{K-1}\left(\frac{2-\gamma}{1-\gamma}R_{\max}\right)
    +\frac{2}{1-\gamma
    }C_5(\varepsilon)(NT)^{-\delta}
    \Big)
    \prod_{j=1}^{I}\mathbb{I}[\Delta^{(j)}=0]\,
    \mathbb{I}[\Delta^{(0)}=1]
    \Bigg].
\end{align}
\end{sizeddisplay}

We now bound the contribution of $\widetilde{\mathrm{UQ}}$ in Equation~\eqref{eq: rkhs_bound_on_term_i_2nd}
under the general RKHS setting. Before doing that, we introduce several definitions and notations that are related to RKHS setting.

First, we assume bounded rewards $|R_t|\le R_{\max}$ and bounded kernels
$k_i(h,h)\le 1$ for all relevant length-$i$ history blocks $h$, which matches the bounded-kernel condition used in
\cite[Assumption~22]{chang2021mitigating}.
Additionally, for a fixed depth $i\in\{0,1,\ldots,n_{K,b}\}$, let
$\mathcal{O}_N^{(i)}=\{z_{i,1},\ldots,z_{i,|\mathcal{O}_N^{(i)}|}\}$
denote the depth-$i$ partition. Let $k_i(\cdot,\cdot)$ be a PSD kernel on length-$i$ history blocks with RKHS
$\mathcal{H}_i$ and feature map $\Phi_i(\cdot)$ such that
$k_i(h,h')=\langle \Phi_i(h),\Phi_i(h')\rangle_{\mathcal{H}_i}$.

Define the Gram matrix $\mathbf{G}_{\mathcal{O}_N^{(i)}}\in\mathbb{R}^{|\mathcal{O}_N^{(i)}|\times|\mathcal{O}_N^{(i)}|}$
by $[\mathbf{G}_{\mathcal{O}_N^{(i)}}]_{ab}=k_i(z_{i,a},z_{i,b})$ and its regularized version
\begin{align}\label{eq: rkhs_reg_gram_def_again}
\Lambda^{(i)}
\;:=\;
\mathbf{G}_{\mathcal{O}_N^{(i)}} + \zeta^2 I,
\end{align}
where $\zeta^2>0$ is the noise variance parameter, consistent with the posterior-variance formulation in
\cite[Example~3]{chang2021mitigating}.%
\footnote{
In the linear case, $\lambda$ served as an algorithmic ridge regularization parameter inside the empirical design matrix
to ensure invertibility and control the confidence width. Here, following \cite{chang2021mitigating}, the analogous role is
played by the GP/KRR noise variance $\zeta^2$ inside $(\mathbf{G}+\zeta^2 I)^{-1}$, which simultaneously regularizes the Gram
matrix and calibrates posterior uncertainty.
}

Then, for any length-$i$ history block $h$, the kernel vector is defined as
\[
k_{\mathcal{O}_N^{(i)}}(h)
:=
\big(k_i(z_{i,1},h),\ldots,k_i(z_{i,|\mathcal{O}_N^{(i)}|},h)\big)^\top\in\mathbb{R}^{|\mathcal{O}_N^{(i)}|}.
\]
Accordingly, we define the (posterior) kernel variance as
\begin{align}\label{eq: rkhs_post_var_again}
k_{i,\mathcal{O}_N^{(i)}}(h,h)
:=
k_i(h,h) - k_{\mathcal{O}_N^{(i)}}(h)^\top(\Lambda^{(i)})^{-1}k_{\mathcal{O}_N^{(i)}}(h),
\end{align}
which is the standard GP/KRR posterior variance (cf. \cite[Example~3]{chang2021mitigating}).

Given these, for each depth $i$ and iteration $k$, the uncertainty quantifier for the RKHS setting is defined as
\begin{align}\label{eq: rkhs_U_single_def_again}
\tilde{U}_{k}^{(i)}\!\big(h^{(t-i):t}\big)
\;:=\;
\tilde{\beta}_k\left[\frac{k_{i,\mathcal{O}_N^{(i)}}\!\big(h^{(t-i):t},h^{(t-i):t}\big)}{\zeta^2}\right]^{1/2},
\end{align}
and thus, for each $i$, $\tilde{U}_{k}^{(i)}$ in Equation~\eqref{eq: rkhs_U_single_def_again} satisfies Definition~\ref{def:UE} as implied by Lemma~14 of \citet{chang2021mitigating}. 

Now we are ready to bound $\widetilde{\mathrm{UQ}}$. First, recall the characterization of $\widetilde{\mathrm{UQ}}$
\begin{sizeddisplay}\footnotesize
    \begin{align*}
        \widetilde{\mathrm{UQ}}(w^{(0)},a^{(0)},\ldots,w^{(i)},a^{(i)})
=
\sum_{k=0}^{K-1}\gamma^k\,
\EE^{\pi^\ast_{n_{K,b}}}\!\left[2
\tilde{U}_{K-k}^{(M_{i+k})}\!\big(H_{(i+k-M_{i+k}):(i+k)}\big)
\ \Big|\ (W_0,A_0\ldots,W_i,A_i)=(w^{(0)},a^{(0)},\ldots,w^{(i)},a^{(i)})
\right].
    \end{align*}
\end{sizeddisplay}

Then for a fixed $i$, we have
\begin{align}\label{eq:UQ rkhs}
\EE_{H^{(i)}\sim d_{\nu,i}^{\pi^*}}\!\big[\tilde{U}_{k}^{(i)}(H^{(i)})\big]
&=
\tilde{\beta}_k\,
\EE_{H^{(i)}\sim d_{\nu,i}^{\pi^*}}\!\left[\sqrt{\frac{k_{i,\mathcal{O}_N^{(i)}}(H^{(i)},H^{(i)})}{\zeta^2}}\right]
\end{align}
where $H^{(i)}$ denotes the generic length-$i$ history block random variable.

Under the bounded-kernel 
assumption and $\zeta^2=\Omega(1)$, \cite[Theorem~25 (Statement~2)]{chang2021mitigating}
applied to the depth-$i$ kernel $k_i$ yields that, with probability at least $1-\epsilon/(n_{K,b}+1)$,
\begin{align}\label{eq:chang_thm25_depth_i}
\EE_{H^{(i)}\sim d_{\nu,i}^{\pi^*}}\!\left[\sqrt{k_{i,\mathcal{O}_N^{(i)}}(H^{(i)},H^{(i)})}\right]
\;\le\;
c_1\sqrt{\frac{ C_{\mathrm{cov}}d^{(i)}_{*}\big(d^{(i)}_{*}+\log(c_2/\epsilon)\big)}{|\mathcal{O}_N^{(i)}|}},
\end{align}
where $d^{(i)}_{*}$ is the (depth-$i$) effective dimension appearing in \cite[Definition~9]{chang2021mitigating}
and $c_1,c_2$ are universal constants.
Since $\zeta^2=\Omega(1)$, we may absorb $1/\zeta$ into $\lesssim$ and obtain
\begin{align}\label{eq:chang_thm25_scaled}
\EE_{H^{(i)}\sim d_{\nu,i}^{\pi^*}}\!\left[\sqrt{\frac{k_{i,\mathcal{O}_N^{(i)}}(H^{(i)},H^{(i)})}{\zeta^2}}\right]
\;\lesssim\;
\sqrt{\frac{ C_{\mathrm{cov}}d^{(i)}_{*}\big(d^{(i)}_{*}+\log(1/\epsilon)\big)}{|\mathcal{O}_N^{(i)}|}}.
\end{align}

Then, by a union bound over $i\in\{0,\ldots,n_{K,b}\}$, Equation \eqref{eq:chang_thm25_scaled} holds simultaneously
for all depths with probability at least $1-\epsilon$, and we denote this event by $\Omega_{MP}$.

Following this, substituting Equation \eqref{eq:chang_thm25_scaled} into Equation \eqref{eq:UQ rkhs}, we obtain
the final bound
\begin{align}\label{eq: rkhs_expected_U_bound_cov_final_again}
\EE_{H^{(i)}\sim d_{\nu,i}^{\pi^*}}\!\big[\tilde{U}_{k}^{(i)}(H^{(i)})\big]
\;\lesssim\;
\tilde{\beta}_k\,
\sqrt{
\frac{ C_{\mathrm{cov}}d^{(i)}_{*}\big(d^{(i)}_{*}+\log(1/\epsilon))}{|\mathcal{O}_N^{(i)}|}
},
\end{align}
which holds for all $i\in\{0,\ldots,n_{K,b}\}$ simultaneously with probability at least $1-\epsilon$.

We also invoke Assumption~\ref{lower bound on partition sizes} which states that there exists $\rho\in(0,1]$ and an event
$\Omega_{\mathrm{str}}$ with $\Pr(\Omega_{\mathrm{str}})\ge 1-\epsilon_s$ such that on $\Omega_{\mathrm{str}}$,
\[
\min_{0\le i\le n_{K,b}}|\mathcal{O}_N^{(i)}|\ \ge\ \rho\,NT.
\]
Accordingly, by a union bound, we have 
\begin{align}
\Pr(\widetilde{\Omega}_P):=\Pr(\Omega_P \cap\Omega^r \cap \Omega_{\mathrm{str}}\cap \Omega_{MP}) \ \ge\ 1-(n_{K,b}+3)\epsilon -\varepsilon.
\end{align}

Next, we apply Equation \eqref{eq: rkhs_expected_U_bound_cov_final_again} to the definition of $\widetilde{\text{UQ}}$ on the event on $\widetilde{\Omega}_P$. This brings
\begin{sizeddisplay}\small
\begin{align}\label{eq: pess rkhs UQ_bound_cov}
\EE^{\pi^\ast_{n_{K,b}}}\!\left[\widetilde{\text{UQ}}(w^{(0)},a^{(0)},\ldots,w^{(i)},a^{(i)})\right]
&\leq
\sum_{k=0}^{K-1}\gamma^k\; 2\;\tilde{\beta}_{K-k}\EE_{M_{i+k} \sim \pi^*_{n_{K,b}}}\left[\sqrt{\frac{ C_{\mathrm{cov}}d^{(M_{i+k})}_{*}\big(d^{(M_{i+k})}_{*}+\log(1/\epsilon))}{|\mathcal{O}_N^{(M_{i+k})}|}}\right]\nonumber\\
&\leq
\frac{2\;\tilde{\beta}_{\max}}{1-\gamma}\sqrt{\frac{ C_{\mathrm{cov}}\max_{0\leq j \leq n_{K,b}}d^{(j)}_{*}\big(d^{(j)}_{*}+\log(1/\epsilon))}{\min_{0\le j\le n_{K,b}}|\mathcal{O}_N^{(j)}|}}\nonumber\\
&\leq
\frac{2\;\tilde{\beta}_{\max}}{1-\gamma}\sqrt{\frac{ C_{\mathrm{cov}}d_{*}\big(d_{*}+\log(1/\epsilon))}{\rho NT}}
\end{align}\end{sizeddisplay}
where we use the fact that
$\min_{j}|\mathcal{O}_N^{(j)}|\ge \rho NT$, $d_* := \max_{0\leq i \leq n_{K,b}}d_*^{(i)}$ and  $\tilde{\beta}_{\max}:=\max_{0\le \ell\le K-1}\tilde{\beta}_\ell$.

Then, plugging Equation \eqref{eq: pess rkhs UQ_bound_cov} into Equation \eqref{eq: rkhs_bound_on_term_i_2nd}, we can obtain
\begin{align}\label{eq: pess rkhs i_bound_plug_cov}
i(\widetilde{\pi}_K)
&\leq
\frac{1-\gamma^{n_{K,b}}}{(1-\gamma)^2}
\left(\frac{2\tilde{\beta}_{\max}}{1-\gamma}\sqrt{\frac{C_{\mathrm{cov}}\,
d_*\big(d_*+\log(1/\epsilon)\big)}{\rho NT}}
+
2\gamma^{K-1}\frac{2-\gamma}{1-\gamma}R_{\max}+
\frac{2}{1-\gamma
    }C_5(\varepsilon)(NT)^{-\delta}
\right)\nonumber\\
&\le
\frac{1}{(1-\gamma)^2}
\left(\frac{2\tilde{\beta}_{\max}}{1-\gamma}\sqrt{\frac{C_{\mathrm{cov}}\,
d_*\big(d_*+\log(1/\epsilon)\big)}{\rho NT}}+
2\gamma^{K-1}\frac{2-\gamma}{1-\gamma}R_{\max}
+
\frac{2}{1-\gamma
    }C_5(\varepsilon)(NT)^{-\delta}
\right),
\end{align}
which holds with probability at least $ 1-(n_{K,b}+3)\epsilon -\varepsilon.$

To state the final rate, we choose $K$ as in the idealized setting
\begin{align}\label{eq: rkhs_K_choice}
K \;=\; \left\lceil \frac{\log(NT)}{2(1-\gamma)} \right\rceil,
\end{align}
so that $\gamma^{K-1}\le (NT)^{-1/2}$ and hence
\[
2\gamma^{K-1}\frac{2-\gamma}{1-\gamma}R_{\max}
\;\le\;
\frac{c_0\,R_{\max}}{1-\gamma}\cdot (NT)^{-1/2}
\]
for an absolute constant $c_0>0$.

Finally, we choose $\tilde{\beta}_{\max}$ in a way consistent with RKHS confidence multipliers,
namely (up to logarithmic factors and constants) as in \citet{chang2021mitigating},
\begin{align}\label{eq: rkhs_beta_choice}
\tilde{\beta}_{\max}
\;\asymp\;
\sqrt{d_{*}\,\Big(d_{*}+\log(1/\epsilon)\Big)}
\end{align}

Then, on event $\widetilde{\Omega}_P$, given Equations~\eqref{eq: rkhs_regret_decomp_pc} and \eqref{eq: pess rkhs i_bound_plug_cov} as well as the selection of $K$ and $\tilde{\beta}_{\max}$, we have 
\begin{align}
\textup{\textbf{Regret}}(\widetilde{\pi}_K)
\;\le\;&
\frac{1}{(1-\gamma)^2}
\left(
\frac{2\tilde{\beta}_{\max}}{1-\gamma}\sqrt{\frac{C_{\mathrm{cov}}\,
d_*\big(d_*+\log(1/\epsilon)\big)}{\rho NT}}
+
\frac{c_0\,R_{\max}}{1-\gamma}\,(NT)^{-1/2}
+
\frac{2}{1-\gamma}C_5(\varepsilon)(NT)^{-\delta}
\right)\nonumber\\
&+\frac{4R_{\max}}{1-\gamma}
\sqrt{\left\lceil \frac{\log(NT)}{2(1-\gamma)} \right\rceil}\;\omega_{NT}\nonumber\\
\;\leq\;&
\frac{1}{(1-\gamma)^3}
\left(2d_*\big(d_*+\log(1/\epsilon)\big)
\sqrt{\frac{C_{\mathrm{cov}}\,}{\rho NT}}\;
+
\frac{c_0R_{\max}}{\sqrt{NT}}
+
2C_5(\varepsilon)(NT)^{-\delta}
\right)\nonumber\\
&+\frac{4R_{\max}}{1-\gamma}
\sqrt{\left\lceil \frac{\log(NT)}{2(1-\gamma)} \right\rceil}\;\omega_{NT}
\end{align}

To simplify the expression, we consider the regime in which the truncation term is dominated by the statistical term, namely without loss of generality, we assume
\begin{align}\label{eq: truncation dom rkhs pfqi}
   c_0R_{\max}\;\le\;d_*\big(d_*+\log(1/\epsilon)\big)\sqrt{\frac{C_{\mathrm{cov}}\,}{\rho}}.
\end{align}

Under this condition, the truncation contribution is absorbed into the leading statistical term. Given this, we further use $\lesssim$ notation which hides absolute constants and logarithmic factors
(e.g., $\gamma$, $R_{\max}$), while keeping $\sqrt{\log(1/\epsilon)}$ explicit. This yields the final bound on $i(\widetilde{\pi}_K)$ and $\textup{\textbf{Regret}}(\widetilde{\pi}_K)$ as follows:
\begin{align}\label{eq: pess rkhs i_final_cov_simplified}
i(\widetilde{\pi}_K)
\;\lesssim\;&d_*\big(d_*+\log(1/\epsilon)\big)
\sqrt{\frac{C_{\mathrm{cov}}\,}{\rho NT}}\;
+
C_5(\varepsilon)(NT)^{-\delta}
\end{align}
and
\begin{align}\label{eq: pess rkhs final regret}
\textup{\textbf{Regret}}(\widetilde{\pi}_K)
\;\lesssim\;&d_*\big(d_*+\log(1/\epsilon)\big)
\sqrt{\frac{C_{\mathrm{cov}}\,}{\rho NT}}\;
+
C_5(\varepsilon)(NT)^{-\delta}
\;+\;
\omega_{NT},
\end{align}
which holds with probability at least $1-(n_{K,b}+3)\epsilon -\varepsilon$.

\subsection{Step 2: Regret for C-FQI under the Relaxation of Assumption~\ref{ass:linear-mdp}}
\label{C-FQI under RKHS}

The proof for $\widehat{\pi}_K$ follows the same steps as in the idealized linear case, with the only change being
that we upper bound the uncertainty-quantifier and pessimism terms using RKHS information-gain arguments.
Recall that in the idealized setting, Assumptions~\ref{assump:bounded-n} and \ref{assump:sufficient-coverage-main-text} hold.
Moreover, $\widehat{\pi}_K \in \Pi_{n_{K,b}}$ by definition. Therefore, we may invoke the reduced form of
Lemma~\ref{lm: regret decom}, which is valid under these two assumptions. Specifically, this corresponds to the
decomposition given in Equation \eqref{eq:general regret decomposition with coverage} in the proof of
Lemma~\ref{lm: regret decom}.
\begin{sizeddisplay}
{\footnotesize}
 \begin{align}\label{eq: rkhs_fqi_regret_1_app}
    \text{Regret}(\widehat \pi_K)
    &\leq
    \underbrace{\EE^{\pi^*} \left[ \sum_{t=0}^{\infty} \gamma^t R_t \right]
    -\mathbb{E}^{\hat{\pi}_K}\left[ \sum_{t=0}^{\infty} \gamma^t R_t \right]
}_{\textstyle i(\widehat{\pi}_K)}
    \;+\;
    \frac{4R_{\max}\sqrt{K-1}}{1-\gamma}\omega_{NT}.
\end{align}
\end{sizeddisplay}
Thus, it remains to upper bound the term $i(\widehat{\pi}_K)$.

Assumptions~\ref{assump:bounded-n} and \ref{assump:sufficient-coverage-main-text} imply that the optimal policy satisfies
$\pi^* \in \Pi_n \subseteq \Pi_{n_{K,b}}$.
Since $\pi^*$ maximizes the value globally and is a valid candidate within the restricted class $\Pi_{n_{K,b}}$,
it must coincide with the $\pi^{\ast}_{n_{K,b}}$ where
$\pi^{\ast}_{n_{K,b}}\in \argmax_{\pi' \in \Pi_{n_{K,b}}}\EE^{\pi'}\!\left[\sum_{t=0}^{\infty}\gamma^tR_t\right]$.
Thus, without loss of generality, we let $\pi^{\ast}_{n_{K,b}} = \pi^*$ as in the idealized setting.
Given this and the fact that $\pi^{\ast}_{n_{K,b}}$ and $\widehat{\pi}_K$ both belong to $\Pi_{n_{K,b}}$,
Lemma~\ref{lm: performance difference} yields
\begin{sizeddisplay}
{\footnotesize}
\begin{align}\label{eq: rkhs_regret_1_app}
         i(\widehat{\pi}_K)
         &=
         \EE^{\pi^{\ast}}\left[\sum_{t=0}^{\infty}\gamma^tR_t\right]
         -\EE^{\hat{\pi}_K}\left[\sum_{t=0}^{\infty}\gamma^tR_t\right]\nonumber\\
         &=
         \EE^{\pi^{\ast}_{n_{K,b}}}\left[\sum_{t=0}^{\infty}\gamma^tR_t\right]
         -\EE^{\hat{\pi}_K}\left[\sum_{t=0}^{\infty}\gamma^tR_t\right]\nonumber\\
         &=
         \frac{1-\gamma^{n_{K,b}}}{(1-\gamma)^2}
         \EE_{(W^{(0)},A^{(0)},\cdots,W^{(I)}, A^{(I)}) \sim d_{\nu,I}^{ \hat{\pi}_K} \times \hat{\pi}_K
    }\!\left[
    -A^{(I)}_{\pi^{\ast}_{n_{K,b}}}(W^{(0)},A^{(0)},\cdots,W^{(I)},A^{(I)})
    \prod_{j=1}^{I}\left\{\mathbb{I}[\Delta^{(j)}=0]\right\}\mathbb{I}[\Delta^{(0)}=1]
    \right]\nonumber\\
    &\leq
    \frac{1-\gamma^{n_{K,b}}}{(1-\gamma)^2}
    \EE_{(W^{(0)},A^{(0)},\cdots,W^{(I)}, A^{(I)}) \sim d_{\nu,I}^{\hat{\pi}_K} }\Bigg[
    \Big(
    \EE^{\pi^{\ast}_{n_{K,b}}}\!\left[\text{UQ}(W^{(0)},A^{(0)},\cdots,W^{(I)},A^{(I)})
    \given (W^{(0)},A^{(0)},\cdots,W^{(I)})\right]\nonumber\\
    &\hspace{6em}
    + \EE^{\hat{\pi}_K}\!\left[ \textup{TAD}^{\hat \pi_K}(W^{(0)},A^{(0)},\cdots,W^{(I)},A^{(I)})
    \given(W^{(0)},A^{(0)},\cdots,W^{(I)})\right]\nonumber\\
    &\hspace{6em}
    + C^{(I)}(W^{(0)},A^{(0)},\cdots,W^{(I)})
    + 2\gamma^{K-1}\left(\frac{2-\gamma}{1-\gamma}R_{\max}\right)
    +\frac{2}{1-\gamma
    }C_5(\varepsilon)(NT)^{-\delta}\nonumber\\
    &\hspace{6em}
    +\textup{AD}^{\hat{\pi}_K}(W^{(0)},A^{(0)},\cdots,W^{(I)})\mathbb{I}[I=n_{K,b}]
    \Big)
    \prod_{j=1}^{I}\left\{\mathbb{I}[\Delta^{(j)}=0]\right\}\mathbb{I}[\Delta^{(0)}=1]
    \Bigg],
\end{align}
\end{sizeddisplay}
which holds with probability at least $1 -(n_{K,b}+1)\epsilon-\varepsilon$ (i.e., on the event  $\Omega \cap\Omega^r$). Observe that this is exactly the same bound as Equation~\eqref{eq: bound on term (i)} from Section~\ref{C-FQI under Idealized Setting}.

Given Equation \eqref{eq: rkhs_regret_1_app}, the components related to the alignment discrepancies are exactly:
\begin{sizeddisplay}
{\scriptsize}
\begin{align*}
    \mathrm{TAP}(\widehat \pi_K)
&:=
\frac{1-\gamma^{n_{K,b}}}{(1-\gamma)^2}
\EE_{H^{(I)} \sim d_{\nu,I}^{\hat \pi_K}}
\!\left[
\EE^{\hat \pi_K}\!\left[\textup{TAD}^{\hat \pi_K}(W^{(0)},A^{(0)},\cdots,W^{(I)},A^{(I)}) \mid (W^{(0)},A^{(0)},\cdots,W^{(I)})\right]
\prod_{j=1}^{I}\mathbb{I}[\Delta^{(j)}=0]\,
\mathbb{I}[\Delta^{(0)}=1]
\right],
\end{align*}
\end{sizeddisplay}
and
\begin{sizeddisplay}
{\footnotesize}
\begin{align*}
   \mathrm{BAP}(\widehat \pi_K)
&:=
\frac{1-\gamma^{n_{K,b}}}{(1-\gamma)^2}
\EE_{H^{(I)} \sim d_{\nu,I}^{\hat \pi_K}}
\!\left[
\textup{AD}^{\hat \pi_K}(W^{(0)},A^{(0)},\cdots,W^{(I)})\,\mathbb{I}[I=n_{K,b}]
\right]. 
\end{align*}
\end{sizeddisplay}
Recall that under Assumption~\ref{ass:TAD} and Assumption~\ref{ass:AD}, we have
$\mathrm{TAP}(\widetilde{\pi}_K) \leq 0$ and $\mathrm{BAP}(\widetilde{\pi}_K)=0$
as shown in Section~\ref{C-FQI under Idealized Setting}.
Therefore, Equation \eqref{eq: rkhs_regret_1_app} reduces to the following:
\begin{sizeddisplay}
{\footnotesize}
\begin{align}\label{eq: rkhs_bound_on_term_i_2nd fqi}
    i(\widehat{\pi}_K)
    &\leq
    \frac{1-\gamma^{n_{K,b}}}{(1-\gamma)^2}
    \EE_{(W^{(0)},A^{(0)},\cdots,W^{(I)}, A^{(I)}) \sim d_{\nu,I}^{\hat{\pi}_K} }\Bigg[
    \Big(
    \EE^{\pi^{\ast}_{n_{K,b}}}\!\left[\text{UQ}(W^{(0)},A^{(0)},\cdots,W^{(I)},A^{(I)})
    \given (W^{(0)},A^{(0)},\cdots,W^{(I)})\right]\nonumber\\
    &\hspace{10em}
    + C^{(I)}(W^{(0)},A^{(0)},\cdots,W^{(I)})
    \nonumber\\
    &\hspace{10em}+ 2\gamma^{K-1}\left(\frac{2-\gamma}{1-\gamma}R_{\max}\right)
    +\frac{2}{1-\gamma
    }C_5(\varepsilon)(NT)^{-\delta}
    \Big)\prod_{j=1}^{I}\left\{\mathbb{I}[\Delta^{(j)}=0]\right\}\mathbb{I}[\Delta^{(0)}=1]
    \Bigg].
\end{align}
\end{sizeddisplay}

We now bound the contribution of $\textup{UQ}$ and $C^{(I)}$ in Equation \eqref{eq: rkhs_bound_on_term_i_2nd fqi}
under the general RKHS setting. Recall the characterization of $\text{UQ}$ and $C^{(I)}$ as follows. 
\begin{sizeddisplay}\scriptsize
    \begin{align*}
        \text{UQ}(w^{(0)},a^{(0)},\ldots,w^{(i)},a^{(i)})
=
\sum_{k=0}^{K-1}\gamma^k\,
\EE^{\pi^\ast_{n_{K,b}}}\!\left[U_{K-k}^{(M_{i+k})}\!\big(H_{(i+k-M_{i+k}):(i+k)}\big)
\ \Big|\ (W_0,A_0\ldots,W_i,A_i)=(w^{(0)},a^{(0)},\ldots,w^{(i)},a^{(i)})
\right],
    \end{align*}
\end{sizeddisplay}

\begin{sizeddisplay}
    \scriptsize
\begin{align}
C^{(i)}(w^{(0)},a^{(0)},\ldots,w^{(i)})
&=
\max_{a\in\mathcal{A}}
\sum_{k=0}^{K-1}\gamma^{k}\,
\EE^{\pi_{K-1},\cdots,\pi_{K-k}}\!\left[
U_{K-k}^{(M_{i+k})}\big(H_{i+k}^{(M_{i+k})}\big)
\ \Big|\ (W_0,\ldots,W_i,A_i)=(w^{(0)},\ldots,w^{(i)},a)
\right],\nonumber
\end{align}
\end{sizeddisplay}
where $M_t
:=\min\Big\{n_{K,b},\ \max\{v\ge 0:\ \Delta_{t-v}=\cdots=\Delta_{t-1}=0,\ \Delta_{t-v-1}=1\}\Big\}$

Next, we introduce several definitions and notations for the RKHS setting, and then provide a finite-rank
projection argument that yields an uncertainty quantifier bound in terms of the effective dimension. 

First, without loss of generality, we assume bounded rewards $|R_t|\le R_{\max}$ and bounded kernels $k_i(h,h)\le 1$ for all relevant length-$i$ history
blocks $h$, matching the bounded-kernel condition used in \cite[Assumption~22]{chang2021mitigating}. Fix a depth
$i\in\{0,1,\ldots,n_{K,b}\}$ and let $\mathcal{O}_N^{(i)}=\{z_{i,1},\ldots,z_{i,n_i}\}$ denote the depth-$i$ partition. Let $k_i(\cdot,\cdot)$ be a PSD kernel on length-$i$ history blocks with RKHS
$\mathcal{H}_i$ and feature map $\Phi_i(\cdot)$ such that
$k_i(h,h')=\langle \Phi_i(h),\Phi_i(h')\rangle_{\mathcal{H}_i}$.

We define the Gram matrix $\mathbf{G}_{\mathcal{O}_N^{(i)}}\in\mathbb{R}^{n_i\times n_i}$ as the matrix whose entries are
$[\mathbf{G}_{\mathcal{O}_N^{(i)}}]_{ab}=k_i(z_{i,a},z_{i,b})$. We then define the regularized Gram matrix as
\begin{align}
\Lambda^{(i)}
\;:=\;
\mathbf{G}_{\mathcal{O}_N^{(i)}} + \zeta^2 I,
\nonumber
\end{align}
where $\zeta^2>0$ is the (GP/KRR) noise variance parameter, consistent with the posterior-variance formulation in
\cite[Example~3]{chang2021mitigating}.

For any length-$i$ history block $h$, we define the kernel vector as
\begin{align}
k_{\mathcal{O}_N^{(i)}}(h)
:=
\big(k_i(z_{i,1},h),\ldots,k_i(z_{i,n_i},h)\big)^\top\in\mathbb{R}^{n_i},
\nonumber
\end{align}
and accordingly the posterior kernel variance is defined as
\begin{align}
k_{i,\mathcal{O}_N^{(i)}}(h,h)
:=
k_i(h,h) - k_{\mathcal{O}_N^{(i)}}(h)^\top(\Lambda^{(i)})^{-1}k_{\mathcal{O}_N^{(i)}}(h).
\nonumber
\end{align}

For each depth $i$ and iteration $k$, the uncertainty quantifier is defined as
\begin{align}
{U}_{k}^{(i)}\!\big(h^{(t-i):t}\big)
\;:=\;
{\beta}_k\left[\frac{k_{i,\mathcal{O}_N^{(i)}}\!\big(h^{(t-i):t},h^{(t-i):t}\big)}{\zeta^2}\right]^{1/2}.
\nonumber
\end{align}

For an arbitrary policy $\pi$, let $H^{(i)}\sim d^\pi$ denote a length-$i$ history block under $d^\pi$. We aim to bound
$\EE_{H^{(i)}\sim d^\pi}[U_k^{(i)}(H^{(i)})]$. To begin, we use Using Jensen’s
inequality which results in the following
\begin{align}
\EE_{H^{(i)}\sim d^\pi}\!\big[U_k^{(i)}(H^{(i)})\big]
&=
\beta_k\;\EE_{H^{(i)}\sim d^\pi}\!\left[\sqrt{\frac{k_{i,\mathcal{O}_N^{(i)}}(H^{(i)},H^{(i)})}{\zeta^2}}\right]
\nonumber\\
&\le
\beta_k\sqrt{\frac{1}{\zeta^2}\,\EE_{H^{(i)}\sim d^\pi}\!\big[k_{i,\mathcal{O}_N^{(i)}}(H^{(i)},H^{(i)})\big]}.
\label{eq:jensen_reduction}
\end{align}
Accordingly, it suffices to upper bound
\begin{align}
\mathcal V_i
\;:=\;
\EE_{H^{(i)}\sim d^\pi}\!\big[k_{i,\mathcal{O}_N^{(i)}}(H^{(i)},H^{(i)})\big].
\nonumber
\end{align}

To do this, without loss of generality, assume $k_i$ admits a Mercer expansion with respect to $d_i^\mu$:
\begin{align}
k_i(h,h')
&=
\sum_{j=1}^\infty \mu_{i,j}\,\psi_{i,j}(h)\psi_{i,j}(h'),
\qquad
\mu_{i,1}\ge \mu_{i,2}\ge \cdots \ge 0,
\nonumber
\end{align}
where $\{\psi_{i,j}\}_{j\ge 1}$ are orthonormal in $L_2(d_i^\mu)$:
\begin{align}
\EE_{H\sim d_i^\mu}\!\big[\psi_{i,j}(H^{(i)})\psi_{i,\ell}(H^{(i)})\big]=\mathcal{I}\{j=\ell\}.
\label{eq:orthonormal}
\end{align}

Given a fix integer truncation rank $D^{(i)}\ge 1$, the truncated feature map is defined as
\begin{align}
\psi_{i,D^{(i)}}(h)
:=
\Big(\sqrt{\mu_{i,1}}\psi_{i,1}(h),\ldots,\sqrt{\mu_{i,D^{(i)}}}\psi_{i,D^{(i)}}(h)\Big)^\top\in\R^{D^{(i)}},
\nonumber
\end{align}
and the truncated kernel is defined as
\begin{align}
k_{i,D^{(i)}}(h,h')
:=
\psi_{i,D^{(i)}}(h)^\top\psi_{i,D^{(i)}}(h').
\nonumber
\end{align}
Moreover, we denote the residual kernel and its diagonal residual as follows
\begin{align}
k_{i,\perp}(h,h')
&:=
k_i(h,h')-k_{i,D^{(i)}}(h,h'),
\nonumber\\
r_{i,D^{(i)}}(h)
&:=
k_{i,\perp}(h,h)
=
\sum_{j>D^{(i)}}\mu_{i,j}\psi_{i,j}(h)^2.
\nonumber
\end{align}

Assume the eigenfunctions are uniformly bounded:
\begin{align}
\sup_h|\psi_{i,j}(h)|\le \kappa_{\psi,i}
\qquad \forall j.
\label{eq:eigfun_bdd}
\end{align}
We define the tail sum as
\begin{align}
B_i(j):=\sum_{k=j}^\infty \mu_{i,k}.
\label{eq:Bi_def}
\end{align}
Then, using Equation~\eqref{eq:eigfun_bdd}, the residual diagonal term satisfies the following
\begin{align}
r_{i,D^{(i)}}(h)
=
\sum_{j>D^{(i)}}\mu_{i,j}\psi_{i,j}(h)^2
\le
\kappa_{\psi,i}^2\sum_{j>D^{(i)}}\mu_{i,j}
=
\kappa_{\psi,i}^2\,B_i(D^{(i)}+1).
\label{eq:tail_uniform}
\end{align}

Next, we decompose the posterior variance into a head term and a tail term. By PSD monotonicity and the decomposition
$k_i=k_{i,D^{(i)}}+k_{i,\perp}$, the posterior variance satisfies, for all $h$,
\begin{align}
k_{i,\mathcal O_N^{(i)}}(h,h)
\le
k_{i,D^{(i)},\mathcal O_N^{(i)}}(h,h)
+
r_{i,D^{(i)}}(h),
\label{eq:var_decomp}
\end{align}
where $k_{i,D^{(i)},\mathcal O_N^{(i)}}(h,h)$ denotes the posterior variance computed using the truncated kernel
$k_{i,D^{(i)}}$ with the same locations $\mathcal O_N^{(i)}$ and the same $\zeta^2$.

Taking expectations of Equation~\eqref{eq:var_decomp} under $H^{(i)}\sim d^\pi$ and then applying
Equation~\eqref{eq:tail_uniform} yields
\begin{align}
\mathcal V_i
\le
\EE_{H^{(i)}\sim d^\pi}\!\big[k_{i,D^{(i)},\mathcal O_N^{(i)}}(H^{(i)},H^{(i)})\big]
+
\kappa_{\psi,i}^2\,B_i(D^{(i)}+1).
\label{eq:V_split}
\end{align}
Consequently, it remains to bound the finite-rank head term.

We now rewrite the head term in a finite-dimensional linear-design form. The empirical covariance is denoted by
\begin{align}
\widehat{\Sigma}^{(i)}_{D^{(i)}}
:=
\frac{1}{|\mathcal{O}_N^{(i)}|}\sum_{\tau=1}^{|\mathcal{O}_N^{(i)}|}\psi_{i,D^{(i)}}(z_{i,\tau})\psi_{i,D^{(i)}}(z_{i,\tau})^\top
\in\R^{D^{(i)}\times D^{(i)}},
\nonumber
\end{align}
and the population covariance under $d_i^\mu$ is defined as
\begin{align}
\Sigma^{\mu}_{i,D^{(i)}}
:=
\EE_{H^{(i)}\sim d^{\mu}_i}\big[\psi_{i,D^{(i)}}(H^{(i)})\psi_{i,D^{(i)}}(H^{(i)})^\top\big].
\nonumber
\end{align}
By Equation~\eqref{eq:orthonormal}, the population covariance satisfies
\begin{align}
\Sigma^{\mu}_{i,D^{(i)}}
=
\mathrm{diag}(\mu_{i,1},\ldots,\mu_{i,D^{(i)}}).
\label{eq:Sigmamu_diag}
\end{align}

For the truncated kernel, the posterior variance identity is
\begin{align}
\frac{1}{\zeta^2}\,k_{i,D^{(i)},\mathcal O_N^{(i)}}(h,h)
=
\psi_{i,D^{(i)}}(h)^\top\Big(|\mathcal{O}_N^{(i)}|\widehat{\Sigma}^{(i)}_{D^{(i)}}+\zeta^2 I\Big)^{-1}\psi_{i,D^{(i)}}(h).
\label{eq:linear_form}
\end{align}
Taking expectations of Equation~\eqref{eq:linear_form} under $d^\pi$ gives
\begin{align}
\EE_{H^{(i)}\sim d^\pi}\!\left[\frac{1}{\zeta^2}k_{i,D^{(i)},\mathcal O_N^{(i)}}(H^{(i)},H^{(i)})\right]
\le
\Big\|\big(|\mathcal{O}_N^{(i)}|\widehat{\Sigma}^{(i)}_{D^{(i)}}+\zeta^2 I\big)^{-1}\Big\|_{\mathrm{op}}\;
\EE_{H^{(i)}\sim d^\pi}\big[\|\psi_{i,D^{(i)}}(H^{(i)})\|_2^2\big],
\label{eq:head_jensen}
\end{align}
where the inequality stems from $x^\top A x \le \|A\|_{\mathrm{op}}\|x\|_2^2$.

Observe that since PSD truncation implies
$k_{i,D^{(i)}}(h,h)\le k_i(h,h)$, and $k_i(h,h)\le 1$, we have for all $h$,
\begin{align*}
\|\psi_{i,D^{(i)}}(h)\|_2^2
=
k_{i,D^{(i)}}(h,h)
\le
k_i(h,h)
\le
1.
\end{align*}
This implies 
\begin{align}
\EE_{H^{(i)}\sim d^\pi}\big[\|\psi_{i,D^{(i)}}(H^{(i)})\|_2^2\big]\le 1.
\label{eq:radius_exp_bdd}
\end{align}
At this point, the only remaining quantity to control in Equation~\eqref{eq:head_jensen} is the term
$\|(|\mathcal{O}_N^{(i)}|\widehat{\Sigma}^{(i)}_{D^{(i)}}+\zeta^2 I)^{-1}\|_{\mathrm{op}}$.

We now control this term using the similar idea from the idealized setting. In particular, we define the following event
\begin{align}
\Omega_{i,D^{(i)}}
:=
\left\{
\widehat{\Sigma}^{(i)}_{D^{(i)}} \succeq \frac12 \Sigma^{\mu}_{i,D^{(i)}}
\right\}.
\label{eq:Omega_def}
\end{align}
Then, by Lemma~\ref{lemma: Matrix bound}, this event holds with probability at least $1-\epsilon/(n_{K,b}+1)$.


Given this, on the event $\Omega_{i,D^{(i)}}$, we have
\begin{align}
|\mathcal{O}_N^{(i)}|\widehat{\Sigma}^{(i)}_{D^{(i)}}+\zeta^2 I
&\succeq
|\mathcal{O}_N^{(i)}|\cdot \frac12 \Sigma^{\mu}_{i,D^{(i)}}+\zeta^2 I \nonumber\\
&\succeq
\left(\frac{|\mathcal{O}_N^{(i)}|}{2}\mu_{i,D^{(i)}}+\zeta^2\right)I,
\label{eq:design_lb_2}
\end{align}
where Equation~\eqref{eq:design_lb_2} uses Equation~\eqref{eq:Sigmamu_diag} and the fact that $\mu_{i,D^{(i)}}$ is the
smallest eigenvalue among the first $D^{(i)}$. Consequently,
\begin{align}
\Big\|\big(|\mathcal{O}_N^{(i)}|\widehat{\Sigma}^{(i)}_{D^{(i)}}+\zeta^2 I\big)^{-1}\Big\|_{\mathrm{op}}
\le
\frac{1}{\frac{|\mathcal{O}_N^{(i)}|}{2}\mu_{i,D^{(i)}}+\zeta^2}.
\label{eq:opnorm}
\end{align}
Substituting Equation~\eqref{eq:opnorm} and Equation~\eqref{eq:radius_exp_bdd} into Equation~\eqref{eq:head_jensen}
yields, on $\Omega_{i,D^{(i)}}$,
\begin{align}
\EE_{H^{(i)}\sim d^\pi}\!\left[\frac{1}{\zeta^2}k_{i,D^{(i)},\mathcal O_N^{(i)}}(H^{(i)},H^{(i)})\right]
\le
\frac{1}{\frac{|\mathcal{O}_N^{(i)}|}{2}\mu_{i,D^{(i)}}+\zeta^2}.
\label{eq:head_bound}
\end{align}

We now combine the head bound with the tail bound. Plugging Equation~\eqref{eq:head_bound} into
Equation~\eqref{eq:V_split} gives, on $\Omega_{i,D^{(i)}}$,
\begin{align}
\frac{1}{\zeta^2}\mathcal V_i
\le
\frac{1}{\frac{|\mathcal{O}_N^{(i)}|}{2}\mu_{i,D^{(i)}}+\zeta^2}
+
\frac{\kappa_{\psi,i}^2}{\zeta^2}\,B_i(D^{(i)}+1).
\label{eq:V_bound_D}
\end{align}
Substituting Equation~\eqref{eq:V_bound_D} into Equation~\eqref{eq:jensen_reduction} yields, on $\Omega_{i,D^{(i)}}$,
\begin{align}
\EE_{H^{(i)}\sim d^\pi}\big[U_k^{(i)}(H^{(i)})\big]
\le
\beta_k\sqrt{
\frac{1}{\frac{|\mathcal{O}_N^{(i)}|}{2}\mu_{i,D^{(i)}}+\zeta^2}
+
\frac{\kappa_{\psi,i}^2}{\zeta^2}\,B_i(D^{(i)}+1)
}.
\label{eq:Uk_bound_D}
\end{align}

We now optimize right hand side of  Equation~\eqref{eq:Uk_bound_D} with respect to $D^{(i)}$. To do that, we define
effective dimension at depth $i$ as
\begin{align}
d_{*}^{(i)}
=
\min\left\{j\in\mathbb N:\; j \ge \frac{|\mathcal{O}_N^{(i)}|}{\zeta^2}\,B_i(j+1)\right\},
\label{eq:deff_def}
\end{align}
which matches the stated form in \citep{chang2021mitigating} with $n_o=|\mathcal{O}_N^{(i)}|$ and $B_i(j)=\sum_{k=j}^\infty \mu_{i,k}$. By the definition in
Equation~\eqref{eq:deff_def}, we have
\begin{align}
B_i(d_{*}^{(i)}+1)\le \frac{d_{*}^{(i)}\,\zeta^2}{|\mathcal{O}_N^{(i)}|}.
\label{eq:tail_effdim}
\end{align}

Next, consider the followng quantity
\[
g_i(D)\;:=\;\frac{|\mathcal{O}_N^{(i)}|}{\zeta^2}\,B_i(D+1).
\]

With in place and disregarding the relatively minor dependence of the head term on $\mu_{i,D^{(i)}}$ in Equation~\eqref{eq:Uk_bound_D}, the right-hand side is dominated by two main contributions inside the square root: the head contribution, which is of order $1/\zeta^2$, and the tail contribution, which is proportional to $g_i(D) := \frac{|\mathcal{O}_N^{(i)}|}{\zeta^2} B_i(D+1)$, the rescaled tail mass $B_i(D+1)$. The goal of choosing the truncation rank $D$ is to prevent the tail from significantly inflating the overall bound beyond what the head alone would suggest. In these settings, the ``complexity'' or ``scale'' introduced by retaining the first $D$ terms (the head) is naturally on the order of $D$ itself which is the effective number of degrees of freedom or rank that we are already paying for in the bound.

Therefore, truncation is most effective when the tail contribution is controlled relative to this head complexity scale-, specifically, when
\[
g_i(D) \lesssim D.
\]
In other words, we want the rescaled tail mass to be no larger than the truncation rank we have already accepted. The smallest integer $D$ that satisfies this is
\[
D \ge g_i(D),
\]
and this is exactly the effective dimension $d_{*}^{(i)}$ defined in Equation~\eqref{eq:deff_def}.  


Setting $D^{(i)}=d_{*}^{(i)}$ in Equation~\eqref{eq:Uk_bound_D} and then applying Equation~\eqref{eq:tail_effdim} gives, on
$\Omega_{i,d_{*}^{(i)}}$,
\begin{align}
\EE_{H^{(i)}\sim d^\pi}\big[U_k^{(i)}(H^{(i)})\big]
&\le
\beta_k\sqrt{
\frac{1}{\frac{|\mathcal{O}_N^{(i)}|}{2}\mu_{i,d_{*}^{(i)}}+\zeta^2}
+
\frac{\kappa_{\psi,i}^2}{\zeta^2}\,B_i(d_{*}^{(i)}+1)
}
\nonumber\\
&\le
\beta_k\sqrt{
\frac{1}{\frac{|\mathcal{O}_N^{(i)}|}{2}\mu_{i,d_{*}^{(i)}}+\zeta^2}
+
\kappa_{\psi,i}^2\frac{d_{*}^{(i)}}{|\mathcal{O}_N^{(i)}|}
}
\label{eq:Uk_eff_2}
\end{align}

Given this, we define the simultaneous event as
\begin{align}
\Omega_M
:=
\bigcap_{i=0}^{n_{K,b}}\Omega_{i,d_{*}^{(i)}}.
\nonumber
\end{align}
Then, by union bound, we have
\begin{align}
\Pr(\Omega_M)\ge 1-\epsilon.
\nonumber
\end{align}
Therefore, on $\Omega_M$, the bound in Equation~\eqref{eq:Uk_eff_2} holds simultaneously for all
$i\in\{0,1,\ldots,n_{K,b}\}$ with probability at least $1-\epsilon$.

Next, as in the linear setting, we invoke Assumption~\ref{lower bound on partition sizes} which states that there exists a constant $\rho\in(0,1]$ and an event $\Omega_{\mathrm{str}}$ such that
$\Pr(\Omega_{\mathrm{str}})\ge 1-\epsilon$ and, on $\Omega_{\mathrm{str}}$,
\[
\min_{0\le i\le n_{K,b}}|\mathcal{O}_N^{(i)}|\ \ge\ \rho\,NT.
\]

Accordingly, by a union bound, we have 
\begin{align}
\Pr(\widetilde{\Omega}):=\Pr(\Omega \cap\Omega^r \cap \Omega_{\mathrm{str}}\cap \Omega_M) \ \ge\ 1-(n_{K,b}+3)\epsilon -\varepsilon.
\end{align}

Then, we can derive an upper bound $\textup{UQ}$ by applying Equation \eqref{eq:Uk_eff_2} on $\widetilde{\Omega}$, as follows
\begin{align}\label{eq:UQ_bound_cov rkhs}
\textup{UQ}(w^{(0)},a^{(0)},\ldots,w^{(i)},a^{(i)})&\leq  \sum_{k=0}^{K-1}\gamma^k\;
\;\beta_{K-k}\EE_{M_{i+k} \sim \pi^*_{n_{K,b}}}\left[\sqrt{
\frac{1}{\frac{|\mathcal{O}_N^{(M_{i+k})}|}{2}\mu_{M_{i+k},d_{*}^{(M_{i+k})}}+\zeta^2}
+
\kappa_{\psi,M_{i+k}}^2\frac{d_{*}^{(M_{i+k})}}{|\mathcal{O}_N^{(M_{i+k})}|}
}\right] \nonumber\\
&\leq \frac{\;\beta_{\max}}{1-\gamma}\sqrt{
\frac{1}{\frac{\mu}{2}\rho NT+\zeta^2}
+
\kappa_{\psi,*}^2\frac{d_{*}}{\rho NT }
}
\end{align}
and similarly for $C^{(i)}$, we have
\begin{align}\label{eq:C_bound_cov rkhs}
C^{(i)}(w^{(0)},a^{(0)},\ldots,w^{(i)})
&\leq\frac{\;\beta_{\max}}{1-\gamma}\sqrt{
\frac{1}{\frac{\mu}{2}\rho NT+\zeta^2}
+
\kappa_{\psi,*}^2\frac{d_{*}}{\rho NT }
}
\end{align}
where we use the fact that
$\min_{j}|\mathcal{O}_N^{(j)}|\ge \rho NT$, $d_* := \max_{0\leq i \leq n_{K,b}}d_*^{(i)}$, $\mu := \min_{0\leq i \leq n_{K,b}}\mu_{i,d_{*}^{(i)}
}$, $\kappa_{\psi,*}^2 := \max_{0\leq i \leq n_{K,b}}\kappa_{\psi,i}^2$ and  $\tilde{\beta}_{\max}:=\max_{0\le \ell\le K-1}\tilde{\beta}_\ell$.

Then, plugging Equations~\eqref{eq:UQ_bound_cov rkhs} and \eqref{eq:C_bound_cov rkhs} into Equation \eqref{eq: rkhs_bound_on_term_i_2nd fqi}, we can obtain
\begin{sizeddisplay}
    \scriptsize
\begin{align}\label{eq:i_bound_plug_cov rkhs}
i(\widehat{\pi}_K)
&\le
\frac{\;\beta_{\max}}{1-\gamma}\sqrt{
\frac{1}{\frac{\mu}{2}\rho NT+\zeta^2}
+
\kappa_{\psi,*}^2\frac{d_{*}}{\rho NT }
}+
\frac{\;\beta_{\max}}{1-\gamma}\sqrt{
\frac{1}{\frac{\mu}{2}\rho NT+\zeta^2}
+
\kappa_{\psi,*}^2\frac{d_{*}}{\rho NT }
}
2\gamma^{K-1}\frac{2-\gamma}{1-\gamma}R_{\max}
+
\frac{2}{1-\gamma}C_5(\varepsilon)(NT)^{-\delta}
\Bigg)\nonumber\\
&\le
\frac{1}{(1-\gamma)^2}
\left(
\frac{2\beta_{\max}}{1-\gamma}\sqrt{
\frac{1}{\frac{\mu}{2}\rho NT+\zeta^2}
+
\kappa_{\psi,*}^2\frac{d_{*}}{\rho NT }
}
+
2\gamma^{K-1}\frac{2-\gamma}{1-\gamma}R_{\max}
+
\frac{2}{1-\gamma}C_5(\varepsilon)(NT)^{-\delta}
\right),
\end{align}\end{sizeddisplay}
which holds with probability at least $1-(n_{K,b}+3)\epsilon-\varepsilon$.

To state the final rate, we choose $K$ as in the idealized setting i.e.,
\begin{align}
K \;=\; \left\lceil \frac{\log(NT)}{2(1-\gamma)} \right\rceil.\nonumber
\end{align}
This implies that $\sqrt{K-1}<\sqrt{\frac{\log(NT)}{2(1-\gamma)}}$ and $\gamma^{K-1}\le (NT)^{-1/2}$.
Therefore, the truncation contribution satisfies
\[
2\gamma^{K-1}\frac{2-\gamma}{1-\gamma}R_{\max}
\;\le\;
\frac{c_0\,R_{\max}}{1-\gamma}\cdot (NT)^{-1/2}
\]
for an absolute constant $c_0>0$.

Furthermore, choose ${\beta}_{\max}$ in a way consistent with RKHS confidence multipliers,
namely (up to logarithmic factors and constants) as in \citet{chang2021mitigating},
\begin{align}
{\beta}_{\max}
\;\asymp\;
\sqrt{d_{*}\,\Big(d_{*}+\log(1/\epsilon)\Big)}\nonumber
\end{align}

Then, on event $\widetilde{\Omega}$, given Equations~\eqref{eq: rkhs_bound_on_term_i_2nd fqi} and \eqref{eq:i_bound_plug_cov rkhs}
as well as the selection of $K$ and $\beta_{\max}$, we have
\begin{sizeddisplay}
    \small
\begin{align}
\textup{\textbf{Regret}}(\widehat{\pi}_K)
\;\le\;&
\frac{1}{(1-\gamma)^3}
\Bigg(
2\,\sqrt{d_*\big(d_*+\log(1/\epsilon)\big)}\;\sqrt{
\frac{1}{\frac{\mu}{2}\rho NT+\zeta^2}
+
\kappa_{\psi,*}^2\frac{d_{*}}{\rho NT }
}
\;+\;
\frac{c_0 R_{\max}}{\sqrt{NT}}
\;+\;
2\,C_5(\varepsilon)\,(NT)^{-\delta}
\Bigg)\nonumber\\
&\;+\;
\frac{4R_{\max}}{1-\gamma}
\sqrt{\left\lceil \frac{\log(NT)}{2(1-\gamma)} \right\rceil}\;\omega_{NT}.
\end{align}\end{sizeddisplay}

To simplify the expression, we consider the regime in which the truncation term is dominated by the statistical term, namely assume
\begin{align}\label{eq: truncation dom rkhs}
c_0 R_{\max}\;\le\;\frac{2\,\kappa_{\psi,*}\,d_*\,\sqrt{d_*+\log(1/\epsilon)}}{\sqrt{\rho}}.
\end{align}

Under this condition, the truncation contribution is absorbed into the leading statistical term.
Given this, we further use $\lesssim$ notation which hides absolute constants and logarithmic factors
(e.g., $\gamma$, $R_{\max}$ and $\mu$), while keeping $\sqrt{\log(1/\epsilon)}$ explicit.

This yields the final bound on $i(\widehat{\pi}_K)$ and $\textup{\textbf{Regret}}(\widehat{\pi}_K)$ as follows:
\begin{align}\label{eq:i_final_cov_simplified rkhs}
i(\widehat{\pi}_K)
\;\lesssim\;
\sqrt{d_*\big(d_*+\log(1/\epsilon)\big)}\;
\sqrt{\frac{1+d_*}{\rho\,NT}}
\;+\;
C_5(\varepsilon)\,(NT)^{-\delta}
\end{align}
and
\begin{align}\label{eq: final regret rkhs}
\textup{\textbf{Regret}}(\widehat{\pi}_K)
\;\lesssim\;&
\sqrt{d_*\big(d_*+\log(1/\epsilon)\big)}\;
\sqrt{\frac{1+d_*}{\rho\,NT}}
\;+\;
C_5(\varepsilon)(NT)^{-\delta}
\;+\;
\omega_{NT},
\end{align}
which holds with probability at least $1-(n_{K,b}+3)\epsilon-\varepsilon$.

\subsection{Step 3: Combined Regret Bound under the Relaxation of Assumption~\ref{ass:linear-mdp}}
\label{Step-3-Combine-RKHS-Template}

We now combine the results from the previous two steps to derive the final bound.
First, recall that under the event
$\widetilde{\Omega}_P$,
Equation~\eqref{eq: pess rkhs final regret} yields
\begin{sizeddisplay}
{\footnotesize}
 \begin{align}\label{eq: combined regret tilde rkhs}
\textup{\textbf{Regret}}(\widetilde{\pi}_K)
\;\lesssim\;&d_*\big(d_*+\log(1/\epsilon)\big)
\sqrt{\frac{C_{\mathrm{cov}}\,}{\rho NT}}\;
+
C_5(\varepsilon)(NT)^{-\delta}
\;+\;
\omega_{NT},
\end{align}
\end{sizeddisplay}
which holds with probability at least $ 1-(n_{K,b}+3)\epsilon -\varepsilon$.

Similarly, under the event
$\widetilde{\Omega}$,
Equation~\eqref{eq: final regret rkhs} gives
\begin{sizeddisplay}
{\footnotesize}
 \begin{align}\label{eq: combined regret hat rkhs}
   \textup{\textbf{Regret}}(\widehat{\pi}_K)
\;\lesssim\;&
\sqrt{d_*\big(d_*+\log(1/\epsilon)\big)}\;
\sqrt{\frac{1+d_*}{\rho\,NT}}
\;+\;
C_5(\varepsilon)(NT)^{-\delta}
\;+\;
\omega_{NT},
\end{align}
\end{sizeddisplay}
which holds with probability at least $ 1-(n_{K,b}+3)\epsilon -\varepsilon$.

Define the simultaneous event
\begin{align}\label{eq: simultaneous event rkhs}
\Omega_{\mathrm{all}}
\;:=\;
\widetilde{\Omega}_P \cap \widetilde{\Omega}.
\end{align}
By a union bound,
\begin{align}\label{eq: prob Omega all rkhs}
\Pr(\Omega_{\mathrm{all}})
\;\ge\;
1
-
\Pr(\widetilde{\Omega}_P^{c})
-
\Pr(\widetilde{\Omega}^{c})
\;\ge\;
1-2\big((n_{K,b}+3)\epsilon+\varepsilon\big),
\end{align}
and on $\Omega_{\mathrm{all}}$ both regret bounds in
Equations~\eqref{eq: combined regret tilde rkhs} and~\eqref{eq: combined regret hat rkhs}
hold simultaneously.

We define the statistical error term $\mathcal{E}_{\text{stat}}(NT, d)$ by
\begin{align}\label{eq: def stat error rkhs}
\overline{\mathcal{E}}_{\text{stat}}(NT, d_*) :=
\max\left(
d_*\big(d_*+\log(1/\epsilon)\big)
\sqrt{\frac{C_{\mathrm{cov}}}{\rho NT}},\ 
\sqrt{d_*\big(d_*+\log(1/\epsilon)\big)}\;
\sqrt{\frac{1+d_*}{\rho\,NT}}
\right)
+
C_5(\varepsilon)(NT)^{-\delta}
+
\omega_{NT}
\end{align}

Therefore, on $\Omega_{\mathrm{all}}$,
\begin{align}\label{eq: max regret final rkhs}
\max\!\big(\textup{\textbf{Regret}}(\widehat{\pi}_K),\ \textup{\textbf{Regret}}(\widetilde{\pi}_K)\big)
\;\lesssim\;
\overline{\mathcal{E}}_{\text{stat}}(NT, d),
\end{align}
which holds with probability at least
$1-2\big((n_{K,b}+3)\epsilon+\varepsilon\big)$. This concludes the proof of Theorem~\ref{cor:ideal}.

\section{Proof of Theorem~\ref{thm:general_regret}} \label{app:proof-general-theorem}

In this section, we provide the proof of Theorem~\ref{thm:general_regret}. This proof synthesizes the analyses of the individual layers presented in Theorem~\ref{cor:ideal} and Corollaries \ref{cor:relaxed_coverage} (Coverage), \ref{cor:relaxed-TAD} (TAD), \ref{cor:relaxed-AD} (BAP), and \ref{cor:relaxed-structure} (Linear Structure).


\subsection{Step 1: Regret for PC-FQI under the Relaxation}\label{Step 1: Regret for PC-FQI under the Relaxation}

In this section, we relax Assumptions~\ref{assump:sufficient-coverage-main-text}-\ref{ass:linear-mdp} all together. Consequently, it is possible that $n > n_{K,b}$, meaning the optimal policy $\pi^*$ may induce censoring streaks longer than those supported by the offline dataset. In this case, we invoke Lemma~\ref{lm: regret decom}, which does not rely on Assumption~\ref{assump:sufficient-coverage-main-text}, together with the fact that the learned policy $\widetilde{\pi}_K \in \Pi_{n_{K,b}}$. Therefore, by Lemma~\ref{lm: regret decom}, the regret of $\widetilde{\pi}_K$ satisfies
\begin{sizeddisplay}
{\footnotesize}
 \begin{align}\label{eq: final corr regret 1_app}
    \text{Regret}(\widetilde{\pi}_K) 
    &\leq\left(\underbrace{\sup_{\pi' \in \Pi_{n_{K,b}}} \EE^{\pi'}\!\left[\sum_{t=0}^{\infty}\gamma^tR_t\right] -\EE^{\tilde{\pi}_K} \!\left[ \sum_{t=0}^{\infty} \gamma^t R_t  \right]}_{i(\tilde{\pi}_K)} 
    +2R_{\max}\frac{\gamma^{K+1}}{1-\gamma}\right)\prob^{\pi^*}(\mathcal{C}(K, n_{K,b}))  \nonumber\\
    &\quad +(K-n_{K,b}+1)(1-\alpha_{\pi^*,1})^{n_{K,b}}\left(\frac{2R_{\max}}{1-\gamma}\right)\nonumber\\
    &\quad+\frac{4R_{\max}\sqrt{K-1}}{1-\gamma}\omega_{NT}. 
\end{align}
\end{sizeddisplay}

Consider the term $i(\widetilde{\pi}_K)$:
\begin{align*}
    i(\widetilde{\pi}_K)
    &=\sup_{\pi' \in \Pi_{n_{K,b}}} \EE^{\pi'}\!\left[\sum_{t=0}^{\infty}\gamma^tR_t\right] -\EE^{\tilde{\pi}_K} \!\left[ \sum_{t=0}^{\infty} \gamma^t R_t  \right]\\
    &=\EE^{\pi^*_{n_{K,b}}}\!\left[\sum_{t=0}^{\infty}\gamma^tR_t\right] -\EE^{\tilde{\pi}_K} \!\left[ \sum_{t=0}^{\infty} \gamma^t R_t  \right],
\end{align*}
where $\pi^*_{n_{K,b}} \in \argmax_{\pi' \in \Pi_{n_{K,b}}} \EE^{\pi'}\!\left[\sum_{t=0}^{\infty}\gamma^tR_t\right]$.

Observe that \(i(\widetilde{\pi}_K)\) has the same structure as the term analyzed in Step~\ref{PC-FQI under Idealized Setting} of the idealized setting (see Equation~\eqref{eq: pes regret 1}, except that \(\pi^*\) is replaced by \(\pi^*_{n_{K,b}}\). Under Assumption~\ref{assump:sufficient-coverage-main-text}, we have \(\pi^* \in \Pi_n \subseteq \Pi_{n_{K,b}}\) and
\[
\pi^*_{n_{K,b}} \in \argmax_{\pi' \in \Pi_{n_{K,b}}} \EE^{\pi'}\!\left[\sum_{t=0}^{\infty}\gamma^t R_t\right].
\]
It follows that \(\pi^*\) is also optimal over \(\Pi_{n_{K,b}}\), and hence \(\pi^*\) and \(\pi^*_{n_{K,b}}\) are equivalent. However, with the relaxation of Assumption~\ref{assump:sufficient-coverage-main-text}, this no longer holds, and the equivalence between \(\pi^*\) and \(\pi^*_{n_{K,b}}\) fail. As a result, we have unlearnable part of $\pi^*$ which is the term $(K-n_{K,b}+1)(1-\alpha_{\pi^*,1})^{n_{K,b}}\left(\frac{2R_{\max}}{1-\gamma}\right)$ in Equation \eqref{eq: final corr regret 1_app}.

However, Lemma~\ref{lm: performance difference} and the analysis in Step~\ref{PC-FQI under Idealized Setting} of the idealized setting only require that both policies belong to $\Pi_{n_{K,b}}$. Hence, the same bound on \(i(\widetilde{\pi}_K)\) continues to hold. The only modifications are: (i) the additional additive terms $\mathrm{TAP}(\widetilde{\pi}_K)$ and $\mathrm{BAP}(\widetilde{\pi}_K)$, and (ii) replacing $d$ by $d_*$ (i.e., effective dimension in RKHS setting), as implied by Equations \eqref{eq: tadrel_i_final_pc}, \eqref{eq: adrel_i_final_pc}, and \eqref{eq: pess rkhs i_final_cov_simplified} from Sections~\ref{Step 1: Regret for PC-FQI under the Relaxation of tad} (i.e., the relaxation of Assumption~\ref{ass:TAD}), \ref{Step 1: Regret for PC-FQI under the Relaxation of ad} (i.e., the relaxation of Assumption~\ref{ass:AD}) and \ref{Step-1-Regret-RKHS} (i.e., the relaxation of  Assumption~\ref{ass:linear-mdp}), respectively.

In particular, by combining Equations  \eqref{eq: tadrel_i_final_pc}, \eqref{eq: adrel_i_final_pc} and \eqref{eq: pess rkhs i_final_cov_simplified}, we obtain that with probability at least  $1-(n_{K,b}+3)\epsilon -\varepsilon$,
\begin{align}\label{eq: final corr_i_bound}
i(\widetilde{\pi}_K)
\;\lesssim\;d_*\big(d_*+\log(1/\epsilon)\big)
\sqrt{\frac{C_{\mathrm{cov}}\,}{\rho NT}}\;
+
C_5(\varepsilon)(NT)^{-\delta}+\mathrm{TAP}(\widetilde{\pi}_K) + \mathrm{BAP}(\widetilde{\pi}_K).
\end{align}

Plugging Equation \eqref{eq: final corr_i_bound} into Equation \eqref{eq: final corr regret 1_app} yields, with probability at least  $1-(n_{K,b}+3)\epsilon -\varepsilon$,
\begin{sizeddisplay}
{\footnotesize}
 \begin{align}\label{eq: final corr regret 1_plug}
    \text{Regret}(\widetilde{\pi}_K) 
    &\lesssim\left(d_*\big(d_*+\log(1/\epsilon)\big)
\sqrt{\frac{C_{\mathrm{cov}}\,}{\rho NT}}\;
+
C_5(\varepsilon)(NT)^{-\delta}+\mathrm{TAP}(\widetilde{\pi}_K) + \mathrm{BAP}(\widetilde{\pi}_K)\right.\nonumber\\
&\left.
    +2R_{\max}\frac{\gamma^{K+1}}{1-\gamma}
    \right)\prob^{\pi^*}(\mathcal{C}(K, n_{K,b}))  \nonumber\\
    &\quad +(K-n_{K,b}+1)(1-\alpha_{\pi^*,1})^{n_{K,b}}\left(\frac{2R_{\max}}{1-\gamma}\right)
    +\omega_{NT}. 
\end{align}
\end{sizeddisplay}
We select $K$ as in the idealized setting:
\begin{align}\label{eq: final corr_K_choice}
K \;=\; \left\lceil \frac{\log(NT)}{2(1-\gamma)} \right\rceil,
\end{align}
so that $\gamma^{K+1}\le \gamma^{K-1}\lesssim (NT)^{-1/2}$. Consequently, the truncation contribution satisfies
\begin{align}\label{eq: final corr_trunc_simplify}
2R_{\max}\frac{\gamma^{K+1}}{1-\gamma}
\;\lesssim\;
\frac{R_{\max}}{1-\gamma}\cdot (NT)^{-1/2}.
\end{align}

To simplify the expression further for illustration purposes, we consider the same regime in which the truncation term is dominated by the statistical term (i.e., Equation~\eqref{eq: truncation dom rkhs pfqi}).


Under this condition, the truncation contribution is absorbed into the leading statistical term in Equation~\eqref{eq: final corr regret 1_plug}. Finally, without loss of generality and for the sake of presentation, we assume that for the chosen $K$ there exists a constant $c_1\in(0,1]$ such that
$\prob^{\pi^\ast}\!\big(\mathcal{C}(K,n_{K,b})\big)\ \ge\ c_1$. This condition ensures that any additive term of order $\omega_{NT}$ can be absorbed into
$\prob^{\pi^\ast}(\mathcal{C}(K,n_{K,b}))$ up to a constant factor $1/c_1$.

Substituting Equations \eqref{eq: final corr_trunc_simplify} into Equation \eqref{eq: final corr regret 1_plug} yields, with probability at least $1-(n_{K,b}+3)\epsilon -\varepsilon$,
\begin{sizeddisplay}
{\footnotesize}
\begin{align}\label{eq: final corr regret 1_final}
\textup{\textbf{Regret}}(\widetilde{\pi}_K)
&\lesssim
\left(d_*\big(d_*+\log(1/\epsilon)\big)
\sqrt{\frac{C_{\mathrm{cov}}\,}{\rho NT}}\;
+
C_5(\varepsilon)(NT)^{-\delta}+\mathrm{TAP}(\widetilde{\pi}_K) + \mathrm{BAP}(\widetilde{\pi}_K)+\omega_{NT}
    \right)\prob^{\pi^*}(\mathcal{C}(K, n_{K,b}))\nonumber\\
&+(K-n_{K,b}+1)(1-\alpha_{\pi^*,1})^{n_{K,b}}\left(\frac{2R_{\max}}{1-\gamma}\right)
\end{align}
\end{sizeddisplay}

\subsection{Step 2: Regret for C-FQI under the Relaxation}
\label{Step 2: Regret for C-FQI under the Relaxation}

In this section, we relax Assumptions~\ref{assump:sufficient-coverage-main-text}-\ref{ass:linear-mdp} all together. Consequently, it is possible that $n>n_{K,b}$, meaning the optimal policy $\pi^*$ may induce censoring streaks longer than those supported by the offline dataset. In this case, we invoke Lemma~\ref{lm: regret decom}, which does not rely on Assumption~\ref{assump:sufficient-coverage-main-text}, together with the fact that the learned policy $\widehat{\pi}_K \in \Pi_{n_{K,b}}$. Therefore, by Lemma~\ref{lm: regret decom}, the regret of $\widehat{\pi}_K$ satisfies
\begin{sizeddisplay}
{\footnotesize}
 \begin{align}\label{eq: final corr regret 2_app}
    \text{Regret}(\widehat{\pi}_K) 
    &\leq\left(\underbrace{\sup_{\pi' \in \Pi_{n_{K,b}}} \EE^{\pi'}\!\left[\sum_{t=0}^{\infty}\gamma^tR_t\right] -\EE^{\hat{\pi}_K} \!\left[ \sum_{t=0}^{\infty} \gamma^t R_t  \right]}_{i(\widehat{\pi}_K)} 
    +2R_{\max}\frac{\gamma^{K+1}}{1-\gamma}\right)\prob^{\pi^*}(\mathcal{C}(K, n_{K,b}))  \nonumber\\
    &\quad +(K-n_{K,b}+1)(1-\alpha_{\pi^*,1})^{n_{K,b}}\left(\frac{2R_{\max}}{1-\gamma}\right)\nonumber\\
    &\quad+\frac{4R_{\max}\sqrt{K-1}}{1-\gamma}\omega_{NT}. 
\end{align}
\end{sizeddisplay}

Consider the term $i(\widehat{\pi}_K)$:
\begin{align*}
    i(\widehat{\pi}_K)
    &=\sup_{\pi' \in \Pi_{n_{K,b}}} \EE^{\pi'}\!\left[\sum_{t=0}^{\infty}\gamma^tR_t\right] -\EE^{\hat{\pi}_K} \!\left[ \sum_{t=0}^{\infty} \gamma^t R_t  \right]\\
    &=\EE^{\pi^*_{n_{K,b}}}\!\left[\sum_{t=0}^{\infty}\gamma^tR_t\right] -\EE^{\hat{\pi}_K} \!\left[ \sum_{t=0}^{\infty} \gamma^t R_t  \right],
\end{align*}
where $\pi^*_{n_{K,b}} \in \argmax_{\pi' \in \Pi_{n_{K,b}}} \EE^{\pi'}\!\left[\sum_{t=0}^{\infty}\gamma^tR_t\right]$.

Observe that \(i(\widehat{\pi}_K)\) has the same structure as the term analyzed in Step~\ref{C-FQI under Idealized Setting} of the idealized setting, except that \(\pi^*\) is replaced by \(\pi^*_{n_{K,b}}\). Under Assumption~\ref{assump:sufficient-coverage-main-text}, we have \(\pi^* \in \Pi_n \subseteq \Pi_{n_{K,b}}\), and it follows that \(\pi^*\) is also optimal over \(\Pi_{n_{K,b}}\). Hence \(\pi^*\) and \(\pi^*_{n_{K,b}}\) are equivalent. However, with the relaxation of Assumption~\ref{assump:sufficient-coverage-main-text}, this equivalence can fail, and the resulting unlearnable part of $\pi^*$ is captured by the additive term $(K-n_{K,b}+1)(1-\alpha_{\pi^*,1})^{n_{K,b}}\left(\frac{2R_{\max}}{1-\gamma}\right)$ in Equation~\eqref{eq: final corr regret 2_app}.

Nevertheless, Lemma~\ref{lm: performance difference} (the performance difference lemma) and the analysis in Step~\ref{C-FQI under Idealized Setting} only require that both policies belong to $\Pi_{n_{K,b}}$. Hence, the same bound on \(i(\widehat{\pi}_K)\) continues to hold. The only modifications are: (i) the additional additive terms $\mathrm{TAP}(\widehat{\pi}_K)$ and $\mathrm{BAP}(\widehat{\pi}_K)$, and (ii) replacing $d$ by $d_*$ with slight change in the composition, as implied by Equations \eqref{eq: tadrel_i_final_c}, \eqref{eq: adrel_i_final_c}, and \eqref{eq:i_final_cov_simplified rkhs} from Sections~\ref{Step 2: Regret for C-FQI under the Relaxation of tad} (i.e., the relaxation of Assumption~\ref{ass:TAD}), \ref{Step 2: Regret for C-FQI under the Relaxation of ad} (i.e., the relaxation of Assumption~\ref{ass:AD}) and \ref{C-FQI under RKHS} (i.e., the relaxation of  Assumption~\ref{ass:linear-mdp}), respectively.

In particular, by combining Equations  \eqref{eq: tadrel_i_final_c}, \eqref{eq: adrel_i_final_c} and \eqref{eq:i_final_cov_simplified rkhs} (i.e.,  under the relaxations of Assumptions~\ref{ass:TAD}, \ref{ass:AD}, and \ref{ass:linear-mdp}), we obtain that with probability at least  $1-(n_{K,b}+3)\epsilon -\varepsilon$,
\begin{align}\label{eq: final corr_i_bound_hat}
i(\widehat{\pi}_K)
\;\lesssim\;
\sqrt{d_*\big(d_*+\log(1/\epsilon)\big)}\;
\sqrt{\frac{1+d_*}{\rho\,NT}}
\;+\;
C_5(\varepsilon)\,(NT)^{-\delta}
+\mathrm{TAP}(\widehat{\pi}_K) + \mathrm{BAP}(\widehat{\pi}_K).
\end{align}

Plugging Equation~\eqref{eq: final corr_i_bound_hat} into Equation~\eqref{eq: final corr regret 2_app} yields, with probability at least  $1-(n_{K,b}+3)\epsilon-\varepsilon$,
\begin{sizeddisplay}
{\footnotesize}
 \begin{align}\label{eq: final corr regret 2_plug}
    \text{Regret}(\widehat{\pi}_K) 
    &\lesssim\left( \sqrt{d_*\big(d_*+\log(1/\epsilon)\big)}\;
\sqrt{\frac{1+d_*}{\rho\,NT}}
\;+\;
C_5(\varepsilon)\,(NT)^{-\delta}
+\mathrm{TAP}(\widehat{\pi}_K) + \mathrm{BAP}(\widehat{\pi}_K)\right.\nonumber\\
&\left.
    +2R_{\max}\frac{\gamma^{K+1}}{1-\gamma}
    \right)\prob^{\pi^*}(\mathcal{C}(K, n_{K,b}))  \nonumber\\
    &\quad +(K-n_{K,b}+1)(1-\alpha_{\pi^*,1})^{n_{K,b}}\left(\frac{2R_{\max}}{1-\gamma}\right)
    +\omega_{NT}. 
\end{align}
\end{sizeddisplay}

We select $K$ as in the idealized analysis:
\begin{align}\label{eq: final corr_K_choice_hat}
K \;=\; \left\lceil \frac{\log(NT)}{2(1-\gamma)} \right\rceil,
\end{align}
so that $\gamma^{K+1}\le \gamma^{K-1}\lesssim (NT)^{-1/2}$. Consequently, the truncation contribution satisfies
\begin{align}\label{eq: final corr_trunc_simplify_hat}
2R_{\max}\frac{\gamma^{K+1}}{1-\gamma}
\;\lesssim\;
\frac{R_{\max}}{1-\gamma}\cdot (NT)^{-1/2}.
\end{align}

To simplify the expression further for illustration purposes, we consider the same regime in which the truncation term is dominated by the statistical term (i.e., Equation~\eqref{eq: truncation dom rkhs}).


Under this condition, the truncation contribution is absorbed into the leading statistical term in Equation~\eqref{eq: final corr regret 1_plug}. Finally, without loss of generality and for the sake of presentation, we assume that for the chosen $K$ there exists a constant $c_1\in(0,1]$ such that
$\prob^{\pi^\ast}\!\big(\mathcal{C}(K,n_{K,b})\big)\ \ge\ c_1$. This condition ensures that any additive term of order $\omega_{NT}$ can be absorbed into
$\prob^{\pi^\ast}(\mathcal{C}(K,n_{K,b}))$ up to a constant factor $1/c_1$.

Substituting Equations~\eqref{eq: final corr_trunc_simplify_hat} into Equation~\eqref{eq: final corr regret 2_plug} yields, with probability at least $1-(n_{K,b}+3)\epsilon-\varepsilon$,
\begin{sizeddisplay}
{\footnotesize}
\begin{align}\label{eq: final corr regret 2_final}
\textup{\textbf{Regret}}(\widehat{\pi}_K)
&\lesssim
\left(\sqrt{d_*\big(d_*+\log(1/\epsilon)\big)}\;
\sqrt{\frac{1+d_*}{\rho\,NT}}
\;+\;
C_5(\varepsilon)\,(NT)^{-\delta}
+\mathrm{TAP}(\widehat{\pi}_K) + \mathrm{BAP}(\widehat{\pi}_K)
+\omega_{NT}
\right)\prob^{\pi^*}(\mathcal{C}(K, n_{K,b})) \nonumber\\
&\quad
+(K-n_{K,b}+1)(1-\alpha_{\pi^*,1})^{n_{K,b}}\left(\frac{2R_{\max}}{1-\gamma}\right).
\end{align}
\end{sizeddisplay}

\subsection{Step 3: Combined Regret Bound under the Relaxation}\label{Step 3: Combined Regret Bound under the Relaxation}

We now combine the results from Equation~\eqref{eq: final corr regret 1_final} (PC-FQI) and
Equation~\eqref{eq: final corr regret 2_final} (C-FQI) under the simultaneous relaxation of
Assumptions~\ref{assump:sufficient-coverage-main-text}--\ref{ass:linear-mdp}.

Recall the definition of the unidentifiable regret term from Equation~\eqref{eq: def Ecov}
\begin{align*}
\mathcal{E}_{\textup{cov}}(K, n_{K,b})
\;:=\;
(K-n_{K,b}+1)(1-\alpha_{\pi^*,1})^{n_{K,b}}\left(\frac{2R_{\max}}{1-\gamma}\right).
\end{align*}

Recall that in the idealized setting (Theorem~\ref{cor:ideal}) statistical error term defined the as
\[
\mathcal{E}_{\text{stat}}(NT, d) :=
\max\left(
d\big(d+\log(1/\epsilon)\big)
\sqrt{\frac{C_{\mathrm{cov}}}{\rho NT}},\ 
d
\sqrt{\frac{\big(d+\log(1/\epsilon)\big)d}{\rho\,NT}}
\right)
+
C_5(\varepsilon)(NT)^{-\delta}
+
\omega_{NT}
\]

In the present relaxed setting, the statistical term replaces the dimension
$d$ by the effective dimension $d_*$. Additionally, there is a slight modification in the regret of C-FQI in RKHS setting compared to idealized setting. Therefore, we define statistical error term in Equation~\eqref{eq: def stat error rkhs} (i.e., in RKHS setting) as
\begin{align}
\overline{\mathcal{E}}_{\text{stat}}(NT, d_*) :=
\max\left(
d_*\big(d_*+\log(1/\epsilon)\big)
\sqrt{\frac{C_{\mathrm{cov}}}{\rho NT}},\ 
\sqrt{d_*\big(d_*+\log(1/\epsilon)\big)}\;
\sqrt{\frac{1+d_*}{\rho\,NT}}
\right)
+
C_5(\varepsilon)(NT)^{-\delta}
+
\omega_{NT}\nonumber
\end{align}

Given this, recall that
Equation~\eqref{eq: final corr regret 1_final} yields
\begin{sizeddisplay}
{\footnotesize}
 \begin{align}\label{eq: combined regret tilde rkhs final}
\textup{\textbf{Regret}}(\widetilde{\pi}_K)
&\lesssim
\left(d_*\big(d_*+\log(1/\epsilon)\big)
\sqrt{\frac{C_{\mathrm{cov}}\,}{\rho NT}}\;
+
C_5(\varepsilon)(NT)^{-\delta}+\mathrm{TAP}(\widetilde{\pi}_K) + \mathrm{BAP}(\widetilde{\pi}_K)+\omega_{NT}
    \right)\prob^{\pi^*}(\mathcal{C}(K, n_{K,b}))\nonumber\\
&+(K-n_{K,b}+1)(1-\alpha_{\pi^*,1})^{n_{K,b}}\left(\frac{2R_{\max}}{1-\gamma}\right)
\end{align}
\end{sizeddisplay}
which holds with probability at least $ 1-(n_{K,b}+3)\epsilon -\varepsilon$.

Similarly,
Equation~\eqref{eq: final corr regret 2_final} gives
\begin{sizeddisplay}
{\footnotesize}
 \begin{align}\label{eq: combined regret hat rkhs final}
   \textup{\textbf{Regret}}(\widehat{\pi}_K)
&\lesssim
\left(\sqrt{d_*\big(d_*+\log(1/\epsilon)\big)}\;
\sqrt{\frac{1+d_*}{\rho\,NT}}
\;+\;
C_5(\varepsilon)\,(NT)^{-\delta}
+\mathrm{TAP}(\widehat{\pi}_K) + \mathrm{BAP}(\widehat{\pi}_K)
+\omega_{NT}
\right)\prob^{\pi^*}(\mathcal{C}(K, n_{K,b})) \nonumber\\
&\quad
+(K-n_{K,b}+1)(1-\alpha_{\pi^*,1})^{n_{K,b}}\left(\frac{2R_{\max}}{1-\gamma}\right).
\end{align}
\end{sizeddisplay}
which holds with probability at least $ 1-(n_{K,b}+3)\epsilon -\varepsilon$.

Then by union bound both regret bounds in
Equations~\eqref{eq: combined regret tilde rkhs final} and~\eqref{eq: combined regret hat rkhs final}
hold simultaneously with probability at least $1-2\big((n_{K,b}+3)\epsilon+\varepsilon\big)$. This yields 
\begin{align}
\max(\textup{\textbf{Regret}}(\widehat \pi_K),\ \textup{\textbf{Regret}}(\widetilde \pi_K))\;\lesssim\; 
&\underbrace{\prob^{\pi^*}(\mathcal{C}(K, n_{K,b}))}_{\text{Learnability Probability}} \cdot \left[ 
\underbrace{\overline{\mathcal{E}}_{\text{stat}}(NT, d_*)}_{\text{Statistical Error}} \nonumber \right. \\
&\left.+ \underbrace{\max(\mathrm{TAP}(\widehat \pi_K),\mathrm{TAP}(\widetilde \pi_K)) + \max(\mathrm{BAP}(\widehat \pi_K),\mathrm{BAP}(\widetilde \pi_K))}_{\text{Alignment Discrepancy}} 
\right] \nonumber \\
&+ \underbrace{\mathcal{E}_{\text{cov}}(K, n_{K,b})}_{\text{Unidentifiable Regret}},
\end{align}
This concludes the proof of Theorem~\ref{thm:general_regret}.

\section{Key Lemmas}\label{appendix: key lemmas}

Before presenting key lemmas, we introduce several related notation and definitions. First of all, throughout this section, let $W_t=(X_t,Y_t,Z_{t-1},\Delta_{t-1})$ and $A_t=(P_t,O_t)$. Also, let $w^{(t)}$ and $a^{(t)}$ be generic realizations of the random vectors $W_t$ and $A_t$, respectively.
\begin{definition}[Effective Dimension]
\label{def:effective_dim}
Consider a continuous positive semidefinite kernel $k(\cdot, \cdot)$ on the space $\mathcal{M}$ such that $k(x,x) \le 1$ for all $x \in \mathcal{M}$. Let $\{(\mu_i, \psi_i)\}_{i=1}^\infty$ denote the pairs of eigenvalues and eigenfunctions of $k$ with respect to the offline data distribution, sorted in non-increasing order such that $\mu_1 \ge \mu_2 \ge \dots$. The effective dimension $d_*$ is defined as the smallest integer $j$ satisfying the following condition:
\begin{equation}
    d_* := \min \left\{ j \in \mathbb{N} : j \ge \frac{n}{\zeta^2} \sum_{k=j+1}^\infty \mu_k \right\},
\end{equation}
where $n$ denotes the sample size of the offline dataset and $\zeta^2$ is the noise variance parameter associated with the transition model. This definition is consistent with \cite{chang2021mitigating}.
\end{definition}

\begin{definition}[Maximum Information Gain]\label{def:info_gain}
Let $\mathcal{D} = \mathcal{S} \times \mathcal{A}$ be the domain of all possible state-action pairs and let $k(\cdot,\cdot)$ be a positive semi-definite kernel. For a sample budget $NT$, the maximum information gain is defined as
\[
\mathcal{I}_{NT} := \max_{\mathcal{Z} \subset \mathcal{D},\, |\mathcal{Z}|=NT} 
\frac{1}{2} \log \det \left( I + \lambda^{-1} \mathbf{G}_{\mathcal{Z}} \right),
\]
where the maximum is taken over all subsets $\mathcal{Z} = \{z_1,\dots,z_{NT}\} \subset \mathcal{D}$ with the cardinality $|\cal Z| = NT$, 
$\mathbf{G}_{\mathcal{Z}} \in \mathbb{R}^{NT \times NT}$ is the Gram matrix with entries 
$[\mathbf{G}_{\mathcal{Z}}]_{ij} = k(z_i, z_j)$, 
$\lambda > 0$ is the regularization parameter, and $I$ is the identity matrix.
\end{definition}

\begin{definition}\label{def: set of all polcies}
    Denote $\Pi$ as the set of all policies, i.e.,
    $$\Pi = \{ \pi = (\pi_0, \cdots, \pi_t, \cdots) \given \pi_t : H^{'}_t \rightarrow \Delta(\calA), \forall t\geq 0  \},$$
    where $H^{'}_t=(W_0,A_0,\cdots,W_t)$. In other words, $\pi_t$ specifies a probability distribution over the action space conditioned on the history of observations at the decision point $t$. Therefore, $\pi_t(a\given H^{'}_t)$ denotes the probability that the action $a$ will be selected at time $t$ given the history. 
\end{definition}
\begin{definition}\label{def: actions leading uncensored state rigorous}
For all $t, n \geq 0$ and $\{W_{t+n}, A_{t+n-1}, W_{t+n-1}, \ldots, A_{t}, W_{t}\}$ such that $\{\Delta_i\}_{i =t}^{t+n-1} = 0$ and $\Delta_{t-1}=1$, define the set, named Uncensoring Pathways (UP), as 
\begin{align*}
    &\textup{UP}^{(n)}(W_{t+n}, A_{t+n-1}, W_{t+n-1}, \ldots, A_{t}, W_{t}) \\
    &= \left\{ a \in \mathcal{A} \mid \prob\left(\Delta_{t+n} = 1 \mid A_{t+n}=a, W_{t+n}, \ldots, A_{t}, W_{t} \right) = 1 \right\}.
\end{align*}
\end{definition}

\begin{definition}\label{def: n_{T',b} and n_{T',*}}
    For any $T'\geq0$, $n_{T',b}$ and $n_{T',\ast}$ are defined as follows:
    \begin{align}
        &n_{T',b} = \sup_{\tau \in \mathcal{T}(\pi^b)} \left( \sup_{t \in \mathbb{N}} \sum_{k=t}^{t+T'-1} (1 - \Delta_k) \right)\nonumber\\
        &n_{T',\ast} = \sup_{\tau \in \mathcal{T}(\pi^\ast)} \left( \sup_{t \in \mathbb{N}} \sum_{k=t}^{t+T'-1} (1 - \Delta_k) \right)\nonumber,
    \end{align}
   where $\mathcal{T}(\pi^b) $ and $\mathcal{T}(\pi^\ast) $ represent the set of all possible trajectories generated by $ \pi^b $ and $\pi^\ast$, respectively.
\end{definition}

\begin{definition}\label{def: ith value func}
For any $ \pi \in \Pi_{n}$ with $n \geq 0$ and $0\leq i\leq n$, the $i^{th}$ state value function conditioned on the event $\mathcal{C}(K,n)$ is defined as 
    \begin{align}
    &V^{(i)}_{\pi.\mathcal{C}}(w^{(0)},a^{(0)},\cdots,w^{(i)})\nonumber\\
=&\mathbb{E}^{\pi}\left[\sum_{t'=i}^{\infty} \gamma^{t'-i} R_{t'} \mid (W_{0},A_0,\cdots,W_{i})= (w^{(0)},a^{(0)},\cdots,w^{(i)}),\mathcal{C}(K,n)\right] \nonumber\\
=&\mathbb{E}^{\pi}\left[\sum_{t'=i}^{\infty} \gamma^{t'-i} R_{t'} \mid (W_{0},A_0,\cdots,W_{i})= (w^{(0)},a^{(0)},\cdots,w^{(i)})\right],
    \end{align}
   where, in the component of $W_{j}$, we have $\Delta_{j-1}=0$ for all $j=1,\cdots,i$, and in the component of $W_{0}$, we have $\Delta_{-1}=1$
\end{definition}
Note that the event $\mathcal{C}(K,n)$ is a sure event under any policy $\pi \in \Pi_{n}$. That is why in the second equality, the condition on the expectation  does not involve $\mathcal{C}(K,n)$.
\begin{definition}\label{def: optimal under pi nkb}
    \begin{align}
    \pi_{n_{K,b}}^*(W_{t},\cdots,W_{t-n_{K,b}},A_{t-n_{K,b}}) = \left\{ \sum_{i=0}^{n_{K,b}} \pi^{(i)}_{n_{K,b},\ast}(a' \mid W_{t}, \cdots, W_{t-i}, A_{t-i}) \prod_{j=1}^{i}\left\{\mathbb{I}[\Delta_{t-j}=0]\right\}\mathbb{I}[\Delta_{t-i-1}=1] \right\}\nonumber,
\end{align}
where
$\forall i =0,\cdots,n_{K,b}-1$ and $\forall t \geq0$
\begin{align}
    \pi^{(i)}_{n_{K,b},\ast}(a' \mid W_{t}, \cdots, W_{t-i}, A_{t-i}) &=\sup_{a \in \calA} \mathbb{E}\left[ R_{t} + \gamma I[\Delta_{t}=1] V^{(0)}_{*}(W_{t+1}) \right.\nonumber\\
    &\left.+ \gamma I[\Delta_{t}=0] V^{(i+1)}_{*}(W_t, A_t,\cdots,W_{t+1}) \right.\nonumber\\
&\left.\quad\mid W_{t-i}, A_{t-i},\cdots,W_{t},a,\right] \nonumber
\end{align}
and for $i=n_{K,b}$
\begin{align}
    \pi^{(n_{K,b})}_{n_{K,b},\ast}(a' \mid W_{t}, \cdots, W_{t-n_{K,b}}, A_{t-n_{K,b}}) &=\sup_{a \in \textup{UP}^{(n_{K,b})}(W_{t},\cdots,A_{t-n_{K,b}},W_{t-n_{K,b}})} \mathbb{E}\left[ R_{t} + \gamma I[\Delta_{t}=1] V^{(0)}_{*}(W_{t+1}) \right.\nonumber\\
&\left.\quad\mid W_{t-n_{K,b}}, A_{t-n_{K,b}},\cdots,W_{t},a\right]. \nonumber
\end{align}
\end{definition}
Now, we are ready state the key Lemmas:
\begin{lemma}\label{lm: identification for censored demand}
	For every $t \geq 0$ and $0 \leq i \leq n_{K,b}$, we have the following 
	\begin{align}\label{eqn: identification for censored demand}
\EE\left[\left(R_{t} -\widetilde{R}_{t}\right)\mathbb{I}[\Delta_{t}=0]|W_{t-i}, A_{t-i},\cdots,W_{t}, A_{t}\right]=0,
	\end{align}
 almost surely. Here, in the component of $W_{t-i+j}$, we have $\Delta_{t-i+j-1}=0$ for all $j=1,\cdots,i$, and in the component of $W_{t}$, we have $\Delta_{t-i-1}=1$.
\end{lemma}
The proof is given in Appendix \ref{subsec:proof of identification for censored demand }

\begin{lemma}\label{lm: adress censored}
	Under Assumption \ref{ass: inventory and demand independency}, 
	\begin{align}\label{eqn: suvival demand}
	\EE\!\left[D_{t}\mid \Delta_{t}=0, H_{(t-i):t} \right]
= Y_{t}
+ \int_{Y_{t}}^{D_{\max}}
\frac{\textup{SF}(c\mid H_{(t-i):t}\setminus Y_{t})}
{\textup{SF}(Y_{t}\mid H_{(t-i):t}\setminus Y_{t})}\,dc, 
	\end{align}
	where $\textup{SF}(c \mid H_{(t-i):t}\setminus Y_{t})$ denotes the conditional survival function $\Pr(D_t > c \mid H_{(t-i):t}\setminus Y_{t})$.
\end{lemma}
See Appendix \ref{subsec: proof of adress censored} for the proof.
\begin{lemma}
    \label{lemma: nkb and action relation}
	For all $t\geq 0$\begin{align}
    &\sup_{\pi \in \Pi_{n_{K,b}}} \mathbb{E}^{\pi}\left[ R_{t+n_{K,b}} + \gamma I[\Delta_{t+n_{K,b}}=1] V^{(0)}_{*}(W_{t+n_{K,b}+1}) \mid H_{t-1}, W_t, A_t, \ldots, W_{t+n_{K,b}} \right] \nonumber \\
    &\leq \sup_{a \in A(W_{t+n_{K,b}}, \ldots, A_{t}, W_{t})} \mathbb{E}\left[ R_{t+n_{K,b}} + \gamma I[\Delta_{t+n_{K,b}}=1] V^{(0)}_{*}(W_{t+n_{K,b}+1}) \mid W_t, A_t, \ldots, W_{t+n_{K,b}}, a \right]\nonumber.
\end{align}
 
\end{lemma}

See Appendix \ref{subsec: nkb and action relation} for the proof.

\begin{lemma}\label{lm: performance difference}
	Let $W^{(0)},A^{(0)},\cdots,W^{(i)},A^{(i)}$ and $\Delta^{(i)}$ be the generic random variables. Consider a discrete random variable $I$ taking values $0,\cdots,n_{K,b}$ with $P(I=i)=\frac{\gamma^i(1-\gamma)}{1-\gamma^{n_{K,b}}}$. Then the difference between state value functions of $\pi_1$ and $\pi_2$, where both belongs to $\Pi_{n_{K,b}}$, can be expressed, $\forall w^{(0)} \in \calW$ such that $\Delta_{-1}=1$, as
	{\footnotesize\begin{align*}  
&V^{(0)}_{\pi_1}(w^{(0)}) -V^{(0)}_{\pi_2 }(w^{(0)})\\ 
&=\frac{1-\gamma^{n_{K,b}}}{(1-\gamma)^2}\EE_{(W^{(0)},A^{(0)},\cdots,W^{(I)}, A^{(I)}) \sim (d_{w^{(0)}}^{\pi_1} \times \pi_1)
    }\left[A^{(I)}_{\pi_2}(W^{(0)},A^{(0)},\cdots,W^{(I)},A^{(I)})\prod_{j=1}^{I}\left\{\mathbb{I}[\Delta^{(j)}=0]\right\}\mathbb{I}[\Delta^{(0)}=1]\right]
\end{align*}}
	where $d_{w^{(0)},I}^{\pi}$ is the state visitation measure under $\pi$ given in Definition \ref{def: generalized discounted visitation}. 
\end{lemma}
See Appendix \ref{subsec: proof of performance difference} for the proof.

\section{Technical Lemmas}\label{app: technical lemmas}

Before presenting key lemmas, we introduce several related notations and definitions. First of all, throughout this section, for the ease of notation, let $(W_t,A_t)=X_t$, and thus, $\calS\times\calW=\calX$. We similarly denote $(w^{(i)},a^{(i)})=x^{(i)}\in \calX$ where $w^{(i)}$ and $a^{(i)}$ are the realizations of random vectors $W_t$ and $A_t$. 
\begin{definition}\label{def: P as conditional exp} For any $0 \leq i \leq n_{K,b}$ and a policy $\pi: (\calX)^{\otimes (i+1)} \times \calW \rightarrow \calA$, define a conditional expectation operator denoted by ${\cal P}^\pi_i$:
{\footnotesize\begin{align*}
    &\calP^\pi_i Q(\bullet)=\EE^{\pi}\left[Q(W_t,A_t,\cdots,W_{t+i+1},A_{t+i+1})\given (W_{t},A_{t},\cdots,W_{t+i},A_{t+i})=\bullet,\mathcal{C}(K,n_{K,b})\right],
\end{align*}}
for any function $Q$ over $(\calX)^{\otimes (i+2)}$.
    
\end{definition}
Note that for any policy $\pi \in \Pi_{n_{K,b}}$, the event $\mathcal{C}(K,n_{K,b})$ is a sure event which can be dropped from the expectation.

\begin{definition}\label{def: bellman operator}The conditional Bellman operator under the policy $\pi$, denoted by \(\calT_{i,\mathcal{C}}^{\pi}\) is defined as:
  {\footnotesize
\begin{align*}
    \calT_{i,\mathcal{C}}^{\pi}Q(w^{(0)},a^{(0)},\cdots, w^{(i)},a^{(i)})&=\EE\left[R_{t+i} + \gamma Q^{(0)}(W_{t+i+1},\pi^{(0)}(W_{t+i+1}))\mathbb{I}[\Delta_{t+i}=1]\right.\\
    &\left. \quad+ \gamma Q^{(i+1)}(W_t,A_t,\cdots,W_{t+i+1},\pi^{(i+1)}(W_{t+i+1},\cdots,W_t,A_t))\mathbb{I}[\Delta_{t+i}=0] \right.\\
    &\left. \quad \given (W_{t},A_{t},\cdots,W_{t+i},A_{t+i})=(w^{(0)},a^{(0)},\cdots, w^{(i)},a^{(i)}),\mathcal{C}(K,n_{K,b})\right].
\end{align*}}  
\end{definition}
Note that for any policy $\pi \in \Pi_n$, the event $\mathcal{C}(K,n_{K,b})$ is a sure event which can be dropped from the expectation. In that case, $\calT_{i,\mathcal{C}}^{\pi}$ becomes $\calT_{i}^{\pi}$.

\begin{definition}[\text{Action Discrepancy for }$\widetilde{\pi}_K\text{ and }\widehat{\pi}_K$]
\label{def:Qhat_action_discrepancy}
For $(w^{(0)},a^{(0)},\ldots,w^{(n_{K,b})})$, define
\begin{sizeddisplay}
{\footnotesize}
\begin{align}
    (i)\quad &
    \hypertarget{AD_tilde_pi}{\textup{AD}^{\tilde{\pi}_K}\!\left(w^{(0)},a^{(0)},\cdots,w^{(n_{K,b})}\right)} \nonumber\\
    &= \max_{a\in\mathcal{A}}
    \widetilde{Q}_{K}^{(n_{K,b})}(w^{(0)},a^{(0)},\cdots,w^{(n_{K,b})},a)
    - 
    \widetilde{Q}_{K}^{(n_{K,b})}(
        w^{(0)},a^{(0)},\cdots,w^{(n_{K,b})},
        \widetilde{\pi}_K^{(n_{K,b})}(w^{(n_{K,b})},\cdots,w^{(0)},a^{(0)})
    ).
    \label{def:AD_tilde_pi}
\end{align}
\begin{align}
    (ii)\quad &
    \hypertarget{AD_hat_pi}{\textup{AD}^{\hat{\pi}_K}(w^{(0)},a^{(0)},\cdots,w^{(n_{K,b})})} \nonumber\\
    &= \max_{a\in\mathcal{A}}
    \widehat{Q}_{K}^{(n_{K,b})}(w^{(0)},a^{(0)},\cdots,w^{(n_{K,b})},a)
    -
    \widehat{Q}_{K}^{(n_{K,b})}(
        w^{(0)},a^{(0)},\cdots,w^{(n_{K,b})},
        \widehat{\pi}_K^{(n_{K,b})}(w^{(n_{K,b})},\cdots,w^{(0)},a^{(0)})
    ).
    \label{def:AD_hat_pi}
\end{align}

\end{sizeddisplay}
\end{definition}

\begin{definition}[\text{Action Discrepancy and Total Action Discrepancy}]
\label{def:action_discrepancy}
For $t\geq0$, let
\begin{sizeddisplay}
{\footnotesize}
\begin{align}
    (i)\quad & 
    \hypertarget{action_discrepancy_i}{\textup{AD}^{\pi^{*}_{n_{K,b}}}(X_t,\cdots,W_{t+n_{K,b}})} \nonumber\\
    &=\max_{a\in\calA}Q_{\pi^{\ast}_{n_{K,b}}}^{(n_{K,b})}(X_t,\cdots,W_{t+n_{K,b}},a)-Q_{\pi^{\ast}_{n_{K,b}}}^{(n_{K,b})}(X_t,\cdots,W_{t+n_{K,b}},\pi_{n_{K,b},*}^{(n_{K,b})}(W_{t+n_{K,b}},\cdots,X_t)).\label{def:action_discrepancy_i}
\end{align}
\begin{align}
    (ii)\quad&\hypertarget{total_action_discrepancy_ii}{\textup{TAD}^{\hat \pi_K}(x^{(0)},\cdots,x^{(i)})} \nonumber\\
    &= \sum_{k=1}^{K-1}\prod_{j=1}^{k}\left(\gamma\calP_{i+j-1}^{\widehat{\pi}_{K-j}}\right)\left\{\left(\textup{AD}^{\pi^{*}_{n_{K,b}}}(X_0,\cdots,W_{i+k})\mathbb{I}[i=n_{K,b}-k]\right)\prod_{l=0}^{i+k-1}\mathbb{I}[\Delta_{l}=0]\mathbb{I}[\Delta_{-1}=1]\right.\nonumber\\
    &\left. +\sum_{v=1}^{\min(k-1,n_{K,b})}\left(\textup{AD}^{\pi^{*}_{n_{K,b}}}(X_{i+k-v},\cdots,W_{i+k})\mathbb{I}[v=n_{K,b}]\right)\prod_{l=i+k-v}^{i+k-1}\mathbb{I}[\Delta_{l}=0]\mathbb{I}[\Delta_{i+k-v-1}=1]\right\}\label{def:total_action_discrepancy_ii}
\end{align}
\begin{align}
   (iii) \quad & \hypertarget{total_action_discrepancy_iii}{\textup{TAD}^{\tilde{\pi}_K}(x^{(0)},\cdots,x^{(i)})} \nonumber\\
   &= \sum_{k=1}^{K-1}\prod_{j=1}^{k}\left(\gamma\calP_{i+j-1}^{\widetilde{\pi}_{K-j}}\right)\left\{\left(\textup{AD}^{\pi^{*}_{n_{K,b}}}(X_0,\cdots,W_{i+k})\mathbb{I}[i=n_{K,b}-k]\right)\prod_{l=0}^{i+k-1}\mathbb{I}[\Delta_{l}=0]\mathbb{I}[\Delta_{-1}=1]\right.\nonumber\\
    &\left. +\sum_{v=1}^{\min(k-1,n_{K,b})}\left(\textup{AD}^{\pi^{*}_{n_{K,b}}}(X_{i+k-v},\cdots,W_{i+k})\mathbb{I}[v=n_{K,b}]\right)\prod_{l=i+k-v}^{i+k-1}\mathbb{I}[\Delta_{l}=0]\mathbb{I}[\Delta_{i+k-v-1}=1]\right\}\label{def:total_action_discrepancy_iii}.
\end{align}
\end{sizeddisplay}
\end{definition}

\begin{definition}[\text{Reward Discrepancy}]\label{def: fqi RD definition}
    {\footnotesize\begin{align}
    \textup{RD}(NT) &=
    \sum_{k=0}^{K-1}\gamma^k C_5(\varepsilon)(NT)^{-\delta}= \frac{1}{1-\gamma
    }C_5(\varepsilon)(NT)^{-\delta}
\end{align}}
\end{definition}

\begin{definition}[\text{Uncertainty Quantification}]
\label{def:uq_definition}
    \begin{sizeddisplay}{\footnotesize}
    \begin{align}
    (i)\quad & 
    \hypertarget{uq_fqi_i}{\textup{UQ}(x^{(0)},\cdots,x^{(i)})} \nonumber\\
    &= \sum_{k=0}^{K-1}\gamma^k \prod_{k'=0}^{k-1}\calP_{i+k'}^{\pi^{\ast}_{n_{K,b}}}\left\{U_{K-k}^{(0)}(X_{i+k})\mathbb{I}[\Delta_{i+k-1}=1]\right.\nonumber\\
    &\left.+U_{K-k}^{(i+k)}(X_{0},\cdots,X_{i+k})\prod_{j=0}^{i+k-1}\mathbb{I}[\Delta_{j}=0]\mathbb{I}[\Delta_{-1}=1]\mathbb{I}[k\leq n_{K,b}-i]\right.\nonumber\\
    &\left.+ \sum_{v=1}^{\min(n_{K,b},k-1)}\left\{U_{K-k}^{(v)}(X_{i+k-v},\cdots,X_{i+k})\prod_{j=i+k-v}^{i+k-1}\mathbb{I}[\Delta_{j}=0]\mathbb{I}[\Delta_{i+k-v-1}=1]\right\} \right\}\nonumber\\
    &=\sum_{k=0}^{K-1}\gamma^k\left(\prod_{k'=0}^{k-1}\calP_{i+k'}^{\pi^{\ast}_{n_{K,b}}}\right)
\Big\{ U_{K-k}^{(M_{i+k})}\big(H_{i+k}^{(M_{i+k})}\big)\Big\}\nonumber\\
&=
\sum_{k=0}^{K-1}\gamma^k\,
\EE^{\pi^\ast_{n_{K,b}}}\!\left[
U_{K-k}^{(M_{i+k})}\!\big(H_{i+k}^{(M_{i+k})}\big)
\ \Big|\ (X_0,\ldots,X_i)=(x^{(0)},\ldots,x^{(i)})
\right].\label{uq_fqi_i}
    \end{align}
   
    \begin{align}
    (ii)\quad & 
    \hypertarget{uq_pfqi_ii}{\widetilde{\textup{UQ}}(x^{(0)},\cdots,x^{(i)})} \nonumber\\
    &= \sum_{k=0}^{K-1}\gamma^k \prod_{k'=0}^{k-1}\calP_{i+k'}^{\pi^{\ast}_{n_{K,b}}}\left\{2\Tilde{U}_{K-k}^{(0)}(X_{i+k})\mathbb{I}[\Delta_{i+k-1}=1]\right.\nonumber\\
    &\left.+2\Tilde{U}_{K-k}^{(i+k)}(X_{0},\cdots,X_{i+k})\prod_{j=0}^{i+k-1}\mathbb{I}[\Delta_{j}=0]\mathbb{I}[\Delta_{-1}=1]\mathbb{I}[k\leq n_{K,b}-i] \right.\nonumber\\
    &\left.+ \sum_{v=1}^{\min(n_{K,b},k-1)}\left\{2\Tilde{U}_{K-k}^{(v)}(X_{i+k-v},\cdots,X_{i+k})\prod_{j=i+k-v}^{i+k-1}\mathbb{I}[\Delta_{j}=0]\mathbb{I}[\Delta_{i+k-v-1}=1]\right\} \right\}\nonumber \\
    &=\sum_{k=0}^{K-1}\gamma^k\left(\prod_{k'=0}^{k-1}\calP_{i+k'}^{\pi^{\ast}_{n_{K,b}}}\right)
\Big\{ 2\Tilde{U}_{K-k}^{(M_{i+k})}\big(H_{(i+k-M_{i+k}):(i+k)}\big)\Big\}\nonumber\\ 
&= \sum_{k=0}^{K-1}\gamma^k\,
\EE^{\pi^\ast_{n_{K,b}}}\!\left[
2\Tilde{U}_{K-k}^{(M_{i+k})}\!\big(H_{(i+k-M_{i+k}):(i+k)}\big)
\ \Big|\ (X_0,\ldots,X_i)=(x^{(0)},\ldots,x^{(i)})
\right].\label{uq_pfqi_ii}
    \end{align}

     where $M_t
:=\min\Big\{n_{K,b},\ \max\{v\ge 0:\ \Delta_{t-v}=\cdots=\Delta_{t-1}=0,\ \Delta_{t-v-1}=1\}\Big\}$
    \end{sizeddisplay}
\end{definition}

\begin{definition}[Price of not being pessimistic]\label{def: price of pessimisim}
  {\footnotesize
 \begin{align}\label{def:def C^{(i)}}
 C^{(i)}(w^{(0)},a^{(0)},\cdots,w^{(i)})&=\max_{a \in \calA}\sum_{k=1}^{K-1}\gamma^{k}\left\{\EE^{\pi_{K-1},\cdots,\pi_{K-k}}\left[U^{(0)}_{K-k}(W_{i+k},A_{i+k})\mathbb{I}[\Delta_{i+k-1}=1]\right.\right.\\
 &\left.\left.+U_{K-k}^{(i+k)}(W_{0},A_{0},\cdots,W_{i+k},A_{i+k})\prod_{l=0}^{i+k-1}\mathbb{I}[\Delta_{l}=0]\mathbb{I}[\Delta_{-1}=1]\mathbb{I}[k\leq n_{K,b}-i]\right.\right.\nonumber\\
 &\left.\left.+\sum_{v=1}^{\min(k-1,n_{K,b})}U_{K-k}^{(v)}(W_{i+k-v},A_{i+k-v},\cdots,W_{i+k},A_{i+k}) \prod_{l=i+k-v}^{i+k-1}\mathbb{I}[\Delta_{l}=0]\right.\right.\nonumber\\
 &\left.\left. \quad \mathbb{I}[\Delta_{i+k-v-1}=1] +U_K^{(i)}(W_0,A_0,\cdots,W_{i},A_{i})\right.\right.\nonumber\\
 &\left.\left.\given (W_0,A_0,\cdots,W_{i},A_{i})=(w^{(0)},a^{(0)},\cdots,w^{(i)},a)\right]\right\}\nonumber\\
 &=\max_{a\in\calA}\sum_{k=0}^{K-1}\gamma^{k}\,
\EE^{\pi_{K-1},\cdots,\pi_{K-k}}\!\left[
U_{K-k}^{(M_{i+k})}\big(H_{i+k}^{(M_{i+k})}\big)
\ \Big|\ (W_0,A_0,\cdots,W_i,A_i)=(w^{(0)},a^{(0)},\cdots,w^{(i)},a)
\right]\nonumber
 \end{align}}  
   where $M_t
:=\min\Big\{n_{K,b},\ \max\{v\ge 0:\ \Delta_{t-v}=\cdots=\Delta_{t-1}=0,\ \Delta_{t-v-1}=1\}\Big\}$
\end{definition}
Now we are ready to state the technical lemmas:
\begin{lemma}\label{lm:upper bound on Qs}
    Under Assumptions~\ref{ass: surrogate outcome} and \ref{ass: UQ ass}, the following upper bound on the difference between $Q^{(i)}_{\pi^{\ast}_{n_{K,b}}}$ and $\overline{Q}_{K}^{(i)}$ $\forall i=0,\cdots,n_{K,b}$ holds with probability at least $1 -(n_{K,b}+1)\epsilon-\varepsilon$:

    {\footnotesize\begin{align}
    Q^{(i)}_{\pi^{\ast}_{n_{K,b}}}(x^{(0)},\cdots,x^{(i)})-\overline{Q}_{K}^{(i)}(x^{(0)},\cdots,x^{(i)})&\leq\gamma^{K-1}\left(\frac{2-\gamma}{1-\gamma}R_{\max}\right)+ \textup{RD}(NT)\nonumber\\
     & + \textup{UQ}(x^{(0)},\cdots,x^{(i)}) +C^{(i)}(x^{(0)},\cdots,w^{(i)}),
\end{align}}where $\textup{RD}(NT)$, $\textup{UQ}(x^{(0)},\cdots,x^{(i)})$ and $C^{(i)}(x^{(0)},\cdots,w^{(i)})$ are given  in Definitions \ref{def: fqi RD definition}, \hyperlink{uq_fqi_i}{~\ref{def:uq_definition} (i)} and \ref{def: price of pessimisim}, respectively. 
\end{lemma}
See Appendix \ref{subsec: proof of upper bound Qs fqi} for details.

\begin{lemma}\label{lm:lower bound on Qs}
    Under Assumptions~\ref{ass: surrogate outcome} and \ref{ass: UQ ass}, the following lower bound on the difference between $Q^{(i)}_{\pi^{\ast}_{n_{K,b}}}$ and $\overline{Q}_{K}^{(i)}$, $\forall i=0,\cdots,n_{K,b}$,  holds with probability at least $1 -(n_{K,b}+1)\epsilon-\varepsilon$: 
    \begin{sizeddisplay}
{\footnotesize}\begin{align} 
    Q^{(i)}_{\pi^{\ast}_{n_{K,b}},}(x^{(0)},\cdots,x^{(i)})-\overline{Q}_{K}^{(i)}(x^{(0)},\cdots,x^{(i)})
    &\geq -\gamma^{K-1}\left(\frac{2-\gamma}{1-\gamma}R_{\max}\right)-\textup{RD}(NT)\nonumber\\
    &-\textup{TAD}^{\hat \pi_K}(x^{(0)},\cdots,x^{(i)}),
\end{align} \end{sizeddisplay}
where the characterizations of $\textup{TAD}^{\hat \pi_K}$ and $\textup{RD}(NT)$ are given in Definitions \hyperlink{total_action_discrepancy_ii}{~\ref{def:action_discrepancy} (ii)} and \ref{def: fqi RD definition} and , respectively. 
\end{lemma}
See Appendix \ref{subsec: proof of lower bound on Qs fqi} for details.

\begin{lemma}\label{lm:upper bound on Qs pessimistic}
    Under Assumptions~\ref{ass: surrogate outcome} and \ref{ass: UQ ass}, the following upper bound on the difference between $Q^{(i)}_{\pi^{\ast}_{n_{K,b}}}$ and $\widetilde{Q}_{K}^{(i)}$, $\forall i=0,\cdots,n$, holds with probability at least $1 -(n_{K,b}+1)\epsilon-\varepsilon$:
    \begin{sizeddisplay}{\footnotesize}\begin{align}\label{eq:pessimistic  final upper bound Q}
   Q^{(i)}_{\pi^{\ast}_{n_{K,b}}}(x^{(0)},\cdots,x^{(i)})-\widetilde{Q}_{K}^{(i)}(x^{(0)},\cdots,x^{(i)})&\leq \gamma^{K-1}\left(\frac{2-\gamma}{1-\gamma}R_{\max}\right)+ \textup{RD}(NT)\nonumber\\
   &+\widetilde{\textup{UQ}}(x^{(0)},\cdots,x^{(i)}),
\end{align}\end{sizeddisplay}
where $\text{RD}(NT)$ and $\widetilde{\textup{UQ}}(x^{(0)},\cdots,x^{(i)})$ are defined in \ref{def: fqi RD definition} and\hyperlink{uq_pfqi_ii}{~\ref{def:uq_definition} (ii)}, respectively.  
\end{lemma}
See Appendix \ref{subsec: proof of upper bound on Qs pessimistic} for details.

\begin{lemma}\label{lm:lower bound on Qs pessimistic}
    Under Assumptions~\ref{ass: surrogate outcome} and \ref{ass: UQ ass}, the following lower bound on the difference between $Q^{(i)}_{\pi^{\ast}_{n_{K,b}}}$ and $\widetilde{Q}_{K}^{(i)}$, $\forall i=0,\cdots,n$, holds with probability at least $1 -(n_{K,b}+1)\epsilon-\varepsilon$
    \begin{sizeddisplay}{\footnotesize}\begin{align}
    Q^{(i)}_{\pi^{\ast}_{n_{K,b}}}(x^{(0)},\cdots,x^{(i)})-\widetilde{Q}_{K}^{(i)}(x^{(0)},\cdots,x^{(i)})
    &\geq -\gamma^{K-1}\left(\frac{2-\gamma}{1-\gamma}R_{\max}\right)-\textup{RD}(NT)\nonumber\\
    &- \textup{TAD}^{\tilde \pi_K}(x^{(0)},\cdots,x^{(i)}),
\end{align}
\end{sizeddisplay}
where $\textup{TAD}^{\tilde \pi_K}$ and $\textup{RD}(NT)$ are given in Definitions\hyperlink{total_action_discrepancy_iii}{~\ref{def:action_discrepancy} (iii)} and \ref{def: fqi RD definition}, respectively.  
\end{lemma}
See Appendix \ref{seubsec: proof of lower bound on Qs pessimistic} for details.

\begin{lemma}[Uniform Gram comparison (adapted Theorem 21 of \cite{chang2021mitigating})]\label{lemma: aux mitigating}
Fix any $i \in \{0,1,\dots,n_{K,b}\}$. Let $\phi_i:\mathcal{H}^{(i)}\to\mathbb{R}^{d_i}$ be a feature map and assume $\|\phi_i(h)\|_2 \le 1$ for all $h\in\mathcal{H}^{(i)}$. Let $\mathcal{O}_N^{(i)}=\{\bar H^{(i)}_1,\dots,\bar H^{(i)}_{|\mathcal{O}_N^{(i)}|}\}$ denote an offline dataset of length-$(i{+}1)$ history windows.

Define the population covariance under $d_i^\mu$ as
\[
\Sigma_i^\mu \;:=\; \mathbb{E}_{H^{(i)}\sim d_i^\mu}\!\big[\phi_i(H^{(i)})\phi_i(H^{(i)})^\top\big],
\qquad
r^{(i)} \;:=\; \mathrm{rank}(\Sigma_i^\mu),
\]
and the regularized empirical Gram matrix as
\[
\Lambda^{(i)} \;:=\; \sum_{j=1}^{|\mathcal{O}_N^{(i)}|}\phi_i(\bar H^{(i)}_j)\phi_i(\bar H^{(i)}_j)^\top \;+\;\lambda I,
\]
for some $\lambda>0$. Furthermore, given the following event $\Omega^{(i)}_{MP}$,
\[
\Omega^{(i)}_{MP}
:=
\left\{
\forall h\in\mathcal{H}^{(i)}:\ 
\phi_i(h)^\top (\Lambda^{(i)})^{-1}\phi_i(h)
\le
c_1\big(r^{(i)}+\log(c_2/\epsilon)\big)\,
\phi_i(h)^\top \big(|\mathcal{O}_N^{(i)}|\,\Sigma_i^\mu+\lambda I\big)^{-1}\phi_i(h)
\right\},
\]
we have \(\mathbb{P}(\Omega^{(i)}_{MP}) \ge 1 - \epsilon / (n_{K,b} + 1)\) for each \(i\). Then, by the union bound, the event \(\Omega_{MP} = \bigcap_i \Omega^{(i)}_{MP}\) holds with probability at least $1 - \epsilon$.
\end{lemma}
See Appendix~\ref{lm proof aux mitig} for details.
\begin{lemma}\label{lemma: Matrix bound}
Let $\Omega^{(i)}_M$ be the event that the empirical covariance concentrates for partition $i$:
\[
\Omega_M^{(i)} \;:=\; \left\{ \widehat{\Sigma}^{(i)} \succeq \tfrac{1}{2} \Sigma_i^\mu \right\}.
\]
By the Matrix Chernoff bound, for any fixed $i$, $\mathbb{P}(\Omega_M^{(i)}) \ge 1 - \frac{\epsilon}{n_{K,b}+1}$ provided that $|\mathcal{O}_N^{(i)}| \gtrsim d^{(i)} \log(\frac{d^{(i)} (n_{K,b}+1)}{\epsilon})$.
Applying a Union Bound over all $i \in \{0, \dots,n_{K,b}\}$:
\[
\prob\left( \Omega_M:=\bigcap_{i=0}^{n_{K,b}} \Omega_M^{(i)}\right) 
\;\ge\; 
1 - \sum_{i=0}^{n_{K,b}} \frac{\epsilon}{n_{K,b}+1} 
\;=\; 
1 - \epsilon,
\]
\end{lemma}
where the result follows directly from Theorem 5.1 (Matrix Chernoff) in \citet{tropp2012user}.

\section{Proof of Propositions and Lemmas}\label{app: Proof of Propositions and Lemmas}
\subsection{Proof of Proposition \ref{prop: high-order}}\label{proof: proposiiton}
\textbf{Proof:} Consider the AR(1) demand process where $D_t$ depends on $D_{t-1}$ via the term $\rho D_{t-1}$. When censoring occurs at $t-1$ (i.e., $\Delta_{t-1}=0$), the observed sales equal the inventory, $Z_{t-1}=Y_{t-1}$, and only the inequality $D_{t-1}>Y_{t-1}$ is observed. In this case, $W_t$ no longer contains the full information about the underlying demand. The hidden value of $D_{t-1}$ still influences the dynamics of $D_t$ through the autoregressive term $\rho D_{t-1}$, but $D_{t-1}$ itself depends on \emph{earlier} variables $(P_{t-1},X_{t-1},D_{t-2})$ that are not included in $W_t$. As a result, the future evolution of $(W_{t+1},R_t)$ depends on information from the past that is not captured by $(W_t,A_t)$, and the conditional independence property fails.

\subsection{Proof of Lemma~\ref{cor:mdp-order}}\label{proof: lemma 1}
Fix any time $t\ge 0$ and any policy $\pi\in\Pi_n$. Recall
\[
\tau_t(n)\;:=\;\max\big\{\,j\in\{t,t-1,\ldots,t-n\}\,:\ \Delta_{j-1}=1\,\big\}.
\]
Because $\Delta_{-1}=1$ and $\pi\in\Pi_n$ rules out censoring streaks longer than $n$, such a $j$ exists almost surely; hence $\tau_t(n)$ is well-defined.

We prove Lemma \eqref{cor:mdp-order} by showing that, for any bounded measurable function $f$,
\begin{equation}\label{eq:ms-proof-goal}
\EE^\pi\!\big[f(W_{t+1},R_t)\mid H_{0:t}\big]
\;=\;
\EE^\pi\!\big[f(W_{t+1},R_t)\mid H_{\tau_t(n):t}\big]
\qquad\text{a.s.}
\end{equation}
This identity is equivalent to the conditional-independence statement
\(
(W_{t+1},R_t)\ \independent\ H_{0:\tau_t(n)-1}\mid H_{\tau_t(n):t}.
\)

Recall the state of the \emph{Underlying Process}, $\{S_u\}_{u\ge 0}$ (the fully observed, uncensored state). By Equation \eqref{eq: markov property}, the Underlying Process is Markov: conditional on $(S_u,A_u)$, the pair $(S_{u+1},R_u)$ is independent of all earlier history.

Now consider the state \emph{Observed Process} $\{W_u\}$. By construction, $W_u$. The key point is:
\emph{at any time $j$ such that $\Delta_{j-1}=1$, the observation at time $j$ reveals the full “demand” state needed to propagate the dependent-demand dynamics.}
In particular, when $\Delta_{j-1}=1$, sales equals demand at $j-1$, so the demand component that drives dependence is observed rather than truncated, and is included in $W_j$.

Consequently, at the time $\tau:=\tau_t(n)$ (for which $\Delta_{\tau-1}=1$ by definition), the latent state $S_\tau$ is determined by the observed information at time $\tau$:
\begin{equation}\label{eq:state-from-uncensored-obs}
S_{\tau}\;=\;\Psi(W_{\tau})
\end{equation}
for some measurable mapping $\Psi$ (the exact form is model-specific, but existence follows from the fact that the relevant demand component is fully observed when uncensored).

Take any bounded measurable function $f$. Start from the conditional expectation given the full observed history $H_{0:t}$. By the law of iterated expectations, we may condition first on the latent state $S_\tau$:
\begin{align}
\EE^\pi\!\big[f(W_{t+1},R_t)\mid H_{0:t}\big]
&=
\EE^\pi\!\Big[
\EE^\pi\!\big[f(W_{t+1},R_t)\mid H_{0:t}, S_\tau\big]
\ \Big|\ H_{0:t}
\Big].
\label{eq:tower-ms}
\end{align}

Next, note that once we fix $S_\tau$ and the post-$\tau$ observed history $H_{\tau:t}$, the sequence of actions $(A_\tau,\ldots,A_t)$ is also fixed (because actions are chosen by $\pi$ as functions of observed history). Therefore, conditional on $(S_\tau,H_{\tau:t})$, the evolution from time $\tau$ to time $t+1$ is entirely governed by the Markov transition of the Underlying Process and those fixed actions. In particular, by the Markov property of the Underlying Process, the distribution of $(W_{t+1},R_t)$ cannot depend on the earlier observed history $H_{0:\tau-1}$ once $(S_\tau,H_{\tau:t})$ is fixed. Formally,
\begin{equation}\label{eq:drop-early-history}
\EE^\pi\!\big[f(W_{t+1},R_t)\mid H_{0:t}, S_\tau\big]
=
\EE^\pi\!\big[f(W_{t+1},R_t)\mid H_{\tau:t}, S_\tau\big].
\end{equation}
Substituting Equation \eqref{eq:drop-early-history} into Equation \eqref{eq:tower-ms} yields
\[
\EE^\pi\!\big[f(W_{t+1},R_t)\mid H_{0:t}\big]
=
\EE^\pi\!\Big[
\EE^\pi\!\big[f(W_{t+1},R_t)\mid H_{\tau:t}, S_\tau\big]
\ \Big|\ H_{0:t}
\Big].
\]

By Equation \eqref{eq:state-from-uncensored-obs}, $S_\tau$ is a function of $W_\tau$, and $W_\tau$ is contained in $H_{\tau:t}$. Thus, conditioning on $(H_{\tau:t},S_\tau)$ is the same as conditioning on $H_{\tau:t}$ alone. This implies the following
\[
\EE^\pi\!\big[f(W_{t+1},R_t)\mid H_{\tau:t}, S_\tau\big]
=
\EE^\pi\!\big[f(W_{t+1},R_t)\mid H_{\tau:t}\big].
\]
Plugging this into the previous display gives
\[
\EE^\pi\!\big[f(W_{t+1},R_t)\mid H_{0:t}\big]
=
\EE^\pi\!\big[f(W_{t+1},R_t)\mid H_{\tau:t}\big],
\]
which is exactly \eqref{eq:ms-proof-goal}. Therefore,
\[
(W_{t+1},R_t)\ \independent\ H_{0:\tau-1}\ \Big|\ H_{\tau:t}.
\]

By construction, $\tau_t(n)\in\{t,t-1,\ldots,t-n\}$, so the conditioning block $H_{\tau_t(n):t}$ contains at most $n{+}1$ observation--action pairs. Hence, under any policy $\pi\in\Pi_n$, the Observed Process admits a Markov representation of order at most $n{+}1$, proving the Lemma ~\ref{cor:mdp-order}.

\subsection{Proof of Lemma \ref{lm: identification for censored demand-main text}}
\label{subsec:proof of identification for censored demand }

Fix an integer $i\ge 0$ and define the observed history window
\[
H_{(t-i):t}:=(W_{t-i},A_{t-i},\ldots,W_t,A_t).
\]
Let
\[
H_{(t-i):t}=(w^{(0)},a^{(0)},\ldots,w^{(i)},a^{(i)}),
\]
where $(w^{(j)},a^{(j)})$ corresponds to $(W_{t-i+j},A_{t-i+j})$. In particular, in the components of
$W_{t-i+j}$ we have $\Delta_{t-i+j-1}=0$ for all $j=1,\ldots,i$, and in the components of $W_{t-i}$
we have $\Delta_{t-i-1}=1$.

Consider
\begin{sizeddisplay}
{\footnotesize
\begin{align}
&\EE\!\left[D_t\mathbb{I}(\Delta_t=0)\,\Big|\,H_{(t-i):t}=(w^{(0)},a^{(0)},\ldots,w^{(i)},a^{(i)})\right] \label{eq: lemma 1 eq1_adj}\\
&=\EE\!\left[D_t\,\Big|\,H_{(t-i):t}=(w^{(0)},a^{(0)},\ldots,w^{(i)},a^{(i)}),\,\Delta_t=0\right]\,
\prob\!\left(\Delta_t=0\,\Big|\,H_{(t-i):t}=(w^{(0)},a^{(0)},\ldots,w^{(i)},a^{(i)})\right),\nonumber
\end{align}}
\end{sizeddisplay}
where the equality follows from the law of iterated expectations.

Similarly,
\begin{sizeddisplay}
{\footnotesize
\begin{align}
&\EE\!\left[\EE\!\left(D_t\mid \Delta_t, H_{(t-i):t}\right)\mathbb{I}(\Delta_t=0)\,\Big|\,H_{(t-i):t}=(w^{(0)},a^{(0)},\ldots,w^{(i)},a^{(i)})\right] \label{eq: lemma 1 eq2_adj}\\
&=\EE\!\left[D_t\,\Big|\,H_{(t-i):t}=(w^{(0)},a^{(0)},\ldots,w^{(i)},a^{(i)}),\,\Delta_t=0\right]\,
\prob\!\left(\Delta_t=0\,\Big|\,H_{(t-i):t}=(w^{(0)},a^{(0)},\ldots,w^{(i)},a^{(i)})\right).\nonumber
\end{align}}
\end{sizeddisplay}

Next, note that
\begin{sizeddisplay}
{\footnotesize
\begin{align}
R_t-\widetilde R_t
= C_{1,t}\Big(\EE\!\left[D_t\mid \Delta_t, H_{(t-i):t}\right]-D_t\Big)\mathbb{I}(\Delta_t=0).
\end{align}}
\end{sizeddisplay}
Therefore,
\begin{sizeddisplay}
{\footnotesize
\begin{align}
&\EE\!\left[\left(R_t-\widetilde R_t\right)\mathbb{I}(\Delta_t=0)\,\Big|\,H_{(t-i):t}=(w^{(0)},a^{(0)},\ldots,w^{(i)},a^{(i)})\right]\nonumber\\
&=C_{1,t}\Bigg(
\EE\!\left[\EE\!\left(D_t\mid \Delta_t, H_{(t-i):t}\right)\mathbb{I}(\Delta_t=0)\,\Big|\,H_{(t-i):t}=(\cdot)\right]
-\EE\!\left[D_t\mathbb{I}(\Delta_t=0)\,\Big|\,H_{(t-i):t}=(\cdot)\right]
\Bigg)\nonumber\\
&=0,
\end{align}}
\end{sizeddisplay}
where the last equality follows from \eqref{eq: lemma 1 eq1_adj} and \eqref{eq: lemma 1 eq2_adj}.
This concludes the proof.

\subsection{Proof of Lemma \ref{lm: adress censored-main text}}
\label{subsec: proof of adress censored}

This lemma follows by changing the order of integration. Fix $i\ge 0$ and work at time $t$ with the
history window $H_{(t-i):t}$. Let $f(\cdot \mid H_{(t-i):t}\setminus Y_t)$ denote the conditional density
of $D_t$ given $H_{(t-i):t}$ excluding $Y_t$, and let $\textup{SF}(\cdot\mid H_{(t-i):t}\setminus Y_t)$
be the corresponding conditional survival function. Then
\begin{align*}
\int_{Y_t}^{D_{\max}}
\frac{\textup{SF}\!\left(c \mid H_{(t-i):t}\setminus Y_t\right)}
{\textup{SF}\!\left(Y_t \mid H_{(t-i):t}\setminus Y_t\right)}\,dc
&=
\int_{Y_t}^{D_{\max}}
\int_{c}^{D_{\max}}
\frac{f\!\left(w \mid H_{(t-i):t}\setminus Y_t\right)}
{\textup{SF}\!\left(Y_t \mid H_{(t-i):t}\setminus Y_t\right)}\,dw\,dc\\
&=
\int_{Y_t}^{D_{\max}}
\left\{\int_{Y_t}^{w}
\frac{f\!\left(w \mid H_{(t-i):t}\setminus Y_t\right)}
{\textup{SF}\!\left(Y_t \mid H_{(t-i):t}\setminus Y_t\right)}\,dc\right\}dw\\
&=
\int_{Y_t}^{D_{\max}}
(w-Y_t)\,
\frac{f\!\left(w \mid H_{(t-i):t}\setminus Y_t\right)}
{\textup{SF}\!\left(Y_t \mid H_{(t-i):t}\setminus Y_t\right)}\,dw\\
&=
\int_{Y_t}^{D_{\max}}
w\,
\frac{f\!\left(w \mid H_{(t-i):t}\setminus Y_t\right)}
{\textup{SF}\!\left(Y_t \mid H_{(t-i):t}\setminus Y_t\right)}\,dw
- Y_t\\
&=
\EE\!\left[D_t \mid \Delta_t=0,\, H_{(t-i):t}\right]-Y_t,
\end{align*}
which concludes the proof.


\subsection{Proof of Lemma \ref{lemma:optimal-policy-Pin (final)}}\label{subsec: proof of  optimalpol under C}
Before the proof, we first introduce several related notations and definitions.
Recall that the state in the underlying MDP is defined as $S_t=(X_t,Y_t,D_{t-1})$. Therefore, $W_t\setminus \Delta_{t-1}=S_t$ if and only if $\Delta_{t-1}=1$. Furthermore, we introduce the following policy definition.

\begin{definition}\label{def: optimal-under-pi-n}
Let $n\ge 0$ and let $H_{(t-n):t}$ denote the history block of length $n$ ending at time $t$.
Define the policy
\begin{equation}
\pi^\ast_n\!\big(\,\cdot\ \big|\ H_{(t-n):t}\big)
\;=\;
\sum_{i=0}^{n}
\pi^{(i)}_{n,\ast}\!\big(\,\cdot\ \big|\ H_{(t-i):t}\big)\;
\prod_{j=1}^{i}\mathbb{I}[\Delta_{t-j}=0]\;\mathbb{I}[\Delta_{t-i-1}=1].
\end{equation}

For all $t\ge 0$ and all $i=0,\ldots,n-1$, each $i$-component is greedy with respect to the corresponding conditional value:
\begin{sizeddisplay}\scriptsize
\begin{align}
\pi^{(i)}_{n,\ast}\!\big(a' \mid H_{(t-i):t}\big)
&=
\begin{cases}
1, & a'\in \argmax_{a\in\calA}\ 
\mathbb{E}\!\Big[
R_{t}
+ \gamma \mathbb{I}[\Delta_{t}=1] V^{(0)}_{\ast}(W_{t+1})
+ \gamma \mathbb{I}[\Delta_{t}=0] V^{(i+1)}_{\ast}\!\big(H_{(t-i):t+1}\big)
\ \Big|\ H_{(t-i):t},\,A_t=a
\Big], \\
0, & \text{otherwise.}
\end{cases}
\end{align}\end{sizeddisplay}

For $i=n$,
\begin{align}
\pi^{(n)}_{n,\ast}\!\big(a' \mid H_{(t-n):t}\big)
&=
\begin{cases}
1, & a'\in \argmax_{a\in \textup{UP}^{(n)}\!\big(H_{(t-n):t}\big)}\ 
\mathbb{E}\!\Big[
R_{t}
+ \gamma \mathbb{I}[\Delta_{t}=1] V^{(0)}_{\ast}(W_{t+1})
\ \Big|\ H_{(t-n):t},\,A_t=a
\Big], \\[6pt]
0, & \text{otherwise.}
\end{cases}
\end{align}
\end{definition}

Observe that for any policy $\pi \in \Pi_{n}$, where the characterization of $\Pi_{n}$ given under Definition \ref{def: set of policies such that at most n cons censoring observed}, the event $\mathcal{C}(K,n)$ is a sure event. This implies $\forall \pi \in \Pi_{n}$:
$$\EE^{\pi}\left[\bullet \given \mathcal{C}(K,n)\right]=\EE^{\pi}\left[\bullet   \right]$$
Following this, for any policy $\pi \in \Pi_{n}$, we define an offset policy $\pi_{(h)} \in \Pi_{n}$ as follows.
\begin{definition}\label{def: offset policy} $\forall \pi \in \Pi_{n}$, a corresponding offset policy $\pi_h = (\pi_{(h_0)}, \cdots, \pi_{(h_t)}, \cdots)$ is defined as that $\forall t\geq 1$ and any $(w^{(t+i)},\cdots,w^{(t)},a^{(t)},h_{t-1}) \in \calH_{t+i}$,
    \begin{align*}
    &\pi_{(h_{t-1})}(A_{i}=\bullet\given (W_{i},\cdots,W_0,A_0)=(w^{(t+i)},\cdots,w^{(t)},a^{(t)}))\\
    =&\pi(A_{t+i}=\bullet \given (W_{t+i},\cdots,W_{t},A_{t},H_{t-1})=(w^{(t+i)},\cdots,w^{(t)},a^{(t)},h_{t-1})).
\end{align*}

\end{definition}
In other words for a given $\pi \in \Pi_{n}$, the offset policy $\pi_{(h)}$  chooses actions based on a trajectory, $\tau$, according to  the same distribution that $\pi$ chooses actions based on the trajectory $(h,\tau)$.

It follows from Definition \ref{def: ith value func} that $i^{th}$ optimal state value function over a subset of policies,
$\Pi_{n}$, can be correspondingly expressed as:
\begin{align}\label{eqn: optimal_v_0, optimal}
    &V^{(i)}_{*,\mathcal{C}}(w^{(0)},a^{(0)},\cdots,w^{(i)})\nonumber \\
    &\triangleq\sup_{\pi \in \Pi_{n}}V^{(i)}_{\pi,\mathcal{C}}(w^{(0)},a^{(0)},\cdots,w^{(i)})\nonumber\\
    &=\sup_{\pi \in \Pi_{n}} \mathbb{E}^{\pi}\left[\sum_{t=i}^{\infty} \gamma^{t-i} R_{t} \mid (W_0,A_0,\cdots,W_{i})= (w^{(0)},a^{(0)},\cdots,w^{(i)}),\mathcal{C}(K,n)\right] \nonumber\\
    &=\sup_{\pi \in \Pi_{n}} \mathbb{E}^{\pi}\left[\sum_{t=i}^{\infty} \gamma^{t-i} R_{t} \mid (W_0,A_0,\cdots,W_{i})= (w^{(0)},a^{(0)},\cdots,w^{(i)})\right]
\end{align}
where, in the component of $W_{j}$, we have $\Delta_{j-1}=0$ for all $j=1,\cdots,i$, and in the component of $W_{0}$, we have $\Delta_{-1}=1$.

Note that the $i$-th state value function depends not only on the current state but also on all previous states and actions. This is because the policy $\pi$ can be history-dependent, requiring the state-value functions to be conditioned on all past information. Additionally, we only need to define the $i$-th state value functions up to $n$, as the Markov property has an order of at most $(n+1)$ throughout the entire horizon, implied by the optimization over $\Pi_{n}$. Due to this property, we will demonstrate that these $(n+1)$ state value functions are sufficient to characterize the optimal policy in this setting.


\textit{Proof}:

First, we show that once conditioned on $(H_{t-1},W_{t},A_{t},\cdots,W_{t+i})= (h,w^{(t)},a^{(t)},\cdots,w^{(t+i)})$, the supremum of the expected discounted future rewards, from time $t+i$ and onwards, is independent of $h$ given the past $i$ states and actions with $\Delta_{t-1}=1$, $\text{ and } \Delta_{t+j-1}=0  \text{ for } j=1,\cdots,i$ (i.e., the past information up to the last non-censored state). 
Specifically, we aim to show the following: 
\begin{align}\label{eq:history independent 1, optimal}
    &\sup_{\pi \in \Pi_{n}} \mathbb{E}^{\pi}\left[\sum_{t'=t+i}^{\infty} \gamma^{t'} R_{t'} \mid (H_{t-1},W_{t},A_{t},\cdots,W_{t+i})= (h,w^{(t)},a^{(t)},\cdots,w^{(t+i)})\right]\nonumber\\
    &= \gamma^{t+i} V^{(i)}_{*,\mathcal{C}}(w^{(t)},a^{(t)},\cdots,w^{(t+i)})\quad \forall \; 0\leq i\leq n.
\end{align}

Observe that since we let $\Delta_{t-1}=1$, we have the following $\forall \pi \in \Pi_{n}$:
\begin{align}
    &\mathbb{E}^{\pi}\left[\sum_{t'=t+i}^{\infty} \gamma^{t'} R_{t'} \mid (H_{t-1},W_{t},A_{t},\cdots,W_{t+i})= (h,w^{(t)},a^{(t)},\cdots,w^{(t+i)})\right]\nonumber\\
    &=\gamma^{t+i}\mathbb{E}^{\pi_{(h)}}\left[\sum_{t'=i}^{\infty} \gamma^{t'-(t+i)} R_{t'} \mid (W_{0},A_0,\cdots,W_{i})= (w^{(t)},a^{(t)},\cdots,w^{(t+i)})\right]\nonumber\\
    &=\gamma^{t+i} V^{(i)}_{\pi_{(h),\mathcal{C}}}(w^{(t)},a^{(t)},\cdots,w^{(t+i)}),
\end{align}
where the first equality is based on Definition \ref{def: offset policy} and the second equality is based on Definition \ref{def: ith value func}.

Additionally, we have that the set $\left\{\pi_{(h)}\given \forall h \in \calH_{t-1} \; \text{and} \; \pi \in \Pi_{n} \right\}$ is equal to itself by definition of $\Pi_{n}$ and $\pi_{(h)}$. This implies that:
\begin{align}\label{eq:history indep 1, optimal}
    &\sup_{\pi \in \Pi_{n}}\mathbb{E}^{\pi}\left[\sum_{t'=t+i}^{\infty} \gamma^{t'} R_{t'} \mid (H_{t-1},W_{t},A_{t},\cdots,W_{t+i})= (h,w^{(t)},a^{(t)},\cdots,w^{(t+i)})\right]\nonumber\\
    &= \sup_{\pi \in \Pi_{n}}\gamma^{t+i} V^{(i)}_{\pi_{(h),\mathcal{C}}}(w^{(t)},a^{(t)},\cdots,w^{(t+i)})\nonumber\\
    &= \sup_{\pi \in \Pi_{n}}\gamma^{t+i} V^{(i)}_{\pi,\mathcal{C}}(w^{(t)},a^{(t)},\cdots,w^{(t+i)})=\gamma^{t+i} V^{(i)}_{*,\mathcal{C}}(w^{(t)},a^{(t)},\cdots,w^{(t+i)}),
\end{align}
which further demonstrates the validity of Equation \eqref{eq:history independent 1, optimal}.

Next, we derive an upper bound on $V^{(0)}_{*,\mathcal{C}}(w^{(0)})$, where $\Delta_{-1}=1$:
\begin{align}
&V^{(0)}_{*,\mathcal{C}}(w^{(0)}) = \sup_{\pi \in \Pi_{n} }\mathbb{E}^{\pi}\left[\sum_{t'=0}^{\infty} \gamma^{t'} R_{t'} \mid W_0=w^{(0)}\right] \nonumber \\
&= \sup_{\pi \in \Pi_{n}} \mathbb{E}^{\pi}\left[R_0 + I[\Delta_0=1] \mathbb{E}^{\pi}\left[\sum_{t'=1}^{\infty} \gamma^{t'} R_{t'} \mid W_0,A_0,W_{1}\right]\right.   \label{eq:a, optimal}  \\
&\quad + \left.I[\Delta_0=0]\mathbb{E}^{\pi}\left[\sum_{t'=1}^{\infty} \gamma^{t'} R_{t'} \mid W_0,A_0,W_{1} \right] \mid W_0=w^{(0)} \right] \label{eq:b, optimal}\\
&\leq \sup_{\pi \in \Pi_{n}} \mathbb{E}^\pi\left[R_0 + \sup_{\pi'\in \Pi_{n}}  I[\Delta_0=1] \mathbb{E}^{\pi'}\left[\sum_{t'=1}^{\infty} \gamma^{t'} R_{t'} \mid W_0,A_0,W_{1} \right]\right. \nonumber \\
&\quad + \left.\sup_{\pi'' \in \Pi_{n}}  I[\Delta_0=0]\mathbb{E}^{\pi''}\left[\sum_{t'=1}^{\infty} \gamma^{t'} R_{t'} \mid W_0,A_0,W_{1} \right] \mid W_0=w^{(0)} \right] \nonumber \\
&= \sup_{a \in A} \mathbb{E}\left[R_0 + \gamma I[\Delta_0=1] V^{(0)}_{*,\mathcal{C}}(W_{1}) + \gamma I[\Delta_0=0] V^{(1)}_{*,\mathcal{C}}(W_0, A_0,W_{1})\mid (W_0, A_0)=(w^{(0)},a)\right] \nonumber\\
&\left.\quad\mid (W_0, A_0)=(w^{(0)},a)\right]    \label{eq:c, optimal}\\
&= \mathbb{E}^{\pi^{(0)}_{n},\ast} \left[ R_0 + \gamma I[\Delta_0=1]\underbrace{ V^{(0)}_{*,\mathcal{C}}(W_{1})}_{\text{(i)}} + \gamma I[\Delta_0=0] \underbrace{V^{(1)}_{*,\mathcal{C}}(W_0, A_0,W_{1})}_{\text{(ii)}}\mid W_0=w^{(0)}\right]  \label{eq:d, optimal},
\end{align}
where steps \eqref{eq:a, optimal} and \eqref{eq:b, optimal}  use the law of iterated expectations and the conditioning argument based on if the next state is censored or not.  Step  \eqref{eq:c, optimal} uses Equations \eqref{eqn: optimal_v_0, optimal} and \eqref{eq:history independent 1, optimal} and dtep  \eqref{eq:d, optimal} follows from the definition of $\pi^{(0)}_{n,*}$ given in Definition \ref{def: optimal-under-pi-n}.

We can generalize the result above for all $i=0,\cdots,n-1$ and $\forall t \geq 0$, assuming, in the component of $W_{t+j}$, we have $\Delta_{t+j-1}=0$ for all $j=1,\cdots,i$, and in the component of $W_{t}$, we have $\Delta_{t-1}=1$. Specifically,
\begin{align}
& V^{(i)}_{*,\mathcal{C}}(w^{(t)},a^{(t)},\cdots,w^{(t+i)})\nonumber\\
&= \sup_{\pi \in \Pi_{n}}\mathbb{E}^{\pi}\left[ R_{t+i} +  I[\Delta_{t+i}=1] \mathbb{E}^{\pi}\left[\sum_{t'=t+i+1}^{\infty} \gamma^{t'-(t+i)} R_{t'} \mid (H_{t-1},W_t,A_t,\cdots,W_{t+i+1})\right]\right.\nonumber\\
&\left.\quad +  I[\Delta_{t+i}=0]\mathbb{E}^{\pi}\left[\sum_{t'=t+i+1}^{\infty} \gamma^{t'-(t+i)} R_{t'} \mid (H_{t-1},W_t,A_t,\cdots,W_{t+i+1}) \right]\right.\nonumber\\
&\left.\quad \mid (H_{t-1},W_{t},A_{t},\cdots,W_{t+i})= (h,w^{(t)},a^{(t)},\cdots,w^{(t+i)})\right] \nonumber \nonumber\\
&\leq \sup_{\pi \in \Pi_{n}} \mathbb{E}\left[  R_{t+i} + \sup_{\pi'\in \Pi_{n}}  \gamma I[\Delta_{t+i}=1] \mathbb{E}^{\pi'}\left[\sum_{t'=t+i+1}^{\infty} \gamma^{t'-(t+i+1)} R_{t'} \mid (H_{t-1},W_t,A_t,\cdots,W_{t+i+1})\right]\right. \nonumber \\
&\quad + \left.\sup_{\pi'' \in \Pi_{n}} \gamma I[\Delta_{t+i}=0]\mathbb{E}^{\pi''}\left[\sum_{t'=t+i+1}^{\infty} \gamma^{t'-(t+i+1)} R_{t'} \mid (H_{t-1},W_t,A_t,\cdots,W_{t+i+1})\right] \right.\nonumber\\
&\left.\quad\mid (H_{t-1},W_{t},A_{t},\cdots,W_{t+i})= (h,w^{(t)},a^{(t)},\cdots,w^{(t+i)}) \right] \nonumber \\
&= \sup_{\pi \in \Pi_{n}} \mathbb{E}^{\pi}\left[ R_{t+i} + \gamma I[\Delta_{t+i}=1] V^{(0)}_{*,\mathcal{C}}(W_{t+i+1}) + \gamma I[\Delta_{t+i}=0] V^{(i+1)}_{*,\mathcal{C}}(W_t, A_t,\cdots,W_{t+i+1}) \right.\nonumber\\
&\left.\quad\mid (H_{t-1},W_t, A_t,\cdots,W_{t+i})=(h,w^{(t)},a^{(t)},\cdots,w^{(t+i)})\right]  \label{eq:optimal Vs, optimal}  \\
&= \sup_{a \in A} \mathbb{E}\left[ R_{t+i} + \gamma I[\Delta_{t+i}=1] V^{(0)}_{*,\mathcal{C}}(W_{t+i+1}) + \gamma I[\Delta_{t+i}=0] V^{(i+1)}_{*,\mathcal{C}}(W_t, A_t,\cdots,W_{t+i+1}) \right.\nonumber\\
&\left.\quad\mid (W_t, A_t,\cdots,W_{t+i},a)=(w^{(t)},a^{(t)},\cdots,w^{(t+i)},a)\right]  \label{eq:indepedence dyn, optimal}  \\
&=\mathbb{E}^{\pi^{(i)}_{n,*}} \left[  R_{t+i} + \gamma I[\Delta_{t+i}=1] V^{(0)}_{*,\mathcal{C}}(W_{t+i+1}) + \gamma I[\Delta_{t+i}=0] V^{(i+1)}_{*,\mathcal{C}}(W_t, A_t,\cdots,W_{t+i+1}) \right.\nonumber\\
&\left. \quad \mid (W_t,A_t,\cdots,W_{t+i})=(w^{(t)},a^{(t)},\cdots,w^{(t+i)})\right], \label{eq:upper bound Vis, optimal}
\end{align}
where Step \eqref{eq:optimal Vs, optimal} is implied by Equation \eqref{eq:history indep 1, optimal} and Step \eqref{eq:indepedence dyn, optimal} is due to the fact that the transition to next state is independent of the $H_{t-1}$ as $\Delta_{t-1}=1$

It is essential that when $i=n$, the next state should be non-censored almost surely under any policy $\pi \in \Pi_{n}$, as it has to result in at most $n$ consecutive censoring throughout the entire horizon. Therefore, we have the following when $i=n$: 
\begin{align}\label{eq: upper bound on V^n before K, optimal}
    V^{(n)}_{*,\mathcal{C}}(w^{(t)},a^{(t)},\cdots,w^{(t+n)})&\leq \sup_{\pi \in \Pi_{n }} \mathbb{E}^{\pi} \left[  R_{t+n} + \gamma I[\Delta_{t+n}=1]V^{(0)}_{*,\mathcal{C}}(W_{t+n+1}) \right.\nonumber\\
    &\left.\mid (H_{t-1},W_t,A_t,\cdots,W_{t+n})=(w^{(t)},a^{(t)},\cdots,w^{(t+n)})\right].
\end{align}

In Equation \eqref{eq: upper bound on V^n before K, optimal}, we cannot condition on $a$ and take the supremum over the entire action space, $\calA$, as previously done. This limitation arises because the next state must be uncensored almost surely, and not all actions in $\calA$ guarantee this outcome. To ensure the next state is uncensored, we must restrict the action space to a subset where, given the last $n$ states are censored, any action taken from this subset will lead to an uncensored state. Such a restricted set is denoted by $\textup{UP}^{(n)}$ and characterized in Definition \ref{def: actions leading uncensored state rigorous}. This restricted set of actions, $\textup{UP}^{(n)}$ along with Lemma \ref{lemma: nkb and action relation} leads to the following conclusion:
\begin{align}\label{eq: upper bound on v nkb}
    &V^{(n)}_{*,\mathcal{C}}(w^{(t)},a^{(t)},\cdots,w^{(t+n)})\nonumber\\
    &\leq \sup_{\pi \in \Pi_{n} } \mathbb{E}^{\pi} \left[  R_{t+n} + \gamma I[\Delta_{t+n}=1]V^{(0)}_{*,\mathcal{C}}(W_{t+n+1}) \right.\nonumber\\
    &\left.\mid (H_{t-1},W_t,A_t,\cdots,W_{t+n})=(w^{(t)},a^{(t)},\cdots,w^{(t+n)})\right].\nonumber\\
    &\leq \sup_{a \in \textup{UP}^{(n)}(W_{t+n}, \ldots, A_{t}, W_{t})} \mathbb{E}\left[ R_{t+n} + \gamma I[\Delta_{t+n}=1] V^{(0)}_{*,\mathcal{C}}(W_{t+n+1})\right.\nonumber\\
    &\left.\mid W_t, A_t, \ldots, W_{t+n}, a \right]\nonumber\\
    &=\mathbb{E}^{\pi^{(n)}_{n,\ast}}\left[ R_{t+n} + \gamma I[\Delta_{t+n}=1] V^{(0)}_{*,\mathcal{C}}(W_{t+n+1})\right.\nonumber\\
    &\left.\mid W_t, A_t, \ldots, W_{t+n} \right]
\end{align}

In the sequel, we use Equation \eqref{eq:upper bound Vis, optimal}  to plug the upper bounds of terms (i) and (ii) in Equation \eqref{eq:d, optimal}:
\begin{align}
    V^{(0)}_{*,\mathcal{C}}(w^{(0)}) &\leq \mathbb{E}^{\pi^{(0)}_{n,*}} \left[ R_0 + \gamma I[\Delta_0=1]\underbrace{ V^{(0)}_{*,\mathcal{C}}(W_{1})}_{\text{(i)}}+ \gamma I[\Delta_0=0]\underbrace{V^{(1)}_{*,\mathcal{C}}(W_0, A_0,W_{1})}_{\text{(ii)}}\mid W_0=w^{(0)} \right]  \nonumber \\
    &\leq \mathbb{E}^{\pi^{(0)}_{n,*}} \bigg[ R_0 + \gamma I[\Delta_0=1] \mathbb{E}^{\pi^{(0)}_{\mathcal{C},*}} \Big[R_{1} + \gamma I[\Delta_{1}=1]V^{(0)}_{*,\mathcal{C}}(W_{2}) \nonumber \\
    &\quad + \gamma I[\Delta_{1}=0]V^{(1)}_{*,\mathcal{C}}(W_{1}, A_{1}, W_{2}) \mid W_{1} \Big] \nonumber \\
    &\quad + \gamma I[\Delta_0=0] \mathbb{E}^{\pi^{(1)}_{\mathcal{C},*}} \Big[R_{1} + \gamma I[\Delta_{1}=1] V^{(0)}_{*,\mathcal{C}}(W_{2})\nonumber\\
    &\quad+\gamma I[\Delta_{1}=0] V^{(2)}_{*,\mathcal{C}}(W_0,A_0,\cdots,W_{2}) \mid W_0, A_0, W_{1}\Big] \mid W_0=w^{(0)} \bigg] \nonumber \\
    & = \mathbb{E}^{\pi^{\ast}_{n}} \left[ R_0 + \gamma R_{1} + \gamma^2 I[\Delta_{1}=1] V^{(0)}_{*,\mathcal{C}}(W_{2})+\gamma^2I[\Delta_{0}=1]I[\Delta_{1}=0]V^{(1)}_{*,\mathcal{C}}(W_{1},A_{1},W_{2})\right. \nonumber\\
    &\left.\quad+ \gamma^2 I[\Delta_{0}=0]I[\Delta_{1}=0] V^{(2)}_{*,\mathcal{C}}(W_{0}, A_{0},\cdots, W_{2})  \mid W_0=w^{(0)}\right].\label{eq:k}
    \end{align}

By iterating the same procedure for $n$ times, we can show the following:
    \begin{align}
    V^{(0)}_{*,\mathcal{C}}(w^{(0)})
    &\leq \mathbb{E}^{\pi^{\ast}_{n}} \left[ R_0 + \gamma R_{1} + \cdots+\gamma^{n-1}R_{n-1} \right. \nonumber\\
    &\left. +\gamma^{n }\left\{\sum_{j=0}^{n}\prod_{k=n-j}^{n-1}I[\Delta_{k}=0]I[\Delta_{n-j-1}=1] V^{(j)}_{*,\mathcal{C}}(W_{n-j},A_{n-j},\cdots,W_{n})\right\} \right.\nonumber\\
    &\left.\mid W_0=w^{(0)}\right].\nonumber
\end{align}

Recall the upper bound on $V^{(n)}_{\ast,\mathcal{C}}$ from Equation \eqref{eq: upper bound on v nkb}. Given that we are operating under the specific class of policies, $\Pi_{n}$, this ensures that after $n$ consecutive censored states, the system will transition to a non-censored state almost surely. Therefore, in the subsequent iteration, we have:
\begin{align}\label{eq:bound up to n+1, optimal}
    V^{(0)}_{*,\mathcal{C}}(w^{(0)})&\leq \mathbb{E}^{\pi^{\ast}_{n}} \left[ R_0 + \gamma R_{1}+ \cdots+\gamma^{n-1}R_{n-1}+\gamma^{n}R_{n}\right. \nonumber\\
    &\left.+ \gamma^{n} \left\{\sum_{j=0}^{n}\prod_{k=n+1-j}^{n}I[\Delta_{k}=0]I[\Delta_{n-j}=1] V^{(j)}_{*,\mathcal{C}}(W_{n+1-j},A_{n+1-j},\cdots,W_{n+1})\right\} \right.\nonumber\\
    &\left.\mid W_0=w^{(0)} \right].
\end{align}
Observe that even though the time index shifts by one (i.e. from $n$ to $n+1$), the value function indexes remain the same, from $0$ to $n$. This is the consequences of the policy class, $\Pi_{n}$, over which the supremum is taken.

Therefore, by applying the same procedure recursively infinite times, it can be shown that:
\begin{align}
     V^{(0)}_{*,\mathcal{C}}(w^{(0)})&\leq \EE^{\pi^{\ast}_{n}}\left[R_0 + \gamma R_1+ \cdots \given W_0=w^{(0)}\right] \nonumber\\
     &=V^{(0)}_{\pi^{\ast}_{n},\mathcal{C}}(w^{(0)}).
\end{align}
However, we also know by definition:$$\sup_{\pi \in \Pi_{n}}V^{(0)}_{\pi,\mathcal{C}}(w^{(0)})=V^{(0)}_{*,\mathcal{C}}(w^{(0)})\geq V^{(0)}_{\pi^{\ast}_{n},\mathcal{C}}(w^{(0)}),$$
which implies that 
$$V^{(0)}_{*,\mathcal{C}}(w^{(0)})= V^{(0)}_{\pi^{\ast}_{n},\mathcal{C}}(w^{(0)}).$$

Therefore, $\pi^{\ast}_{n} \in \argmax_{\pi \in \Pi_{n}}V^{(0)}_{\pi,\mathcal{C}}(w^{(0)})$. This result can be easily generalized to $\forall i \in \left\{0,\cdots,n\right\}$ by following exactly the same steps. Therefore, we have $\forall i \in \left\{0,\cdots,n\right\}$:
$$\pi^{\ast}_{n} \in \argmax_{\pi \in \Pi_{n}}V^{(i)}_{\pi,\mathcal{C}}(w^{(0)},a^{(0)},\cdots,w^{(i)}),$$

Therefore, given $\pi^{\ast}$ produces at most $n$ consecutive censoring, then it maximizes $\EE^{\pi'}[\sum_{t=0}^{\infty} \gamma^t R_t]$ over $\Pi_{n}$ and has the following form:
\begin{equation}
\pi^\ast\!\big(H'_{(t-n):t}\big)
\;=\;
\sum_{i=0}^{n}
\pi^{(i)}_{\ast}\!\big(\,\cdot\ \big|\ H'_{(t-i):t}\big)\;
\prod_{j=1}^{i}\mathbb{I}[\Delta_{t-j}=0]\;\cdot\;\mathbb{I}[\Delta_{t-i-1}=1],
\end{equation}
Specifically, each $i$-component selects a greedy action with respect to $Q^{(i)}_{\ast}$:
\begin{align}
\pi^{(i)}_{\ast}\!\big(a'\mid H'_{(t-i):t}\big)
&=
\begin{cases}
1, & a'\in\argmax_{a\in\mathcal{A}}\ Q^{(i)}_{\ast}\!\left(H'_{(t-i):t},a\right), \quad i=0,\ldots,n{-}1, \\[6pt]
1, & a'\in\argmax_{a\in\,\textup{UP}^{(n)}(H'_{(t-n):t})}\ Q^{(n)}_{\ast}\!\left(H'_{(t-n):t},a\right), \quad i=n, \\[4pt]
0, & \text{otherwise.}
\end{cases}
\end{align}
where  \[\qquad
V_\ast^{(i)}\!\big(H'_{(t-i):t}\big)
:=\max_{a\in\mathcal{A}_i\big((H'_{(t-i):t}\big)}
Q^{(i)}_{\ast}\big(H'_{(t-i):t},a\big) \]
with 
$$
\mathcal{A}_i\!\big(H'_{(t-i):t}\big)
:=\begin{cases}
\calA, & i< n,\\[4pt]
\textup{UP}^{( n)}\!\big(H'_{(t-n):t}\big), & i= n,
\end{cases}$$

This concludes the proof of Lemma \ref{lemma:optimal-policy-Pin (final)}. Observe that this proof is valid for any policy class $\Pi_{n^{'}}$ where $n^{'} \geq 0$. Therefore, by following the same procedure, we can show the optimal policy $\pi^*_{n_{K,b}}$ for policy class $\Pi_{n_{K,b}}$ takes the following form.
\begin{align}
    \pi_{n_{K,b}}^*(W_{t},\cdots,W_{t-n_{K,b}},A_{t-n_{K,b}}) = \left\{ \sum_{i=0}^{n_{K,b}} \pi^{(i)}_{n_{K,b},\ast}(a' \mid W_{t}, \cdots, W_{t-i}, A_{t-i}) \prod_{j=1}^{i}\left\{\mathbb{I}[\Delta_{t-j}=0]\right\}\mathbb{I}[\Delta_{t-i-1}=1] \right\}\nonumber,
\end{align}
where each $i$-component selects a greedy action with respect to its own optimal Q-function

\subsection{Proof of Lemma \ref{lm: performance difference}}\label{subsec: proof of performance difference}
 
Before the proof, we introduce several related notations and definitions. Firstly, assume that $\pi_1$ and $ \pi_2 \in \Pi_{n_{K,b}}$ where the characterizations of the policy classes are given under Definition \ref{def: set of policies such that at most n cons censoring observed}, respectively. Then, for $i=0,\cdots,n_{K,b}$, let
{\footnotesize\begin{align}\label{eq: pd lemma first eq}
    &dV^{(i)}_{\pi_1,\pi_2,\mathcal{C}}(W_t,A_{t},\cdots,W_{t+i}) = V^{(i)}_{\pi_1,\mathcal{C}}(W_t,A_{t},\cdots,W_{t+i}) - V^{(i)}_{\pi_2,\mathcal{C}}(W_t,A_{t},\cdots,W_{t+i}).\nonumber\\
    &Q^{(i)}_{\pi_1,\mathcal{C}}(W_t,A_{t},\cdots,W_{t+i},A_{t+i})-V^{(i)}_{\pi_1,\mathcal{C}}(W_t,A_{t},\cdots,W_{t+i})=A^{(i)}_{\pi_1,\mathcal{C}}(W_t,A_{t},\cdots,W_{t+i})\nonumber\\
    &Q^{(i)}_{\pi_2,\mathcal{C}}(W_t,A_{t},\cdots,W_{t+i},A_{t+i})-V^{(i)}_{\pi_2,\mathcal{C}}(W_t,A_{t},\cdots,W_{t+i})=A^{(i)}_{\pi_2,\mathcal{C}}(W_t,A_{t},\cdots,W_{t+i})\nonumber
\end{align}}
 Note that the event $\mathcal{C}(K,n_{K,b})$ occurs almost surely under any policy $\pi \in \Pi_{n_{K,b}}$. Consequently, conditioning on $\mathcal{C}(K,n_{K,b})$ does not alter any conditional distributions induced by such policies. In particular, for any $i \le n_{K,b}$ and any history block $(W_t,A_t,\ldots,W_{t+i},A_{t+i})$, we have
\begin{align}
Q^{(i)}_{\pi,\mathcal{C}}(W_t,A_t,\ldots,W_{t+i},A_{t+i})
&:= \mathbb{E}^{\pi}\!\left[\sum_{t'=i}^{\infty} \gamma^{t'-i} R_{t'} \;\middle|\; (W_t,A_t,\ldots,W_{t+i},A_{t+i}),\,\mathcal{C}(K,n_{K,b})\right] \nonumber\\
&= \mathbb{E}^{\pi}\!\left[\sum_{t'=i}^{\infty} \gamma^{t'-i} R_{t'} \;\middle|\; (W_t,A_t,\ldots,W_{t+i},A_{t+i})\right] \nonumber\\
&=: Q^{(i)}_{\pi}(W_t,A_t,\ldots,W_{t+i},A_{t+i}).
\end{align}

Therefore, throughout the analysis we may interchangeably use $Q^{(i)}_{\pi,\mathcal{C}}$ and $Q^{(i)}_{\pi}$ for all $\pi \in \Pi_{n_{K,b}}$ and $i \le n_{K,b}$, without ambiguity.

Next, for policies that belong to $\Pi_{n_{K,b}}$, we introduce the visitation measures \( d^{\pi}_{w, i} \)  \(\forall i=0,\cdots,n_{K,b} \) as follows:

\begin{definition}\label{def: generalized discounted visitation}[Generalized Discounted Visitation Measures]
Given a policy $\pi \in \Pi_{n_{K,b}}$, the generalized discounted visitation measure \(d_{w,i}^{\pi}(w^{(0)},a^{(0)},\cdots,w^{(i)})\)is defined as follows:
\[
d_{w,i}^{\pi}(w^{(0)},a^{(0)},\cdots,w^{(i)}) = (1-\gamma)\sum_{h=0}^{\infty}\gamma^h\mathbb{P}^{\pi}_{h,i}(w^{(0)},a^{(0)},\cdots,w^{(i)} \mid w)
\]
where $\mathbb{P}^{\pi}_{h,i}(w^{(0)},a^{(0)},\cdots,w^{(i)} \mid w)$ denotes the probability density of observing the trajectory segment $(w^{(0)},a^{(0)},\cdots,w^{(i)})$ occurring at time steps $t=h$ through $t=h+i$ (i.e., $W_h=w^{(0)}, \dots, W_{h+i}=w^{(i)}$), conditioned on the episode initializing at state $W_0 = w$.
\end{definition}

A natural extension of Definition \ref{def: generalized discounted visitation} is to include the final action $a^{(i)}$. In order to do that, we can multiply the previous visitation by the policy’s probability of taking that action given the history, which leads to the following extended visitation measure definition.
\begin{definition}\label{def: generalized discounted visitation2}
Given a policy $\pi \in \Pi_{n_{K,b}}$, the generalized discounted visitation measure
is defined as
\[
d_{w,i}^{\pi}\bigl(w^{(0)},a^{(0)},\ldots,w^{(i)},a^{(i)}\bigr)
:= (1-\gamma)\sum_{h=0}^{\infty}\gamma^h\,
\mathbb{P}^{\pi}_{h,i}\bigl(w^{(0)},a^{(0)},\ldots,w^{(i)},a^{(i)} \mid w\bigr),
\]
where $\mathbb{P}^{\pi}_{h,i}(w^{(0)},a^{(0)},\ldots,w^{(i)},a^{(i)} \mid w)$ denotes the probability density of observing the sequence $(w^{(0)},a^{(0)},\ldots,w^{(i)},a^{(i)})$ occurring at time steps $t=h$ through $t=h+i$ (i.e., $W_h=w^{(0)}, \ldots, A_{h+i}=a^{(i)}$), conditioned on the episode initializing at state $W_0 = w$. Here, the tuple $(w^{(0)},a^{(0)},\ldots,w^{(i)},a^{(i)})$ represents a fixed history window. Specifically, $w^{(0)}$ corresponds to a fully observed (non-censored) state, whereas for $j=1,\ldots,i$, the states $w^{(j)}$ correspond to censored observations.
\end{definition}

We have $\left\{\Delta_{t-j}=0\right\}_{j=1}^{i}$ and $\Delta_{t-i-1}=1$
Furthermore, define the following random variable $I$ as:
\begin{definition}\label{def: random var. I} $I$ is a discrete random variable with the support $\{0,1,\cdots,n_{K,b}\}$ and the probability distribution given below.
$$\prob(I=i)=\frac{\gamma^i(1-\gamma)}{1-\gamma^{n_{K,b}}}$$
\end{definition}

Finally, without loss of generality, we assume that $\Delta_{-1}=1$, i.e there is no censoring at the beginning and we let $\prod_{j=1}^{i}\left\{\mathbb{I}[\Delta_{t-j}=0]\right\}=1$ whenever $i< j$. Now, we are ready to show the proof of Lemma \ref{lm: performance difference}.

\textit{Proof:}

Consider two policies, denoted as $\pi_1$ and $\pi_2$ both of which belong to the policy class $\Pi_{n_{K,b}}$. This implies that under either $\pi_1$ or $\pi_2$, we can observe at most $n_{K,b}$ consecutive censoring throughout the entire horizon. Consequently, for any $i = 0, \ldots, n_{K,b}$ and $t \geq 0$, the following Bellman equation holds for all $\pi \in \Pi_{n_{K,b}}$:
\begin{sizeddisplay}
\footnotesize
\begin{align}
    V^{(i)}_{\pi,\mathcal{C}}(w^{(0)},a^{(0)},\cdots,w^{(i)})&=\mathbb{E}^{\pi}\left[ R_{t+i} + \gamma V^{(0)}_{\pi,\mathcal{C}}(W_{t+i+1})\mathbb{I}[\Delta_{t+i}=1]\right.\nonumber\\
    &\left.+ \gamma V^{(i+1)}_{\pi,\mathcal{C}}(W_{t},A_{t},\cdots,W_{t+i+1})\mathbb{I}[\Delta_{t+i}=0]  \right.\nonumber\\
    &\left. \mid (W_{t},A_{t},\cdots,W_{t+i})=(w^{(t)},a^{(t)},\cdots,w^{(t+i)})\right],
\end{align}\end{sizeddisplay}where, in the component of $W_{t+j}$, we have $\Delta_{t+j-1}=0$ for all $j=1,\cdots,i$, and in the component of $W_{t}$, we have $\Delta_{t-1}=1$. 

Following this, for $i=0$, the difference between state-value functions of the policies $\pi_1$ and $\pi_2$ can be written as:

{\footnotesize
\begin{align}\label{eq: dv 0}
dV^{(0)}_{\pi_1,\pi_2,\mathcal{C}}(w^{(0)}) &= V^{(0)}_{\pi_1,\mathcal{C}}(w^{(0)}) - V^{(0)}_{\pi_2,\mathcal{C}}(w^{(0)}) \nonumber\\
 &= V^{(0)}_{\pi_1,\mathcal{C}}(w^{(0)}) - \mathbb{E}^{\pi_1}\left[ R_0 + \gamma V^{(0)}_{\pi_2,\mathcal{C}}(W_1)\mathbb{I}[\Delta_{0}=1] \right.\nonumber \\
 &\quad \left. + \gamma V^{(1)}_{\pi_2,\mathcal{C}}(W_0,A_0,W_1)\mathbb{I}[\Delta_{0}=0]\mathbb{I}[\Delta_{-1}=1]  \mid W_0=w^{(0)} \right] \nonumber\\
 &\quad+\mathbb{E}^{\pi_1}\left[ R_0 + \gamma V^{(0)}_{\pi_2,\mathcal{C}}(W_1)\mathbb{I}[\Delta_{0}=1] \right. \nonumber\\
 &\quad \left. + \gamma V^{(1)}_{\pi_2,\mathcal{C}}(W_0,A_0,W_1)\mathbb{I}[\Delta_{0}=0]\mathbb{I}[\Delta_{-1}=1]  \mid W_0=w^{(0)} \right] -V^{(0)}_{\pi_2,\mathcal{C}}(w^{(0)}) \nonumber\\
&=\gamma\mathbb{E}^{\pi_1}\left[dV^{(0)}_{\pi_1,\pi_2,\mathcal{C}}(W_1)\mathbb{I}[\Delta_{0}=1] + dV^{(1)}_{\pi_1,\pi_2,\mathcal{C}}(W_0,A_0,W_1)\mathbb{I}[\Delta_{0}=0]\mathbb{I}[\Delta_{-1}=1] \mid W_0=w^{(0)}\right]\nonumber\\
&\quad+\mathbb{E}^{\pi_1}\left[Q_{\pi_2,\mathcal{C}}^{(0)}(W_0,A_0) - V^{(0)}_{\pi_2,\mathcal{C}}(W_0) \mid W_0=w^{(0)} \right]\nonumber\\
&=\gamma\mathbb{E}^{\pi_1}\left[dV^{(0)}_{\pi_1,\pi_2,\mathcal{C}}(W_1)\mathbb{I}[\Delta_{0}=1] + dV^{(1)}_{\pi_1,\pi_2,\mathcal{C}}(W_0,A_0,W_1)\mathbb{I}[\Delta_{0}=0]\mathbb{I}[\Delta_{-1}=1] \mid W_0=w^{(0)}\right]\nonumber\\
&\quad +\mathbb{E}^{A_0 \sim \pi_1}\left[ A^{(0)}_{\pi_2,\mathcal{C}}(W_0,A_0)\mid W_0=w^{(0)}\right].
\end{align}}

The equation above can be generalized for $i=1,\cdots,n_{K,b}-1$ by following the same procedure given below.
\begin{sizeddisplay}
{\footnotesize}
\begin{align*}
dV^{(i)}_{\pi_1,\pi_2,\mathcal{C}}(w^{(0)},a^{(0)},\cdots,w^{(i)})&=\mathbb{E}^{\pi_1}\left[\gamma dV^{(0)}_{\pi_1,\pi_2,\mathcal{C}}(W_{i+1})\mathbb{I}[\Delta_{i}=1] \right.\\
&\left.\quad +\gamma dV^{(i+1)}_{\pi_1,\pi_2,\mathcal{C}}(W_{0},A_{0},\cdots,W_{i+1})\prod_{j=0}^{i}\left\{\mathbb{I}[\Delta_{j}=0]\right\}\mathbb{I}[\Delta_{-1}=1] \right.\nonumber\\
&\quad \left. + A^{(i)}_{\pi_2,\mathcal{C}}(W_{0},A_{0},\cdots,W_{i},A_{i})\given (W_{0},A_{0},\cdots,W_{i})=(w^{(0)},a^{(0)},\cdots,w^{(i)})\right].
\end{align*}
\end{sizeddisplay}

By using the recursive relationship, Equation \eqref{eq: dv 0} becomes:
\begin{sizeddisplay}
{\footnotesize}
\begin{align*}
    dV^{(0)}_{\pi_1,\pi_2,\mathcal{C}}(w^{(0)}) 
&=\gamma^2\mathbb{E}^{\pi_1}\left[dV^{(0)}_{\pi_1,\pi_2,\mathcal{C}}(W_2)\mathbb{I}[\Delta_{1}=1] +dV^{(1)}_{\pi_1,\pi_2,\mathcal{C}}(W_1,A_1,W_2)\mathbb{I}[\Delta_{1}=0]\mathbb{I}[\Delta_{0}=1] \right.\\
&\left. \quad+ dV^{(2)}_{\pi_2,\mathcal{C}}(W_0,A_0,W_1,A_1,W_2)\mathbb{I}[\Delta_{1}=0]\mathbb{I}[\Delta_{0}=0]\mathbb{I}[\Delta_{-1}=1] \given W_0=w^{(0)}\right] \\
&\quad + \gamma\mathbb{E}^{\pi_1}\left[A^{(0)}_{\pi_2,\mathcal{C}}(W_1,A_1)\mathbb{I}[\Delta_{0}=1] + A^{(1)}_{\pi_2,\mathcal{C}}(W_0,A_0,W_1,A_1)\mathbb{I}[\Delta_{0}=0]\mathbb{I}[\Delta_{-1}=1] \given W_0=w^{(0)},\mathcal{C}(K,n) \right] \\
&\quad + \EE^{\pi_1}\left[ A^{(0)}_{\pi_2,\mathcal{C}}(W_0,A_0)\given W_0=w^{(0)}\right].
\end{align*}
\end{sizeddisplay}

By doing the same steps recursively for $n_{K,b}$ times, we can show the following:

{\footnotesize\begin{align}
    dV^{(0)}_{\pi_1,\pi_2,\mathcal{C}}(w^{(0)}) &=\gamma^{n_{K,b}}\EE^{\pi}\left[\sum_{i=0}^{n_{K,b}}dV^{(i)}_{\pi_1,\pi_2,\mathcal{C}}(W_{n_{K,b}-i},A_{n_{K,b}-i},\cdots,W_{n_{K,b}})\prod_{j=0}^{i-1}\left\{\mathbb{I}[\Delta_{n_{K,b}-i+j}=0]\right\}\mathbb{I}[\Delta_{n_{K,b}-i-1}=1]\right.\nonumber\\
    &\left.\quad\given W_0=w^{(0)}\right]\nonumber\\
&\quad + \sum_{h=0}^{n_{K,b}-1}\gamma^h\EE^{\pi}\left[  \sum_{i=0}^{\min(h,n_{K,b})}A^{(i)}_{\pi_2,\mathcal{C}}(W_{h-i},A_{h-i},\cdots,W_{h},A_{h})\prod_{j=0}^{i-1}\left\{\mathbb{I}[\Delta_{h-i+j}=0]\right\}\mathbb{I}[\Delta_{h-i-1}=1] \right.\nonumber\\
&\left.\quad\given W_0=w^{(0)}\right].
\end{align}}

Observe that when $i=n_{K,b}$, we have the following Bellman equation
\begin{sizeddisplay}
{\footnotesize}
\begin{align}
Q^{(n_{K,b})}_{\pi_2,\mathcal{C}}(w^{(t)},a^{(t)},\cdots,w^{(t+n_{K,b})},a^{(t+n_{K,b})})&=\EE\left[R_{t+n_{K,b}} +\gamma V_{\pi_2,\mathcal{C}}^{(0)}(W_{t+n_{K,b}+1})\mathbb{I}[\Delta_{t+n_{K,b}}=1]\right.\nonumber\\
    &\quad \left.\given (W_{t},A_{t},\cdots, W_{t+n_{K,b}},A_{t+n_{K,b}})=(w^{(t)},a^{(t)},\cdots,w^{(t+n_{K,b})},a^{(t+n_{K,b})})\right] \nonumber\\
    &=\EE\left[R_{t+n_{K,b}} +\gamma V_{\pi_2,\mathcal{C}}^{(0)}(W_{t+n_{K,b}+1})\right.\nonumber\\
    &\left.\quad\given (W_{t},A_{t},\cdots, W_{t+n_{K,b}},A_{t+n_{K,b}})=(w^{(t)},a^{(t)},\cdots,w^{(t+n_{K,b})},a^{(t+n_{K,b})})\right]\nonumber,
\end{align}
\end{sizeddisplay}which is only valid $\forall a^{(t+n_{K,b})} \in \textup{UP}^{(n_{K,b})}(w^{(t+n_{K,b})},\cdots,a^{(t)},w^{(t)})$. In other words, Bellman equation is not well-defined for all $A_{t+n_{K,b}}=a \in \calA$ when $i=n_{K,b}$. This is due to the fact that $\pi_2$ belongs to $\Pi_{n_{K,b}}$, and thus, cannot produce more consecutive censoring instances than $n_{K,b}$.

As a result, the following equation holds because $\pi_1$, by definition, selects an action from the set $\textup{UP}^{(n_{K,b}}(W_{t+n_{K,b}}, A_{t+n_{K,b}}, \ldots, W_{t})$ when faced with $n_{K,b}$ consecutive censored states:
\begin{sizeddisplay}
{\footnotesize}
\begin{align}
    &\EE^{\pi_1}\left[A^{(n_{K,b})}_{\pi_2,\mathcal{C}}(W_{t},A_{t},\cdots, W_{t+n_{K,b}},A_{t+n_{K,b}})\given (W_{t},A_{t},\cdots, W_{t+n_{K,b}})=(w^{(t)},a^{(t)},\cdots,w^{(t+n_{K,b})})\right]\nonumber\\
    &=\mathbb{E}^{\pi_1}\left[Q_{\pi_2,\mathcal{C}}^{(0)}(W_{t},A_{t},\cdots, W_{t+n_{K,b}},A_{t+n_{K,b}}) - V^{(0)}_{\pi_2,\mathcal{C}}(W_{t},A_{t},\cdots, W_{t+n_{K,b}}) \right.\nonumber\\
    &\left.\given (W_{t},A_{t},\cdots, W_{t+n_{K,b}})=(w^{(t)},a^{(t)},\cdots,w^{(t+n_{K,b})}) \right].
\end{align}
\end{sizeddisplay}
This further implies the following:
\begin{sizeddisplay}
{\footnotesize} 
\begin{align}
    &dV^{(n_{K,b})}_{\pi_1,\pi_2,\mathcal{C}}(w^{(t)},a^{(t)},\cdots,w^{(t+n_{K,b})})\nonumber\\
&= \mathbb{E}^{\pi_1}\left[\gamma dV^{(0)}_{\pi_1,\pi_2,\mathcal{C}}(W_{t+n_{K,b}+1})\mathbb{I}[\Delta_{t+n_{K,b}}=1]\right.\nonumber\\
&\quad \left. + A^{(n_{K,b})}_{\pi_2,\mathcal{C}}(W_{t},A_{t},\cdots,W_{t+n_{K,b}},A_{t+n_{K,b}}) \given (W_{t},A_{t},\cdots,W_{t+n_{K,b}})=(w^{(t)},a^{(t)},\cdots,w^{(t+n_{K,b})})\right].\nonumber
\end{align}
\end{sizeddisplay}

Therefore, by recursively applying the same procedure, we can demonstrate the that
\begin{sizeddisplay}
{\footnotesize}
\begin{align}\label{eq: relation suboptimal and pi final 0}
&dV^{(0)}_{\pi_1,\pi_2,\mathcal{C}}(w^{(0)})\nonumber\\
    &=\sum_{h=0}^{\infty}\gamma^h\left\{\sum_{i=0}^{n_{K,b}}\gamma^i\EE^{\pi_1}\left[A^{(i)}_{\pi_2,\mathcal{C}}(W_{h},A_{h},\cdots,W_{h+i},A_{h+i})\prod_{j=0}^{i-1}\left\{\mathbb{I}[\Delta_{h+j}=0]\right\}\mathbb{I}[\Delta_{h-1}=1]\given W_0=w^{(0)}\right] \right\}\nonumber\\
&=\sum_{h=0}^{\infty}\gamma^h\left\{\sum_{i=0}^{n_{K,b}}\gamma^i\EE_{\substack{W_{h},A_{h},\cdots W_{h+i} \sim P_{h,i}^{\pi_1}(\cdot \mid w^{(0)}) \\ A_{h+i} \sim \pi_1}}\left[A^{(i)}_{\pi_2,\mathcal{C}}(W_{h},A_{h},\cdots,W_{h+i},A_{h+i})\prod_{j=0}^{i-1}\left\{\mathbb{I}[\Delta_{h+j}=0]\right\}\mathbb{I}[\Delta_{h-1}=1]\right]\right\}
\end{align}
\end{sizeddisplay}

With some abuse of notation, assume that $W^{(i)}$ and $A^{(i)}$ with associated censoring indicator $\Delta^{(i)}$ represent generic random vectors. Therefore, under the Definitions \ref{def: generalized discounted visitation} and \ref{def: random var. I}, Equation \eqref{eq: relation suboptimal and pi final 0}  can be written as follows:
\begin{sizeddisplay}
{\footnotesize}
\begin{align}\label{eq: performance difference final eq0}   
&dV^{(0)}_{\pi_1,\pi_2,\mathcal{C}}(w^{(0)}) \\
&=\frac{1}{1-\gamma}\left\{\sum_{i=0}^{n_{K,b}}\gamma^i\EE_{\substack{W^{(0)},A^{(0)},\cdots,W^{(i)} \sim d_{w^{(0)},i}^{\pi_1}\nonumber\\ A^{(i)} \sim \pi_1}}\left[A^{(i)}_{\pi_2,\mathcal{C}}(W^{(0)},A^{(0)},\cdots,W^{(i)},A^{(i)})\prod_{j=1}^{i}\left\{\mathbb{I}[\Delta^{(j)}=0]\right\}\mathbb{I}[\Delta^{(0)}=1]\right] \right\}\nonumber\\
&=\frac{1-\gamma^{n_{K,b}}}{(1-\gamma)^2}\EE_{(W^{(0)},A^{(0)},\cdots,W^{(I)}, A^{(I)}) \sim (d_{w^{(0)},I}^{\pi_1} \times \pi_1)
    }\left[A^{(I)}_{\pi_2,\mathcal{C}}(W^{(0)},A^{(0)},\cdots,W^{(I)},A^{(I)})\prod_{j=1}^{I}\left\{\mathbb{I}[\Delta^{(j)}=0]\right\}\mathbb{I}[\Delta^{(0)}=1]\right]\nonumber.
\end{align}
\end{sizeddisplay}

By further take the expectation with respect to the initial state $w^{(0)}$ which follows $\nu$, Equation \eqref{eq: performance difference final eq0} can be expressed as:

\begin{sizeddisplay}
{\footnotesize}
\begin{align}\label{eq: performance difference final eq}   
&\EE^{\pi_1}\left[\sum_{t=0}^{\infty}\gamma^{t}R_t\right]-\EE^{\pi_2}\left[\sum_{t=0}^{\infty}\gamma^{t}R_t\right] \nonumber\\
&=\frac{1-\gamma^{n_{K,b}}}{(1-\gamma)^2}\EE_{(W^{(0)},A^{(0)},\cdots,W^{(I)}, A^{(I)}) \sim (d_{\nu,I}^{\pi_1} \times \pi_1)
    }\left[A^{(I)}_{\pi_2,\mathcal{C}}(W^{(0)},A^{(0)},\cdots,W^{(I)},A^{(I)})\prod_{j=1}^{I}\left\{\mathbb{I}[\Delta^{(j)}=0]\right\}\mathbb{I}[\Delta^{(0)}=1]\right],
\end{align}
\end{sizeddisplay}
which concludes our proof of Lemma \ref{lm: performance difference}.


\subsection{Proof of Lemma \ref{lm:upper bound on Qs}}\label{subsec: proof of upper bound Qs fqi}
Before the proof, we introduce several related notations and definitions.
For $i=0,\cdots,n_{K,b}$, let 
\begin{align}\label{def: tilde Q} \overline{Q}^{(i)}_j(x^{(0)}, \ldots, x^{(i)}) = \begin{cases}
        \widehat{Q}^{(i)}_j(x^{(0)}, \ldots, x^{(i)}) - U_j^{(i)}(x^{(0)}, \ldots, x^{(i)}), & j = 1, \ldots, K-1\\
        \widehat{Q}^{(i)}_K(x^{(0)}, \ldots, x^{(i)}) - C^{(i)}(x^{(0)}, \ldots,x^{(i-1)}, w^{(i)}), & j = K
    \end{cases}
\end{align}

Observe that the greedy policy derived from the pairs $\widehat{Q}^{(i)}_K(x^{(0)},\cdots,x^{(i)})$ and $\overline{Q}^{(i)}_K(x^{(0)},\cdots,x^{(i)})$ is the same. 

Next, $\forall j=1,\cdots,K$ and $\forall i=0,\cdots,n_{K,b}$, define:
\begin{align}\label{eq: error FQI}
    \rho_j^{(i)}(x^{(0)},\cdots,x^{(i)}) = \calT^{\hat{\pi}_{j-1}}_{i,\mathcal{C}} \widehat{Q}_{j-1}(x^{(0)},\cdots,x^{(i)}) - \overline{Q}^{(i)}_{j}(x^{(0)},\cdots,x^{(i)}),
\end{align} 
where $\calT^{\hat{\pi}_{j-1}}_{i,\mathcal{C}}$ is defined under Definition \ref{def: bellman operator}.

Furthermore, note that the event $\mathcal{C}(K,n_{K,b})$ occurs almost surely under any policy $\pi \in \Pi_{n_{K,b}}$. Consequently, conditioning on $\mathcal{C}(K,n_{K,b})$ does not alter any conditional distributions induced by such policies. In particular, for any $i \le n_{K,b}$ and any history block $(W_t,A_t,\ldots,W_{t+i},A_{t+i})$, we have
\begin{align}
Q^{(i)}_{\pi,\mathcal{C}}(W_t,A_t,\ldots,W_{t+i},A_{t+i})
&:= \mathbb{E}^{\pi}\!\left[\sum_{t'=i}^{\infty} \gamma^{t'-i} R_{t'} \;\middle|\; (W_t,A_t,\ldots,W_{t+i},A_{t+i}),\,\mathcal{C}(K,n_{K,b})\right] \nonumber\\
&= \mathbb{E}^{\pi}\!\left[\sum_{t'=i}^{\infty} \gamma^{t'-i} R_{t'} \;\middle|\; (W_t,A_t,\ldots,W_{t+i},A_{t+i})\right] \nonumber\\
&=: Q^{(i)}_{\pi}(W_t,A_t,\ldots,W_{t+i},A_{t+i}).
\end{align}

Therefore, throughout the analysis we may interchangeably use $Q^{(i)}_{\pi,\mathcal{C}}$ and $Q^{(i)}_{\pi}$ for all $\pi \in \Pi_{n_{K,b}}$ and $i \le n_{K,b}$, without ambiguity. For instance, since $\pi_{n_{K,b}} \in \Pi_{n_{K,b}}$, this result will imply that $$Q^{(i)}_{\pi^{\ast}_{n_{K,b}}}(x^{(0)},\cdots,x^{(i)})=Q^{(i)}_{\pi^{\ast}_{n_{K,b},\mathcal{C}}}(x^{(0)},\cdots,x^{(i)}).$$

Additionally, recall that we denote uncertainty quantifier for C-FQI by ${U}_{k}^{(i)}$. Under Assumption~\ref{ass: UQ ass} which asserts the existence of an event $\Omega^{(i)}$ with $\Pr(\Omega^{(i)})\ge 1-\epsilon$ where ${U}_{k}^{(i)}$ satisfies Definition~\ref{def:UE} and is therefore a valid Uncertainty Quantifier. Because the proof requires $U_{k}^{(i)}$ to be a valid uncertainty quantifier $\forall i=0,\cdots, n_{K,b}$, we work on the intersection event
\[
\Omega \;:=\;\bigcap_{i=0}^{n_{K,b}}\Omega^{(i)}.
\]
which holds with probability at least $1-(n_{K,b}+1)\epsilon$. And on $\Omega$, we have $\forall i=0,\cdots, n_{K,b}$ and $\forall j=1,\cdots,K-1$ 
\begin{align}\label{eq: upper bound rho fqi}
    {\rho}_j^{ (i)}(x^{(0)},\cdots,x^{(i)}) 
    &= \calT^{\hat{\pi}_{j-1}}_{i,\mathcal{C}} \widehat{Q}_{j-1}(x^{(0)},\cdots,x^{(i)}) - \overline{Q}^{(i)}_{j}(x^{(0)},\cdots,x^{(i)})\nonumber\\
    &= \calT^{\hat{\pi}_{j-1}}_{i,\mathcal{C}} \widehat{Q}_{j-1}(x^{(0)},\cdots,x^{(i)}) - (\widehat{Q}^{(i)}_{j}(x^{(0)},\cdots,x^{(i)}) - {U}^{(i)}_j(x^{(0)},\cdots,x^{(i)}))\nonumber\\
    &\leq  2 {U}^{(i)}
_j(x^{(0)},\cdots,x^{(i)}).
\end{align}
and for $j=K$, we have
\begin{align}\label{eq: upper bound rho2 fqi}
    {\rho}_K^{ (i)}(x^{(0)},\cdots,x^{(i)}) 
    &= \calT^{\hat{\pi}_{K-1}}_{i,\mathcal{C}} \widehat{Q}_{K-1}(x^{(0)},\cdots,x^{(i)}) - \overline{Q}^{(i)}_{K}(x^{(0)},\cdots,x^{(i)})\nonumber\\
    &= \calT^{\hat{\pi}_{K-1}}_{i,\mathcal{C}} \widehat{Q}_{K-1}(x^{(0)},\cdots,x^{(i)}) - (\widehat{Q}^{(i)}_{K}(x^{(0)},\cdots,x^{(i)}) -  C^{(i)}(x^{(0)}, \ldots,x^{(i-1)}, w^{(i)})\nonumber\\
    &\leq   C^{(i)}(x^{(0)}, \ldots,x^{(i-1)}, w^{(i)})+ {U}^{(i)}
_K(x^{(0)},\cdots,x^{(i)}).
\end{align}

These identities will be used in the following proof.

In addition, under Assumption~\ref{ass: surrogate outcome}, we work on the event $\Omega^r$ on which the surrogate reward relation holds. In particular, we have $\Omega^r$,
\begin{align}\label{eq: reward dis fqi upper}
&\EE\left[\left(\widehat{R}_{i}\right)\prod_{j=0}^{i}\mathbb{I}[\Delta_{j}=0]\mathbb{I}[\Delta_{-1}=1]\given X_{0},\cdots,X_{i},\mathcal{C}(K,n_{K,b})\right]\nonumber\\
    &\geq\EE\left[\left(\widetilde{R}_{i} - C_5(\varepsilon)(NT)^{-\delta}\right)\prod_{j=0}^{i}\mathbb{I}[\Delta_{j}=0]\mathbb{I}[\Delta_{-1}=1]\given X_{0},\cdots,X_{i},\mathcal{C}(K,n_{K,b})\right],
\end{align}
which we use to account for the estimation error in the surrogate reward.

Accordingly, Lemma \ref{lm:upper bound on Qs pessimistic} is proved on the joint event $\Omega_P \cap \Omega^r$, i.e., under the simultaneous validity of the uncertainty quantification bounds across all depths and the surrogate reward approximation. By union bound, we have $\prob(\Omega_P \cap \Omega^r) \geq 1-(n_{K,b}+1)\epsilon-\varepsilon$.

Now, we are ready to show the proof.

\textit{Proof:}

We focus on deriving an upper bound on the difference between $Q^{(i)}_{\pi^{\ast}_{n_{K,b}},\mathcal{C}}(x^{(0)},\cdots,x^{(i)})$ and $\overline{Q}_{K}^{(i)}(x^{(0)},\cdots,x^{(i)})$ $\forall i=0,\cdots,n_{K,b}$:
\begin{sizeddisplay}
{\footnotesize}
\begin{align}\label{eq: fqi bound Q first equation}
I_{i,K}
(x^{(0)},\cdots,x^{(i)})&
\triangleq Q^{(i)}_{\pi^{\ast}_{n_{K,b}},\mathcal{C}}(x^{(0)},\cdots,x^{(i)})-\overline{Q}_{K}^{(i)}(x^{(0)},\cdots,x^{(i)}) \nonumber\\
&=Q^{(i)}_{\pi^{\ast}_{n_{K,b}},\mathcal{C}}(x^{(0)},\cdots,x^{(i)}) - \calT^{\hat{\pi}_{K-1}}_{i,\mathcal{C}}\widehat{Q}_{K-1}(x^{(0)},\cdots,x^{(i)}) + \rho_K^{(i)}(x^{(0)},\cdots,x^{(i)})\nonumber\\
    &= Q^{(i)}_{\pi^{\ast}_{n_{K,b}},\mathcal{C}}(x^{(0)},\cdots,x^{(i)}) - \calT^{\pi^{\ast}_{n_{K,b}}}_{i,\mathcal{C}}\widehat{Q}_{K-1}(x^{(0)},\cdots,x^{(i)})+ \rho_K^{(i)}(x^{(0)},\cdots,x^{(i)})\nonumber\\
    &+ \underbrace{\calT^{\pi^{\ast}_{n_{K,b}}}_{i,\mathcal{C}}\widehat{Q}_{K-1}(x^{(0)},\cdots,x^{(i)})-\calT^{\hat{\pi}_{K-1}}_{i,\mathcal{C}}\widehat{Q}_{K-1}(x^{(0)},\cdots,x^{(i)})}_{(1)}, 
\end{align}
\end{sizeddisplay}
where we use Equation \eqref{eq: error FQI} to derive the equality.

Observe that given the optimality of $\widehat{\pi}^{(i)}_{j}$ with respect to $\widehat{Q}_{j}^{(i)}$ for all $i=0,\cdots,n_{K,b}$ and for all $j=1,\cdots, K-1$, the following bound holds for all $i=0,\cdots,n_{K,b}$ and $j=1,\cdots, K-1$:

\begin{sizeddisplay}{\footnotesize}
    \begin{align}\label{eq: (1) 1}
(1)&=\calT^{\pi^{\ast}_{n_{K,b}}}_{i,\mathcal{C}}\widehat{Q}_{j}(x^{(0)},\cdots,x^{(i)}) - \calT^{\hat{\pi}_{j}}_{i,\mathcal{C}}\widehat{Q}_{j}(x^{(0)},\cdots,x^{(i)})\nonumber\\
    &=\EE\left[\gamma\left(\widehat{Q}_{j}^{(0)}(W_{i+1},\pi_{n_{K,b},*}^{(0)}(W_{i+1}))-\widehat{Q}_{j}^{(0)}(W_{i+1},\widehat{\pi}^{(0)}_{j}(W_{i+1})\right)\mathbb{I}[\Delta_{i}=1] \right.\nonumber\\
    &\left.  + \gamma\left(\widehat{Q}_{j}^{(i+1)}(X_0,\cdots,W_{i+1},\pi^{(i+1)}_{n_{K,b,\ast}}(W_{i+1},\cdots,X_0))-\widehat{Q}_{j}^{(i+1)}(X_0,\cdots,W_{i+1},\widehat{\pi}^{(i+1)}_{j}(W_{i+1},\cdots,X_0))\right)\mathbb{I}[\Delta_{i}=0]\right.\nonumber\\
    &\quad\left.\given (X_0,\cdots,X_{i})=(x^{(0)},\cdots,x^{(i)}),\mathcal{C}(K,n_{K,b})\right]\nonumber\\
    &\leq 0.
\end{align}
\end{sizeddisplay}
Then Equation \eqref{eq: fqi bound Q first equation} can be upper bounded as:
\begin{sizeddisplay}
{\footnotesize}\begin{align}\label{eq:ffqi upper bound on Q}
    I_{i,K}(x^{(0)},\cdots,x^{(i)})  &\leq Q^{(i)}_{\pi^{\ast}_{n_{K,b}},\mathcal{C}}(x^{(0)},\cdots,x^{(i)}) - \calT^{\pi^{\ast}_{n_{K,b}}}_{i,\mathcal{C}}\widehat{Q}_{K-1}(x^{(0)},\cdots,x^{(i)}) + \rho_K^{(i)}(x^{(0)},\cdots,x^{(i)}) \nonumber\\
    &= \calT^{\pi^{\ast}_{n_{K,b}}}_{i,\mathcal{C}}Q_{\pi^{\ast}_{n_{K,b}},\mathcal{C}}(x^{(0)},\cdots,x^{(i)}) - \calT^{\pi^{\ast}_{n_{K,b}}}_{i,\mathcal{C}}\widehat{Q}_{K-1}(x^{(0)},\cdots,x^{(i)})+\rho_K^{(i)}(x^{(0)},\cdots,x^{(i)})\nonumber\\
    &=\gamma \calP^{\pi^{\ast}_{n_{K,b}}}_i\left(\big(Q^{(0)}_{\pi^{\ast}_{n_{K,b}},\mathcal{C}}(X_{i+1}) - \widehat{Q}_{K-1}^{(0)}(X_{i+1})\big)\mathbb{I}[\Delta_{i}=1] \right.\nonumber\\
    & \left.+ \big(Q^{(i+1)}_{\pi^{\ast}_{n_{K,b}},\mathcal{C}}(X_0,\cdots,X_{i+1}) - \widehat{Q}_{K-1}^{(i+1)}(X_0,\cdots,X_{i+1})\big)\prod_{j=0}^{i}\mathbb{I}[\Delta_{j}=0]\mathbb{I}[\Delta_{-1}=1]\right)\nonumber\\
    &+\EE\left[\left(R_{i}-\widehat{R}_{i}\right)\prod_{j=0}^{i}\mathbb{I}[\Delta_{j}=0]\mathbb{I}[\Delta_{-1}=1]\given X_0,\cdots,X_{i})=(x^{(0)},\cdots,x^{(i)}),\mathcal{C}(K,n_{K,b})\right]\nonumber\\
    &+\rho_K^{(i)}(x^{(0)},\cdots,x^{(i)}),
\end{align}
\end{sizeddisplay}
where the characterization of $\calP^{\pi^{\ast}_{n_{K,b}}}_i$ given in Definition \ref{def: P as conditional exp}.

By Lemma \ref{lm: identification for censored demand} and on event $\Omega \cap \Omega^r$, the following holds
\begin{sizeddisplay}
{\footnotesize}
\begin{align*}
    &\EE\left[\left(R_{i}-\widehat{R}_{i}\right)\prod_{j=0}^{i}\mathbb{I}[\Delta_{j}=0]\mathbb{I}[\Delta_{-1}=1]\given X_{0},\cdots,X_{i},\mathcal{C}(K,n_{K,b})\right]\\
    &\leq\EE\left[\left(R_{i}-\widetilde{R}_{i} + C_5(\varepsilon)(NT)^{-\delta}\right)\prod_{j=0}^{i}\mathbb{I}[\Delta_{j}=0]\mathbb{I}[\Delta_{-1}=1]\given X_{0},\cdots,X_{i}\right]\\
    &\leq C_5(\varepsilon)(NT)^{-\delta},
\end{align*}\end{sizeddisplay}
where the first inequality is implied by Equation~\eqref{eq: reward dis fqi upper}.
Then, Equation \eqref{eq:ffqi upper bound on Q} becomes
{\footnotesize\begin{align}\label{eq: (1) (2)}
    I_{i,K}(x^{(0)},\cdots,x^{(i)})  &\leq \gamma \calP^{\pi^{\ast}_{n_{K,b}}}_i\left(\big(\underbrace{Q^{(0)}_{\pi^{\ast}_{n_{K,b}},\mathcal{C}}(X_{i+1}) - \widehat{Q}_{K-1}^{(0)}(X_{i+1})}_{(2)}\big)\mathbb{I}[\Delta_{i}=1] \right.\nonumber\\
    & \left.+ \big(\underbrace{Q^{(i+1)}_{\pi^{\ast}_{n_{K,b}},\mathcal{C}}(X_0,\cdots,X_{i+1}) - \widehat{Q}_{K-1}^{(i+1)}(X_0,\cdots,X_{i+1})}_{(3)}\big)\prod_{j=0}^{i}\mathbb{I}[\Delta_{j}=0]\mathbb{I}[\Delta_{-1}=1]\right)\nonumber\\
    &+C_5(\varepsilon)(NT)^{-\delta}+ \rho_K^{(i)}(x^{(0)},\cdots,x^{(i)}).
\end{align}}
Observe that 
{\footnotesize\begin{align*}
    (2)+U_{K-1}^{(0)}(X_{i+1})=I_{0,K-1}(X_{i+1})
\end{align*}}
{\footnotesize\begin{align*}
    (3)+U_{K-1}^{(i+1)}(X_0,\cdots,X_{i+1})=I_{i+1,K-1}(X_0,\cdots,X_{i+1}).
\end{align*}}

Therefore, Equation \eqref{eq: (1) (2)} becomes
{\footnotesize\begin{align}\label{eq:fqi upper bound on Q}
    I_{i,K}(x^{(0)},\cdots,x^{(i)})  
    &\leq\gamma \calP^{\pi^{\ast}_{n_{K,b}}}_i\left(\big(I_{0,K-1}(X_{i+1})-U_{K-1}(X_{i+1})\big)\mathbb{I}[\Delta_{i}=1] \right.\nonumber\\
    & \left.+ \big(I_{i+1,K-1}(X_0,\cdots,X_{i+1})-U_{K-1}^{(i+1)}(X_0,\cdots,X_{i+1})\big)\prod_{j=0}^{i}\mathbb{I}[\Delta_{j}=0]\mathbb{I}[\Delta_{-1}=1]\right)\nonumber\\
    & +C_5(\varepsilon)(NT)^{-\delta}+ \rho_K^{(i)}(x^{(0)},\cdots,x^{(i)}).
\end{align}}

Next, we use Equation \eqref{eq: (1) (2)} to further upper bound the right hand side of Equation \eqref{eq:fqi upper bound on Q} in an iterative manner as follows:
\begin{sizeddisplay}
 {\footnotesize}
\begin{align*}
I_{i,K}(x^{(0)},\cdots,x^{(i)}) &\leq \gamma \calP^{\pi^{\ast}_{n_{K,b}}}_i\left\{\left(\gamma \calP^{\pi^{\ast}_{n_{K,b}}}_0\left(\big(Q^{(0)}_{\pi^{\ast}_{n_{K,b}},\mathcal{C}}(X_{i+2}) - \widehat{Q}_{K-2}^{(0)}(X_{i+2})\big)\mathbb{I}[\Delta_{i+1}=1] \right.\right.\right.\\
&\left.\left.\left.+ \big(Q^{(1)}_{\pi^{\ast}_{n_{K,b}},\mathcal{C}}(X_{i+1},X_{i+2}) - \widehat{Q}_{K-2}^{(1)}(X_{i+1},X_{i+2})\big)\mathbb{I}[\Delta_{i+1}=0]\mathbb{I}[\Delta_{i}=1]\right) \right.\right.\\
&\left.\left.+C_5(\varepsilon)(NT)^{-\delta}+ \rho_{K-1}^{(0)}(X_{i+1})-U_{K-1}^{(0)}(X_{i+1})\right)\mathbb{I}[\Delta_{i}=1]\right.\\
&\left.+\left(\gamma\calP^{\pi^{\ast}_{n_{K,b}}}_{i+1}\left(\big(Q^{(0)}_{\pi^{\ast}_{n_{K,b}},\mathcal{C}}(X_{i+2}) - \widehat{Q}_{K-2}^{(0)}(X_{i+2})\big)\mathbb{I}[\Delta_{i+1}=1] \right.\right.\right.\\
&\left.\left.\left.+ \big(Q^{(i+2)}_{\pi^{\ast}_{n_{K,b}},\mathcal{C}}(X_0,\cdots,X_{i+2}) - \widehat{Q}_{K-2}^{(i+2)}(X_0,\cdots X_{i+2})\big)\prod_{j=0}^{i+1}\mathbb{I}[\Delta_{j}=0]\mathbb{I}[\Delta_{-1}=1]\right)\right.\right.\nonumber\\
&\left.\left.+C_5(\varepsilon)(NT)^{-\delta}+\rho_{K-1}^{(i+1)}(X_0,\cdots,X_{i+1})\right.\right.\\
&\left.\left.-U_{K-1}^{(i+1)}(X_0,\cdots,X_{i+1})\right)\prod_{j=0}^{i}\mathbb{I}[\Delta_{j}=0]\mathbb{I}[\Delta_{-1}=1]\right\}\nonumber\\
&+C_5(\varepsilon)(NT)^{-\delta}+ \rho_K^{(i)}(x^{(0)},\cdots,x^{(i)}).
 \end{align*}
 \end{sizeddisplay}

After simplification, we obtain the following:
 {\footnotesize \begin{align*}
 I_{i,K}(x^{(0)},\cdots,x^{(i)})&\leq \gamma^2 (\calP^{\pi^{\ast}_{n_{K,b}}}_i\calP^{\pi^{\ast}_{n_{K,b}}}_{i+1})\left\{\left(I_{0,K-2}(X_{i+2})-U_{K-2}^{(0)}(X_{i+2})\right)\mathbb{I}[\Delta_{i+1}=1] \right.\\
    &\quad \left. + \left(I_{1,K-2}(X_{i+1},X_{i+2})-U_{K-2}^{(1)}(X_{i+1},X_{i+2})\right)\mathbb{I}[\Delta_{i+1}=0]\mathbb{I}[\Delta_{i}=1] \right.\\
    &\left. \quad + \left(I_{i+2,K-2}(X_{0},\cdots,X_{i+2})-U_{K-2}^{(i+2)}(X_{0},\cdots,X_{i+2})\right)\prod_{j=0}^{i+1}\mathbb{I}[\Delta_{j}=0]\mathbb{I}[\Delta_{-1}=1]\right\} \\
    &\quad + \gamma\calP^{\pi^{\ast}_{n_{K,b}}}_{i}\left\{\left(C_5(\varepsilon)(NT)^{-\delta}+\rho_{K-1}^{(0)}(X_{i+1})-U_{K-1}^{(0)}(X_{i+1})\right)\mathbb{I}[\Delta_{i}=1] \right.\\
    &\quad\left.+ \left(C_5(\varepsilon)(NT)^{-\delta}+\rho_{K-1}^{(i+1)}(X_{0},\cdots,X_{i+1})-U_{K-1}^{(i+1)}(X_{0},\cdots,X_{i+1}) \right)\prod_{j=0}^{i}\mathbb{I}[\Delta_{j}=0]\mathbb{I}[\Delta_{-1}=1] \right\} \\
 &\quad +C_5(\varepsilon)(NT)^{-\delta} + \rho_K^{(i)}(x^{(0)},\cdots,x^{(i)}).
 \end{align*}}

For the remaining part of the proof, we use $I_{i,K}$ to denote $I_{i,K}(x^{(0)},\cdots,x^{(i)})$. Then, by following the same procedure,  we reach the following after $K-1$ steps:

\begin{sizeddisplay}
{\footnotesize}\begin{align*}
     I_{i,K}
     &\leq\gamma^{K-1} \left(\prod_{k=0}^{K-2}{\calP_{i+k}^{\pi^{\ast}_{n_{K,b}}}}\right) \left\{\sum_{k=0}^{n_{K,b}}\left(Q^{(k)}_{\pi^{\ast}_{n_{K,b}},\mathcal{C}}(X_{i+K-1-k},\cdots,X_{i+K-1}) - \widehat{R}_{i+K-1}\right)\right.\\
     &\left.\prod_{j=i+K-1
    -k}^{i+K-2}\mathbb{I}[\Delta_{j}=0]\mathbb{I}[\Delta_{i+K-2-k}=1] \right\}\\
     &+\sum_{k=1}^{K-1}\gamma^k \left(\prod_{k'=0}^{k-1}\calP_{i+k'}^{\pi^{\ast}_{n_{K,b}}}\right)\left\{\left(C_5(\varepsilon)(NT)^{-\delta}+\rho_{K-k}^{(0)}(X_{i+k})-U_{K-k}^{(0)}(X_{i+k})\right)\mathbb{I}[\Delta_{i+k-1}=1] \right.\\
     &\left.+\left(C_5(\varepsilon)(NT)^{-\delta}+\rho_{K-k}^{(i+k)}(X_{0},\cdots,X_{i+k})-U_{K-k}^{(i+k)}(X_{0},\cdots,X_{i+k})\right)\prod_{j=0}^{i+k-1}\mathbb{I}[\Delta_{j}=0]\mathbb{I}[\Delta_{-1}=1] \mathbb{I}[k\leq n_{K,b}-i] \right.\\
     &\left.+ \sum_{v=1}^{\min(n_{K,b},k-1)}\left\{\left(C_5(\varepsilon)(NT)^{-\delta}+\rho_{K-k}^{(v)}(X_{i+k-v},\cdots,X_{i+k})-U_{K-k}^{(v)}(X_{i+k-v},\cdots,X_{i+k})\right)\prod_{j=i+k-v}^{i+k-1}\mathbb{I}[\Delta_{j}=0]\right.\right.\\
     &\left.\left.\mathbb{I}[\Delta_{i+k-v-1}=1]\right\}\right\}\\
     &+C_5(\varepsilon)(NT)^{-\delta}+ \rho_K^{(i)}(x^{(0)},\cdots,x^{(i)}).
\end{align*}\end{sizeddisplay}
The introduction of $\min(n_{K,b},k-1)$ as the upper limit in the last summation is implied by the fact that  $\pi^{\ast}_{n_{K,b}} \in \Pi_{n_{K,b}}$, which indicates  $\rho_{K-k}^{(v)}$ cannot exist for any $v>n_{K,b}$. 

On the event $\Omega \cap \Omega^r$, Equations~\eqref{eq: upper bound rho fqi} and \eqref{eq: upper bound rho2 fqi} imply that, for all $i \in \{0,\ldots,n_{K,b}\}$, we have:
\begin{sizeddisplay}{\footnotesize}
\begin{align}\label{eq: FQI error and UQ}
	& \rho_j^{(i)}(x^{(0)},\cdots,x^{(i)}) \leq 2 U_j^{(i)}(x^{(0)},\cdots,x^{(i)}) \qquad \forall j=1,\cdots,K-1\\\
 & \rho_K^{(i)}(x^{(0)},\cdots,x^{(i)}) \leq C^{(i)}(x^{(0)},\cdots,w^{(i)}) + U_K^{(i)}(x^{(0)},\cdots,x^{(i)}),\nonumber 
\end{align}\end{sizeddisplay}
which holds for all $(x^{(0)},\cdots, x^{(i)}) \in \calX^{\otimes (i+1)} $  and with probability at least $1-\epsilon(n_{K,b}+1)$. Additionally,  $|Q^{(i)}_{\pi^{\ast}_{n_{K,b}},\mathcal{C}}|$ is uniformly bounded by $\frac{R_{\max}}{1-\gamma}$ $\forall i=0,\cdots,n_{K,b}$ and $|\widehat{R}_t|$ is uniformly bounded by $R_{\max}$ $\forall t\geq0$. Together, they imply:
\begin{sizeddisplay}{\footnotesize}
    \begin{align}\label{eq:fqi  final upper bound Q}
    I_{i,K}&\leq \gamma^{K-1}\left(\frac{2-\gamma}{1-\gamma}R_{\max}\right)+\sum_{k=1}^{K-1}\gamma^k \left(\prod_{k'=0}^{k-1}\calP_{i+k'}^{\pi^{\ast}_{n_{K,b}}}\right)\left\{\left(C_5(\varepsilon)(NT)^{-\delta}+U_{K-k}^{(0)}(X_{i+k})\right)\mathbb{I}[\Delta_{i+k-1}=1]\right.\nonumber\\
     &\left.+\left(C_5(\varepsilon)(NT)^{-\delta}+U_{K-k}^{(i+k)}(X_{0},\cdots,X_{i+k})\right)\prod_{j=0}^{i+k-1}\mathbb{I}[\Delta_{j}=0]\mathbb{I}[\Delta_{-1}=1]\mathbb{I}[k\leq n_{K,b}-i] \right.\nonumber\\
     &\left.+ \sum_{v=1}^{\min(n_{K,b},k-1)}\left\{\left(C_5(\varepsilon)(NT)^{-\delta}+U_{K-k}^{(v)}(X_{i+k-v},\cdots,X_{i+k})\right)\prod_{j=i+k-v}^{i+k-1}\mathbb{I}[\Delta_{j}=0]\mathbb{I}[\Delta_{i+k-v-1}=1]\right\}\right\}\nonumber\\
     &+C_5(\varepsilon)(NT)^{-\delta}+U_{K}^{(i)}(x^{(0)},\cdots,x^{(i)})+C^{(i)}(x^{(0)},\cdots,w^{(i)})\nonumber\\
     &=\gamma^{K-1}\left(\frac{2-\gamma}{1-\gamma}R_{\max}\right)+ \textup{RD}(NT)\nonumber\\
     & + \textup{UQ}(x^{(0)},\cdots,x^{(i)}) +C^{(i)}(x^{(0)},\cdots,w^{(i)}),
\end{align}
\end{sizeddisplay}
where the characterizations of $\textup{RD}(NT)$ and $\textup{UQ}(x^{(0)},\cdots,x^{(i)})$ are given in Definitions \ref{def: fqi RD definition} and\hyperlink{uq_fqi_i}{~\ref{def:uq_definition} (i)}, respectively. This  concludes the proof of Lemma \ref{lm:upper bound on Qs}.

\subsection{Proof of Lemma \ref{lm:lower bound on Qs}}\label{subsec: proof of lower bound on Qs fqi}
Before the proof, we introduce several related notations and definitions.
For $i=0,\cdots,n_{K,b}$, let 
\begin{align}\overline{Q}^{(i)}_j(x^{(0)}, \ldots, x^{(i)}) = \begin{cases}
        \widehat{Q}^{(i)}_j(x^{(0)}, \ldots, x^{(i)}) - U_j^{(i)}(x^{(0)}, \ldots, x^{(i)}), & j = 1, \ldots, K-1\\
        \widehat{Q}^{(i)}_K(x^{(0)}, \ldots, x^{(i)}) - C^{(i)}(x^{(0)}, \ldots, ,x^{(i-1)},w^{(i)}), & j = K
    \end{cases}
\end{align}

Observe that the greedy policy derived from the pairs $\widehat{Q}^{(i)}_K(x^{(0)},\cdots,x^{(i)})$ and $\overline{Q}^{(i)}_K(x^{(0)},\cdots,x^{(i)})$ is the same. 

Next, $\forall j=1,\cdots,K$ and $\forall i=0,\cdots,n_{K,b}$, define:
\begin{align}\label{eq: error FQI lower}
    \rho_j^{(i)}(x^{(0)},\cdots,x^{(i)}) = \calT^{\hat{\pi}_{j-1}}_{i,\mathcal{C}} \widehat{Q}_{j-1}(x^{(0)},\cdots,x^{(i)}) - \overline{Q}^{(i)}_{j}(x^{(0)},\cdots,x^{(i)}),
\end{align} 
where $\calT^{\hat{\pi}_{j-1}}_{i,\mathcal{C}}$ is defined under Definition \ref{def: bellman operator}.

Additionally, recall that we denote uncertainty quantifier for PC-FQI by $U_{k}^{(i)}$. Under Assumption~\ref{ass: UQ ass} which asserts the existence of an event $\Omega^{(i)}$ with $\Pr(\Omega^{(i)})\ge 1-\epsilon$ where $\tilde{U}_{k}^{(i)}$ satisfies Definition~\ref{def:UE} and is therefore a valid uncertainty quantifier. Because the proof requires ${U}_{k}^{(i)}$ to be a valid Uncertainty Quantifier $\forall i=0,\cdots, n_{K,b}$, we work on the intersection event
\[
\Omega \;:=\;\bigcap_{i=0}^{n_{K,b}}\Omega^{(i)}.
\]
which holds with probability at least $1-(n_{K,b}+1)\epsilon$. And on $\Omega$, we have $\forall i=0,\cdots, n_{K,b}$ and $\forall j=1,\cdots,K-1$ 
\begin{align}\label{eq: lower bound rho fqi}
    {\rho}_j^{ (i)}(x^{(0)},\cdots,x^{(i)}) 
    &= \calT^{\hat{\pi}_{j-1}}_{i,\mathcal{C}} \widehat{Q}_{j-1}(x^{(0)},\cdots,x^{(i)}) - \overline{Q}^{(i)}_{j}(x^{(0)},\cdots,x^{(i)})\nonumber\\
    &= \calT^{\hat{\pi}_{j-1}}_{i,\mathcal{C}} \widehat{Q}_{j-1}(x^{(0)},\cdots,x^{(i)}) - (\widehat{Q}^{(i)}_{j}(x^{(0)},\cdots,x^{(i)}) - {U}^{(i)}_j(x^{(0)},\cdots,x^{(i)}))\nonumber\\
    &\geq  0
\end{align}
and for $j=K$ we have
\begin{align}\label{eq: lower bound rho2 fqi}
    {\rho}_K^{ (i)}(x^{(0)},\cdots,x^{(i)}) 
    &= \calT^{\hat{\pi}_{K-1}}_{i,\mathcal{C}} \widehat{Q}_{K-1}(x^{(0)},\cdots,x^{(i)}) - \overline{Q}^{(i)}_{K}(x^{(0)},\cdots,x^{(i)})\nonumber\\
    &= \calT^{\hat{\pi}_{K-1}}_{i,\mathcal{C}} \widehat{Q}_{K-1}(x^{(0)},\cdots,x^{(i)}) - (\widehat{Q}^{(i)}_{K}(x^{(0)},\cdots,x^{(i)}) - {C}^{(i)}(x^{(0)},\cdots,x^{(i-1)},w^{(i)})\nonumber\\
    &\geq  C^{(i)}(x^{(0)},\cdots,w^{(i)})-U_K^{(i)}(x^{(0)},\cdots,x^{(i)})
\end{align}
These identities will be used in the following proof.

In addition, under Assumption~\ref{ass: surrogate outcome}, we work on the event $\Omega^r$ on which the surrogate reward relation holds. In particular, we have $\Omega^r$,
\begin{align}\label{eq: reward dis fqi lower}
&\EE\left[\left(\widehat{R}_{i}\right)\prod_{j=0}^{i}\mathbb{I}[\Delta_{j}=0]\mathbb{I}[\Delta_{-1}=1]\given X_{0},\cdots,X_{i},\mathcal{C}(K,n_{K,b})\right]\nonumber\\
    &\leq\EE\left[\left(\widetilde{R}_{i} + C_5(\varepsilon)(NT)^{-\delta}\right)\prod_{j=0}^{i}\mathbb{I}[\Delta_{j}=0]\mathbb{I}[\Delta_{-1}=1]\given X_{0},\cdots,X_{i},\mathcal{C}(K,n_{K,b})\right],
\end{align}
which we use to account for the estimation error in the surrogate reward.

Accordingly, Lemma \ref{lm:upper bound on Qs pessimistic} is proved on the joint event $\Omega \cap \Omega^r$, i.e., under the simultaneous validity of the uncertainty quantification bounds across all depths and the surrogate reward approximation. By union bound, we have $\prob(\Omega \cap \Omega^r) \geq 1-(n_{K,b}+1)\epsilon-\varepsilon$.

Now, we are ready to show the proof.

\textit{Proof:}

We focus on deriving a lower bound on the difference between $Q^{(i)}_{\pi^{\ast}_{n_{K,b}},\mathcal{C}}(x^{(0)},\cdots,x^{(i)})$ and $\overline{Q}_{K}^{(i)}(x^{(0)},\cdots,x^{(i)})$ $\forall i=0,\cdots,n_{K,b}$:
\begin{sizeddisplay}
{\footnotesize}\begin{align}\label{eq: FQI lower first eq}
    I_{i,K}
(x^{(0)},\cdots,x^{(i)})&
\triangleq Q^{(i)}_{\pi^{\ast}_{n_{K,b}}    ,\mathcal{C}}(x^{(0)},\cdots,x^{(i)})-\overline{Q}_{K}^{(i)}(x^{(0)},\cdots,x^{(i)})\nonumber\\&=Q_{\pi^{\ast}_{n_{K,b}},\mathcal{C}}^{(i)}(x^{(0)},\cdots,x^{(i)}) - \calT_{i,\mathcal{C}}^{\hat{\pi}_{K-1}}\widehat{Q}_{K-1}(x^{(0)},\cdots,x^{(i)}) + \rho_K^{(i)}(x^{(0)},\cdots,x^{(i)})\nonumber\\
    &= Q_{\pi^{\ast}_{n_{K,b}},\mathcal{C}}^{(i)}(x^{(0)},\cdots,x^{(i)}) - \calT^{\hat{\pi}_{K-1}}_{i,\mathcal{C}}Q_{\pi^{\ast}_{n_{K,b}},\mathcal{C}}(x^{(0)},\cdots,x^{(i)})\nonumber\\
    &+ \calT^{\hat{\pi}_{K-1}}_{i,\mathcal{C}}Q_{\pi^{\ast}_{n_{K,b}},\mathcal{C}}(x^{(0)},\cdots,x^{(i)})-\calT^{\hat{\pi}_{K-1}}_{i,\mathcal{C}}\widehat{Q}_{K-1}(x^{(0)},\cdots,x^{(i)})\nonumber \\
    &+ \rho_K^{(i)}(x^{(0)},\cdots,x^{(i)})\nonumber\\
    &=\underbrace{\calT^{\pi^{\ast}_{n_{K,b}}}_{i,\mathcal{C}}Q_{\pi^{\ast}_{n_{K,b}},\mathcal{C}}(x^{(0)},\cdots,x^{(i)}) - \calT^{\hat{\pi}_{K-1}}_{i,\mathcal{C}}Q_{\pi^{\ast}_{n_{K,b}},\mathcal{C}}(x^{(0)},\cdots,x^{(i)})}_{(1)}\nonumber\\
    &+ \calT^{\hat{\pi}_{K-1}}_{i,\mathcal{C}}Q_{\pi^{\ast}_{n_{K,b}},\mathcal{C}}^{(i)}(x^{(0)},\cdots,x^{(i)})
    - \calT^{\hat{\pi}_{K-1}}_{i,\mathcal{C}}\widehat{Q}_{K-1}(x^{(0)},\cdots,x^{(i)})\nonumber\\
    &+ \rho_K^{(i)}(x^{(0)},\cdots,x^{(i)}).
\end{align}\end{sizeddisplay}Consider the term (1) above $\forall i=0,\cdots,n_{K,b}$:

\begin{sizeddisplay}
{\footnotesize}
\begin{align*}
    (1)&=\calT^{\pi^{\ast}_{n_{K,b}}}_{i,\mathcal{C}}Q_{\pi^{\ast}_{n_{K,b}},\mathcal{C}}^{(i)}(x^{(0)},\cdots,x^{(i)}) - \calT^{\hat{\pi}_{K-1}}_{i,\mathcal{C}}Q_{\pi^{\ast}_{n_{K,b}},\mathcal{C}}^{(i)}(x^{(0)},\cdots,x^{(i)})\nonumber\\
    &=\EE\left[\gamma\left(Q_{\pi^{\ast}_{n_{K,b}},\mathcal{C}}^{(0)}(W_{i+1},\pi^{(0)}_{n_{K,b},*}(W_{i+1}))-Q_{\pi^{\ast}_{n_{K,b}},\mathcal{C}}^{(0)}(W_{i+1},\widehat{\pi}^{(0)}_{K-1}(W_{i+1}))\right)\mathbb{I}[\Delta_{i}=1] \right.\nonumber\\
    &\left.  + \gamma\left(Q_{\pi^{\ast}_{n_{K,b}},\mathcal{C}}^{(i+1)}(X_0,\cdots,W_{i+1},\pi_{n_{K,b},*}^{(i+1)}(W_{i+1},\cdots,X_0))-Q_{\pi^{\ast}_{n_{K,b}},\mathcal{C}}^{(i+1)}(X_0,\cdots,W_{i+1},\widehat{\pi}^{(i+1)}_{K-1}(W_{i+1},\cdots,X_0))\right)\mathbb{I}[\Delta_{i}=0]\right.\nonumber\\
    &\left.\quad\given (X_0,\cdots,X_{i})=(x^{(0)},\cdots,x^{(i)}),\mathcal{C}(K,n_{K,b})\right]\nonumber,
\end{align*}\end{sizeddisplay}
where the first equality stems from the definition of $\calT^{\pi}_{i,\mathcal{C}}$ given in Definition \ref{def: bellman operator}. Recall the characterization of policy $\pi^{\ast}_{n_{K,b}}$ given under Lemma \ref{lemma: nkb and action relation}. For $i=0,\cdots,n_{K,b}-1$, $\pi^{\ast}_{n_{K,b}}$ takes the point-wise maximum of $Q^{(i)}_{\pi^{\ast}_{n_{K,b}},\mathcal{C}}$ over all possible actions given $(W_{i},\cdots,X_0)$. This implies $\forall i=0,\cdots,n_{K,b}-1$:
\begin{sizeddisplay}
    {\footnotesize}
    \begin{align}\label{eq: action discrepancy lower bound fqi}
        Q_{\pi^{\ast}_{n_{K,b}},\mathcal{C}}^{(i)}(X_0,\cdots,W_{i},\pi_{n_{K,b},*}^{(i)}(W_{i},\cdots,X_0))-Q_{\pi^{\ast}_{n_{K,b}},\mathcal{C}}^{(i)}(X_0,\cdots,W_{i},\widehat{\pi}^{(i)}_{K-1}(W_{i},\cdots,X_0))\geq 0 
    \end{align}
\end{sizeddisplay}
When $i=n_{K,b}$, $\pi^{\ast}_{n_{K,b}}$ takes the point-wise maximum of $Q^{(i)}_{\pi^{\ast}_{n_{K,b}},\mathcal{C}}$ over a restricted set of actions denoted by $\textup{UP}^{n_{K,b}}(W_{n_{K,b}},X_{n_{K,b}-1},\cdots,X_{0})$. Therefore, it cannot be ensured that Equation \eqref{eq: action discrepancy lower bound fqi} holds when $i=n_{K,b}$. To tackle this issue, we make use of Definition\hyperlink{action_discrepancy_i}{~\ref{def:action_discrepancy} (i)}:
\begin{sizeddisplay}
    {\footnotesize}
    \begin{align}
        &Q_{\pi^{\ast}_{n_{K,b}},\mathcal{C}}^{(n_{K,b})}(X_0,\cdots,W_{n_{K,b}},\pi_{n_{K,b},*}^{(n_{K,b})}(W_{n_{K,b}},\cdots,X_0))-Q_{\pi^{\ast}_{n_{K,b}},\mathcal{C}}^{(n_{K,b})}(X_0,\cdots,W_{n_{K,b}},\widehat{\pi}^{(n_{K,b})}_{K-1}(W_{n_{K,b}},\cdots,X_0))\nonumber\\
        &\geq  Q_{\pi^{\ast}_{n_{K,b}},\mathcal{C}}^{(n_{K,b})}(X_0,\cdots,W_{n_{K,b}},\pi_{n_{K,b},*}^{(n_{K,b})}(W_{n_{K,b}},\cdots,X_0))-\max_{a \in \calA}Q_{\pi^{\ast}_{n_{K,b}},\mathcal{C}}^{(n_{K,b})}(X_0,\cdots,W_{n_{K,b}},a)\nonumber\\
        &=-\textup{AD}^{\pi^{*}_{n_{K,b}}}(X_0,\cdots,W_{n_{K,b}})
    \end{align}
\end{sizeddisplay}
Consequently, it can be concluded that:
\begin{sizeddisplay}
    {\footnotesize}
    \begin{align}
&\calT^{\pi^{\ast}_{n_{K,b}}}_{i,\mathcal{C}}Q_{\pi^{\ast}_{n_{K,b}},\mathcal{C}}^{(i)}(x^{(0)},\cdots,x^{(i)}) - \calT^{\hat{\pi}_{K-1}}_{i,\mathcal{C}}Q_{\pi^{\ast}_{n_{K,b}},\mathcal{C}}^{(i)}(x^{(0)},\cdots,x^{(i)})\\
        &\geq -\EE\left[\gamma\textup{AD}^{\pi^{*}_{n_{K,b}}}(X_0,\cdots,W_{i+1})\mathbb{I}[\Delta_{i}=0]\given (X_0,\cdots,X_{i})=(x^{(0)},\cdots,x^{(i)}),\mathcal{C}(K,n_{K,b})]\right]\mathbb{I}[i=n_{K,b}-1],\nonumber
    \end{align}
\end{sizeddisplay}
which further implies the following:
{\footnotesize\begin{align}\label{eq:fqi lower bound on Q}
    I_{i,K}(x^{(0)},\cdots,x^{(i)}) 
    &\geq \calT^{\hat{\pi}_{K-1}}_{i,\mathcal{C}}Q_{\pi^{\ast}_{n_{K,b}},\mathcal{C}}^{(i)}(x^{(0)},\cdots,x^{(i)})
    - \calT^{\hat{\pi}_{K-1}}_{i,\mathcal{C}}\widehat{Q}_{K-1}(x^{(0)},\cdots,x^{(i)}) \nonumber\\
    & -\EE\left[\gamma\textup{AD}^{\pi^{*}_{n_{K,b}}}(X_0,\cdots,W_{i+1})\mathbb{I}[\Delta_{i}=0]\given (X_0,\cdots,X_{i})=(x^{(0)},\cdots,x^{(i)}),\mathcal{C}(K,n_{K,b})]\right]\mathbb{I}[i=n_{K,b}-1]\nonumber\\
    &+ \rho_K^{(i)}(x^{(0)},\cdots,x^{(i)}) 
\end{align}}

By extending the right hand side of Equation \eqref{eq:fqi lower bound on Q}, we obtain:
{\footnotesize\begin{align}\label{eq:fqi lower bound on Q extended}
    I_{i,K}(x^{(0)},\cdots,x^{(i)}) 
    &\geq\gamma \calP_{i}^{\widehat{\pi}_{K-1}}\left\{\left(\underbrace{Q^{(0)}_{\pi^{\ast}_{n_{K,b}},\mathcal{C}}(X_{i+1}) - \widehat{Q}_{K-1}^{(0)}(X_{i+1})+U_{K-1}^{(0)}(X_{i+1})}_{I_{0,K-1}(X_{i+1})}-U_{K-1}^{(0)}(X_{i+1})\right)\mathbb{I}[\Delta_{i}=1]\right.\nonumber\\
    &\left. +\left(\underbrace{Q^{(i+1)}_{\pi^{\ast}_{n_{K,b}},\mathcal{C}}(X_0,\cdots,X_{i+1}) - \widehat{Q}_{K-1}^{(i+1)}(X_0,\cdots,X_{i+1})+U_{K-1}^{(i+1)}(X_0,\cdots,X_{i+1})}_{I_{i+1,K-1}(X_0,\cdots,X_{i+1})}\right.\right.\nonumber\\
    &\left.\left.-U_{K-1}^{(i+1)}(X_0,\cdots,X_{i+1})-\textup{AD}^{\pi^{*}_{n_{K,b}}}(X_0,\cdots,W_{i+1})\mathbb{I}[i=n_{K,b}-1]\right)\mathbb{I}[\Delta_{i}=0]\right\}\nonumber\\
    &+\EE\left[\left(R_{i}-\widehat{R}_{i}\right)\prod_{j=0}^{i}\mathbb{I}[\Delta_{j}=0]\mathbb{I}[\Delta_{-1}=1]\given (X_{0},\cdots,X_{i}) = (x^{(0)},\cdots,x^{(i)}),\mathcal{C}(K,n)\right]\nonumber\\
    &+\rho_K^{(i)}(x^{(0)},\cdots,x^{(i)}),
\end{align}}where characterization of $\calP_{i}^{\hat{\pi}_{K-1}}$ given in Definition \ref{def: P as conditional exp}. Additionally, we add and subtract $U_{K-1}^{(i)}$ to obtain the recursive relation.

Observe that By Lemma \ref{lm: identification for censored demand} and on event $\Omega_P \cap \Omega^r$, the following holds 
\begin{sizeddisplay}
{\footnotesize}
\begin{align*}
    &\EE\left[\left(R_{i}-\widehat{R}_{i}\right)\prod_{j=0}^{i}\mathbb{I}[\Delta_{j}=0]\mathbb{I}[\Delta_{-1}=1]\given (X_{0},\cdots,X_{i}) = (x^{(0)},\cdots,x^{(i)}),\mathcal{C}(K,n)\right]\\
    &\geq \EE\left[\left(R_{i}-\widetilde{R}_{i}-C_5(\varepsilon)(NT)^{-\delta}\right)\prod_{j=0}^{i}\mathbb{I}[\Delta_{j}=0]\mathbb{I}[\Delta_{-1}=1]\given (X_{0},\cdots,X_{i}) = (x^{(0)},\cdots,x^{(i)}),\mathcal{C}(K,n)\right]\\
    &=\EE\left[-C_5(\varepsilon)(NT)^{-\delta}\prod_{j=0}^{i}\mathbb{I}[\Delta_{j}=0]\mathbb{I}[\Delta_{-1}=1]\given (X_{0},\cdots,X_{i}) = (x^{(0)},\cdots,x^{(i)}),\mathcal{C}(K,n)\right]\\ 
    &\geq-C_5(\varepsilon)(NT)^{-\delta},
\end{align*}\end{sizeddisplay}
where the first inequality is implied by Equation~\eqref{eq: reward dis fqi lower}.

Therefore, Equation \eqref{eq:fqi lower bound on Q extended} becomes:
\begin{sizeddisplay}
{\footnotesize}\begin{align}\label{eq:fqi lower bound on Q2}
    I_{i,K}(x^{(0)},\cdots,x^{(i)}) &\geq \gamma \calP_{i}^{\hat{\pi}_{K-1}}\left\{\left(I_{0,K-1}(X_{i+1})-U_{K-1}^{(0)}(X_{i+1})\right)\mathbb{I}[\Delta_{i}=1]\right.\nonumber\\
    &\left. +\left(I_{i+1,K-1}(X_0,\cdots,X_{i+1})-U_{K-1}^{(i+1)}(X_0,\cdots,X_{i+1})\right.\right.\nonumber\\
    &\left.\left.-\textup{AD}^{\pi^{*}_{n_{K,b}}}(X_0,\cdots,W_{i+1})\mathbb{I}[i=n_{K,b}-1]\right)\mathbb{I}[\Delta_{i}=0]\right\}\nonumber\\
    &+\rho_K^{(i)}(x^{(0)},\cdots,x^{(i)}) -C_5(\varepsilon)(NT)^{-\delta}.
\end{align}
\end{sizeddisplay}
Next, we apply the lower bounds on $I_{0,K-1}$ and $ I_{i+1,K-1}$
in an iterative manner:{\footnotesize\begin{align*}
    I_{i,K}(x^{(0)},\cdots,x^{(i)})
    &\geq \gamma \calP_{i}^{\hat{\pi}_{K-1}}\left\{\left(\gamma \calP_{0}^{\hat{\pi}_{K-2}}\left(\big(Q^{(0)}_{\pi^{\ast}_{n_{K,b}},\mathcal{C}}(X_{i+2}) - \widehat{Q}_{K-2}^{(0)}(X_{i+2})\big)\mathbb{I}[\Delta_{i+1}=1]\right.\right.\right.\\
    &\left.\left.\left.+\big(Q^{(1)}_{\pi^{\ast}_{n_{K,b}},\mathcal{C}}(X_{i+1},X_{i+2}) - \widehat{Q}_{K-2}^{(1)}(X_{i+1},X_{i+2})\big)\mathbb{I}[\Delta_{i+1}=0]\right)\right.\right.\\
    &\left.\left.+\rho_{K-1}^{(0)}(X_{i+1}) -C_5(\varepsilon)(NT)^{-\delta} -U_{K-1}^{(0)}(X_{i+1}) \right)\mathbb{I}[\Delta_{i}=1]\right.\\
    &\left. +\left(\gamma \calP_{i+1}^{\hat{\pi}_{K-2}}\left(\big(Q^{(0)}_{\pi^{\ast}_{n_{K,b}},\mathcal{C}}(X_{i+2}) - \widehat{Q}_{K-2}^{(0)}(X_{i+2})\big)\mathbb{I}[\Delta_{i+1}=1] \right.\right.\right.\\
    &\left.\left.+\left.\left(Q^{(i+2)}_{\pi^{\ast}_{n_{K,b}},\mathcal{C}}(X_0,\cdots,X_{i+2})- \widehat{Q}_{K-2}^{(i+2)}(X_0,\cdots,X_{i+2})\right.\right.\right.\right.\\
    &\left.\left.\left.-\textup{AD}^{\pi^{*}_{n_{K,b}}}(X_0,\cdots,W_{i+2})\mathbb{I}[i=n_{K,b}-2]\right)\mathbb{I}[\Delta_{i+1}=0]\right)\right.\\
    &\left.\left.+ \rho_{K-1}^{(i+1)}(X_{0},\cdots,X_{i+1})-C_5(\varepsilon)(NT)^{-\delta}-U_{K-1}^{(i+1)}(X_{0},\cdots,X_{i+1})\right.\right.\\
    &\left.\left.-\textup{AD}^{\pi^{*}_{n_{K,b}}}(X_0,\cdots,W_{i+1})\mathbb{I}[i=n_{K,b}-1]\right)\mathbb{I}[\Delta_{i}=0] \right\}+\rho_K^{(i)}(x^{(0)},\cdots,x^{(i)})  -C_5(\varepsilon)(NT)^{-\delta},\end{align*}}
where the right hand side, after simplification, is equivalent to:
\begin{sizeddisplay}
{\footnotesize}\begin{align*}  I_{i,K}(x^{(0)},\cdots,x^{(i)})
    &\geq \gamma^2\calP_{i}^{\hat{\pi}_{K-1}}\calP_{i+1}^{\hat{\pi}_{K-2}}\left(\left(I_{0,K-2}(X_{i+2})-U_{K-2}^{(0)}(X_{i+2})\right)\mathbb{I}[\Delta_{i+1}=1]\right.\\
    &\left.+\left(I_{1,K-2}(X_{i+1},X_{i+2})-U_{K-2}^{(1)}(X_{i+1},X_{i+2})\right)\mathbb{I}[\Delta_{i+1}=0]\mathbb{I}[\Delta_{i}=1]\right.\\
    &\left.+\left({I_{i+2,K-2}(X_0,\cdots,X_{i+2})}-U_{K-2}^{(i+2)}(X_0,\cdots,X_{i+2})\right.\right.\\
    &\left.\left.-\textup{AD}^{\pi^{*}_{n_{K,b}}}(X_0,\cdots,W_{i+2})\mathbb{I}[i=n_{K,b}-2]\right)\mathbb{I}[\Delta_{i+1}=0]\mathbb{I}[\Delta_{i}=0]\right)\\
    &+\gamma\calP_{i}^{\hat{\pi}_{K-1}}\left(\left(\rho_{K-1}^{(0)}(X_{i+1})-C_5(\varepsilon)(NT)^{-\delta}-U_{K-1}^{(0)}(X_{i+1})\right) \mathbb{I}[\Delta_{i}=1] \right.\\
    &\left.+ \left(\rho_{K-1}^{(i+1)}(X_{0},\cdots,X_{i+1})-C_5(\varepsilon)(NT)^{-\delta}-U_{K-1}^{(i+1)}(X_{0},\cdots,X_{i+1})\right.\right.\\
    &\left.\left.-\textup{AD}^{\pi^{*}_{n_{K,b}}}(X_0,\cdots,W_{i+1})\mathbb{I}[i=n_{K,b}-1]\right)\mathbb{I}[\Delta_{i}=0]\right)\\
    &+\rho_K^{(i)}(x^{(0)},\cdots,x^{(i)}) -C_5(\varepsilon)(NT)^{-\delta}.
\end{align*}\end{sizeddisplay}
For the remaining part of the proof, we use $I_{i,K}$ to denote $I_{i,K}(x^{(0)},\cdots,x^{(i)})$. Then, by pursuing the same procedure $(K-1)$ times, we can obtain the following:
\begin{sizeddisplay}
{\footnotesize}\begin{align}\label{eq:fqi final lower bound on Q}
    I_{i,K}&\geq \prod_{j=1}^{K-1}\left(\gamma\calP_{i+j-1}^{\hat{\pi}_{K-j}}\right)\left\{\sum_{k=0}^{n_{K,b}}\left(Q^{(k)}_{\pi^{\ast}_{n_{K,b}},\mathcal{C}}(X_{i+K-1-k},\cdots,X_{i+K-1})- \widehat{R}_{i+K-1}\right) \prod_{l=i+K-1-k}^{i+K-2}\mathbb{I}[\Delta_{l}=0]\mathbb{I}[\Delta_{i+K-2-k}=1]\right\}\nonumber\\
    &+\sum_{k=1}^{K-1}\prod_{j=1}^{k}\left(\gamma\calP_{i+j-1}^{\hat{\pi}_{K-j}}\right)\left\{\left(\rho_{K-k}^{(0)}(X_{i+k})-C_5(\varepsilon)(NT)^{-\delta}-U_{K-k}^{(0)}(X_{i+k})\right)\mathbb{I}[\Delta_{i+k-1}=1]\right.\nonumber\\
    &\left.+\left(\rho_{K-k}^{(i+k)}(X_{0},\cdots,X_{i+k})-C_5(\varepsilon)(NT)^{-\delta}-U_{K-k}^{(i+k)}(X_{0},\cdots,X_{i+k})\right.\right.\nonumber\\
    &\left.\left.-\textup{AD}^{\pi^{*}_{n_{K,b}}}(X_0,\cdots,W_{i+k})\mathbb{I}[i=n_{K,b}-k]\right)\prod_{l=0}^{i+k-1}\mathbb{I}[\Delta_{l}=0]\mathbb{I}[\Delta_{-1}=1]\mathbb{I}[k\leq n_{K,b}-i]\right.\nonumber\\
    &\left. +\sum_{v=1}^{\min(k-1,n_{K,b})}\left(\rho_{K-k}^{(v)}(X_{i+k-v},\cdots,X_{i+k})-C_5(\varepsilon)(NT)^{-\delta}\right.\right.\nonumber\\
    &\left.\left.-U_{K-k}^{(v)}(X_{i+k-v},\cdots,X_{i+k})-\textup{AD}^{\pi^{*}_{n_{K,b}}}(X_{i+k-v},\cdots,W_{i+k})\mathbb{I}[v=n_{K,b}]\right) \prod_{l=i+k-v}^{i+k-1}\mathbb{I}[\Delta_{l}=0]\mathbb{I}[\Delta_{i+k-v-1}=1]\right\}\nonumber\\
    &+\rho_K^{(i)}(x^{(0)},\cdots,x^{(i)})-C_5(\varepsilon)(NT)^{-\delta}. \nonumber
\end{align}
\end{sizeddisplay}
On the event $\Omega_P \cap \Omega^r$, Equations~\eqref{eq: lower bound rho fqi} and \eqref{eq: lower bound rho2 fqi} imply that, for all $i \in \{0,\ldots,n_{K,b}\}$, we have
\begin{sizeddisplay}
{\footnotesize}
\begin{align}
	&0 \leq \rho_j^{(i)}(x^{(0)},\cdots,x^{(i)})\qquad \forall j=1,\cdots,K-1 \nonumber\\
 &C^{(i)}(x^{(0)},\cdots,w^{(i)})-U_K^{(i)}(x^{(0)},\cdots,x^{(i)}) \leq \rho_K^{(i)}(x^{(0)},\cdots,x^{(i)})  \nonumber 
\end{align}
\end{sizeddisplay}
which holds for all $(x^{(0)},\cdots, x^{(i)}) \in \calX^{\otimes (i+1)}$. Additionally,  $|Q^{(i)}_{\pi^{*}_{n_{K,b}},\mathcal{C}}|$ is uniformly bounded by $\frac{R_{\max}}{1-\gamma}$ $\forall i=0,\cdots,n$ and $|\widehat{R}_t|\leq R_{\max}$ $\forall t\geq0$. Together, they imply:
\begin{sizeddisplay}
{\footnotesize}
\begin{align}
 I_{i,K}&\geq -\gamma^{K-1}\left(\frac{2-\gamma}{1-\gamma}R_{\max}\right)\nonumber\\
    &+\sum_{k=1}^{K-1}\prod_{j=1}^{k}\left(\gamma\calP_{i+j-1}^{\hat{\pi}_{K-j}}\right)\left\{-C_5(\varepsilon)(NT)^{-\delta} -U_{K-k}^{(0)}(X_{i+k})\mathbb{I}[\Delta_{i+k-1}=1]\right.\nonumber\\
    &\left.\left(-C_5(\varepsilon)(NT)^{-\delta} -U_{K-k}^{(i+k)}(X_{0},\cdots,X_{i+k})\right.\right.\nonumber\\
    &\left.\left.-\textup{AD}^{\pi^{*}_{n_{K,b}}}(X_0,\cdots,W_{i+k})\mathbb{I}[i=n_{K,b}-k]\right)\prod_{l=0}^{i+k-1}\mathbb{I}[\Delta_{l}=0]\mathbb{I}[\Delta_{-1}=1]\mathbb{I}[k\leq n_{K,b}-i]\right.\nonumber\\
    &\left. +\sum_{v=1}^{\min(k-1,n_{K,b})}\left(-C_5(\varepsilon)(NT)^{-\delta} -U_{K-k}^{(v)}(X_{i+k-v},\cdots,X_{i+k})\right.\right.\nonumber\\
    &\left.\left.-\textup{AD}^{\pi^{*}_{n_{K,b}}}(X_{i+k-v},\cdots,W_{i+k})\mathbb{I}[v=n_{K,b}]\right)\prod_{l=i+k-v}^{i+k-1}\mathbb{I}[\Delta_{l}=0]\mathbb{I}[\Delta_{i+k-v-1}=1]\right\}\nonumber\\
    &+C^{(i)}(x^{(0)},\cdots,w^{(i)})-U_K^{(i)}(x^{(0)},\cdots,x^{(i)}) -C_5(\varepsilon)(NT)^{-\delta}. 
\end{align}
\end{sizeddisplay}

By using Definitions\hyperlink{total_action_discrepancy_ii}{~\ref{def:action_discrepancy} (ii)}  and \ref{def: fqi RD definition}, Equation \eqref{eq:fqi final lower bound on Q} can be expressed as follows:
\begin{sizeddisplay}
{\footnotesize}\begin{align} 
 I_{i,K}&\geq -\gamma^{K-1}\left(\frac{2-\gamma}{1-\gamma}R_{\max}\right)\nonumber\\
    &+\sum_{k=1}^{K-1}\prod_{j=1}^{k}\left(\gamma\calP_{i+j-1}^{\hat{\pi}_{K-j}}\right)\left\{ -U_{K-k}^{(0)}(X_{i+k})\mathbb{I}[\Delta_{i+k-1}=1]\right.\nonumber\\
    &\left.\left( -U_{K-k}^{(i+k)}(X_{0},\cdots,X_{i+k})\right)\prod_{l=0}^{i+k-1}\mathbb{I}[\Delta_{l}=0]\mathbb{I}[\Delta_{-1}=1]\mathbb{I}[k\leq n_{K,b}-i]\right.\nonumber\\
    &\left. +\sum_{v=1}^{\min(k-1,n_{K,b})}\left( -U_{K-k}^{(v)}(X_{i+k-v},\cdots,X_{i+k}) \right)\prod_{l=i+k-v}^{i+k-1}\mathbb{I}[\Delta_{l}=0]\mathbb{I}[\Delta_{i+k-v-1}=1]\right\}\nonumber\\
    &+C^{(i)}(x^{0},\cdots,w^{(i)})-U_K^{(i)}(x^{(0)},\cdots,x^{(i)})  -RD(NT) \nonumber\\
    & =-\gamma^{K-1}\left(\frac{2-\gamma}{1-\gamma}R_{\max}\right)-\textup{RD}(NT)-\textup{TAD}^{\hat \pi_K}(x^{0},\cdots,x^{(i)}),
\end{align}\end{sizeddisplay}which concludes the proof of Lemma \ref{lm:lower bound on Qs}.

\subsection{Proof of Lemma \ref{lm:upper bound on Qs pessimistic}}\label{subsec: proof of upper bound on Qs pessimistic}
Before the proof, we introduce several related notations and definitions.

For all $ j=1,\cdots,K$ and $ i=0,\cdots,n_{K,b}$, define:
\begin{align}\label{eq: error P-FQI}
    \Tilde{\rho}_j^{ (i)}(x^{(0)},\cdots,x^{(i)}) = \calT^{\widetilde{\pi}_{j-1}}_{i,\mathcal{C}} \widetilde{Q}_{j-1}(x^{(0)},\cdots,x^{(i)}) - \widetilde{Q}^{(i)}_{j}(x^{(0)},\cdots,x^{(i)}),
\end{align} 
where $\calT^{\tilde{\pi}_{j-1}}_{i,\mathcal{C}}$ is defined under Definition \ref{def: bellman operator}.

Furthermore, note that the event $\mathcal{C}(K,n_{K,b})$ occurs almost surely under any policy $\pi \in \Pi_{n_{K,b}}$. Consequently, conditioning on $\mathcal{C}(K,n_{K,b})$ does not alter any conditional distributions induced by such policies. In particular, for any $i \le n_{K,b}$ and any history block $(W_t,A_t,\ldots,W_{t+i},A_{t+i})$, we have
\begin{align}
Q^{(i)}_{\pi,\mathcal{C}}(W_t,A_t,\ldots,W_{t+i},A_{t+i})
&:= \mathbb{E}^{\pi}\!\left[\sum_{t'=i}^{\infty} \gamma^{t'-i} R_{t'} \;\middle|\; (W_t,A_t,\ldots,W_{t+i},A_{t+i}),\,\mathcal{C}(K,n_{K,b})\right] \nonumber\\
&= \mathbb{E}^{\pi}\!\left[\sum_{t'=i}^{\infty} \gamma^{t'-i} R_{t'} \;\middle|\; (W_t,A_t,\ldots,W_{t+i},A_{t+i})\right] \nonumber\\
&=: Q^{(i)}_{\pi}(W_t,A_t,\ldots,W_{t+i},A_{t+i}).
\end{align}

Therefore, throughout the analysis we may interchangeably use $Q^{(i)}_{\pi,\mathcal{C}}$ and $Q^{(i)}_{\pi}$ for all $\pi \in \Pi_{n_{K,b}}$ and $i \le n_{K,b}$, without ambiguity. For instance, since $\pi_{n_{K,b}} \in \Pi_{n_{K,b}}$, this result will imply that $$Q^{(i)}_{\pi^{\ast}_{n_{K,b}}}(x^{(0)},\cdots,x^{(i)})=Q^{(i)}_{\pi^{\ast}_{n_{K,b},\mathcal{C}}}(x^{(0)},\cdots,x^{(i)})$$

Additionally, recall that we denote uncertainty quantifier for PC-FQI by $\tilde{U}_{k}^{(i)}$. Under Assumption~\ref{ass: UQ ass} which asserts the existence of an event $\Omega_{P}^{(i)}$ with $\Pr(\Omega_P^{(i)})\ge 1-\epsilon$ where $\tilde{U}_{k}^{(i)}$ satisfies Definition~\ref{def:UE} and is therefore a valid Uncertainty Quantifier. Because the proof requires $\tilde{U}_{k}^{(i)}$ to be a valid Uncertainty Quantifier $\forall i=0,\cdots, n_{K,b}$, we work on the intersection event
\[
\Omega_P \;:=\;\bigcap_{i=0}^{n_{K,b}}\Omega_P^{(i)}.
\]
which holds with probability at least $1-(n_{K,b}+1)\epsilon$. And on $\Omega_P$, we have $\forall i=0,\cdots, n_{K,b}$ and $\forall j=1,\cdots,K$ 
\begin{align}\label{eq: upper bound rho pess}
    \Tilde{\rho}_j^{ (i)}(x^{(0)},\cdots,x^{(i)}) 
    &= \calT^{\widetilde{\pi}_{j-1}}_{i,\mathcal{C}} \widetilde{Q}_{j-1}(x^{(0)},\cdots,x^{(i)}) - \widetilde{Q}^{(i)}_{j}(x^{(0)},\cdots,x^{(i)})\nonumber\\
    &= \calT^{\widetilde{\pi}_{j-1}}_{i,\mathcal{C}} \widetilde{Q}_{j-1}(x^{(0)},\cdots,x^{(i)}) - (\widehat{Q}^{(i)}_{j}(x^{(0)},\cdots,x^{(i)}) - \tilde{U}^{(i)}_j(x^{(0)},\cdots,x^{(i)}))\nonumber\\
    &\leq  2 \tilde{U}^{(i)}
_j(x^{(0)},\cdots,x^{(i)}).
\end{align}
This identity will be used in the following proof.

In addition, under Assumption~\ref{ass: surrogate outcome}, we work on the event $\Omega^r$ on which the surrogate reward relation holds. In particular, we have $\Omega^r$,
\begin{align}\label{eq: reward dis pess upper}
&\EE\left[\left(\widehat{R}_{i}\right)\prod_{j=0}^{i}\mathbb{I}[\Delta_{j}=0]\mathbb{I}[\Delta_{-1}=1]\given X_{0},\cdots,X_{i},\mathcal{C}(K,n_{K,b})\right]\nonumber\\
    &\geq\EE\left[\left(\widetilde{R}_{i} - C_5(\varepsilon)(NT)^{-\delta}\right)\prod_{j=0}^{i}\mathbb{I}[\Delta_{j}=0]\mathbb{I}[\Delta_{-1}=1]\given X_{0},\cdots,X_{i},\mathcal{C}(K,n_{K,b})\right],
\end{align}
which we use to account for the estimation error in the surrogate reward.

Accordingly, Lemma \ref{lm:upper bound on Qs pessimistic} is proved on the joint event $\Omega_P \cap \Omega^r$, i.e., under the simultaneous validity of the uncertainty quantification bounds across all depths and the surrogate reward approximation. By union bound, we have $\prob(\Omega_P \cap \Omega^r) \geq 1-(n_{K,b}+1)\epsilon-\varepsilon$.

Now, we are ready to show the proof.

\textit{Proof:}

We focus on deriving an upper bound on the difference between $Q^{(i)}_{\pi^{\ast}_{n_{K,b}},\mathcal{C}}(x^{(0)},\cdots,x^{(i)})$ and $\widetilde{Q}_{K}^{(i)}(x^{(0)},\cdots,x^{(i)})$ $\forall i=0,\cdots,n_{K,b}$

\begin{sizeddisplay}
{\footnotesize}
\begin{align}\label{eq: pfqi bound Q first equation}
\Tilde{I}_{i,K}
(x^{(0)},\cdots,x^{(i)})&
\triangleq Q^{(i)}_{\pi^{\ast}_{n_{K,b}},\mathcal{C}}(x^{(0)},\cdots,x^{(i)})-\widetilde{Q}_{K}^{(i)}(x^{(0)},\cdots,x^{(i)})\nonumber\\
&= Q^{(i)}_{\pi^{\ast}_{n_{K,b}},\mathcal{C}}(x^{(0)},\cdots,x^{(i)}) - \calT^{\tilde{\pi}_{K-1}}_{i,\mathcal{C}}\widetilde{Q}_{K-1}(x^{(0)},\cdots,x^{(i)}) + \Tilde{\rho}_K^{(i)}(x^{(0)},\cdots,x^{(i)})\nonumber\\
    &= Q^{(i)}_{\pi^{\ast}_{n_{K,b}},\mathcal{C}}(x^{(0)},\cdots,x^{(i)}) - \calT^{\pi^{\ast}_{n_{K,b}}}_{i,\mathcal{C}}\widetilde{Q}_{K-1}(x^{(0)},\cdots,x^{(i)})+\Tilde{\rho}_K^{(i)}(x^{(0)},\cdots,x^{(i)})\nonumber\\
    &+ \underbrace{\calT^{\pi^{\ast}_{n_{K,b}}}_{i,\mathcal{C}}\widetilde{Q}_{K-1}(x^{(0)},\cdots,x^{(i)})-\calT^{\tilde{\pi}_{K-1}}_{i,\mathcal{C}}\widetilde{Q}_{K-1}(x^{(0)},\cdots,x^{(i)})}_{(1)},
\end{align}
\end{sizeddisplay}
where we use Equation \eqref{eq: error P-FQI} to derive the equality. 

Observe that given the optimality of $\widetilde{\pi}^{(i)}_{j}$ with respect to $\widetilde{Q}_{j}^{(i)}$ for all $i=0,\cdots,n_{K,b}$ and for all $j=1,\cdots, K-1$, the following bound holds for all $i=0,\cdots,n_{K,b}$ and $j=1,\cdots, K-1$:
\begin{sizeddisplay}
{\footnotesize}\begin{align}
    (1)&=\calT^{\pi^{\ast}_{n_{K,b}}}_{i,\mathcal{C}}\widetilde{Q}_{j}(x^{(0)},\cdots,x^{(i)}) - \calT^{\tilde{\pi}_{j}}_{i,\mathcal{C}}\widetilde{Q}_{j}(x^{(0)},\cdots,x^{(i)})\nonumber\\
    &=\EE\left[\gamma\left(\widetilde{Q}_{j}^{(0)}(W_{i+1},\pi_{n_{K,b},*}^{(0)}(W_{i+1}))-\widetilde{Q}_{j}^{(0)}(W_{i+1},\widetilde{\pi}^{(0)}_{j}(W_{i+1})\right)\mathbb{I}[\Delta_{i}=1] \right.\nonumber\\
    &\left.  + \gamma\left(\widetilde{Q}_{j}^{(i+1)}(X_t,\cdots,W_{i+1},\pi^{(i+1)}_{n_{K,b},*}(W_{i+1},\cdots,X_0))-\widetilde{Q}_{j}^{(i+1)}(X_0,\cdots,W_{i+1},\widetilde{\pi}^{(i+1)}_{j}(W_{i+1},\cdots,X_0))\right)\mathbb{I}[\Delta_{i}=0]\right.\nonumber\\
    &\quad\left.\given (X_0,\cdots,X_{i})=(x^{(0)},\cdots,x^{(i)}),\mathcal{C}(K,n_{K,b})\right]\nonumber\\
    &\leq 0,
\end{align}\end{sizeddisplay}
which implies that Equation \eqref{eq: pfqi bound Q first equation} can be upper bounded as:
\begin{sizeddisplay}
{\footnotesize}\begin{align}\label{eq:pffqi upper bound on Q}
    \Tilde{I}_{i,K}(x^{(0)},\cdots,x^{(i)})  &\leq Q^{(i)}_{\pi^{\ast}_{n_{K,b}},\mathcal{C}}(x^{(0)},\cdots,x^{(i)}) - \calT^{\pi^{\ast}_{n_{K,b}}}_{i,\mathcal{C}}\widetilde{Q}_{K-1}(x^{(0)},\cdots,x^{(i)}) + \Tilde{\rho}_K^{(i)}(x^{(0)},\cdots,x^{(i)}) \nonumber\\
    &= \calT^{\pi^{\ast}_{n_{K,b}}}_{i,\mathcal{C}}Q_{\pi^{\ast}_{n_{K,b}},\mathcal{C}}(x^{(0)},\cdots,x^{(i)}) - \calT^{\pi^{\ast}_{n_{K,b}}}_{i,\mathcal{C}}\widetilde{Q}_{K-1}(x^{(0)},\cdots,x^{(i)})+\Tilde{\rho}_K^{(i)}(x^{(0)},\cdots,x^{(i)})\nonumber\\
    &=\gamma \calP^{\pi^{\ast}_{n_{K,b}}}_i\left(\big(Q^{(0)}_{\pi^{\ast}_{n_{K,b}},\mathcal{C}}(X_{i+1}) - \widetilde{Q}_{K-1}^{(0)}(X_{i+1})\big)\mathbb{I}[\Delta_{i}=1] \right.\nonumber\\
    &\quad \left.+ \big(Q^{(i+1)}_{\pi^{\ast}_{n_{K,b}},\mathcal{C}}(X_0,\cdots,X_{i+1}) - \widetilde{Q}_{K-1}^{(i+1)}(X_0,\cdots,X_{i+1})\big)\prod_{j=0}^{i}\mathbb{I}[\Delta_{j}=0]\mathbb{I}[\Delta_{-1}=1]\right)\nonumber\\
    &+\EE\left[\left(R_{i}-\widehat{R}_{i}\right)\prod_{j=0}^{i}\mathbb{I}[\Delta_{j}=0]\mathbb{I}[\Delta_{-1}=1]\given (X_{0},\cdots,X_{i}) = (x^{(0)},\cdots,x^{(i)}),\mathcal{C}(K,n_{K,b})\right]\nonumber\\
    &+ \Tilde{\rho}_K^{(i)}(x^{(0)},\cdots,x^{(i)}),
\end{align}
\end{sizeddisplay}
where the characterization of $\calP^{\pi^{\ast}_{n_{K,b}}}_i$ given in Definition \ref{def: P as conditional exp}.

By Lemma \ref{lm: identification for censored demand} and on event $\Omega_P \cap \Omega^r$, the following holds:
\begin{sizeddisplay}
{\footnotesize}
\begin{align*}
    &\EE\left[\left(R_{i}-\widehat{R}_{i}\right)\prod_{j=0}^{i}\mathbb{I}[\Delta_{j}=0]\mathbb{I}[\Delta_{-1}=1]\given X_{0},\cdots,X_{i},\mathcal{C}(K,n_{K,b})\right]\\
    &\leq\EE\left[\left(R_{i}-\widetilde{R}_{i} + C_5(\varepsilon)(NT)^{-\delta}\right)\prod_{j=0}^{i}\mathbb{I}[\Delta_{j}=0]\mathbb{I}[\Delta_{-1}=1]\given X_{0},\cdots,X_{i},\mathcal{C}(K,n_{K,b})\right]\\
    &\leq C_5(\varepsilon)(NT)^{-\delta},
\end{align*}\end{sizeddisplay}
where the first inequality is implied by Equation~\eqref{eq: reward dis pess upper}.
Then, Equation \eqref{eq:pffqi upper bound on Q} becomes
\begin{sizeddisplay}
{\footnotesize}\begin{align}\label{eq:p (1) (2)}
    \Tilde{I}_{i,K}(x^{(0)},\cdots,x^{(i)})  &\leq \gamma \calP^{\pi^{\ast}_{n_{K,b}}}_i\left(\big(\underbrace{Q^{(0)}_{\pi^{\ast}_{n_{K,b}},\mathcal{C}}(X_{i+1}) - \widetilde{Q}_{K-1}^{(0)}(X_{i+1})}_{(2)}\big)\mathbb{I}[\Delta_{i}=1] \right.\nonumber\\
    &\quad \left.+ \big(\underbrace{Q^{(i+1)}_{\pi^{\ast}_{n_{K,b}},\mathcal{C}}(X_0,\cdots,X_{i+1}) - \widetilde{Q}_{K-1}^{(i+1)}(X_0,\cdots,X_{i+1})}_{(3)}\big)\prod_{j=0}^{i}\mathbb{I}[\Delta_{j}=0]\mathbb{I}[\Delta_{-1}=1]\right)\nonumber\\
    &\quad+C_5(\varepsilon)(NT)^{-\delta}+ \Tilde{\rho}_K^{(i)}(x^{(0)},\cdots,x^{(i)}).
\end{align}
\end{sizeddisplay}
Observe that 
{\footnotesize\begin{align*}
    (2)=\Tilde{I}_{0,K-1}(X_{i+1})
\end{align*}}
{\footnotesize\begin{align*}
    (3)=\Tilde{I}_{i+1,K-1}(X_0,\cdots,X_{i+1}).
\end{align*}}

Therefore, Equation \eqref{eq:p (1) (2)} can be expressed as:
{\footnotesize\begin{align}\label{eq:pfqi upper bound on Q}
    \Tilde{I}_{i,K}(x^{(0)},\cdots,x^{(i)})  
    &\leq\gamma \calP^{\pi^{\ast}_{n_{K,b}}}_i\left(\big(\Tilde{I}_{0,K-1}(X_{i+1})\big)\mathbb{I}[\Delta_{i}=1] \right.\nonumber\\
    &\quad \left.+ \big(\Tilde{I}_{i+1,K-1}(X_0,\cdots,X_{i+1})\big)\prod_{j=0}^{i}\mathbb{I}[\Delta_{j}=0]\mathbb{I}[\Delta_{-1}=1]\right)\nonumber\\
    &\quad +C_5(\varepsilon)(NT)^{-\delta}+ \Tilde{\rho}_K^{(i)}(x^{(0)},\cdots,x^{(i)}).
\end{align}}

Next, we use Equation \eqref{eq:p (1) (2)} to further upper bound the right hand side of Equation \eqref{eq:pfqi upper bound on Q} in an iterative manner as follows: 

\begin{sizeddisplay}
 {\footnotesize}
\begin{align*}
\Tilde{I}_{i,K}(x^{(0)},\cdots,x^{(i)}) &\leq \gamma \calP^{\pi^{\ast}_{n_{K,b}}}_i\left\{\left(\gamma \calP^{\pi^{\ast}_{n_{K,b}}}_0\left(\big(Q^{(0)}_{\pi^{\ast}_{n_{K,b}},\mathcal{C}}(X_{i+2}) - \widetilde{Q}_{K-2}^{(0)}(X_{i+2})\big)\mathbb{I}[\Delta_{i+1}=1] \right.\right.\right.\\
&\left.\left.\left.\quad+ \big(Q^{(1)}_{\pi^{\ast}_{n_{K,b}},\mathcal{C}}(X_{i+1},X_{i+2}) - \widetilde{Q}_{K-2}^{(1)}(X_{i+1},X_{i+2})\big)\mathbb{I}[\Delta_{i+1}=0]\mathbb{I}[\Delta_{i}=1]\right) \right.\right.\\
&.\left.\left.\quad +C_5(\varepsilon)(NT)^{-\delta}+ \Tilde{\rho}_{K-1}^{(0)}(X_{i+1})\right)\mathbb{I}[\Delta_{i}=1]\right.\\
&\left. \quad+\left(\gamma\calP^{\pi^{\ast}_{n_{K,b}}}_{i+1}\left(\big(Q^{(0)}_{\widetilde{\pi}^*,\mathcal{C}}(X_{i+2}) - \widetilde{Q}_{K-2}^{(0)}(X_{i+2})\big)\mathbb{I}[\Delta_{i+1}=1] \right.\right.\right.\\
&\left.\left.\left.\quad+ \big(Q^{(i+2)}_{\pi^{\ast}_{n_{K,b}},\mathcal{C}}(X_0,\cdots,X_{i+2}) - \widetilde{Q}_{K-2}^{(i+2)}(X_0,\cdots X_{i+2})\big)\prod_{j=0}^{i+1}\mathbb{I}[\Delta_{j}=0]\mathbb{I}[\Delta_{-1}=1]\right)\right.\right.\\
&\quad\left.\left. +C_5(\varepsilon)(NT)^{-\delta}+\Tilde{\rho}_{K-1}^{(i+1)}(X_0,\cdots,X_{i+1})\right)\prod_{j=0}^{i}\mathbb{I}[\Delta_{j}=0]\mathbb{I}[\Delta_{-1}=1]\right\}\\
&\quad+C_5(\varepsilon)(NT)^{-\delta}+ \Tilde{\rho}_K^{(i)}(x^{(0)},\cdots,x^{(i)}).
 \end{align*}\end{sizeddisplay}

 After simplification, we obtain the following:
 {\footnotesize \begin{align*}
 \Tilde{I}_{i,K}(x^{(0)},\cdots,x^{(i)})&\leq \gamma^2 (\calP^{\pi^{\ast}_{n_{K,b}}}_i\calP^{\pi^{\ast}_{n_{K,b}}}_{i+1})\left\{\left(\Tilde{I}_{0,K-2}(X_{i+2})\right)\mathbb{I}[\Delta_{i+1}=1] \right.\\
    &\quad \left. + \left(\Tilde{I}_{1,K-2}(X_{i+1},X_{i+2})\right)\mathbb{I}[\Delta_{i+1}=0]\mathbb{I}[\Delta_{i}=1] \right.\\
    &\left. \quad + \left(\Tilde{I}_{i+2,K-2}(X_{0},\cdots,X_{i+2})\right)\prod_{j=0}^{i+1}\mathbb{I}[\Delta_{j}=0]\mathbb{I}[\Delta_{-1}=1]\right\} \\
    &\quad + \gamma\calP^{\pi^{\ast}_{n_{K,b}}}_{i}\left\{\left(C_5(\varepsilon)(NT)^{-\delta}+\Tilde{\rho}_{K-1}^{(0)}(X_{i+1})\right)\mathbb{I}[\Delta_{i}=1] \right.\\
    &\quad\left.+ \left(C_5(\varepsilon)(NT)^{-\delta}+\Tilde{\rho}_{K-1}^{(i+1)}(X_{0},\cdots,X_{i+1}) \right)\prod_{j=0}^{i}\mathbb{I}[\Delta_{j}=0]\mathbb{I}[\Delta_{-1}=1] \right\} \\
 &\quad +C_5(\varepsilon)(NT)^{-\delta} + \Tilde{\rho}_K^{(i)}(x^{(0)},\cdots,x^{(i)}).
 \end{align*}}

 For the remaining part of the proof, we use $I_{i,K}$ to denote $I_{i,K}(x^{(0)},\cdots,x^{(i)})$. Then, by pursuing the same procedure,  we reach to the following after $K-1$ steps:

\begin{sizeddisplay}
{\footnotesize}\begin{align*}
     \Tilde{I}_{i,K}
     &\leq\gamma^{K-1}\left( \prod_{k=0}^{K-2}{\calP_{i+k}^{\pi^{\ast}_{n_{K,b}}}}\right) \left\{\sum_{k=0}^{n_{K,b}}\left(\left(Q^{(k)}_{\pi^{\ast}_{n_{K,b}},\mathcal{C}}(X_{i+K-1-k},\cdots,X_{i+K-1}) - \widehat{R}_{i+K-1}\right)\right.\right.\\
     &\left.\prod_{j=i+K-1-k}^{i+K-2}\mathbb{I}[\Delta_{j}=0]\mathbb{I}[\Delta_{i+K-2-k}=1] \right\}\\
     &+\sum_{k=1}^{K-1}\gamma^k \left(\prod_{k'=0}^{k-1}\calP_{i+k'}^{\pi^{\ast}_{n_{K,b}}}\right)\left\{\left(C_5(\varepsilon)(NT)^{-\delta}+\Tilde{\rho}_{K-k}^{(0)}(X_{i+k})\right)\mathbb{I}[\Delta_{i+k-1}=1] \right.\\
     &\left.+\left(C_5(\varepsilon)(NT)^{-\delta}+\Tilde{\rho}_{K-k}^{(i+k)}(X_{0},\cdots,X_{i+k})\right)\prod_{j=0}^{i+k-1}\mathbb{I}[\Delta_{j}=0]\mathbb{I}[\Delta_{-1}=1]\mathbb{I}[k\leq n_{K,b}-i] \right.\\
     &\left.+ \sum_{v=1}^{\min(n_{K,b},k-1)}\left\{\left(C_5(\varepsilon)(NT)^{-\delta}+\Tilde{\rho}_{K-k}^{(v)}(X_{i+k-v},\cdots,X_{i+k})\right)\prod_{j=i+k-v}^{i+k-1}\mathbb{I}[\Delta_{j}=0]\mathbb{I}[\Delta_{i+k-v-1}=1]\right\}\right\}\\
     &+C_5(\varepsilon)(NT)^{-\delta}+ \Tilde{\rho}_K^{(i)}(x^{(0)},\cdots,x^{(i)}).
\end{align*}\end{sizeddisplay}
The introduction of $\min(n_{K,b},k-1)$ as the upper limit in the last summation is implied by the fact that $\pi^{\ast}_{n_{K,b}}\in \Pi_{n_{K,b}}$, which indicates  $\Tilde{\rho}_{K-k}^{(v)}$ cannot exist for any $v>n_{K,b}$.  

On the event $\Omega_P \cap \Omega^r$, Equation~\eqref{eq: upper bound rho pess} implies that, for all $i \in \{0,\ldots,n_{K,b}\}$, we have:
\begin{sizeddisplay}
{\footnotesize}
\begin{align}\label{eq: PFQI error and UQ}
	 \Tilde{\rho}_j^{(i)}(x^{(0)},\cdots,x^{(i)}) \leq 2 \tilde{U}_j^{(i)}(x^{(0)},\cdots,x^{(i)}) \qquad \forall j=1,\cdots,K
\end{align}\end{sizeddisplay}
holds for all $(x^{(0)},\cdots, x^{(i)}) \in \calX^{\otimes (i+1)}$. Additionally,  $|Q^{(i)}_{\pi^{*}_{n_{K,b}},\mathcal{C}}|$ is uniformly bounded by $\frac{R_{\max}}{1-\gamma}$ $\forall i=0,\cdots,n_{K,b}$ and $|\widehat{R}_t|$ is uniformly bounded by $R_{\max}$ $\forall t\geq 0$. Together, they imply:
\begin{sizeddisplay}
{\footnotesize}\begin{align}\label{eq:pfqi  final upper bound Q}
    \Tilde{I}_{i,K}&\leq \gamma^{K-1}\left(\frac{2-\gamma}{1-\gamma}R_{\max}\right)+\sum_{k=1}^{K-1}\gamma^k \left(\prod_{k'=0}^{k-1}\calP_{i+k'}^{\pi^{\ast}_{n_{K,b}}}\right)\left\{\left(C_5(\varepsilon)(NT)^{-\delta}+2\Tilde{U}_{K-k}^{(0)}(X_{i+k})\right)\mathbb{I}[\Delta_{i+k-1}=1]\right.\nonumber\\
     &\left.+\left(C_5(\varepsilon)(NT)^{-\delta}+2\Tilde{U}_{K-k}^{(i+k)}(X_{0},\cdots,X_{i+k})\right)\prod_{j=0}^{i+k-1}\mathbb{I}[\Delta_{j}=0]\mathbb{I}[\Delta_{-1}=1]\mathbb{I}[k\leq n_{K,b}-i] \right.\nonumber\\
     &\left.+ \sum_{v=1}^{\min(n_{K,b},k-1)}\left\{\left(C_5(\varepsilon)(NT)^{-\delta}+2\Tilde{U}_{K-k}^{(v)}(X_{i+k-v},\cdots,X_{i+k})\right)\prod_{j=i+k-v}^{i+k-1}\mathbb{I}[\Delta_{j}=0]\mathbb{I}[\Delta_{i+k-v-1}=1]\right\}\right\}\nonumber\\
     &+C_5(\varepsilon)(NT)^{-\delta}+2\Tilde{U}_{K}^{(i)}(x^{(0)},\cdots,x^{(i)})\nonumber\\
     &=\gamma^{K-1}\left(\frac{2-\gamma}{1-\gamma}R_{\max}\right)+ \text{RD}(NT)+\widetilde{\text{UQ}}(x^{(0)},\cdots,x^{(i)}),
\end{align}\end{sizeddisplay}
where the characterizations of $\text{RD}(NT)$ and $\widetilde{\text{UQ}}(x^{(0)},\cdots,x^{(i)})$ are given in Definitions \ref{def: fqi RD definition}  and\hyperlink{uq_fqi_ii}{~\ref{def:uq_definition} (ii)}. This concludes the proof of Lemma \ref{lm:upper bound on Qs pessimistic}.

\subsection{Proof of Lemma \ref{lm:lower bound on Qs pessimistic}}\label{seubsec: proof of lower bound on Qs pessimistic}
In this section, we provide the proof of Lemma \ref{lm:lower bound on Qs pessimistic}.
Before the proof, we introduce several related notations and definitions.

For all $ j=1,\cdots,K$ and $ i=0,\cdots,n_{K,b}$, define:
\begin{align}\label{eq: error P-FQI lower}
    \Tilde{\rho}_j^{ (i)}(x^{(0)},\cdots,x^{(i)}) = \calT^{\tilde{\pi}_{j-1}}_{i,\mathcal{C}} \widetilde{Q}_{j-1}(x^{(0)},\cdots,x^{(i)}) - \widetilde{Q}^{(i)}_{j}(x^{(0)},\cdots,x^{(i)}),
\end{align} 
where $\calT^{\tilde{\pi}_{j-1}}_{i,\mathcal{C}}$ is defined under Definition \ref{def: bellman operator}.

Additionally, recall that we denote uncertainty quantifier for PC-FQI by $\tilde{U}_{k}^{(i)}$. Under Assumption~\ref{ass: UQ ass} which asserts the existence of an event $\Omega_{P}^{(i)}$ with $\Pr(\Omega_P^{(i)})\ge 1-\epsilon$ where $\tilde{U}_{k}^{(i)}$ satisfies Definition~\ref{def:UE} and is therefore a valid Uncertainty Quantifier. Because the proof requires $\tilde{U}_{k}^{(i)}$ to be a valid Uncertainty Quantifier $\forall i=0,\cdots, n_{K,b}$, we work on the intersection event
\[
\Omega_P \;:=\;\bigcap_{i=0}^{n_{K,b}}\Omega_P^{(i)}.
\]
which holds with probability at least $1-(n_{K,b}+1)\epsilon$. And on $\Omega_P$, we have $\forall i=0,\cdots, n_{K,b}$ and $\forall j=1,\cdots,K$ 
\begin{align}\label{eq: lower bound rho pess}
    \Tilde{\rho}_j^{ (i)}(x^{(0)},\cdots,x^{(i)}) 
    &= \calT^{\tilde{\pi}_{j-1}}_{i,\mathcal{C}} \widetilde{Q}_{j-1}(x^{(0)},\cdots,x^{(i)}) - \widetilde{Q}^{(i)}_{j}(x^{(0)},\cdots,x^{(i)})\nonumber\\
    &= \calT^{\tilde{\pi}_{j-1}}_{i,\mathcal{C}} \widetilde{Q}_{j-1}(x^{(0)},\cdots,x^{(i)}) - (\widehat{Q}^{(i)}_{j}(x^{(0)},\cdots,x^{(i)}) - \tilde{U}^{(i)}_j(x^{(0)},\cdots,x^{(i)}))\nonumber\\
    &\geq  0
\end{align}
This identity will be used in the following proof.

In addition, under Assumption~\ref{ass: surrogate outcome}, we work on the event $\Omega^r$ on which the surrogate reward relation holds. In particular, we have $\Omega^r$,
\begin{align}\label{eq: reward dis pess lower}
&\EE\left[\left(\widehat{R}_{i}\right)\prod_{j=0}^{i}\mathbb{I}[\Delta_{j}=0]\mathbb{I}[\Delta_{-1}=1]\given X_{0},\cdots,X_{i},\mathcal{C}(K,n_{K,b})\right]\nonumber\\
    &\leq\EE\left[\left(\widetilde{R}_{i} + C_5(\varepsilon)(NT)^{-\delta}\right)\prod_{j=0}^{i}\mathbb{I}[\Delta_{j}=0]\mathbb{I}[\Delta_{-1}=1]\given X_{0},\cdots,X_{i},\mathcal{C}(K,n_{K,b})\right],
\end{align}
which we use to account for the estimation error in the surrogate reward.

Accordingly, Lemma \ref{lm:upper bound on Qs pessimistic} is proved on the joint event $\Omega_P \cap \Omega^r$, i.e., under the simultaneous validity of the uncertainty quantification bounds across all depths and the surrogate reward approximation. By union bound, we have $\prob(\Omega_P \cap \Omega^r) \geq 1-(n_{K,b}+1)\epsilon-\varepsilon$.

Now, we are ready to show the proof.

\textit{Proof:}

We focus on deriving a lower bound on the difference between $ Q_{\pi^{\ast}_{n_{K,b}},\mathcal{C}}^{(i)}(x^{(0)},\cdots,x^{(i)})$ and $\widetilde{Q}_{K}^{(i)}(x^{(0)},\cdots,x^{(i)})$ $\forall i=0,\cdots,n_{K,b}$.
\begin{sizeddisplay}
{\footnotesize}\begin{align*}
    \Tilde{I}_{i,K}
(x^{(0)},\cdots,x^{(i)})  &\triangleq Q_{\pi^{\ast}_{n_{K,b}},\mathcal{C}}^{(i)}(x^{(0)},\cdots,x^{(i)})-\widetilde{Q}_{K}^{(i)}(x^{(0)},\cdots,x^{(i)})\\&=Q_{\pi^{\ast}_{n_{K,b}},\mathcal{C}}^{(i)}(x^{(0)},\cdots,x^{(i)}) - \calT_{i,\mathcal{C}}^{\tilde{\pi}_{K-1}}\widetilde{Q}_{K-1}(x^{(0)},\cdots,x^{(i)}) + \Tilde{\rho}_K^{(i)}(x^{(0)},\cdots,x^{(i)})\\
    &= Q_{\pi^{\ast}_{n_{K,b}},\mathcal{C}}^{(i)}(x^{(0)},\cdots,x^{(i)}) - \calT^{\tilde{\pi}_{K-1}}_{i,\mathcal{C}}Q_{\pi^{\ast}_{n_{K,b}},\mathcal{C}}(x^{(0)},\cdots,x^{(i)})\\
    &+ \calT^{\tilde{\pi}_{K-1}}_{i,\mathcal{C}}Q_{\pi^{\ast}_{n_{K,b}},\mathcal{C}}(x^{(0)},\cdots,x^{(i)})-\calT^{\tilde{\pi}_{K-1}}_{i,\mathcal{C}}\widetilde{Q}_{K-1}(x^{(0)},\cdots,x^{(i)})\\
    &+ \Tilde{\rho}_K^{(i)}(x^{(0)},\cdots,x^{(i)})\\
    &=\underbrace{\calT^{\pi^{\ast}_{n_{K,b}}}_{i,\mathcal{C}}Q_{\pi^{\ast}_{n_{K,b}},\mathcal{C}}(x^{(0)},\cdots,x^{(i)}) - \calT^{\tilde{\pi}_{K-1}}_{i,\mathcal{C}}Q_{\pi^{\ast}_{n_{K,b}},\mathcal{C}}(x^{(0)},\cdots,x^{(i)})}_{(1)}\\
    &+ \calT^{\tilde{\pi}_{K-1}}_{i,\mathcal{C}}Q_{\pi^{\ast}_{n_{K,b}},\mathcal{C}}^{(i)}(x^{(0)},\cdots,x^{(i)})
    - \calT^{\tilde{\pi}_{K-1}}_{i,\mathcal{C}}\widetilde{Q}_{K-1}(x^{(0)},\cdots,x^{(i)})\\
    &+ \Tilde{\rho}_K^{(i)}(x^{(0)},\cdots,x^{(i)}) 
\end{align*}\end{sizeddisplay}
Consider the term (1) above $\forall i=0,\cdots,n_{K,b}$:
\begin{sizeddisplay}
{\footnotesize}
\begin{align*}
    (1)&=\calT^{\pi^{\ast}_{n_{K,b}}}_{i,\mathcal{C}}Q_{\pi^{\ast}_{n_{K,b}},\mathcal{C}}^{(i)}(x^{(0)},\cdots,x^{(i)}) - \calT^{\tilde{\pi}_{K-1}}_{i,\mathcal{C}}Q_{\pi^{\ast}_{n_{K,b}},\mathcal{C}}^{(i)}(x^{(0)},\cdots,x^{(i)})\nonumber\\
    &=\EE\left[\gamma\left(Q_{\pi^{\ast}_{n_{K,b}},\mathcal{C}}^{(0)}(W_{i+1},\pi^{(0)}_{n_{K,b},*}(W_{i+1}))-Q_{\pi^{\ast}_{n_{K,b}},\mathcal{C}}^{(0)}(W_{i+1},\widetilde{\pi}^{(0)}_{K-1}(W_{i+1}))\right)\mathbb{I}[\Delta_{i}=1] \right.\nonumber\\
    &\left.  + \gamma\left(Q_{\pi^{\ast}_{n_{K,b}},\mathcal{C}}^{(i+1)}(X_0,\cdots,W_{i+1},\pi_{n_{K,b},*}^{(i+1)}(W_{i+1},\cdots,X_0))-Q_{\pi^{\ast}_{n_{K,b}},\mathcal{C}}^{(i+1)}(X_0,\cdots,W_{i+1},\widetilde{\pi}^{(i+1)}_{K-1}(W_{i+1},\cdots,X_0))\right)\mathbb{I}[\Delta_{i}=0]\right.\nonumber\\
    &\left.\quad\given (X_0,\cdots,X_{i})=(x^{(0)},\cdots,x^{(i)}),\mathcal{C}(K,n_{K,b})\right]\nonumber,
\end{align*}\end{sizeddisplay}
where the first equality stems from the definition of $\calT^{\pi}_{i,\mathcal{C}}$ given in Definition \ref{def: bellman operator}. Recall the characterization of policy $\pi^{\ast}_{n_{K,b}}$ given under Lemma \ref{lemma: nkb and action relation}. For $i=0,\cdots,n_{K,b}-1$, $\pi^{\ast}_{n_{K,b}}$ takes the point-wise maximum of $Q^{(i)}_{\pi^{\ast}_{n_{K,b}},\mathcal{C}}$ over all possible actions given $(W_{i},\cdots,X_0)$. This implies $\forall i=0,\cdots,n_{K,b}-1$:
\begin{sizeddisplay}
    {\footnotesize}
    \begin{align}\label{eq: action discrepancy lower bound}
        Q_{\pi^{\ast}_{n_{K,b}},\mathcal{C}}^{(i)}(X_0,\cdots,W_{i},\pi_{n_{K,b},*}^{(i)}(W_{i},\cdots,X_0))-Q_{\pi^{\ast}_{n_{K,b}},\mathcal{C}}^{(i)}(X_0,\cdots,W_{i},\widetilde{\pi}^{(i)}_{K-1}(W_{i},\cdots,X_0))\geq 0 
    \end{align}
\end{sizeddisplay}
When $i=n_{K,b}$, $\pi^{\ast}_{n_{K,b}}$ takes the point-wise maximum of $Q^{(i)}_{\pi^{\ast}_{n_{K,b}},\mathcal{C}}$ over a restricted set of actions denoted by $\textup{UP}^{(n_{K,b)}}(W_{n_{K,b}},X_{n_{K,b}-1},\cdots,X_{0})$. Therefore, it cannot be ensured that Equation \eqref{eq: action discrepancy lower bound} holds when $i=n_{K,b}$. To tackle this issue, we make use of Definition\hyperlink{action_discrepancy_i}{~\ref{def:action_discrepancy} (i)}:
\begin{sizeddisplay}
    {\footnotesize}
    \begin{align}
        &Q_{\pi^{\ast}_{n_{K,b}},\mathcal{C}}^{(n_{K,b})}(X_0,\cdots,W_{n_{K,b}},\pi_{n_{K,b},*}^{(n_{K,b})}(W_{n_{K,b}},\cdots,X_0))-Q_{\pi^{\ast}_{n_{K,b}},\mathcal{C}}^{(n_{K,b})}(X_0,\cdots,W_{n_{K,b}},\widehat{\pi}^{(n_{K,b})}_{K-1}(W_{n_{K,b}},\cdots,X_0))\nonumber\\
        &\geq  Q_{\pi^{\ast}_{n_{K,b}},\mathcal{C}}^{(n_{K,b})}(X_0,\cdots,W_{n_{K,b}},\pi_{n_{K,b},*}^{(n_{K,b})}(W_{n_{K,b}},\cdots,X_0))-\max_{a \in \calA}Q_{\pi^{\ast}_{n_{K,b}},\mathcal{C}}^{(n_{K,b})}(X_0,\cdots,W_{n_{K,b}},a)\nonumber\\
        &=-\textup{AD}^{\pi^{*}_{n_{K,b}}}(X_0,\cdots,W_{n_{K,b}})
    \end{align}
\end{sizeddisplay}
Consequently, it can be concluded that:
\begin{sizeddisplay}
    {\footnotesize}
    \begin{align}
&\calT^{\pi^{\ast}_{n_{K,b}}}_{i,\mathcal{C}}Q_{\pi^{\ast}_{n_{K,b}},\mathcal{C}}^{(i)}(x^{(0)},\cdots,x^{(i)}) - \calT^{\tilde{\pi}_{K-1}}_{i,\mathcal{C}}Q_{\pi^{\ast}_{n_{K,b}},\mathcal{C}}^{(i)}(x^{(0)},\cdots,x^{(i)})\\
        &\geq -\EE\left[\gamma\textup{AD}^{\pi^{*}_{n_{K,b}}}(X_0,\cdots,W_{i+1})\mathbb{I}[\Delta_{i}=0]\given (X_0,\cdots,X_{i})=(x^{(0)},\cdots,x^{(i)}),\mathcal{C}(K,n_{K,b})]\right]\mathbb{I}[i=n_{K,b}-1],\nonumber
    \end{align}
\end{sizeddisplay}
which further implies the following:
\begin{sizeddisplay}
{\footnotesize}\begin{align}\label{eq:pfqi lower bound on Q}
    \Tilde{I}_{i,K}(x^{(0)},\cdots,x^{(i)}) 
    &\geq \calT^{\tilde{\pi}_{K-1}}_{i,\mathcal{C}}Q_{\pi^{\ast}_{n_{K,b}},\mathcal{C}}^{(i)}(x^{(0)},\cdots,x^{(i)})
    - \calT^{\tilde{\pi}_{K-1}}_{i,\mathcal{C}}\widetilde{Q}_{K-1}(x^{(0)},\cdots,x^{(i)}) \nonumber\\
    &-\EE\left[\gamma\textup{AD}^{\pi^{*}_{n_{K,b}}}(X_0,\cdots,W_{i+1})\mathbb{I}[\Delta_{i}=0]\given (X_0,\cdots,X_{i})=(x^{(0)},\cdots,x^{(i)}),\mathcal{C}(K,n_{K,b})]\right]\mathbb{I}[i=n_{K,b}-1]\nonumber\\
    &+\Tilde{\rho}_K^{(i)}(x^{(0)},\cdots,x^{(i)})
\end{align}
\end{sizeddisplay}
By extending the right hand side of Equation \eqref{eq:pfqi lower bound on Q}, we obtain:
\begin{sizeddisplay}
{\footnotesize}\begin{align}\label{eq:pessimistic lower bound on Q}
    \Tilde{I}_{i,K}(x^{(0)},\cdots,x^{(i)}) 
    &\geq \gamma \calP_{i}^{\tilde{\pi}_{K-1}}\left\{\left(\underbrace{Q^{(0)}_{\pi^{\ast}_{n_{K,b}},\mathcal{C}}(X_{i+1}) - \widetilde{Q}_{K-1}^{(0)}(X_{i+1})}_{I_{0,K-1}(X_{i+1})}\right)\mathbb{I}[\Delta_{i}=1]\right.\nonumber\\
    &\left. +\left(\underbrace{Q^{(i+1)}_{\pi^{\ast}_{n_{K,b}},\mathcal{C}}(X_0,\cdots,X_{i+1}) - \widetilde{Q}_{K-1}^{(i+1)}(X_0,\cdots,X_{i+1})}_{I_{i+1,K-1}(X_0,\cdots,X_{i+1})}\right.\right.\nonumber\\
    &\left.\left.-\textup{AD}^{\pi^{*}_{n_{K,b}}}(X_0,\cdots,W_{i+1})\mathbb{I}[i=n_{K,b}-1]\right)\mathbb{I}[\Delta_{i}=0]    \right\}\nonumber\\
    &+\EE\left[\left(R_{i}-\widehat{R}_{i}\right)\prod_{j=t}^{t+i}\mathbb{I}[\Delta_{j}=0]\mathbb{I}[\Delta_{-1}=1]\given (X_{0},\cdots,X_{i}) = (x^{(0)},\cdots,x^{(i)}),\mathcal{C}(K,n_{K,b})\right]\nonumber\\
    &+\Tilde{\rho}_K^{(i)}(x^{(0)},\cdots,x^{(i)}),
\end{align}\end{sizeddisplay}
where characterization of $\calP_{i}^{\tilde{\pi}_{K-1}}$ given in Definition \ref{def: P as conditional exp}.

Observe that By Lemma \ref{lm: identification for censored demand} and on event $\Omega_P \cap \Omega^r$, the following holds
\begin{sizeddisplay}
{\footnotesize}
\begin{align*}
    &\EE\left[\left(R_{i}-\widehat{R}_{i}\right)\prod_{j=0}^{i}\mathbb{I}[\Delta_{j}=0]\mathbb{I}[\Delta_{-1}=1]\given (X_{0},\cdots,X_{i}) = (x^{(0)},\cdots,x^{(i)}),\mathcal{C}(K,n)\right]\\
    &\geq \EE\left[\left(R_{i}-\widetilde{R}_{i}-C_5(\varepsilon)(NT)^{-\delta}\right)\prod_{j=0}^{i}\mathbb{I}[\Delta_{j}=0]\mathbb{I}[\Delta_{-1}=1]\given (X_{0},\cdots,X_{i}) = (x^{(0)},\cdots,x^{(i)}),\mathcal{C}(K,n)\right]\\
    &=\EE\left[-C_5(\varepsilon)(NT)^{-\delta}\prod_{j=0}^{i}\mathbb{I}[\Delta_{j}=0]\mathbb{I}[\Delta_{-1}=1]\given (X_{0},\cdots,X_{i}) = (x^{(0)},\cdots,x^{(i)}),\mathcal{C}(K,n)\right]\\ 
    &\geq-C_5(\varepsilon)(NT)^{-\delta},
\end{align*}\end{sizeddisplay}
where the first inequality is implied by Equation~\eqref{eq: reward dis pess lower}. 

Therefore, Equation \eqref{eq:pessimistic lower bound on Q} becomes:
\begin{sizeddisplay}
{\footnotesize}\begin{align}\label{eq:pessimistic lower bound on Q2}
    \Tilde{I}_{i,K}(x^{(0)},\cdots,x^{(i)}) &\geq \gamma \calP_{i}^{\tilde{\pi}_{K-1}}\left\{\left(\underbrace{Q^{(0)}_{\pi^{\ast}_{n_{K,b}},\mathcal{C}}(X_{i+1}) - \widetilde{Q}_{K-1}^{(0)}(X_{i+1})}_{I_{0,K-1}(X_{i+1})}\right)\mathbb{I}[\Delta_{i}=1]\right.\nonumber\\
    &\left. +\left(\underbrace{Q^{(i+1)}_{\pi^{\ast}_{n_{K,b}},\mathcal{C}}(X_0,\cdots,X_{i+1}) - \widetilde{Q}_{K-1}^{(i+1)}(X_0,\cdots,X_{i+1})}_{I_{i+1,K-1}(X_0,\cdots,X_{i+1})}\right.\right.\nonumber\\
    &\left.\left.-\textup{AD}^{\pi^{*}_{n_{K,b}}}(X_0,\cdots,W_{i+1})\mathbb{I}[i=n_{K,b}-1]\right)\mathbb{I}[\Delta_{i}=0]    \right\}\nonumber\\
    &+\Tilde{\rho}_K^{(i)}(x^{(0)},\cdots,x^{(i)})  -C_5(\varepsilon)(NT)^{-\delta}.
\end{align}\end{sizeddisplay}
Next, we apply the lower bounds on $\Tilde{I}_{0,K-1}$ and $ \Tilde{I}_{i+1,K-1}$
in an iterative manner:\begin{sizeddisplay}{\footnotesize}\begin{align*}
    \Tilde{I}_{i,K}(x^{(0)},\cdots,x^{(i)})
    &\geq \gamma \calP_{i}^{\tilde{\pi}_{K-1}}\left(\left(\gamma \calP_{0}^{\tilde{\pi}_{K-2}}\left(\big(Q^{(0)}_{\pi^{\ast}_{n_{K,b}},\mathcal{C}}(X_{i+2}) - \widetilde{Q}_{K-2}^{(0)}(X_{i+2})\big)\mathbb{I}[\Delta_{i+1}=1]\right.\right.\right.\\
    &\left.\left.\left.+\big(Q^{(1)}_{\pi^{\ast}_{n_{K,b}},\mathcal{C}}(X_{i+1},X_{i+2}) - \widetilde{Q}_{K-2}^{(1)}(X_{i+1},X_{i+2})\big)\mathbb{I}[\Delta_{i+1}=0]\right)\right.\right.\\
    &\left.\left.+\Tilde{\rho}_{K-1}^{(0)}(X_{i+1}) -C_5(\varepsilon)(NT)^{-\delta} \right)\mathbb{I}[\Delta_{i}=1]\right.\\
    &\left. +\left(\gamma \calP_{i+1}^{\tilde{\pi}_{K-2}}\left(\big(Q^{(0)}_{\pi^{\ast}_{n_{K,b}},\mathcal{C}}(X_{i+2}) - \widetilde{Q}_{K-2}^{(0)}(X_{i+2})\big)\mathbb{I}[\Delta_{i+1}=1] \right.\right.\right.\\
    &\left.\left.\left.\left.+(Q^{(i+2)}_{\pi^{\ast}_{n_{K,b}},\mathcal{C}}(X_0,\cdots,X_{i+2})- \widetilde{Q}_{K-2}^{(i+2)}(X_0,\cdots,X_{i+2})\right.\right.\right.\right.\\
    &\left.\left.\left.\left.-\textup{AD}^{\pi^{*}_{n_{K,b}}}(X_0,\cdots,W_{i+2})\mathbb{I}[i=n_{K,b}-2]\right)\mathbb{I}[\Delta_{i+1}=0]\right)\right.\right. \\
    &\left.\left.+ \Tilde{\rho}_{K-1}^{(i+1)}(X_{0},\cdots,X_{i+1})-C_5(\varepsilon)(NT)^{-\delta}-\textup{AD}^{\pi^{*}_{n_{K,b}}}(X_0,\cdots,W_{i+1})\mathbb{I}[i=n_{K,b}-1]\right)\mathbb{I}[\Delta_{i}=0] \right)\\
&+\Tilde{\rho}_{K}^{(i)}(x^{(0)},\cdots,x^{(i)})  -C_5(\varepsilon)(NT)^{-\delta},
\end{align*}
\end{sizeddisplay}
where the right hand side, after simplification, is equivalent to:
\begin{sizeddisplay}{\footnotesize}\begin{align*}  \Tilde{I}_{i,K}(x^{(0)},\cdots,x^{(i)})
    &\geq \gamma^2\calP_{i}^{\tilde{\pi}_{K-1}}\calP_{i+1}^{\tilde{\pi}_{K-2}}\left(\left(\Tilde{I}_{0,K-2}(X_{i+2})\right)\mathbb{I}[\Delta_{i+1}=1]\right.\\
    &\left.+\left(\Tilde{I}_{1,K-2}(X_{i+1},X_{i+2})\right)\mathbb{I}[\Delta_{i+1}=0]\mathbb{I}[\Delta_{i}=1]\right.\\
    &\left.+\left({\Tilde{I}_{i+2,K-2}(X_0,\cdots,X_{i+2})}-\textup{AD}^{\pi^{*}_{n_{K,b}}}(X_0,\cdots,W_{i+2})\mathbb{I}[i=n_{K,b}-2]\right)\mathbb{I}[\Delta_{i+1}=0]\mathbb{I}[\Delta_{i}=0]\right)\\
    &+\gamma\calP_{i}^{\tilde{\pi}_{K-1}}\left(\left(\Tilde{\rho}_{K-1}^{(0)}(X_{i+1})-C_5(\varepsilon)(NT)^{-\delta}\right) \mathbb{I}[\Delta_{i}=1] \right.\\
    &\left.+ \left(\Tilde{\rho}_{K-1}^{(i+1)}(X_{0},\cdots,X_{i+1})-C_5(\varepsilon)(NT)^{-\delta}-\textup{AD}^{\pi^{*}_{n_{K,b}}}(X_0,\cdots,W_{i+1})\mathbb{I}[i=n_{K,b}-1]\right)\mathbb{I}[\Delta_{i}=0]\right)\\
    &+\Tilde{\rho}_K^{(i)}(x^{(0)},\cdots,x^{(i)}) -C_5(\varepsilon)(NT)^{-\delta}.
\end{align*}\end{sizeddisplay}
For the remaining part of the proof, we use $I_{i,K}$ to denote $I_{i,K}(x^{(0)},\cdots,x^{(i)})$. Then, by pursuing the same procedure $K-1$ times, we can obtain the following:
\begin{sizeddisplay}
{\footnotesize}\begin{align}\label{eq:pfqi final lower bound on Q}
    \Tilde{I}_{i,K}&\geq \prod_{j=1}^{K-1}\left(\gamma\calP_{i+j-1}^{\tilde{\pi}_{K-j}}\right)\left\{\sum_{k=0}^{n_{K,b}}\left(Q^{(k)}_{\pi^{*}_{n_{K,b}},\mathcal{C}}(X_{i+K-1-k},\cdots,X_{i+K-1})- \widehat{R}_{i+K-1}\right) \prod_{l=i+K-1-k}^{i+K-2-k}\mathbb{I}[\Delta_{l}=0]\mathbb{I}[\Delta_{i+K-2-k}=1]\right\}\nonumber\\
    &+\sum_{k=1}^{K-1}\prod_{j=1}^{k}\left(\gamma\calP_{i+j-1}^{\tilde{\pi}_{K-j}}\right)\left\{\left(\Tilde{\rho}_{K-k}^{(0)}(X_{i+k})-C_5(\varepsilon)(NT)^{-\delta}\right)\mathbb{I}[\Delta_{i+k-1}=1]\right.\nonumber\\
    &\left.+\left(\Tilde{\rho}_{K-k}^{(i+k)}(X_{0},\cdots,X_{i+k})-C_5(\varepsilon)(NT)^{-\delta}\right.\right.\\
    &\left.\left.-\textup{AD}^{\pi^{*}_{n_{K,b}}}(X_0,\cdots,W_{i+k})\mathbb{I}[i=n_{K,b}-k]\right)\prod_{l=0}^{i+k-1}\mathbb{I}[\Delta_{l}=0]\mathbb{I}[\Delta_{-1}=1]\mathbb{I}[k\leq n_{K,b}-i]\right.\nonumber\\
    &\left. +\sum_{v=1}^{\min(k-1,n)}\left(\Tilde{\rho}_{K-k}^{(v)}(X_{i+k-v},\cdots,X_{i+k})-C_5(\varepsilon)(NT)^{-\delta}\right.\right.\\
    &\left.\left.-\textup{AD}^{\pi^{*}_{n_{K,b}}}(X_{i+k-v},\cdots,W_{i+k})\mathbb{I}[v=n_{K,b}]\right) \prod_{l=i+k-v}^{i+k-1}\mathbb{I}[\Delta_{l}=0]\mathbb{I}[\Delta_{i+k-v-1}=1]\right\}\nonumber\\
    &+\Tilde{\rho}_K^{(i)}(x^{(0)},\cdots,x^{(i)})-C_5(\varepsilon)(NT)^{-\delta}. \nonumber
\end{align}
\end{sizeddisplay}

On the event $\Omega_P \cap \Omega^r$, Equation~\eqref{eq: lower bound rho pess} implies that, for all $i \in \{0,\ldots,n_{K,b}\}$, we have
\begin{sizeddisplay}
{\footnotesize}
\begin{align}\label{eq:lower PFQI error and UQ}
	&0 \leq \Tilde{\rho}_j^{(i)}(x^{(0)},\cdots,x^{(i)}) \qquad \forall j=1,\cdots,K
\end{align}\end{sizeddisplay}
which holds for all $(x^{(0)},\cdots, x^{(i)}) \in \calX^{\otimes (i+1)} $. Additionally,  $|Q^{(i)}_{\pi^{*}_{n_{K,b}},\mathcal{C}}|$ is uniformly bounded by $\frac{R_{\max}}{1-\gamma}$ $\forall i=0,\cdots,n$ and $|\widehat{R}_t|\leq R_{\max}$ $\forall t\geq0$. Together, they imply:
\begin{sizeddisplay}
{\footnotesize}\begin{align}\label{eq:pfqi final lower bound on Q 1}  
 \Tilde{I}_{i,K}&\geq -\gamma^{K-1}\left(\frac{2-\gamma}{1-\gamma}R_{\max}\right)\nonumber\\
    &+\sum_{k=1}^{K-1}\prod_{j=1}^{k}\left(\gamma\calP_{i+j-1}^{\tilde{\pi}_{K-j}}\right)\left\{\left(-C_5(\varepsilon)(NT)^{-\delta} \right)\mathbb{I}[\Delta_{i+k-1}=1]\right.\nonumber\\
    &\left.\left(-C_5(\varepsilon)(NT)^{-\delta} -\textup{AD}^{\pi^{*}_{n_{K,b}}}(X_0,\cdots,W_{i+k})\mathbb{I}[i=n_{K,b}-k]\right)\prod_{l=0}^{i+k-1}\mathbb{I}[\Delta_{l}=0]\mathbb{I}[\Delta_{-1}=1]\mathbb{I}[k\leq n_{K,b}-i]\right.\nonumber\\
    &\left. +\sum_{v=1}^{\min(k-1,n_{K,b})}\left(-C_5(\varepsilon)(NT)^{-\delta}-\textup{AD}^{\pi^{*}_{n_{K,b}}}(X_{i+k-v},\cdots,W_{i+k})\mathbb{I}[v=n_{K,b}] \right)\prod_{l=i+k-v}^{i+k-1}\mathbb{I}[\Delta_{l}=0]\mathbb{I}[\Delta_{i+k-v-1}=1]\right\}\nonumber\\
    &-C_5(\varepsilon)(NT)^{-\delta}. 
\end{align}
\end{sizeddisplay}
By using Definitions\hyperlink{total_action_discrepancy_iii}{~\ref{def:action_discrepancy} (iii)} and \ref{def: fqi RD definition}, Equation \eqref{eq:pfqi final lower bound on Q 1} can be expressed as follows:

\begin{sizeddisplay}
{\footnotesize}\begin{align}\label{eq:pessimistic final lower bound on Q}
    \Tilde{I}_{i,K}(x^0,\cdots,x^i)
    &\geq -\gamma^{K-1}\left(\frac{2-\gamma}{1-\gamma}R_{\max}\right)-\textup{RD}(NT)-\textup{TAD}^{\tilde{\pi}_K}(x^{0},\cdots,x^{(i)}),
\end{align}
\end{sizeddisplay}
This concludes the proof of Lemma \ref{lm:lower bound on Qs pessimistic}.

\subsection{Proof of Lemma \ref{lemma: aux mitigating}}\label{lm proof aux mitig}

Fix $i$. For any positive definite matrix $M$ and any vector $v$, the quadratic form admits the variational representation
\[
v^\top M^{-1}v \;=\; \sup_{a:\ a^\top M a \le 1}(a^\top v)^2.
\]
Applying this with $M=\Lambda^{(i)}$ and $v=\phi_i(h)$ yields
\[
\phi_i(h)^\top (\Lambda^{(i)})^{-1}\phi_i(h)
\;=\;
\sup_{a:\ a^\top \Lambda^{(i)} a \le 1}\big(a^\top \phi_i(h)\big)^2.
\]
Write $f_a(\cdot):=a^\top \phi_i(\cdot)$. Since $\Lambda^{(i)}\succeq \lambda I$, any $a$ satisfying $a^\top \Lambda^{(i)}a\le 1$ also satisfies $\|a\|_2^2 \le 1/\lambda$. In addition, since $\|\phi_i(\cdot)\|_2\le 1$, we have the uniform bound $\|f_a\|_\infty \le \|a\|_2 \le 1/\sqrt{\lambda}$. Following the same restriction step as in the proof of Theorem~21(1) in \cite{chang2021mitigating}, we consider the localized linear class of bounded functions
\[
\mathcal{F}_i
:=
\left\{f_a:\ a^\top\Lambda^{(i)}a\le 1,\ \|f_a\|_\infty \le 1/\sqrt{\lambda}\right\},
\]
and note that the supremum above is taken over $f_a\in\mathcal{F}_i$.

We next control empirical squared norms by population squared norms uniformly over $\mathcal{F}_i$. Applying Lemma~36 of \cite{chang2021mitigating} (a uniform law of large numbers with localization for squared loss classes), there exists a quantity $\delta^0_{|\mathcal{O}_N^{(i)}|}$ such that, with probability at least $1-\delta$ over $\mathcal{O}_N^{(i)}$, for all $f\in\mathcal{F}_i$,
\[
\frac{1}{|\mathcal{O}_N^{(i)}|}\sum_{j=1}^{|\mathcal{O}_N^{(i)}|} f(\bar H^{(i)}_j)^2
\;\ge\;
\frac12\,\mathbb{E}_{H^{(i)}\sim d_i^\mu}\!\big[f(H^{(i)})^2\big]
\;-\;\frac12\big(\delta^0_{|\mathcal{O}_N^{(i)}|}\big)^2.
\]
Lemma~37 of \cite{chang2021mitigating} upper bounds the critical radius of linear function classes in terms of the effective dimension (here $\mathrm{rank}(\Sigma_i^\mu)=r^{(i)}$), which implies
\[
|\mathcal{O}_N^{(i)}|\big(\delta^0_{|\mathcal{O}_N^{(i)}|}\big)^2
\;\le\;
c\big(r^{(i)}+\log(c'/\delta)\big)
\]
for universal constants $c,c'>0$ (absorbing the boundedness constants into $c,c'$).

Now fix any $a$ with $a^\top\Lambda^{(i)}a\le 1$. Expanding the constraint gives
\[
\lambda\|a\|_2^2 + \sum_{j=1}^{|\mathcal{O}_N^{(i)}|} f_a(\bar H^{(i)}_j)^2 \;\le\; 1.
\]
On the above high-probability event, applying the uniform inequality to $f_a$ yields
\[
\sum_{j=1}^{|\mathcal{O}_N^{(i)}|} f_a(\bar H^{(i)}_j)^2
\;\ge\;
\frac{|\mathcal{O}_N^{(i)}|}{2}\,\mathbb{E}_{d_i^\mu}\!\big[f_a(H^{(i)})^2\big]
\;-\;\frac{|\mathcal{O}_N^{(i)}|}{2}\big(\delta^0_{|\mathcal{O}_N^{(i)}|}\big)^2,
\]
and therefore
\[
\lambda\|a\|_2^2 + \frac{|\mathcal{O}_N^{(i)}|}{2}\,\mathbb{E}_{d_i^\mu}\!\big[(a^\top\phi_i(H^{(i)}))^2\big]
\;\le\;
1 + \frac{|\mathcal{O}_N^{(i)}|}{2}\big(\delta^0_{|\mathcal{O}_N^{(i)}|}\big)^2.
\]
Using $\mathbb{E}_{d_i^\mu}[(a^\top\phi_i(H^{(i)}))^2]=a^\top \Sigma_i^\mu a$, we obtain
\[
a^\top\big(\lambda I + |\mathcal{O}_N^{(i)}|\,\Sigma_i^\mu\big)a
\;\le\;
2 + |\mathcal{O}_N^{(i)}|\big(\delta^0_{|\mathcal{O}_N^{(i)}|}\big)^2.
\]
Combining this with the variational characterization, for every $h$ we have
\[
\sup_{a:\ a^\top\Lambda^{(i)}a\le 1}(a^\top\phi_i(h))^2
\;\le\;
\sup_{a:\ a^\top(\lambda I + |\mathcal{O}_N^{(i)}|\Sigma_i^\mu)a\le 2+|\mathcal{O}_N^{(i)}|(\delta^0_{|\mathcal{O}_N^{(i)}|})^2}(a^\top\phi_i(h))^2.
\]
The right-hand side is a quadratic maximization and equals
\[
\big(2+|\mathcal{O}_N^{(i)}|\big(\delta^0_{|\mathcal{O}_N^{(i)}|}\big)^2\big)\,
\phi_i(h)^\top\big(\lambda I + |\mathcal{O}_N^{(i)}|\Sigma_i^\mu\big)^{-1}\phi_i(h).
\]
Finally, substituting the bound $|\mathcal{O}_N^{(i)}|(\delta^0_{|\mathcal{O}_N^{(i)}|})^2 \le c(r^{(i)}+\log(c'/\delta))$ and absorbing constants into universal $c_1,c_2$ yields, uniformly over $h$,
\[
\phi_i(h)^\top (\Lambda^{(i)})^{-1}\phi_i(h)
\;\le\;
c_1\big(r^{(i)}+\log(c_2/\delta)\big)\,
\phi_i(h)^\top\big(|\mathcal{O}_N^{(i)}|\,\Sigma_i^\mu+\lambda I\big)^{-1}\phi_i(h),
\]
which holds with probability at least $1-\delta$ for a fix $i$. Then by letting $\delta=\epsilon / (n_{K,b} + 1)$ and using union bound, the bound holds for each $i \in \{0,\cdots,n_{K,b}\}$ with probability at least $1-\epsilon$.

\subsection{Proof of Lemma \ref{lemma: nkb and action relation}}\label{subsec: nkb and action relation}
Throughout this proof, we assume that, in the component of $W_{t+j}$, we have $\Delta_{t+j-1}=0$ for all $j=1,\cdots,n$, and in the component of $W_{t}$, we have $\Delta_{t-1}=1$. Additionally, to streamline the derivations, we assume, without loss of generality, that all policies in the policy class $\Pi_n$ are deterministic where the characterization of $\Pi_n$ is given in Definition \ref{def: set of policies such that at most n cons censoring observed}. This assumption is made for simplicity and does not affect the generality of the results.




\textit{Proof:}

We aim to show that for all $t \geq 0$,
\begin{align}
    &\sup_{\pi \in \Pi_n} \mathbb{E}^{\pi}\left[ R_{t+n} + \gamma I[\Delta_{t+n}=1] V^{(0)}_{*,\mathcal{C}}(W_{t+n+1}) \mid H_{t-1}, W_t, A_t, \ldots, W_{t+n} \right] \nonumber \\
    &\leq \sup_{a \in \textup{UP}^{(n)}(W_{t+n}, \ldots, A_{t}, W_{t})} \mathbb{E}\left[ R_{t+n} + \gamma I[\Delta_{t+n}=1] V^{(0)}_{*,\mathcal{C}}(W_{t+n+1}) \mid W_t, A_t, \ldots, W_{t+n}, a \right],
\end{align}
where the characterization of $\textup{UP}^{(n)}$ is given in Definition \ref{def: actions leading uncensored state rigorous}. The rationale for discarding $H_{t-1}$ in the expectation above lies in the fact that, under the observed process, the transition to the next state and the immediate reward depend solely on the sequence of past states and actions up to the most recent non-censored time point. This follows from the conditional independence structure of the underlying MDP.

Next, we decompose the set $\Pi_n$ into the union of two disjoint sets:
$$\Pi_n = \Pi^{H}_{n} \cup \Pi^{M}_{n},$$
where the superscript $H$ stands for history-dependent policies and $M$ stands for Markovian policies. This implies the following for all $t \geq 0$:
\begin{align*}
    &\sup_{\pi \in \Pi_n} \mathbb{E}^{\pi}\left[ R_{t+n} + \gamma I[\Delta_{t+n}=1] V^{(0)}_{*,\mathcal{C}}(W_{t+n+1}) \mid H_{t-1}, W_t, A_t, \ldots, W_{t+n} \right] \\
    &\triangleq \sup_{\pi \in \Pi_n} f^{\pi}(H_{t-1}, W_t, A_t, \ldots, W_{t+n}) \\
    &= \max\left(\sup_{\pi \in \Pi^{M}_{n}} f^{\pi}(H_{t-1}, W_t, A_t, \ldots, W_{t+n}), \sup_{\pi \in \Pi^{H}_{n}} f^{\pi}(H_{t-1}, W_t, A_t, \ldots, W_{t+n})\right).
\end{align*}

Any $\pi \in \Pi_{n}^{M}$ uses all the history up to the last non-censored time point. Therefore, we have for all $t \geq 0$ and $\pi \in \Pi_n$,
\begin{align}
    &f^{\pi}(H_{t-1}, W_t, A_t, \ldots, W_{t+n}) \nonumber \\
    &= \mathbb{E}^{\pi}\left[ R_{t+n} + \gamma I[\Delta_{t+n}=1] V^{(0)}_{*,\mathcal{C}}(W_{t+n+1}) \mid H_{t-1}, W_t, A_t, \ldots, W_{t+n} \right] \nonumber \\
    &= \mathbb{E}\left[ R_{t+n} + \gamma I[\Delta_{t+n}=1] V^{(0)}_{*,\mathcal{C}}(W_{t+n+1}) \mid H_{t-1}, W_t, A_t, \ldots, W_{t+n}, \pi(W_{t+n}, \ldots, A_t, W_t) \right] \nonumber \\
    &= \mathbb{E}\left[ R_{t+n} + \gamma I[\Delta_{t+n}=1] V^{(0)}_{*,\mathcal{C}}(W_{t+n+1}) \mid W_t, A_t, \ldots, W_{t+n}, \pi(W_{t+n}, \ldots, A_t, W_t) \right] \nonumber \\
    &= f^{\pi}(W_t, A_t, \ldots, W_{t+n}).
\end{align}

Furthermore, for any $\pi \in \Pi_{n}^{M}$, $\pi$ will output an action that belongs to the set $\textup{UP}^{(n)}(W_{t+n}, \ldots, A_{t}, W_{t})$. Therefore,
\begin{align}\label{eq: new lemma-1}
    \sup_{\pi \in \Pi^{M}_{n}} f^{\pi}(H_{t-1}, W_t, A_t, \ldots, W_{t+n})
    &= \sup_{\pi \in \Pi^{M}_{n}} f^{\pi}(W_t, A_t, \ldots, W_{t+n}) \nonumber \\
    &= \sup_{a \in \textup{UP}^{(n)}(W_{t+n}, \ldots, A_{t}, W_{t})} f(W_t, A_t, \ldots, W_{t+n}, a).
\end{align}

Next, let $\pi^{*}_{H} \in \argmax_{\pi \in \Pi^{H}_{n}} f^{\pi}(H_{t-1}, W_t, A_t, \ldots, W_{t+n})$. This policy, $\pi^{*}_{H}$, uses all the history to output an action as $\pi^{*}_{H} \in \Pi_{n}^{H}$. Given that $W_t, A_t, \ldots, W_{t+n}$ are fixed, for each realization of $H_{t-1}$, $\pi^{*}_{H}$ outputs an action that belongs to $\textup{UP}^{(n)}(W_{t+n}, \ldots, A_{t}, W_{t})$. Let $\calH_{t-1}$ denote the space of all possible realizations of $H_{t-1}$. Correspondingly, we can define the following set:
\begin{align*}
    \textup{UP}^{(n)}_{H}(W_{t+n}, \ldots, A_{t}, W_{t}) = \left\{ \pi^{*}_{H}(h, W_t, A_t, \ldots, W_{t+n}), \forall h \in \calH_{t-1} \right\}.
\end{align*}

Observe that $\textup{UP}^{(n)}_{H}(W_{t+n}, \ldots, A_{t}, W_{t}) \subseteq \textup{UP}^{(n)}(W_{t+n}, \ldots, A_{t}, W_{t})$. This is implied by the fact that any $\pi \in \Pi^{H}_{n}$ will output an action in $A(W_{t+n}, A_{t+n-1}, W_{t+n-1}, \ldots, A_{t}, W_{t})$. Therefore,
\begin{align}\label{eq: new lemma-2}
    &\sup_{\pi \in \Pi^{H}_{n}} f^{\pi}(H_{t-1}, W_t, A_t, \ldots, W_{t+n}) \nonumber \\
    &= f^{\pi^{*}_{H}}(H_{t-1}, W_t, A_t, \ldots, W_{t+n}) \nonumber \\
    &= \sup_{a \in \textup{UP}^{(n)}_{H}(W_{t+n}, \ldots, A_{t}, W_{t})} \mathbb{E}\left[ R_{t+n} + \gamma I[\Delta_{t+n}=1] V^{(0)}_{*,\mathcal{C}}(W_{t+n+1}) \mid H_{t-1}, W_t, A_t, \ldots, W_{t+n}, a \right] \nonumber \\
    &= \sup_{a \in \textup{UP}^{(n)}_{H}(W_{t+n}, \ldots, A_{t}, W_{t})} \mathbb{E}\left[ R_{t+n} + \gamma I[\Delta_{t+n}=1] V^{(0)}_{*,\mathcal{C}}(W_{t+n+1}) \mid W_t, A_t, \ldots, W_{t+n}, a \right] \nonumber \\
    &\leq \sup_{a \in \textup{UP}^{(n)}(W_{t+n}, \ldots, A_{t}, W_{t})} \mathbb{E}\left[ R_{t+n} + \gamma I[\Delta_{t+n}=1] V^{(0)}_{*,\mathcal{C}}(W_{t+n+1}) \mid W_t, A_t, \ldots, W_{t+n}, a \right] \nonumber \\
    &= \sup_{a \in \textup{UP}^{(n)}(W_{t+n}, \ldots, A_{t}, W_{t})} f(W_t, A_t, \ldots, W_{t+n}, a).
\end{align}

Consequently, combining Equations \eqref{eq: new lemma-1} and \eqref{eq: new lemma-2} implies:
\begin{align*}
    &\sup_{\pi \in \Pi_n} \mathbb{E}^{\pi}\left[ R_{t+n} + \gamma I[\Delta_{t+n}=1] V^{(0)}_{*,\mathcal{C}}(W_{t+n+1}) \mid H_{t-1}, W_t, A_t, \ldots, W_{t+n} \right] \\
    &= \max\left(\sup_{\pi \in \Pi^{H}_{n}} f^{\pi}(H_{t-1}, W_t, A_t, \ldots, W_{t+n}), \sup_{\pi \in \Pi^{M}_{n}} f^{\pi}(H_{t-1}, W_t, A_t, \ldots, W_{t+n}) \right) \\
    &\leq \sup_{a \in \textup{UP}^{(n)}(W_{t+n}, \ldots, A_{t}, W_{t})} f(W_t, A_t, \ldots, W_{t+n}, a) \\
    &= \sup_{a \in \textup{UP}^{(n)}(W_{t+n}, \ldots, A_{t}, W_{t})} \mathbb{E}\left[ R_{t+n} + \gamma I[\Delta_{t+n}=1] V^{(0)}_{*,\mathcal{C}}(W_{t+n+1}) \mid W_t, A_t, \ldots, W_{t+n}, a \right],
\end{align*}
which concludes the  proof of Lemma \ref{lemma: nkb and action relation}.

\clearpage

%
%
%



\bibliographystyle{agsm}

\bibliography{cleaned_reference} 

@article{feng2014dynamic,
  title={Dynamic inventory--pricing control under backorder: Demand estimation and policy optimization},
  author={Feng, Qi and Luo, Sirong and Zhang, Dan},
  journal={Manufacturing \& Service Operations Management},
  volume={16},
  number={1},
  pages={149--160},
  year={2014},
  publisher={INFORMS}
}

@inproceedings{jin2020provably,
  title={Provably efficient reinforcement learning with linear function approximation},
  author={Jin, Chi and Yang, Zhuoran and Wang, Zhaoran and Jordan, Michael I},
  booktitle={Conference on learning theory},
  pages={2137--2143},
  year={2020},
  organization={PMLR}
}

@article{xie2024vc,
  title={Vc theory for inventory policies},
  author={Xie, Yaqi and Ma, Will and Xin, Linwei},
  journal={arXiv preprint arXiv:2404.11509},
  year={2024}
}

@article{gong2024bandits,
  title={Bandits atop reinforcement learning: Tackling online inventory models with cyclic demands},
  author={Gong, Xiao-Yue and Simchi-Levi, David},
  journal={Management Science},
  volume={70},
  number={9},
  pages={6139--6157},
  year={2024},
  publisher={INFORMS}
}

@inproceedings{agrawal2019learning,
  title={Learning in structured mdps with convex cost functions: Improved regret bounds for inventory management},
  author={Agrawal, Shipra and Jia, Randy},
  booktitle={Proceedings of the 2019 ACM Conference on Economics and Computation},
  pages={743--744},
  year={2019}
}

@article{chen2006optimal,
  title={Optimal pricing and inventory control policy in periodic-review systems with fixed ordering cost and lost sales},
  author={Chen, Youhua and Ray, Saibal and Song, Yuyue},
  journal={Naval Research Logistics (NRL)},
  volume={53},
  number={2},
  pages={117--136},
  year={2006},
  publisher={Wiley Online Library}
}

@article{tropp2012user,
  title={User-friendly tail bounds for sums of random matrices},
  author={Tropp, Joel A},
  journal={Foundations of computational mathematics},
  volume={12},
  number={4},
  pages={389--434},
  year={2012},
  publisher={Springer}
}

@inproceedings{shi2022pessimistic,
  title={Pessimistic q-learning for offline reinforcement learning: Towards optimal sample complexity},
  author={Shi, Laixi and Li, Gen and Wei, Yuting and Chen, Yuxin and Chi, Yuejie},
  booktitle={International conference on machine learning},
  pages={19967--20025},
  year={2022},
  organization={PMLR}
}

@article{huh2011adaptive,
  title={Adaptive data-driven inventory control with censored demand based on Kaplan-Meier estimator},
  author={Huh, Woonghee Tim and Levi, Retsef and Rusmevichientong, Paat and Orlin, James B},
  journal={Operations Research},
  volume={59},
  number={4},
  pages={929--941},
  year={2011},
  publisher={INFORMS}
}

@article{lyu2024ucb,
  title={Ucb-type learning algorithms with kaplan--meier estimator for lost-sales inventory models with lead times},
  author={Lyu, Chengyi and Zhang, Huanan and Xin, Linwei},
  journal={Operations Research},
  volume={72},
  number={4},
  pages={1317--1332},
  year={2024},
  publisher={INFORMS}
}

@inproceedings{yang2020reinforcement,
  title={Reinforcement learning in feature space: Matrix bandit, kernels, and regret bound},
  author={Yang, Lin and Wang, Mengdi},
  booktitle={International Conference on Machine Learning},
  pages={10746--10756},
  year={2020},
  organization={PMLR}
}

@inproceedings{jiang2017contextual,
  title={Contextual decision processes with low bellman rank are pac-learnable},
  author={Jiang, Nan and Krishnamurthy, Akshay and Agarwal, Alekh and Langford, John and Schapire, Robert E},
  booktitle={International Conference on Machine Learning},
  pages={1704--1713},
  year={2017},
  organization={PMLR}
}

@article{chen2017efficient,
  title={Efficient algorithms for the dynamic pricing problem with reference price effect},
  author={Chen, Xin and Hu, Peng and Hu, Zhenyu},
  journal={Management Science},
  volume={63},
  number={12},
  pages={4389--4408},
  year={2017},
  publisher={INFORMS}
}

@article{chen2020data,
  title={Data-based dynamic pricing and inventory control with censored demand and limited price changes},
  author={Chen, Boxiao and Chao, Xiuli and Wang, Yining},
  journal={Operations Research},
  volume={68},
  number={5},
  pages={1445--1456},
  year={2020},
  publisher={INFORMS}
}

@book{porteus2002foundations,
  title={Foundations of stochastic inventory theory},
  author={Porteus, Evan L},
  year={2002},
  publisher={Stanford University Press}
}

@article{chen2004coordinating,
  title={Coordinating inventory control and pricing strategies with random demand and fixed ordering cost: The finite horizon case},
  author={Chen, Xin and Simchi-Levi, David},
  journal={Operations research},
  volume={52},
  number={6},
  pages={887--896},
  year={2004},
  publisher={INFORMS}
}

@article{sethi1997optimality,
  title={Optimality of (s, S) policies in inventory models with Markovian demand},
  author={Sethi, Suresh P and Cheng, Feng},
  journal={Operations Research},
  volume={45},
  number={6},
  pages={931--939},
  year={1997},
  publisher={INFORMS}
}

@article{araman2009dynamic,
  title={Dynamic pricing for nonperishable products with demand learning},
  author={Araman, Victor F and Caldentey, Ren{\'e}},
  journal={Operations research},
  volume={57},
  number={5},
  pages={1169--1188},
  year={2009},
  publisher={INFORMS}
}

@article{petruzzi2002dynamic,
  title={Dynamic pricing and inventory control with learning},
  author={Petruzzi, Nicholas C and Dada, Maqbool},
  journal={Naval Research Logistics (NRL)},
  volume={49},
  number={3},
  pages={303--325},
  year={2002},
  publisher={Wiley Online Library}
}

@article{coemans2022bias,
  title={Bias by censoring for competing events in survival analysis},
  author={Coemans, Maarten and Verbeke, Geert and D{\"o}hler, Bernd and S{\"u}sal, Caner and Naesens, Maarten},
  journal={bmj},
  volume={378},
  year={2022},
  publisher={British Medical Journal Publishing Group}
}

@article{qi2022offline,
  title={Offline Feature-Based Pricing under Censored Demand: A Causal Inference Approach},
  author={Qi, Zhengling and Tang, Jingwen and Fang, Ethan and Shi, Cong},
  journal={Available at SSRN 4040305},
  year={2022}

}

@article{dabrowska1989uniform,
  title={Uniform consistency of the kernel conditional Kaplan-Meier estimate},
  author={Dabrowska, Dorota M},
  journal={The Annals of Statistics},
  pages={1157--1167},
  year={1989},
  publisher={JSTOR}
}

@article{khardani2014nonparametric,
  title={Nonparametric conditional density estimation for censored data based on a recursive kernel},
  author={Khardani, Salah and Semmar, Sihem},
  journal={Electronic Journal of Statistics},
  volume={8},
  pages={2541--2556},
  year={2014}
}

@article{chang2021mitigating,
  title={Mitigating covariate shift in imitation learning via offline data with partial coverage},
  author={Chang, Jonathan and Uehara, Masatoshi and Sreenivas, Dhruv and Kidambi, Rahul and Sun, Wen},
  journal={Advances in Neural Information Processing Systems},
  volume={34},
  pages={965--979},
  year={2021}
}

@inproceedings{riedmiller2005neural,
  title={Neural fitted Q iteration--first experiences with a data efficient neural reinforcement learning method},
  author={Riedmiller, Martin},
  booktitle={Machine learning: ECML 2005: 16th European conference on machine learning, Porto, Portugal, October 3-7, 2005. proceedings 16},
  pages={317--328},
  year={2005},
  organization={Springer}
}

@article{qin2019data,
  title={Data-Driven Approximation Schemes for Joint Pricing and Inventory Control Models},
  author={Qin, Hanzhang and Simchi-Levi, David and Wang, Li},
  journal={Available at SSRN 3354358},
  year={2019}
}

@article{levi2007provably,
  title={Provably near-optimal sampling-based policies for stochastic inventory control models},
  author={Levi, Retsef and Roundy, Robin O and Shmoys, David B},
  journal={Mathematics of Operations Research},
  volume={32},
  number={4},
  pages={821--839},
  year={2007},
  publisher={INFORMS}
}

@article{cheung2019sampling,
  title={Sampling-based approximation schemes for capacitated stochastic inventory control models},
  author={Cheung, Wang Chi and Simchi-Levi, David},
  journal={Mathematics of Operations Research},
  volume={44},
  number={2},
  pages={668--692},
  year={2019},
  publisher={INFORMS}
}

@article{ban2019big,
  title={The big data newsvendor: Practical insights from machine learning},
  author={Ban, Gah-Yi and Rudin, Cynthia},
  journal={Operations Research},
  volume={67},
  number={1},
  pages={90--108},
  year={2019},
  publisher={INFORMS}
}

@article{bu2023offline,
  title={Offline pricing and demand learning with censored data},
  author={Bu, Jinzhi and Simchi-Levi, David and Wang, Li},
  journal={Management Science},
  volume={69},
  number={2},
  pages={885--903},
  year={2023},
  publisher={INFORMS}
}

@inproceedings{fujimoto2019off,
  title={Off-policy deep reinforcement learning without exploration},
  author={Fujimoto, Scott and Meger, David and Precup, Doina},
  booktitle={International conference on machine learning},
  pages={2052--2062},
  year={2019},
  organization={PMLR}
}

@article{liu2020provably,
  title={Provably good batch off-policy reinforcement learning without great exploration},
  author={Liu, Yao and Swaminathan, Adith and Agarwal, Alekh and Brunskill, Emma},
  journal={Advances in neural information processing systems},
  volume={33},
  pages={1264--1274},
  year={2020}
}

@article{rashidinejad2021bridging,
  title={Bridging offline reinforcement learning and imitation learning: A tale of pessimism},
  author={Rashidinejad, Paria and Zhu, Banghua and Ma, Cong and Jiao, Jiantao and Russell, Stuart},
  journal={Advances in Neural Information Processing Systems},
  volume={34},
  pages={11702--11716},
  year={2021}
}

@article{zanette2021provable,
  title={Provable benefits of actor-critic methods for offline reinforcement learning},
  author={Zanette, Andrea and Wainwright, Martin J and Brunskill, Emma},
  journal={Advances in neural information processing systems},
  volume={34},
  pages={13626--13640},
  year={2021}
}

@inproceedings{zhan2022offline,
  title={Offline reinforcement learning with realizability and single-policy concentrability},
  author={Zhan, Wenhao and Huang, Baihe and Huang, Audrey and Jiang, Nan and Lee, Jason},
  booktitle={Conference on Learning Theory},
  pages={2730--2775},
  year={2022},
  organization={PMLR}
}

@article{buckman2020importance,
  title={The importance of pessimism in fixed-dataset policy optimization},
  author={Buckman, Jacob and Gelada, Carles and Bellemare, Marc G},
  journal={arXiv preprint arXiv:2009.06799},
  year={2020}
}

@article{fu2022offline,
  title={Offline reinforcement learning with instrumental variables in confounded markov decision processes},
  author={Fu, Zuyue and Qi, Zhengling and Wang, Zhaoran and Yang, Zhuoran and Xu, Yanxun and Kosorok, Michael R},
  journal={arXiv preprint arXiv:2209.08666},
  year={2022}
}

@article{schulman2017proximal,
  title={Proximal policy optimization algorithms},
  author={Schulman, John and Wolski, Filip and Dhariwal, Prafulla and Radford, Alec and Klimov, Oleg},
  journal={arXiv preprint arXiv:1707.06347},
  year={2017}
}

@article{wang2020statistical,
	author = {Wang, Ruosong and Foster, Dean P and Kakade, Sham M},
	journal = {arXiv preprint arXiv:2010.11895},
	title = {What are the Statistical Limits of Offline RL with Linear Function Approximation?},
	year = {2020}}

@article{ernst2005tree,
  title={Tree-based batch mode reinforcement learning},
  author={Ernst, Damien and Geurts, Pierre and Wehenkel, Louis},
  journal={Journal of Machine Learning Research},
  volume={6},
  pages={503--556},
  year={2005},
  publisher={Microtome Publishing}
}

@article{levine2020offline,
	author = {Levine, Sergey and Kumar, Aviral and Tucker, George and Fu, Justin},
	journal = {arXiv preprint arXiv:2005.01643},
	title = {Offline reinforcement learning: Tutorial, review, and perspectives on open problems},
	year = {2020}}

@article{kaplan1958nonparametric,
  title={Nonparametric estimation from incomplete observations},
  author={Kaplan, Edward L and Meier, Paul},
  journal={Journal of the American statistical association},
  volume={53},
  number={282},
  pages={457--481},
  year={1958},
  publisher={Taylor \& Francis}
}

@article{xie2021bellman,
  title={Bellman-consistent Pessimism for Offline Reinforcement Learning},
  author={Xie, Tengyang and Cheng, Ching-An and Jiang, Nan and Mineiro, Paul and Agarwal, Alekh},
  journal={arXiv preprint arXiv:2106.06926},
  year={2021}
}

@book{kleinbaum2010survival,
  title={Survival analysis},
  author={Kleinbaum, David G and Klein, Mitchel},
  volume={3},
  year={2010},
  publisher={Springer}
}

@book{bertsekas1995dynamic,
	author = {Bertsekas, Dimitri P},

	publisher = {Athena scientific Belmont, MA},
	title = {Dynamic programming and optimal control},
	volume = {1},
	year = {1995}}

@article{ernst_tree-based_2005,
	author = {Ernst, Damien and Geurts, Pierre and Wehenkel, Louis and Littman, L.},
	journal = {Journal of Machine Learning Research},
	pages = {503--556},
	title = {Tree-based batch mode reinforcement learning},
	volume = {6},
	year = {2005}}

@book{reiss1997statistical,
  title={Statistical analysis of extreme values},
  author={Reiss, Rolf-Dieter and Thomas, Michael and Reiss, RD},
  volume={2},
  year={1997},
  publisher={Springer}
}

@book{coles2001introduction,
  title={An introduction to statistical modeling of extreme values},
  author={Coles, Stuart and Bawa, Joanna and Trenner, Lesley and Dorazio, Pat},
  volume={208},
  year={2001},
  publisher={Springer}
}

@book{sutton2018reinforcement,
	title={Reinforcement learning: An introduction},
	author={Sutton, Richard S and Barto, Andrew G},
	year={2018},
	publisher={MIT press}
}

@article{feng2022developing,
  title={Developing operations management data analytics},
  author={Feng, Qi and Shanthikumar, J George},
  journal={Production and Operations Management},
  volume={31},
  number={12},
  pages={4544--4557},
  year={2022},
  publisher={SAGE Publications Sage CA: Los Angeles, CA}
}

@article{miao2023personalized,
  title={Personalized pricing with invalid instrumental variables: Identification, estimation, and policy learning},
  author={Miao, Rui and Qi, Zhengling and Shi, Cong and Lin, Lin},
  journal={arXiv preprint arXiv:2302.12670},
  year={2023}
}

@article{wang2023estimation,
  title={Estimation of high-dimensional contextual pricing models with nonparametric price confounders},
  author={Wang, Yining and Liu, Quanquan},
  journal={Available at SSRN 4482748},
  year={2023}
}

@article{lin2022data,
  title={Data-driven newsvendor problems regularized by a profit risk constraint},
  author={Lin, Shaochong and Chen, Youhua and Li, Yanzhi and Shen, Zuo-Jun Max},
  journal={Production and Operations Management},
  volume={31},
  number={4},
  pages={1630--1644},
  year={2022},
  publisher={SAGE Publications Sage CA: Los Angeles, CA}
}

@article{harsha2021prescriptive,
  title={A prescriptive machine-learning framework to the price-setting newsvendor problem},
  author={Harsha, Pavithra and Natarajan, Ramesh and Subramanian, Dharmashankar},
  journal={Informs Journal on Optimization},
  volume={3},
  number={3},
  pages={227--253},
  year={2021},
  publisher={INFORMS}
}

@article{liu2023solving,
  title={Solving data-driven newsvendor pricing problems with decision-dependent effect},
  author={Liu, Wenxuan and Zhang, Zhihai},
  journal={arXiv preprint arXiv:2304.13924},
  year={2023}
}

@article{kumar2020conservative,
  title={Conservative q-learning for offline reinforcement learning},
  author={Kumar, Aviral and Zhou, Aurick and Tucker, George and Levine, Sergey},
  journal={arXiv preprint arXiv:2006.04779},
  year={2020}
}

@inproceedings{jin2021pessimism,
  title={Is Pessimism Provably Efficient for Offline RL?},
  author={Jin, Ying and Yang, Zhuoran and Wang, Zhaoran},
  booktitle={International Conference on Machine Learning},
  pages={5084--5096},
  year={2021},
  organization={PMLR}
}




\end{document}